\documentclass[10pt,twocolumn,letterpaper]{article}

\usepackage{3dv}
\usepackage{times}
\usepackage{epsfig}
\usepackage{graphicx}
\usepackage{amsmath}
\usepackage{amssymb}
\usepackage{comment}
\usepackage[dvipsnames]{xcolor}
\usepackage{booktabs}
\usepackage{enumitem}
\usepackage{multirow}
\usepackage{arydshln}
\usepackage{subcaption}
\usepackage{authblk}

\usepackage{slashbox}
\usepackage{stfloats}

\usepackage[pagebackref=true,breaklinks=true,letterpaper=true,colorlinks,bookmarks=false]{hyperref}

\threedvfinalcopy 

\def\threedvPaperID{40} 

\ifthreedvfinal\fi

\graphicspath{{figures/}}

\newcommand{\boldparagraph}[1]{\vspace{0.2em}\noindent{\bf #1} }

\definecolor{highlight2nd}{rgb}{0.19, 0.55, 0.91}  

\newcommand{\kaplan}{KAPLAN}

\let\oldnocite\nocite
\makeatletter
\renewcommand*{\nocite}[1]{\oldnocite{#1}\Hy@backout{#1}}
\makeatother

\newcommand{\vect}[1]{\boldsymbol{#1}}

\newcommand{\nR}{\mathbb{R}}                

\newcommand{\nCompPC}{\mathcal{P}_c}          
\newcommand{\nIncompPC}{\mathcal{P}_i}        
\newcommand{\nMissPC}{\mathcal{P}_m}          

\newcommand{\numQuery}{Q} 

\newcommand{\nK}{\mathit{K}}                 
\newcommand{\nRes}{R}                        
\newcommand{\nNumChannelsPlane}{\textit{c}}  
\newcommand{\nThreshHeurist}{\tau}   

\newcommand{\nLevelCoarse}{\boldsymbol{\ell^0}}     
\newcommand{\nLevelMedium}{\boldsymbol{\ell^1}}     
\newcommand{\nLevelFine}{\boldsymbol{\ell^2}}       

\newcommand{\nLoss}{\mathcal{L}}                
\newcommand{\nLossD}{\mathcal{L}_d}             
\newcommand{\nLossVf}{\mathcal{L}_{v}}          
\newcommand{\nLossN}{\mathcal{L}_n}             
\newcommand{\nWeightD}{\mathrm{\lambda}_d}      
\newcommand{\nWeightVf}{\mathrm{\lambda}_{v}}   
\newcommand{\nWeightN}{\mathrm{\lambda}_n}      
\newcommand{\nValidPix}{\mathit{j}}             

\newcommand{\nPredNormal}{\vect{\hat{N}}}       
\newcommand{\nGtNormal}{\vect{N}}               

\begin{document}

\title{KAPLAN: A 3D Point Descriptor for Shape Completion}

\author[1]{Audrey Richard}
\author[1]{Ian Cherabier}
\author[1]{Martin R. Oswald}

\author[1,2]{Marc Pollefeys}
\author[1]{Konrad Schindler}

\affil[1]{ETH Z{\"u}rich, $^{2}$Microsoft Mixed Reality \& AI Z{\"u}rich Lab}

\renewcommand\Authands{ and }


\maketitle

\thispagestyle{plain}
\pagestyle{plain}
\ifthreedvfinal\thispagestyle{empty}\fi

\begin{abstract}
  We present a novel 3D shape completion method that operates directly on unstructured point clouds, thus avoiding resource-intensive data structures like voxel grids.
  To this end, we introduce KAPLAN, a 3D point descriptor that aggregates local shape information via a series of 2D convolutions.
  The key idea is to project the points in a local neighborhood onto multiple planes with different orientations. In each of those planes, point properties like normals or point-to-plane distances are aggregated into a 2D grid and abstracted into a feature representation with an efficient 2D convolutional encoder.
Since all planes are encoded jointly, the resulting representation nevertheless can capture their correlations and retains knowledge about the underlying 3D shape, without expensive 3D convolutions. 
Experiments on public datasets show that KAPLAN
achieves state-of-the-art performance for 3D shape completion.

%
 
\end{abstract}

\section{Introduction}
%
Common 3D sensing technologies like range cameras, laser scanners or multi-view stereo systems deliver \emph{3D point clouds}.
These point clouds then constitute the input for downstream computer vision and graphics tasks across a wide range of applications.
As such, irregular collections (``clouds'') of 3D points have become a fundamental 3D data representation.
Due to deficiencies of the acquisition process (occlusions, specularities, low albedo, matching ambiguities, etc.), real point clouds are typically incomplete: they have ``holes'' -- local regions where no point samples of the object surface are available.

\emph{Shape completion} is the task to fill such holes, so as to obtain a complete representation of an object's shape.
The aim of our work is to perform shape completion directly on the point cloud, without having to transform it into a memory-demanding volumetric shape representation (e.g., a global voxel grid or signed distance function).
In other words, we must learn the shape statistics of local surface patches, so that we can then sample points on the expected surface and fill the holes.
With the natural decision to center the patches on existing 3D points, this is equivalent to predicting \emph{3D point descriptors} from incomplete data.


\begin{figure}[t]
  \vspace{-5pt}
  \centering
  \scriptsize
  \setlength{\tabcolsep}{0.5mm}
  \newcommand{\sz}{0.12}
  \newcommand{\insz}{0.10}
  \begin{tabular}{ccccc}
  \includegraphics[width=\sz\textwidth]{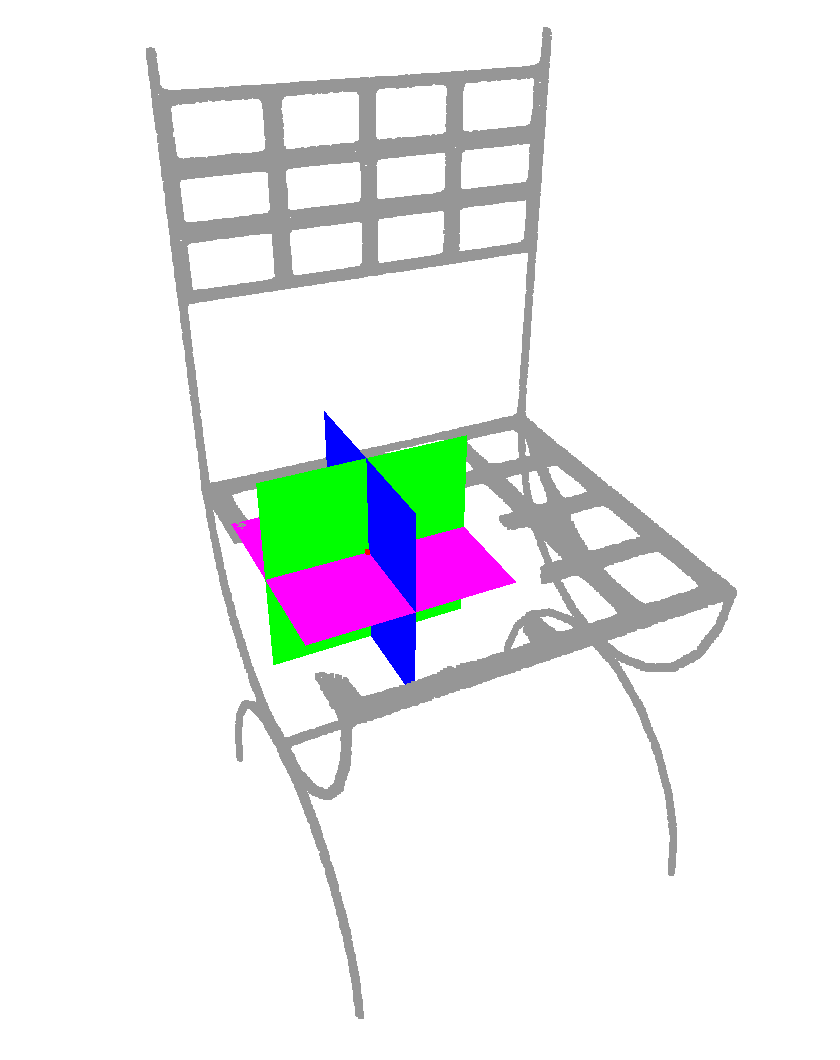} &
  \includegraphics[width=\sz\textwidth]{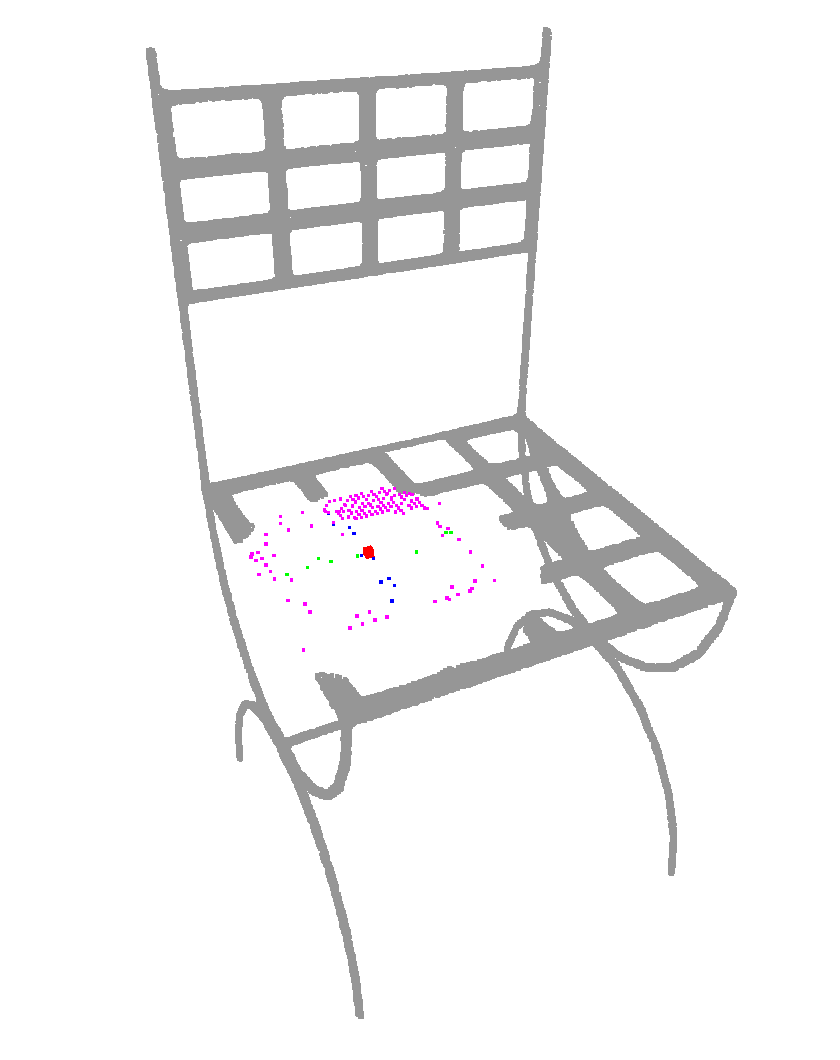} &
  \includegraphics[width=\sz\textwidth]{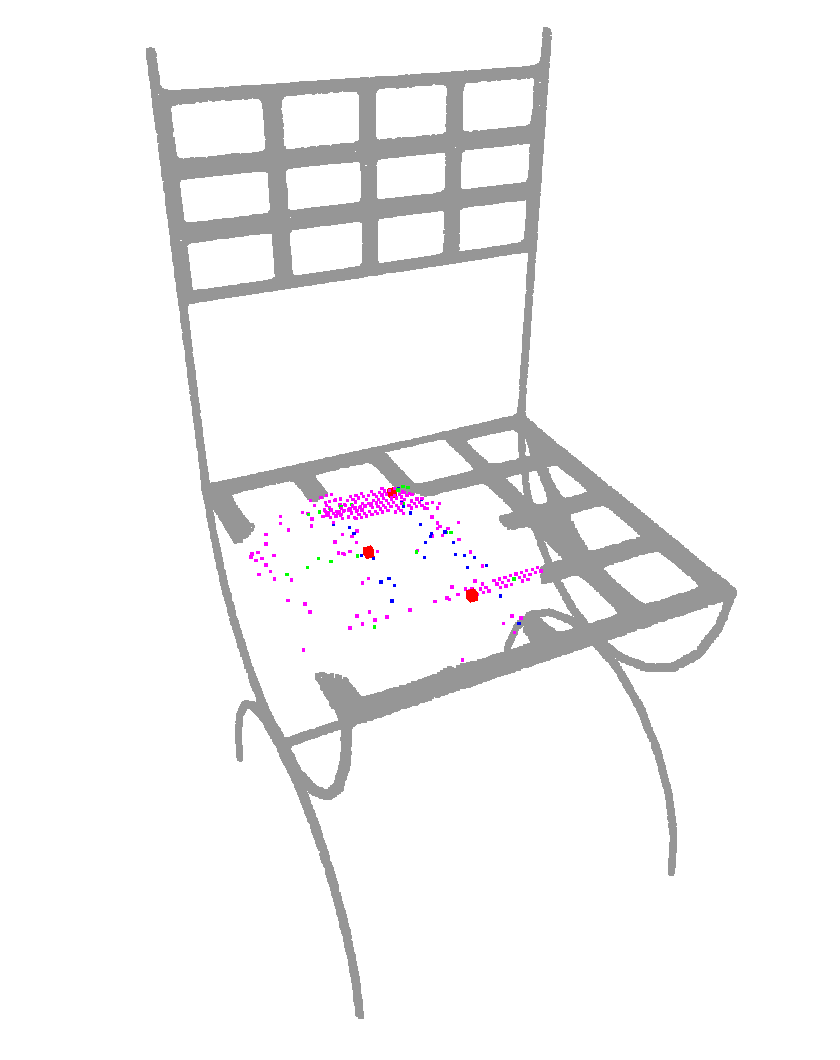} &
  \includegraphics[width=\sz\textwidth]{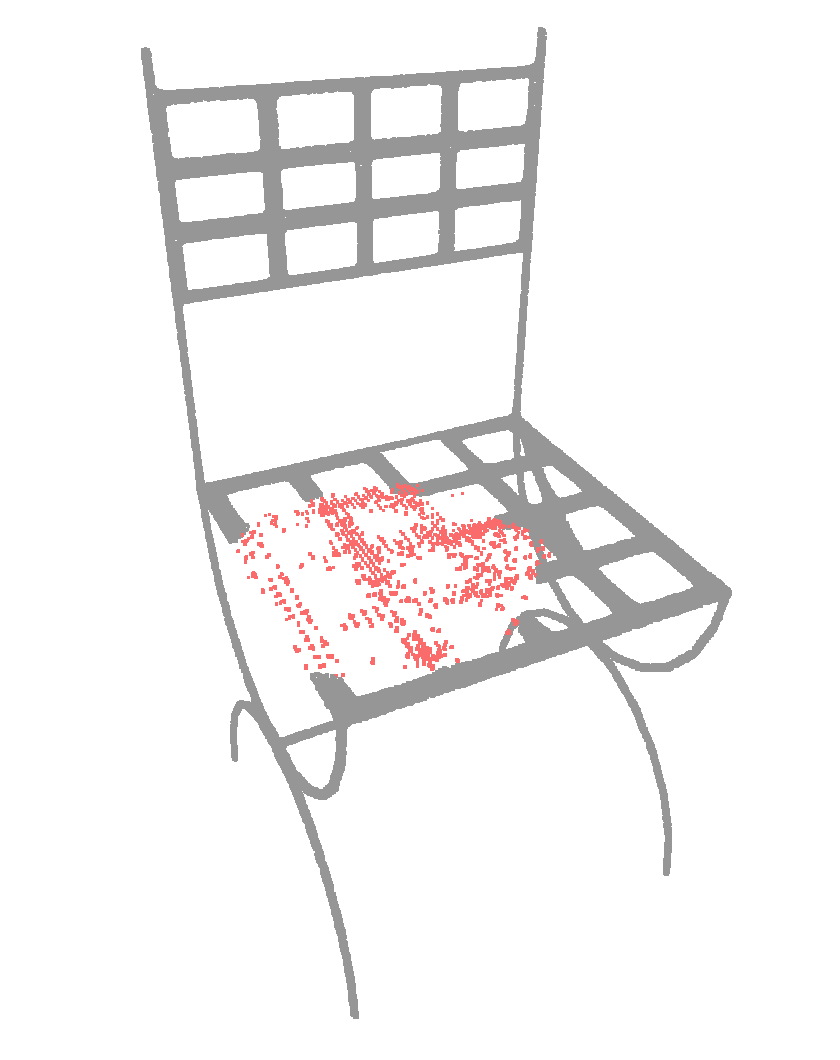} 
  \\
  \includegraphics[width=\insz\textwidth]{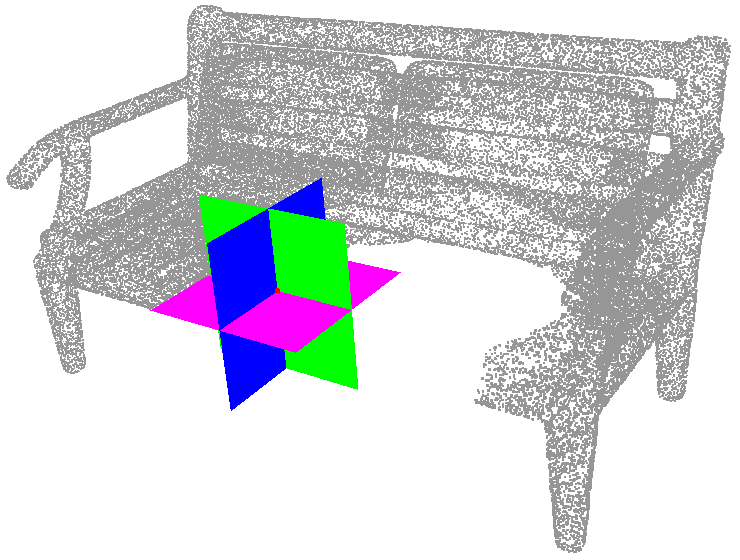} &
  \includegraphics[width=\insz\textwidth]{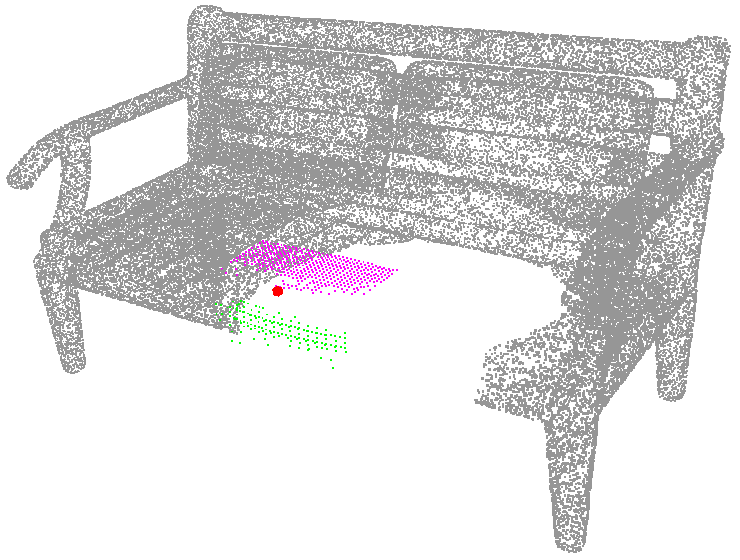} &
  \includegraphics[width=\insz\textwidth]{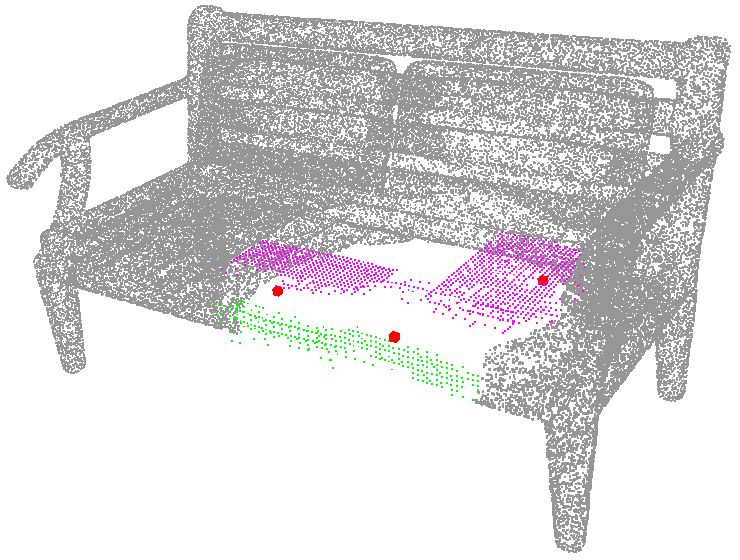} &
  \includegraphics[width=\insz\textwidth]{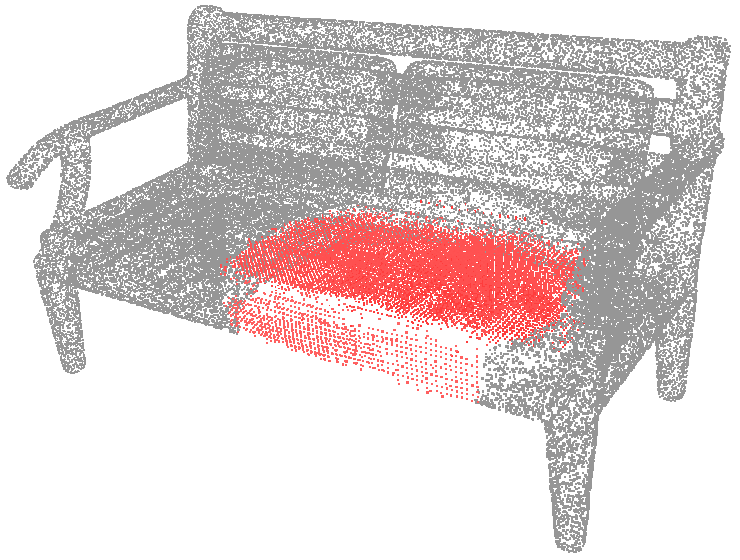}
  \\
  (a) & (b) & (c) & (d)
  \end{tabular}
  \vspace{-2mm}
  \caption{\textbf{Shape completion with the KAPLAN descriptor.} (a) Structure of a single KAPLAN with 3 canonical planes with (b) the corresponding point predictions coloured per plane.
  (c) KAPLAN predictions at $3$ locations (d) all final predictions for the whole object.}
  \label{fig:teaser}
\end{figure}

The state-of-the-art for local descriptors are learned, convolutional encodings.
These can be easily implemented on voxel grids using 3D convolutions, e.g.~\cite{Dai-et-al-CVPR-2017-shape-completion,Dai-et-al-CVPR-2018}, but are memory-hungry.
Another way to deal with unordered points is the approach of PointNet~\cite{Qi-et-al-CVPR-2017, Qi-et-al-NIPS-2017}, to extract per-point features and aggregate them with permutation-invariant operations. With that representation it is, however, not straight-forward to predict new points for shape completion.
It has also been proposed to deal with the orderless structure of point clouds by locally learning 2D descriptors~\cite{Tatarchenko-et-al-CVPR-2018}.
%
This is achieved by finding the tangent plane of the object surface, projecting the object points into it, and performing convolutional encoding in the 2D tangent plane.
Unfortunately, this only works well if the tangent plane approximation is valid and unique, i.e., when there is a single, smooth surface; but fails to capture high-curvature structures like edges and fine surface details.
Moreover, the tangent approximation only makes sense over small receptive fields, which contradicts a main strength of convolutional networks, to learn long-range, object-level context.

To efficiently capture 3D structure without the limitation to a single (tangent) plane, we bring back an old idea in computer vision and computational geometry: project the 3D geometry along multiple directions to obtain a set of 2D ``images'', which together form the descriptor (e.g., \cite{cyr2001_3d}).
The core contribution of our work is a new shape descriptor called \kaplan{} (see Fig.~\ref{fig:teaser}), which combines that classical idea with modern convolutional feature encoding. Instead of a single 2D projection or expensive 3D feature aggregation, we use $\nK$ projections with different orientations and concatenate them, such that the feature extractor can use standard 2D operations.
Importantly, all $\nK$ projections are encoded jointly, such that the descriptor is still optimised for 3D ``awareness'', not independently per projection.
Our contributions can be summarised as follows:
\begin{itemize}[itemsep=1pt,topsep=1pt,leftmargin=*]
\item We present a novel shape completion approach that fills in 3D points without costly convolutions on volumetric 3D grids. The approach operates both locally and globally via a multi-scale pyramid.
\item To that end, we design \kaplan{}, a 3D descriptor that combines projections from 3D space to multiple 2D planes with a modern convolutional feature extractor to obtain an efficient, scalable 3D representation.
\item The combination of \kaplan{} and our multi-scale pyramid approach automatically detects the missing region, and allows to perform shape completion without needing to regenerate the whole object.
\end{itemize}
While our target application is shape completion, \kaplan{} can potentially be used for a range of other point cloud analysis tasks. It is simple, computationally efficient, and easy to implement; but still leverages the power of deep convolutional learning to obtain expressive visual representations.

\section{Related Work}

\boldparagraph{Traditional shape completion approaches.}
Early works fill holes either directly on a triangle mesh \cite{Chui-et-al-CAGD-2000}, or they use a volumetric grid and intrinsically minimize the surface area \cite{Davis-et-al-3DPVT-2002}, curvature \cite{Kazhdan-et-al-SGP-2006,Kazhdan-Hoppe-SIGGRAPH-2013} or exploit object symmetries~\cite{Terzopoulos-et-al-IJCV-1987,Speciale-et-al-ECCV-2016}. 
%
Pauly et al.~\cite{Pauly-et-al-SGP-2005} perform example-based completion using a shape database.
These works capture \textit{local} surface properties and mostly perform interpolation.


\boldparagraph{Learned shape completion approaches.}
Semantic scene completion on a single depth image has recently been proposed~\cite{Song-et-al-CVPR-2017, Wang-et-al-ICCV-19}.
Dai et al.~~\cite{Dai-et-al-CVPR-2017-shape-completion} present a 3D encoder-predictor network (3D-EPN) which learns shape priors on a dense voxel grid.
ScanComplete~\cite{Dai-et-al-CVPR-2018} uses three sequentially trained networks to jointly perform shape completion and semantic labeling.
Han et al.~\cite{Han-et-al-ICCV-2017} perform geometry refinement with local 3D patches in an iterative manner, to fill larger surface holes.
In \cite{Zhang-et-al-ECCV-2018}, sparse convolutions improve scalability and \cite{Dai-et-al-CVPR-2020} uses sparse convolutions with a self-supervised loss.
3D-SIC~\cite{Hou-et-al-Arxiv-2019} jointly performs semantic instance segmentation and shape completion.
Stutz and Geiger~\cite{Stutz-Geiger-CVPR-2018} propose a weakly-supervised learning approach for completing laser scans.
Most of these methods operate on 3D voxel grids and require expensive 3D convolutional operations.
In contrast, AtlasNet~\cite{Groueix-et-al-CVPR-2018} approximates the surface with a series of deformed surface patches.
\cite{Han-et-al-CVPR-2019} inpaints 2D views guided by the volumetric completion output of SSCNet~\cite{Song-et-al-CVPR-2017} with a reinforcement learning scheme.
In \cite{Hu-et-al-AAAI-2020}, a method is proposed that enforces geometric consistency across multiple views.
%

%
%

\boldparagraph{Learned shape representations.}
Several works proposed to learn implicit shape representations that can be leveraged to complete shapes.
Ladicky et al.~\cite{Ladicky-et-al-ICCV-2017} estimate an iso-surface from an point cloud with a random forest.
%
Similarly, deep neural networks have been used to predict occupancies \cite{Mescheder-et-al-CVPR-2019,Chen_CVPR_2019} or signed distance functions~\cite{Park-et-al-CVPR-2019}.
The aforementioned methods rely on global feature extraction, often at the cost of local structures.
DISN~\cite{Xu-et-al-NIPS-2019} addresses this by adding a local feature extractor module.
While these works need explicit supervision, Sitzman~\etal~\cite{Sitzmann-et-al-NIPS-2019} proposed an unsupervised  approach using images with known poses.
%

\boldparagraph{Point cloud-based learning and descriptors.}
A series of works studied how neural networks can be applied to unstructured data like point clouds.
We only highlight some relevant works and refer to~\cite{Guo-et-al-Arxiv-2019,Bello-et-al-Arxiv-2020} for more comprehensive surveys. 
As one of the first methods, PointNet~\cite{Qi-et-al-CVPR-2017} uses per-point multi-layer perceptrons and max-pooling to compute features on point clouds.
PointNet++~\cite{Qi-et-al-NIPS-2017} extends that idea with the introduction of a hierarchical structure that gradually aggregates features.
Both networks aggregate information via spatial max-pooling which negatively affects the localisation of features.
Recently, Hu \etal \cite{Hu-et-al-CVPR-2020} proposed a new local feature aggregation scheme with increasing receptive field.
Coupled with random point sampling, this allows to preserve geometric details. 
These methods do not easily admit transpose convolution, which limits their application domain.
Tangent Convolutions~\cite{Tatarchenko-et-al-CVPR-2018} estimate tangent image patches which then hold a histogram of surface distances as a local shape descriptor, and are used for point cloud segmentation.
The same idea has also been shown to work well for point cloud denoising~\cite{Kripasindhu-et-al-SGP-2018}.
Our approach follows a similar idea, but generalizes the concept to a larger number of arbitrarily aligned patches. 
SPLATNet~\cite{Su-et-al-CVPR-2018} projects features of the point cloud onto a high-dimensional lattice where convolutions are performed. 
The network jointly aggregates 2D and 3D features and uses sparse representation for efficiency.
PointCNN~\cite{Li-et-al-NIPS-2018} follows a hierarchical network design and  lifts the points to a higher-dimensional space before computing convolutions on local point neighborhoods.
SparseConvNet~\cite{Graham-et-al-CVPR-2018} computes 3D convolutions efficiently by means of a sparse implementation of the voxel grid, binning the input points into a hash-table.
While sufficiently rich for tasks like segmentation, these features seem less suitable for shape completion, since they deliberately avoid filling empty voxels to preserve sparsity.
PointConv~\cite{Wu-et-al-CVPR-2019} introduces a learned, weighted convolutional kernel which is used to define learning operations on a 3D point cloud.
KPConv~\cite{Thomas-et-al-ICCV-2019} defines the concept of kernel point convolutions for direct feature computation on point clouds.
FPConv~\cite{Lin-et-al-Arxiv-2020} reduces the feature computation on point clouds to 2D convolutions, via a learned flattening operation.
Similar to our work, Point-PlaneNet~\cite{Peyghambarzadeh-et-al-DSP-2020} considers local and global feature aggregation using planes, however, they learn plane orientation and their feature aggregation is inspired by PointNet~\cite{Qi-et-al-CVPR-2017}.
In \cite{Yuan-et-al-3DV-2018} an encoder-decoder architecture is used to extract global features from the incomplete data and decode them into a complete object that does not retain the input points.
Similarly, in \cite{Achlioptas-et-al-ICML-2018} an autoencoder is designed which, if trained appropriately,
will synthesise a complete model when fed an incomplete one. 
The related \cite{Sarmad-et-al-CVPR-2019} additionally uses reinforcement learning to better control the (adversarial) loss function.
In~\cite{Chen-et-al-ICLR-20}, separate encoders are trained for complete and incomplete scans, which are then used to train a GAN that complete point clouds.
Encoders are not a necessity, as has been shown in~\cite{Tchapmi-et-al-CVPR-19} where the network consists of a single decoder.
%
%
Strategies were proposed that focus on the holes, without altering the observed parts of the point cloud \cite{Wang-et-al-CVPR-2020, Huang-et-al-CVPR-2020}.
Unlike our approach via a new 3D point descriptor, they directly feed 3D points into custom network architectures trained in an adversarial manner.
In~\cite{Minghua-et-al-AAAI-20}, a coarse completion is first extracted, and then fused with the input cloud via sampling.
A Skip Attention Network was proposed in~\cite{Wen-et-al-CVPR-20} to also address this issue.





\section{Method}
\label{sec:method}

Our method starts from an (incomplete) set of 3D points. 
We assume that for each point a normal vector is either available or can be estimated, noting that the method is generic and can be readily adapted to use other point attributes such as colour or laser intensity.
The core of our shape completion is \kaplan{}, an efficient learnable descriptor to represent 3D shapes based on irregularly sampled points.
Since the descriptor is learnable, it can be trained to always output an encoding of the complete local object geometry, even if fed with an incomplete point cloud. Thus, the output descriptor represents the \emph{expected} geometry, rather than the actually observed one.
This autoencoder-like behaviour makes it possible to use \kaplan{} for shape completion, by inverting the descriptor back to a point cloud.
The shape completion pipeline is embedded in an explicit multi-resolution framework. 
In this way, the method starts by filling large holes using coarse, global context and gradually refines and densifies the point cloud with more fine-grained, local context.
In the following, we detail the descriptor and associated network architecture.
The overall process is depicted in Fig.~\ref{overview_pipeline}.


\subsection{From 3D points to 2D images}
\label{sec:method_descriptor}


This part corresponds to the ``Point Cloud Processing'' section in Fig.~\ref{overview_pipeline}.
Our aim is an efficiently computable descriptor, hence we aim to avoid volumetric computations that involve convolutions on 4D tensors.
A simple, but effective trick is to project the 3D geometry onto a 2D plane along some ``canonical'' direction.
One implementation of that principle are Tangent Convolutions~\cite{Tatarchenko-et-al-CVPR-2018} which use the local surface normal as the canonical direction.
Note that the projection does not greatly increase computational cost, as the normals and the associated projections are fixed and can be precomputed.
A disadvantage of a single projection from 3D space onto the tangent plane is the implicit assumption that the object geometry can locally be parametrised as a single function over a 2D domain. 
When this condition is not fulfilled, information will be lost.
A classical trick to mitigate the loss of information is to project onto multiple different planes, thus creating a ``multi-view'' representation of the 3D geometry.
This strategy used to be popular in the early days of 3D object representation and retrieval, e.g.,~\cite{cyr2001_3d}. 
More recently, it has been used in the context of deep learning, for semantic segmentation of 3D point clouds, e.g.,~\cite{boulch2018snapnet}.

\kaplan{} combines the two ideas: the geometry is projected along $\nK$ directions to minimize the loss of 3D information, the set of projections then serves as input for a convolutional architecture that encodes it into a feature representation.
%
We point out a further, more subtle advantage of using multiple projections: contrary to~\cite{Tatarchenko-et-al-CVPR-2018}, it is no longer necessary to align the projection plane with the local surface, the descriptor therefore is less affected by inaccurate point-wise normals. 
Normals can still be used as input, but are no longer crucial to the method.
This is a considerable benefit in our application, as normal estimation is unstable on the border of data gaps, where there are no supporting 3D points.
Also note that $\nK\geq 3$ projections are enough to avoid problems at sharp crease edges and $90^{\circ}$ corners, which are frequent on man-made objects.
In our implementation, we align the \kaplan{} coordinate system with the gravity direction, which is known for most 3D point clouds.


\begin{figure*}[th!]
	\vspace{-6mm}
	\centerline{\includegraphics[scale=0.175]{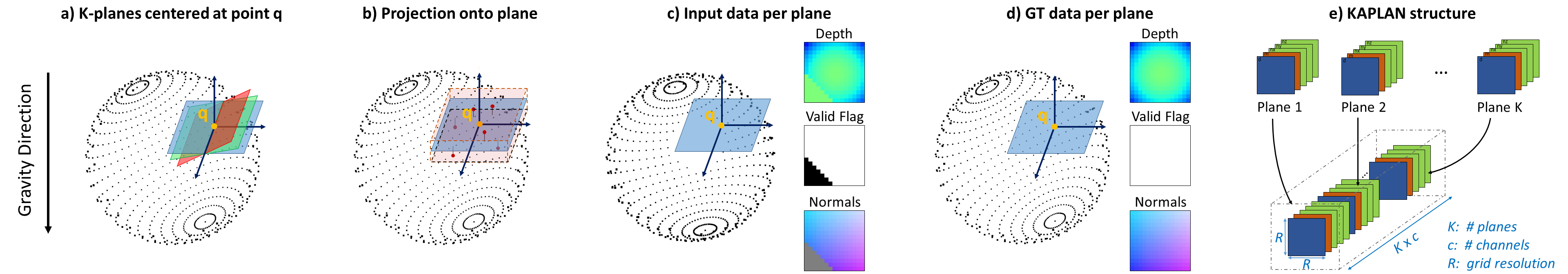}}
	\vspace{-3mm}
	\caption{\label{kaplan_descriptor} \textbf{Computation of KAPLAN for a query point $q$.} (a) Instantiate $\nK$ planes (chosen for illustration purposes); (b) project points within a box neighborhood onto each plane; (c) record point features, including a binary valid flag for empty (missing data) cells; (d) for training data, do the same with complete point clouds; (e) concatenate channels from all planes.} 
\end{figure*}

Fig.~\ref{kaplan_descriptor} illustrates the accumulation of 3D point information into 2D projection planes.
Empirically, already the minimal configuration with $\nK=3$ planes in canonical orientation provides a good representation. 
If necessary one can increase $\nK$, either systematically (e.g., rotating planes around each coordinate axis in fixed steps) or by randomly drawing plane normals.
For each plane, nearby points are selected according to a box constraint to preserve locality (red box in Fig.~\ref{kaplan_descriptor}b) and projected orthogonally onto the plane to form a depth image and a normal image.
The 2D projection plane is discretised into a grid of $\nRes \times \nRes$ cells (``pixels''), if multiple points are projected into the same cell their normals and depth values are averaged.
Additionally, we record a \emph{valid flag} image that indicates whether any points have been projected to a given cell (value $1$) or whether the cell is empty (value $0$).
%
%
%
The $\nNumChannelsPlane=5$ channels in each plane (depth $\times 1$, valid flag $\times 1$, normal $\times 3$) are stacked into a tensor $\mathcal{K}^0$ of dimension $(\nRes \times \nRes \times \nK \nNumChannelsPlane)$.
Note that, while \kaplan{} can in principle be constructed at every 3D point, in practice we only sample descriptors at $\numQuery$ \emph{query points}, since nearby descriptors are highly redundant.

\subsection{Descriptor Network}
\label{sec:method_network}


To map the raw point properties in $\mathcal{K}^0$ into a shape representation that preserves the geometric layout, we pass it through a convolutional autoencoder, as shown in the ``K-Planes Processing'' section in Fig.~\ref{overview_pipeline}.
In our implementation, we chose a variant of U-net with two rounds of pooling in the encoder and the corresponding two rounds of unpooling in the decoder.
%
We use MISH activations~\cite{Misra-Arxiv-2019} after every convolution layer instead of ReLUs for better accuracy.
The decoder has three separate heads (with identical layer structure) for the three modalities depth, normal and valid flag, whose outputs are stacked back together into the final descriptor $\mathcal{K}$ as illustrated in Fig.~\ref{kaplan_descriptor}e.

Since we apply \kaplan{} for shape completion, the main task of the network is to fill empty descriptor cells whereas the other cells remain unchanged.
We therefore add a skip connection directly from the input $\mathcal{K}^0$ to the output $\mathcal{K}$, such that the network can focus on learning to fill in the empty cells.
Note that the predictions for the valid flag are continuous values in $[0, 1]$, for further processing we threshold them at $0.5$.
If a \kaplan{} cell was empty in the input, but is predicted as non-empty, we reconstruct the corresponding 3D point from the predicted depth value.



\subsection{Coarse-to-fine Approach}
\label{sec:method_coarsefine}

\begin{figure*}[t!]
	\vspace{-3mm}
	\centerline{\includegraphics[scale=0.23]{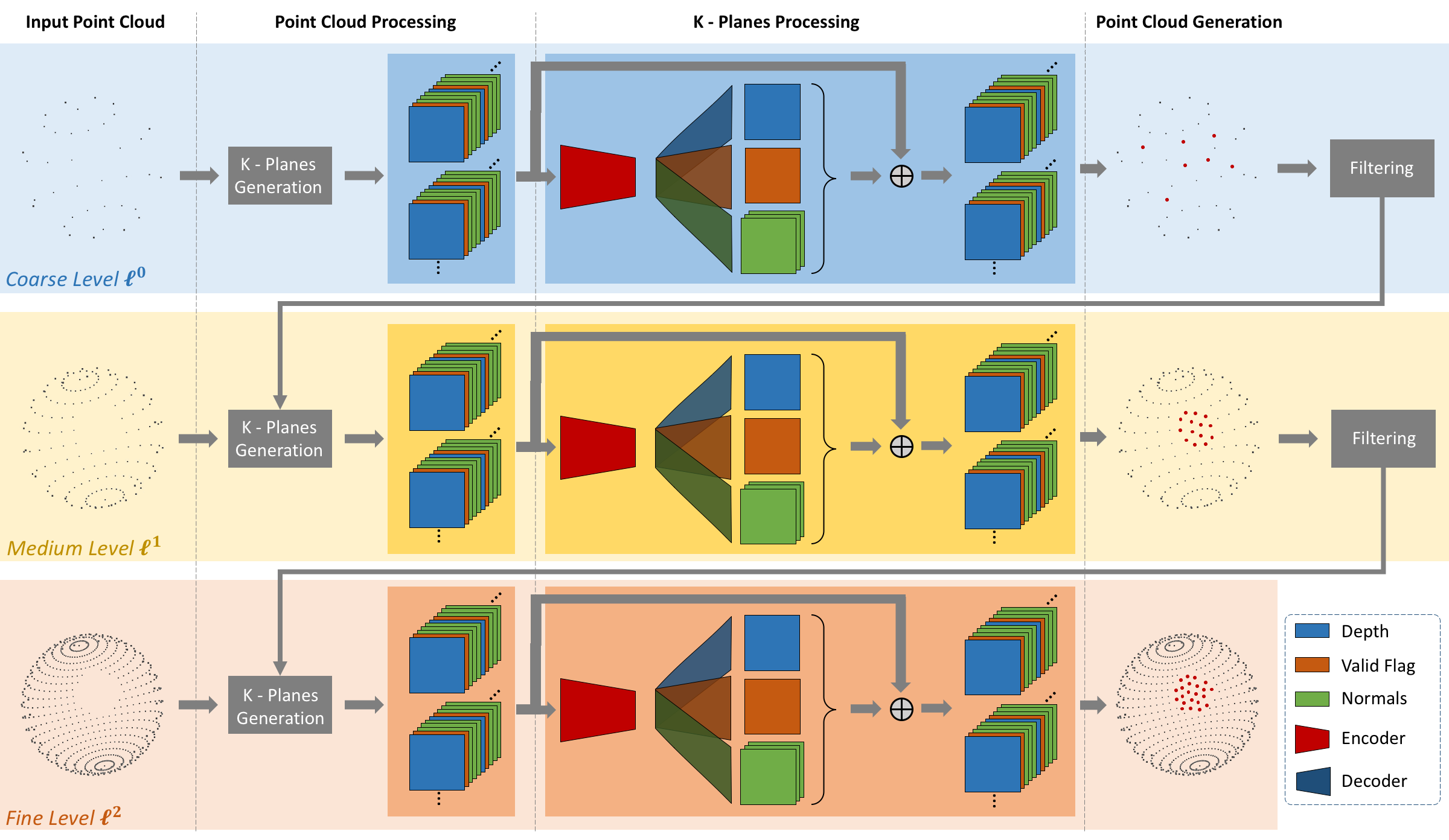}}
	\vspace{-2mm}
	\caption{\label{overview_pipeline} \textbf{Overview of our multi-scale pipeline}. It exhibits the three levels: \textit{coarse level} $\nLevelCoarse$, \textit{medium level} $\nLevelMedium$ and \textit{fine level} $\nLevelFine$. Each level consists of the same autoencoder: a single shared encoder and three separate decoders for each modality. 
	}
\end{figure*}

\kaplan{} encodes \emph{local} shape information. In order to employ long-range shape context and fill large holes, we embed the shape completion in a coarse-to-fine scheme.

We start at an initial, coarse level $\nLevelCoarse$, where the size of the \kaplan{} neighbourhood (defined by the box constraint) is the bounding box size of the full object.
At this level, the descriptors capture global context.
As at most one point per descriptor cell is reconstructed, the first level will result in a point cloud that still has low density in the newly completed regions, but that is \emph{complete}, in the sense that a minimum point density is ensured everywhere on the surface.
That initial reconstruction can now guide the next hierarchy level: the newly generated points serve as query points to compute \kaplan{} descriptors at the next finer scale and further densify the point cloud.
The refinement can be iterated until the desired point density has been reached. 
In our implementation, we use three levels (see Fig.~\ref{overview_pipeline}).
In practice, instantiating a \kaplan{} at every newly created point from the previous level would be highly redundant, so we subsample the query points with a spatial filtering block (see Sec.~\ref{sec:impl_filter} and \textit{supplementary material}). 

We train the coarse-to-fine scheme sequentially, starting at the coarsest level.
Back-propagation across hierarchy levels is not possible, because the instantiation of new points from the \kaplan{} projections is not differentiable.
\subsection{Loss Function}

The learnable parameters of \kaplan{} at each hierarchy level are the autoencoder weights.
To train them we minimize the following loss function: 
%
%
\begin{equation}
\nLoss = \nWeightVf \nLossVf(V,\hat{V}) + \nWeightD \nLossD(D,\hat{D}) + \nWeightN \nLossN(\nGtNormal, \nPredNormal)
\label{eq:global_loss}
\end{equation}
The loss is composed of three terms, one for each modality. 
%
%
For the valid flag, we use a standard $L^{2}$ regression loss between the predictions $\hat{V}$ and the ground truth $V$, computed over all pixels of the projection plane $i \in \Omega \subset \nR^2$:
\begin{equation}
\nLossVf(V,\hat{V}) = \frac{1}{\nK} \cdot \frac{1}{|\Omega|} \; \sum_{k\in[1,\nK]} \sum_{i\in\Omega} \|V_{i} - \hat{V}_{i}\|_2
\end{equation}
For the depth, we compute a masked $L^{2}$ loss between the predictions $\hat{D}$ and the ground truth $D$, only at non-empty cells $\nValidPix \in \Omega_m \subseteq \Omega$ according to the ground truth valid flag:

\begin{equation}
\nLossD(D,\hat{D}) = \frac{1}{\nK} \cdot \frac{1}{|\Omega_m|} \; \sum_{k\in[1,\nK]} \sum_{j\in\Omega_m} \|D_j - \hat{D}_j\|_2
\end{equation}
For the normals, we penalize the angular deviation between the predictions $\nPredNormal$ and the ground truth $\nGtNormal$, following \cite{Ramamonjisoa-et-al-ICCV19}. 
As for the depth, the loss is computed only over non-empty cells using the ground truth valid flag:
\begin{equation}
\nLossN(\nGtNormal, \nPredNormal) = \frac{1}{\nK} \cdot \frac{1}{|\Omega_m|} \; \sum_{k\in[1,\nK]} \sum_{j\in\Omega_m} \left(1 - \frac{\langle \nGtNormal_j , \nPredNormal_j \rangle}{\|\nGtNormal_j\| \|\nPredNormal_j\|}\right)
\end{equation}
In practice, we use $\nWeightVf = 0.75$, $\nWeightD = 1$ and $\nWeightN = 0.01$.

\section{Implementation Details}
\label{sec:implementation}

This section elaborates on implementation details of \kaplan{} (see also the \textit{supplementary material}).
%
\subsection{KAPLAN Precomputation (3D to 2D)}\label{precomputation_3d_2d}
During \kaplan{} generation all points within the bounding box are projected into cells on the plane (\cf Sec.~\ref{sec:method_descriptor}) and attributes like depth and normals are averaged per cell.
However, even with the box constraint it can still happen that projected points originate from different surfaces.
In that case naive averaging would entail information loss, e.g., the depths of two distinct surfaces cannot be recovered from their average.
To prevent that we apply a simple heuristic that bins the points into distinct surfaces and projects only the points on the surface with the lowest depth.

%

\subsection{Point Prediction (2D to 3D)}\label{prediction_2d_3d}
The network predicts a complete \kaplan{}, which is a set of 2D image channels.
To convert the descriptor back to 3D points, one simply lifts the center point of each cell to the appropriate depth and transforms the resulting 3D point back to the global coordinate system.
Our goal is to complete missing regions, whereas we aim to avoid oversampling regions already covered by existing 3D points.
To decide where to introduce new points, the simplest criterion is to look for \kaplan{} cells where the valid flag was $0$ in the input and got switched to $1$.
In practice this criterion is too strict, as even cells in regions of missing geometry will sometimes not be completely empty. 
For instance, if another surface is in the range of the descriptor, its points will project into the 2D image of the hole, which then would not not be recognised.
%
Thus, we additionally instantiate new 3D points for cells that were marked \emph{valid} in both the input and the output, but where the predicted depth significantly differs from the input depth.

\subsection{Prediction Filtering}\label{sec:impl_filter}
As described in Sec.~\ref{sec:method_coarsefine}, we filter output points of a coarser hierarchy level before passing them to the next-finer level, to avoid redundant points and to remove outliers.
%
%
To integrate information across multiple \kaplan{} descriptors and ensure consistency, we use a volumetric representation only \emph{locally} around the predicted hole regions.
%
%
Note that the ``voxels'' serve only to collect points in a small 3D region -- the filtering looks at them one-by-one, so one need not store an explicit voxel space.

\boldparagraph{Inter-\kaplan{} consistency.}
In order to remove outliers, we keep track which query point (and corresponding \kaplan{}) spawned each predicted point.
Points not supported by a second prediction from a different query point (falling into the same voxel) are discarded.

\boldparagraph{Representative points.}
To ensure an even density of generated points, we only take one point per voxel.
We compute a weighted average of all predictions falling into a voxel with Gaussian weights inversely proportional to a point's \kaplan{} depth (giving predictions with lower depth higher confidence).
We retain the prediction closest to the average and discard all other points in the voxel.


\section{Experiments}
\label{sec:experiments}


We implemented \kaplan{} in Tensorflow and run it on a GTX1080Ti GPU (12GB RAM).
We use the Adam optimiser~\cite{Kingma-Ba-ICLR-2015} with learning rate $10^{-6}$ and batch size $128$.
By default, we use the KAPLAN configuration ($\nK\!=\!3$, $\nRes\!=\!35$) 
found via an ablation study (see \textit{supplementary material}). 


\boldparagraph{Data generation.}
We use ShapeNet~\cite{shapenet2015} to create our dataset of input-output pairs, i.e., incomplete ($\nIncompPC$) and complete ($\nCompPC$) versions of the same point clouds.
We use the pre-processed data of~\cite{Mescheder-et-al-CVPR-2019}.
Each object comes in the form of a gravity aligned point cloud consisting of 100k points with normals, sampled on the surface, and normalized to a bounding box with largest dimension 1.



These constitute our ground truth at the finest level,  $\nCompPC^2$.
We then generate incomplete point clouds with different sizes of holes, by removing $\{2\%,5\%,10\%,20\%,30\%\}$ points.
A point is picked randomly as the ``center'' of the hole, then the appropriate number of surrounding points $\nMissPC^2$ is found with a kd-tree search and cut out to obtain $\nIncompPC^2$, such that $\nCompPC^2 = \nIncompPC^2 \cup \nMissPC^2$.
In order to create data for the coarser level $\nLevelMedium$, it is not advisable to downsample $\nCompPC^2$ and repeat the same procedure, as this may lead to inconsistencies between the sampling of the fine and coarser levels.
Instead, we individually dowsample $\nIncompPC^2$ and $\nMissPC^2$ to obtain $\nIncompPC^1$ and $\nMissPC^1$. Their union yields the downsampled version of the complete point cloud $\nCompPC^1 = \nIncompPC^1 \cup \nMissPC^1$.
The procedure is repeated once more to obtain the coarsest level $\nLevelCoarse$ (see Fig.~\ref{overview_pipeline}).

We select five object categories \textit{plane}, \textit{chair}, \textit{lamp}, \textit{sofa} and \textit{table}, and split them into training and test sets following~\cite{Choy-et-al-ECCV16}.
Since our goal is to learn expressive shape priors, not only local surface interpolation, we train a dedicated network for each category, as in \cite{Park-et-al-CVPR-2019}.


\boldparagraph{Baselines.}
We compare our method against five baselines: 
\begin{enumerate}[itemsep=1pt,topsep=1pt,leftmargin=*]
	\item Poisson Surface Reconstruction (PSR)~\cite{Kazhdan-et-al-SGP-2006} reconstructs a mesh from a point cloud. It can interpolate smooth surfaces with small holes, but not larger missing areas.
	\item Point Completion Network (PCN)~\cite{Yuan-et-al-3DV-2018} is a learning-based method which also performs shape completion directly on the incomplete point cloud.
	\item Cascaded Refinement~\cite{Wang-et-al-CVPR-2020} combines a global feature with local details preservation via skip connections.
	\item Occupancy Networks~\cite{Mescheder-et-al-CVPR-2019} estimate the occupancy at sampled points in the 3D bounding volume (similar to a binary classifier) to reconstruct an implicit surface.
	\item DeepSDF~\cite{Park-et-al-CVPR-2019} learns to encode a signed distance function of one or multiple shapes in a latent space.
\end{enumerate}
The baselines follow two different principles for shape completion.
The first three methods directly reconstruct the missing surface points, without explicitly extracting a shape representation. 
%
%
%
The last two methods focus on learning a (complete) shape representation, then extract the missing points by feeding the incomplete input into the representation and decoding it into a complete model. 


For all baselines, we used implementation and trained models provided by the authors, except for DeepSDF, which we retrained following the authors' instructions.
The code available to us did not include the version used for shape completion, \ie we did not have access to the exact values of the SDF sampling distance as well as the weight of the free space loss.
This explains why our results differ from those presented in their paper.


\subsection{Results}\label{sec:exp_results}
We conduct different experiments to evaluate the shape completion performance of our \kaplan{}. 
First, we compare it against several baselines, for different levels of incompleteness. 
We then perform ablation and parameter studies at coarse level $\nLevelCoarse$ to analyse different \kaplan{} configurations.
For all quantitative experiments, we used two error metrics: the Chamfer distance (CD) between prediction and ground truth, and the $F1$-score between predicted and true points. Details about both metrics can be found in the \textit{supplementary material}.

\begin{figure}[t]
  \centering
  \scriptsize
  \newcommand{\sz}{1.16cm}
  \newcommand{\insz}{0.75cm}
  \newcommand{\rd}{5pt}
  \newcommand{\rulesep}{\unskip\ \vrule\ }
  \setlength{\tabcolsep}{1pt}
  \vspace{-1em}
  \begin{tabular}{cccc @{\hspace{6\tabcolsep}}c@{\hspace{1.0\tabcolsep}}cccc}
  &
  \textit{Input} & 
  \textit{Prediction} & 
  \textit{GT} &&
  \textit{Input} &
  \textit{Prediction} &
  \textit{GT} & \\[1pt]
  \rotatebox{90}{\hspace{2pt}\textit{Valid Flag}} & 
  \includegraphics[height=\sz]{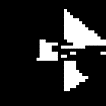} &
  \includegraphics[height=\sz]{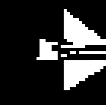} &
  \includegraphics[height=\sz]{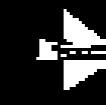} &&
  \includegraphics[height=\sz]{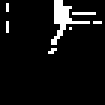} &
  \includegraphics[height=\sz]{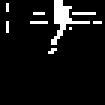} &
  \includegraphics[height=\sz]{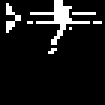} &
  \\[\tabcolsep]
  \rotatebox{90}{\hspace{7pt}\textit{Depth}}  &
  \includegraphics[height=\sz]{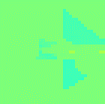} &
  \includegraphics[height=\sz]{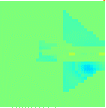} &
  \includegraphics[height=\sz]{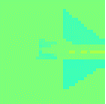} &&
  \includegraphics[height=\sz]{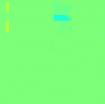} &
  \includegraphics[height=\sz]{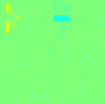} &
  \includegraphics[height=\sz]{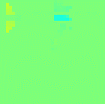}  &
  \includegraphics[height=\sz]{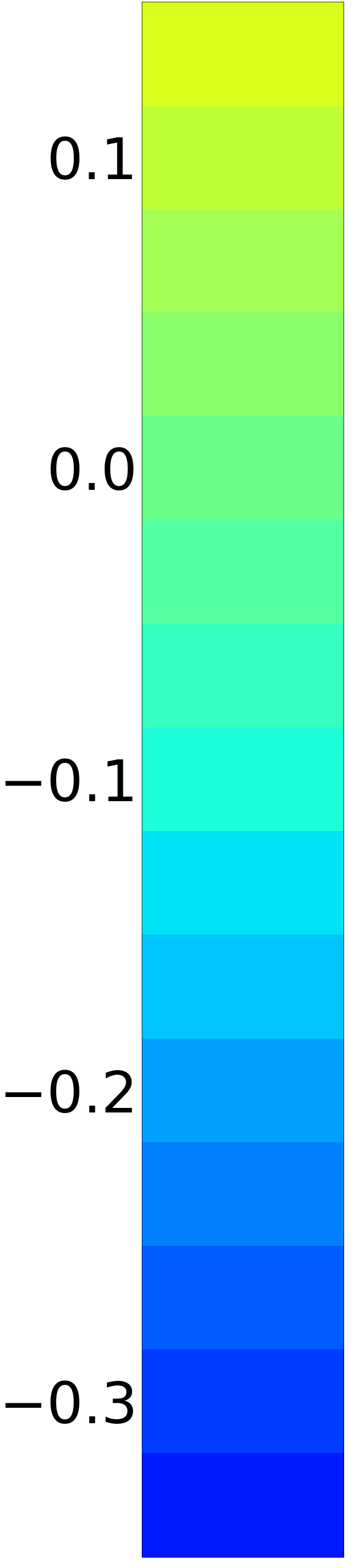}
  \\[\tabcolsep]
  \rotatebox{90}{\hspace{6pt}\textit{Normal}} & 
  \includegraphics[height=\sz]{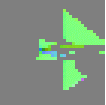} &
  \includegraphics[height=\sz]{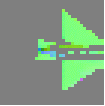} &
  \includegraphics[height=\sz]{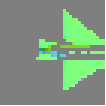} &&
  \includegraphics[height=\sz]{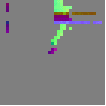} &
  \includegraphics[height=\sz]{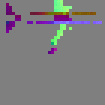} &
  \includegraphics[height=\sz]{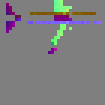} & \\[-0.15cm]
  \end{tabular}
  \caption{\textbf{Illustration of \kaplan{} predictions at $\nLevelCoarse$.}
  Shown are estimated properties on a top view of two airplanes. Note that depth values can be positive and negative.}
  \label{fig:hole_detection}
  \vspace{-3mm}
\end{figure}

\boldparagraph{Missing region detection.}
%
Contrary to some other works, e.g., \cite{Han-et-al-ICCV-2017}, we do
not rely on a separate detection step to find points on the boundary
of a hole.
Instead, our approach detects the missing data regions during descriptor computation, via the prediction of the valid flags (at the coarsest level).
In Fig.~\ref{fig:hole_detection} we show planes from different \kaplan{} descriptors of an airplane, extracted at the coarsest level $\nLevelCoarse$.
Note how the descriptor captures the global shape, for instance in the example on the left, where the projection plane is aligned with the airplane's wings.
The network uses the pixel-wise valid flags to determine where points need to be added, as can be seen from the predicted flags, where the hole has been filled correctly.
The predicted normals and depth at those pixels are then used to instantiate new points.

\begin{figure*}[b]
	\vspace{-3mm}
	\centerline{\includegraphics[scale=0.48]{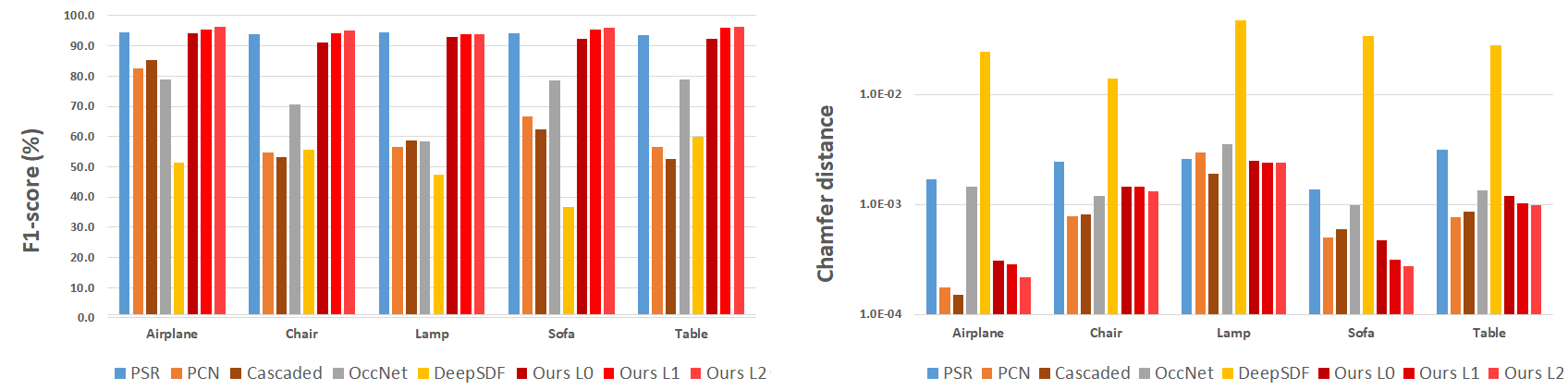}}
	\vspace{-0.1cm}
	\caption{\label{fig:f1_chamfer_graphs} \textbf{Quantitative comparison to the state-of-the-art methods}. \textbf{Left:} Chamfer distance and \textbf{Right:} F1-score for different categories. Our results are shown with a gradation of red, to further distinguish the different coarse-to-fine levels. 
	}
\end{figure*}

\begin{figure*}[th]
	\centering
	\scriptsize
	\setlength{\tabcolsep}{0.1mm}
	\newcommand{\sz}{0.125}
	\newcommand{\insz}{0.09}
	\begin{tabular}{cccccccccc}
		& \textbf{Input} 
		& \textbf{PSR~\cite{Kazhdan-et-al-SGP-2006}} 
		& \textbf{PCN~\cite{Yuan-et-al-3DV-2018}}  
		& \textbf{Cascaded~\cite{Wang-et-al-CVPR-2020}}
		& \textbf{OccNet~\cite{Mescheder-et-al-CVPR-2019}} 
		& \textbf{DeepSDF~\cite{Park-et-al-CVPR-2019}} 
		& \textbf{Ours} 
		& \textbf{GT} 
		\\[-1pt]
		\multirow{2}{*}[30pt]{\rotatebox{90}{\textbf{Plane}}}
		& \includegraphics[width=\sz\textwidth]{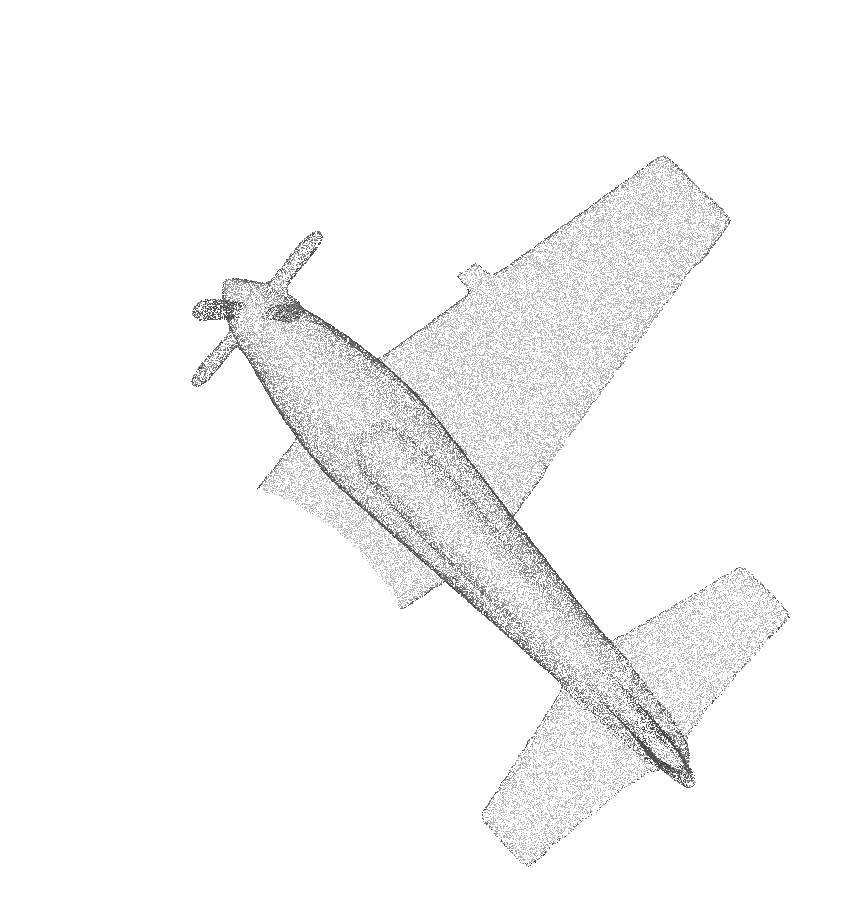} 
		& \includegraphics[width=\sz\textwidth]{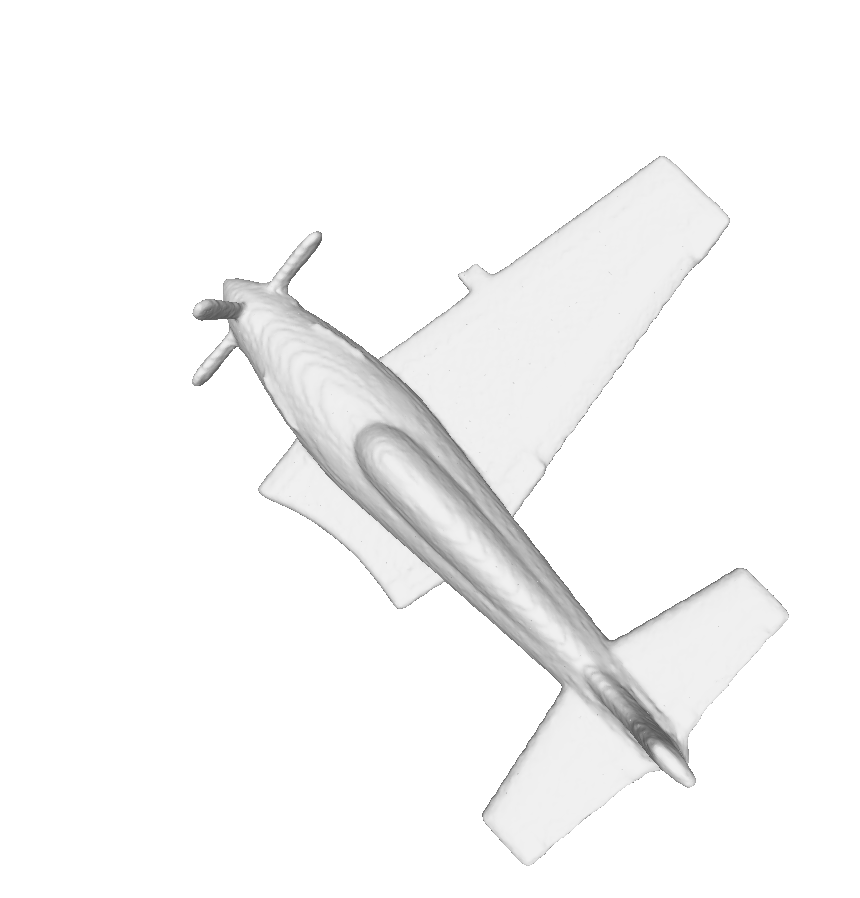} 
		& \includegraphics[width=\sz\textwidth]{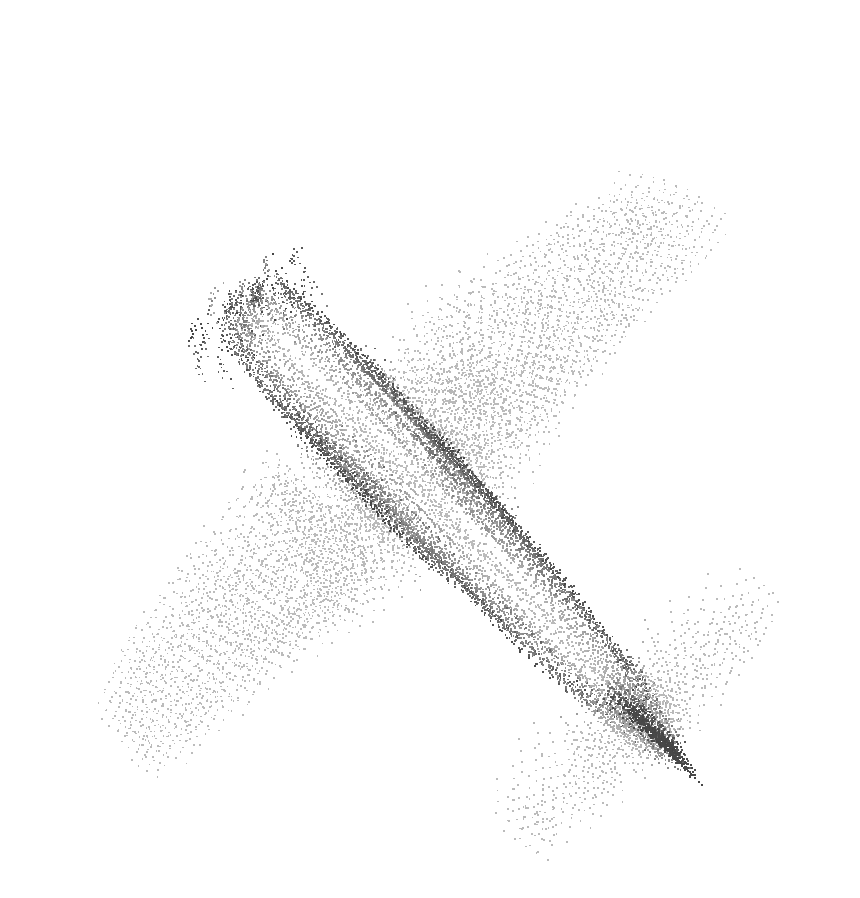} 
		& \includegraphics[width=\sz\textwidth]{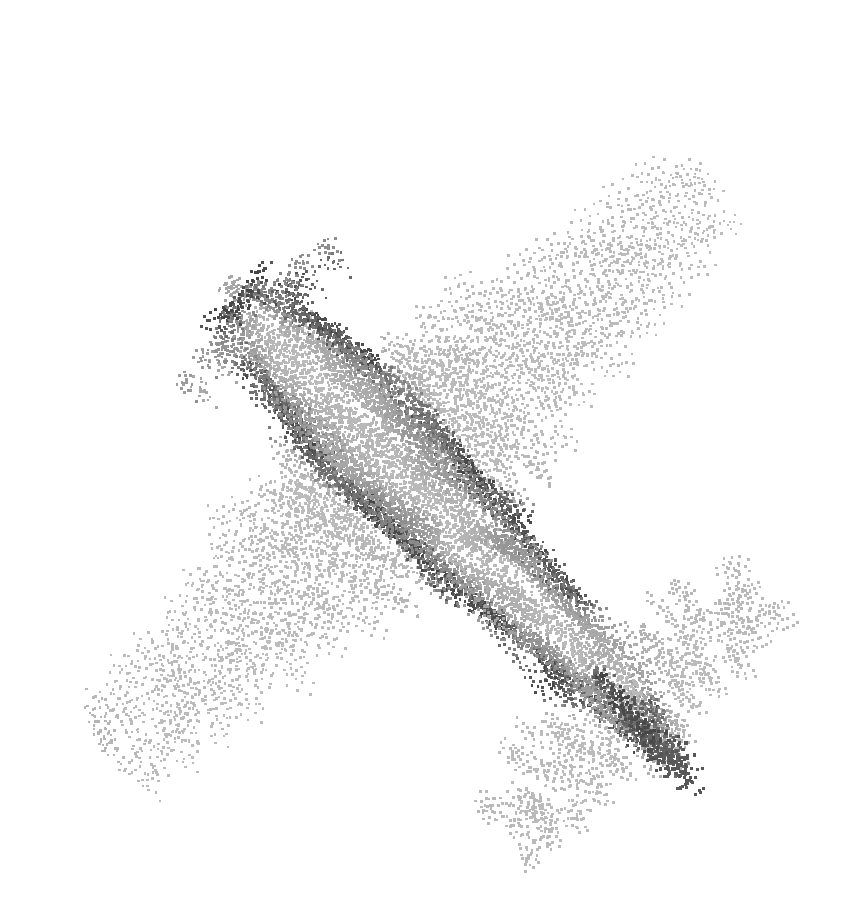} 
		& \includegraphics[width=\sz\textwidth]{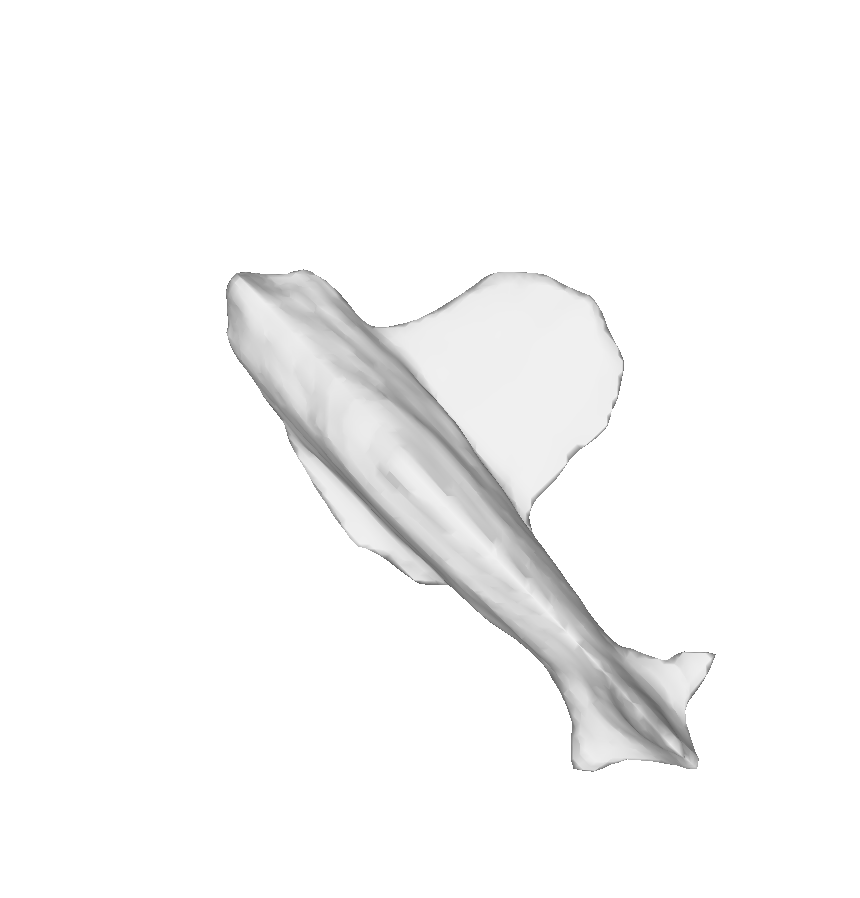} 
		& \includegraphics[width=\sz\textwidth]{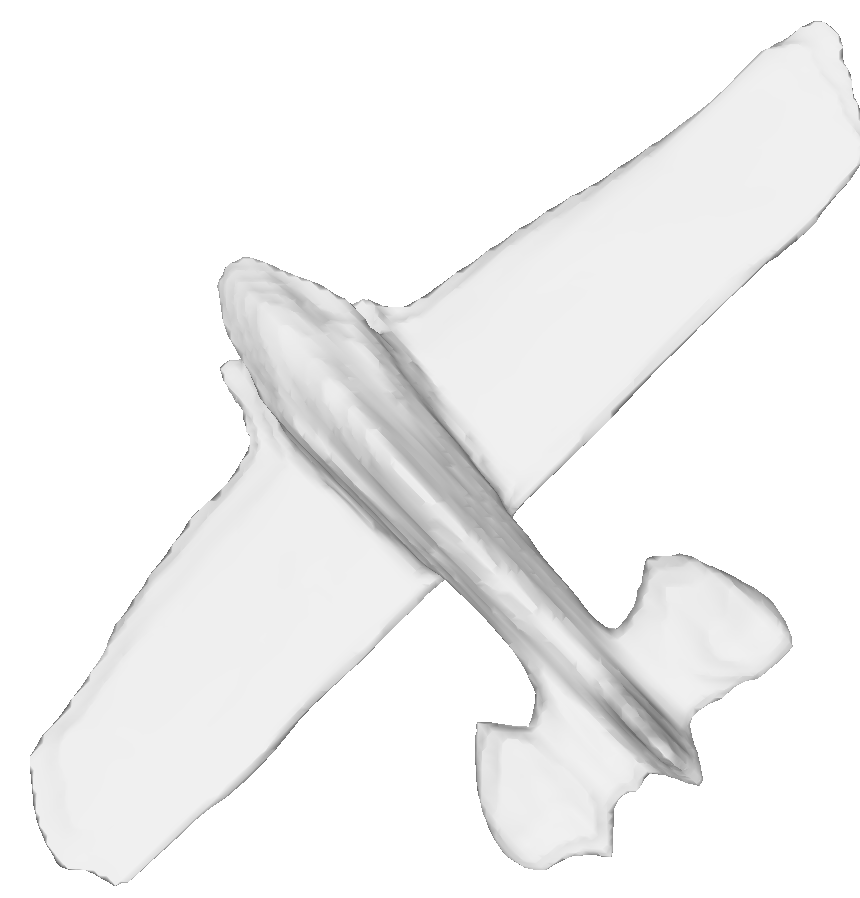} 
		& \includegraphics[width=\sz\textwidth]{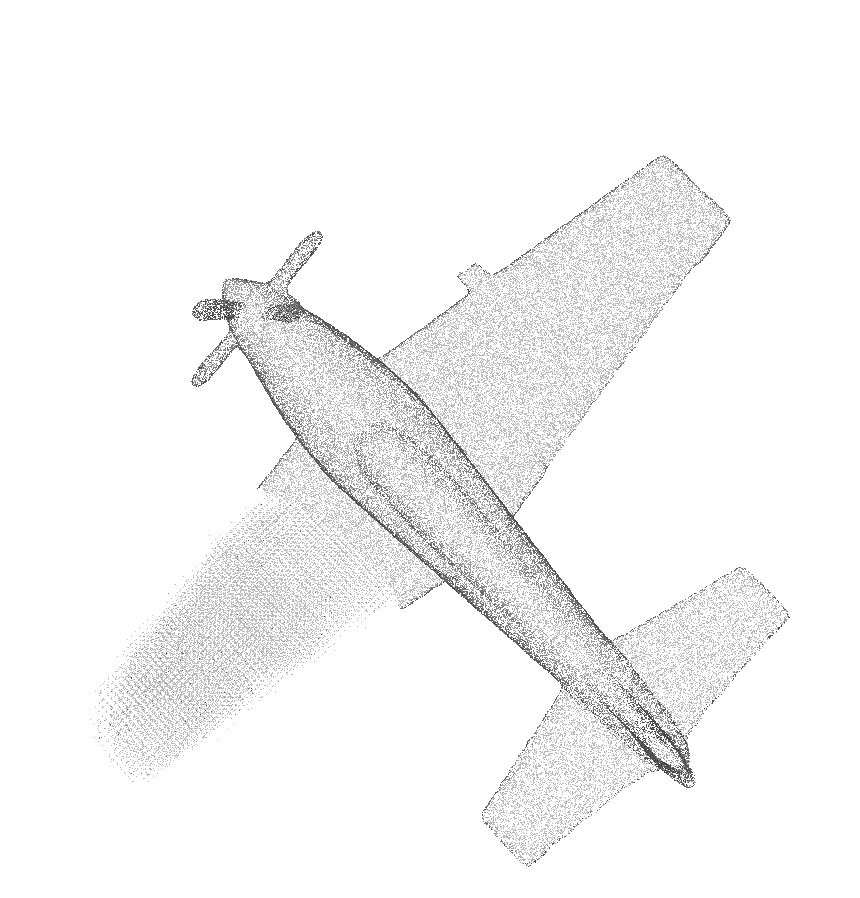} 
		& \includegraphics[width=\sz\textwidth]{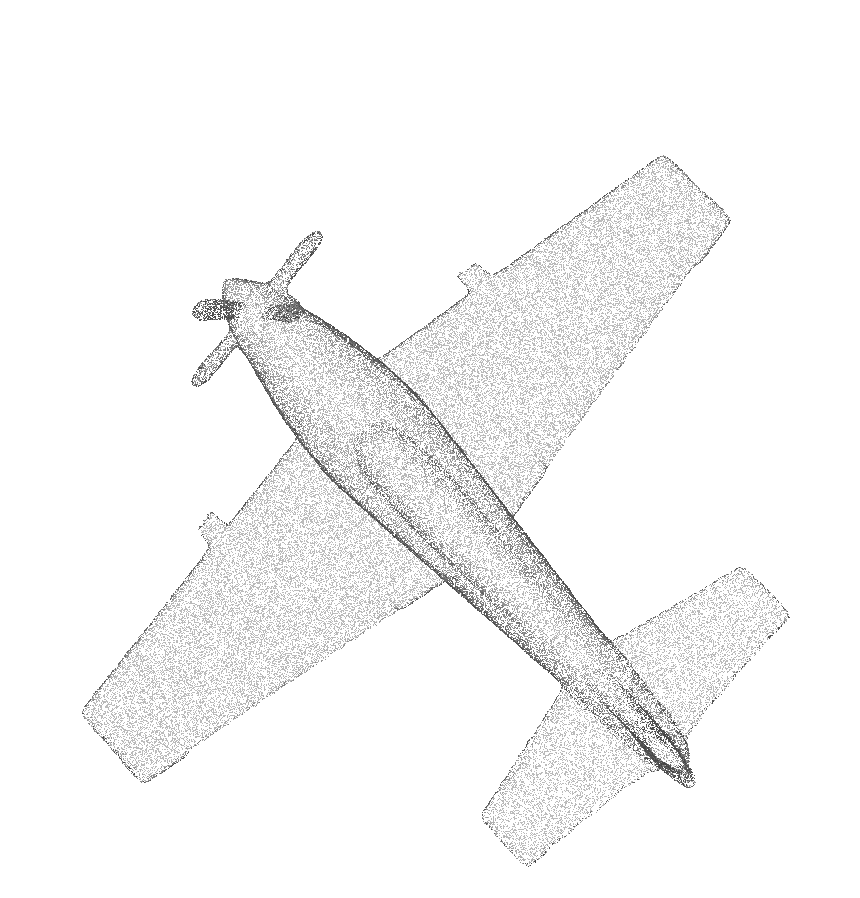} 
		\\[-3pt]
		&
		& $1.69  \;\vert\; 94.51$ 
		& ${\color{highlight2nd} 0.175} \;\vert\; {\color{highlight2nd}82.45}$ 
		& ${\bf 0.151}  \;\vert\; 85.45$ 
		& $1.46  \;\vert\; 78.87$ 
		& $24.5  \;\vert\; 51.41$ 
		& $ 0.219 \;\vert\; {\bf 96.40}$
		\\[3pt]
		\hdashline\\[-6pt]
		\multirow{2}{*}[30pt]{\rotatebox{90}{\textbf{Chair}}} 
		& \includegraphics[width=\insz\textwidth]{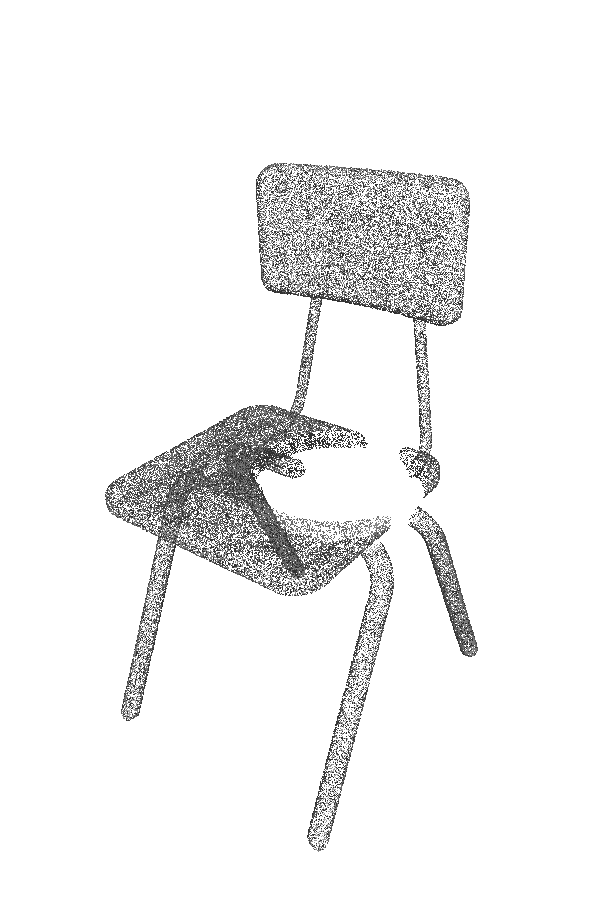} 
		& \includegraphics[width=\insz\textwidth]{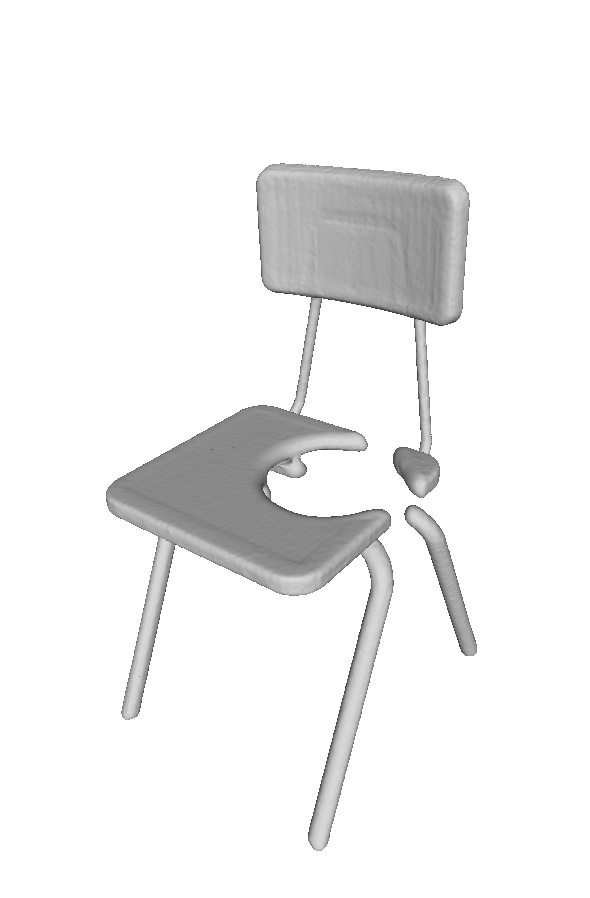} 
		& \includegraphics[width=\insz\textwidth]{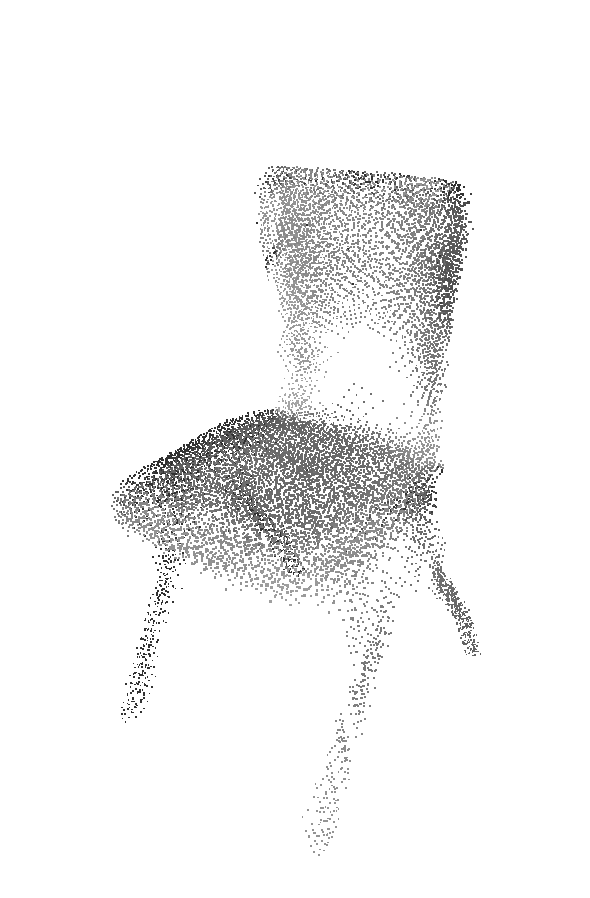} 
		& \includegraphics[width=\insz\textwidth]{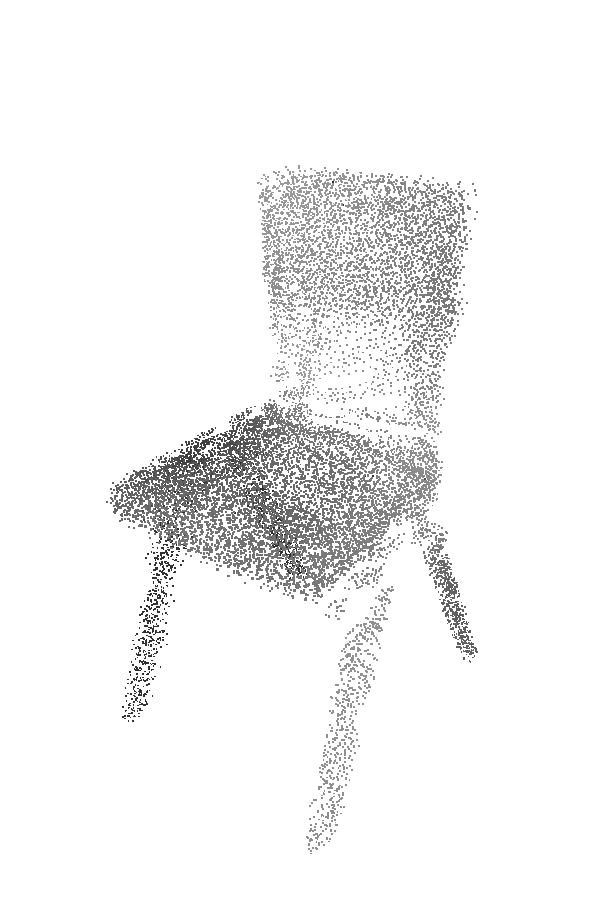} 
		& \includegraphics[width=\insz\textwidth]{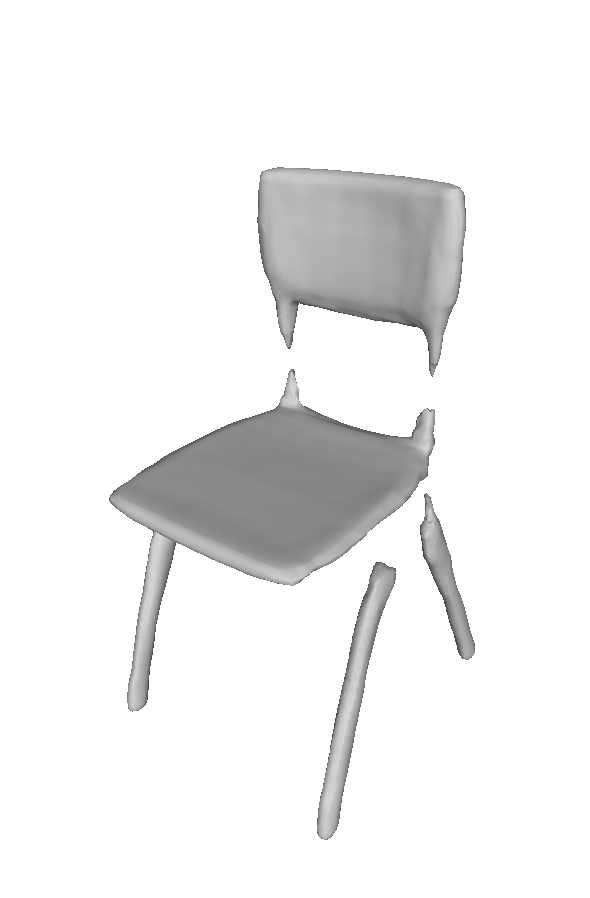} 
		& \includegraphics[width=\insz\textwidth]{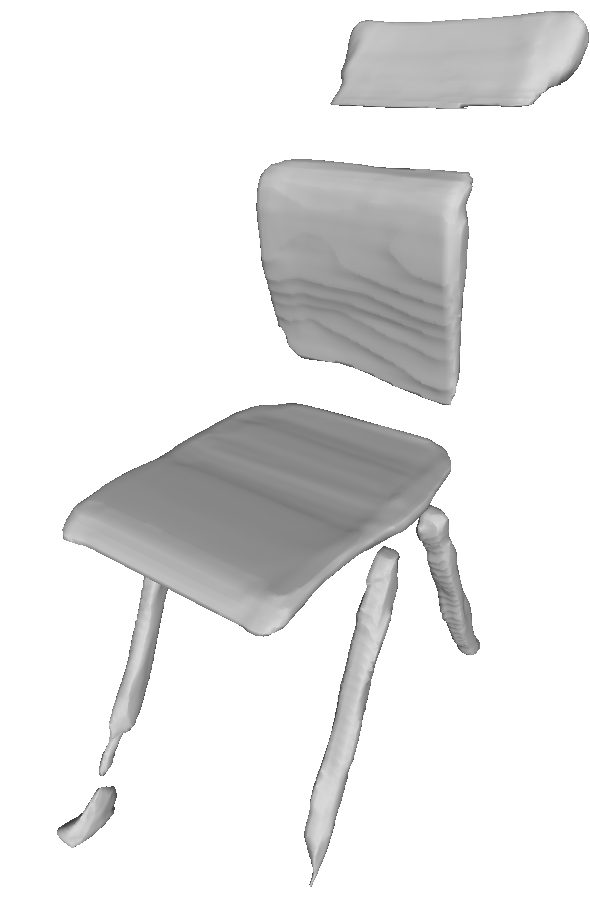} 
		& \includegraphics[width=\insz\textwidth]{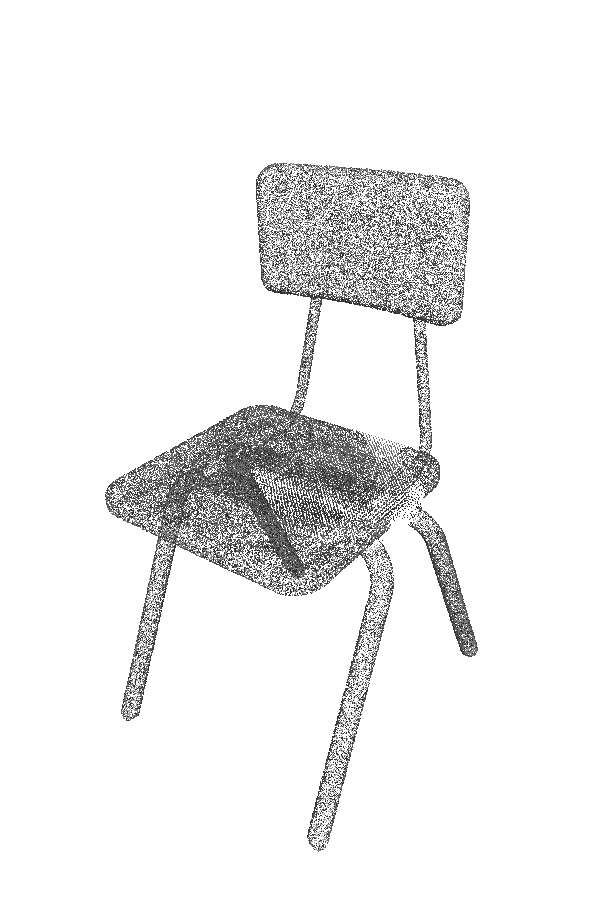} 
		& \includegraphics[width=\insz\textwidth]{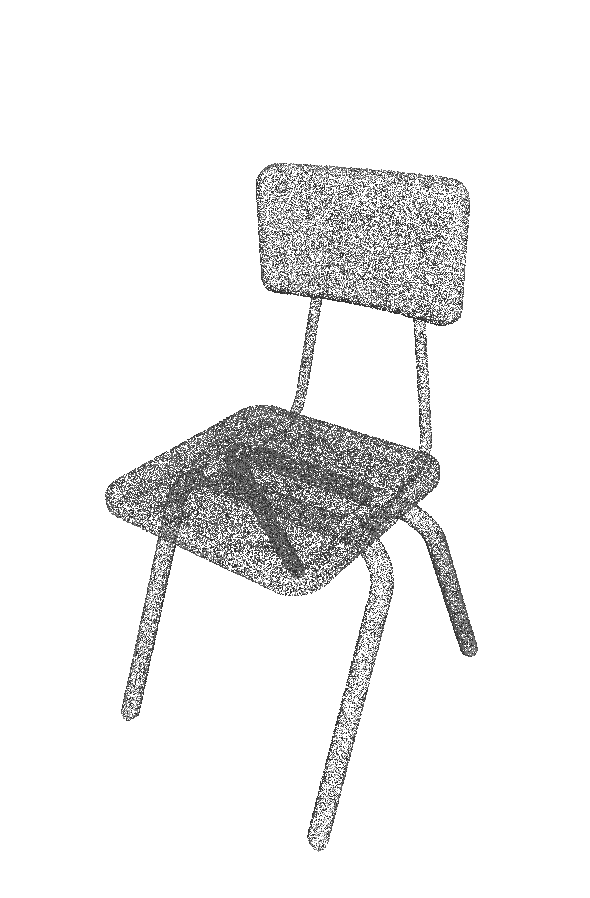} 
		\\[-3pt]
		&   
		& $2.45  \;\vert\; {\color{highlight2nd} 93.87}$ 
		& ${\bf 0.783} \;\vert\; 54.69$ 
		& ${\color{highlight2nd}0.815}  \;\vert\; 53.26$ 
		& $1.20  \;\vert\; 70.77$ 
		& $13.9  \;\vert\; 55.75$ 
		& $1.32  \;\vert\; {\bf 95.02}$
		\\[3pt]
		\hdashline\\[-6pt]
		\multirow{2}{*}[30pt]{\rotatebox{90}{\textbf{Lamp}}} 
		& \includegraphics[width=\sz\textwidth]{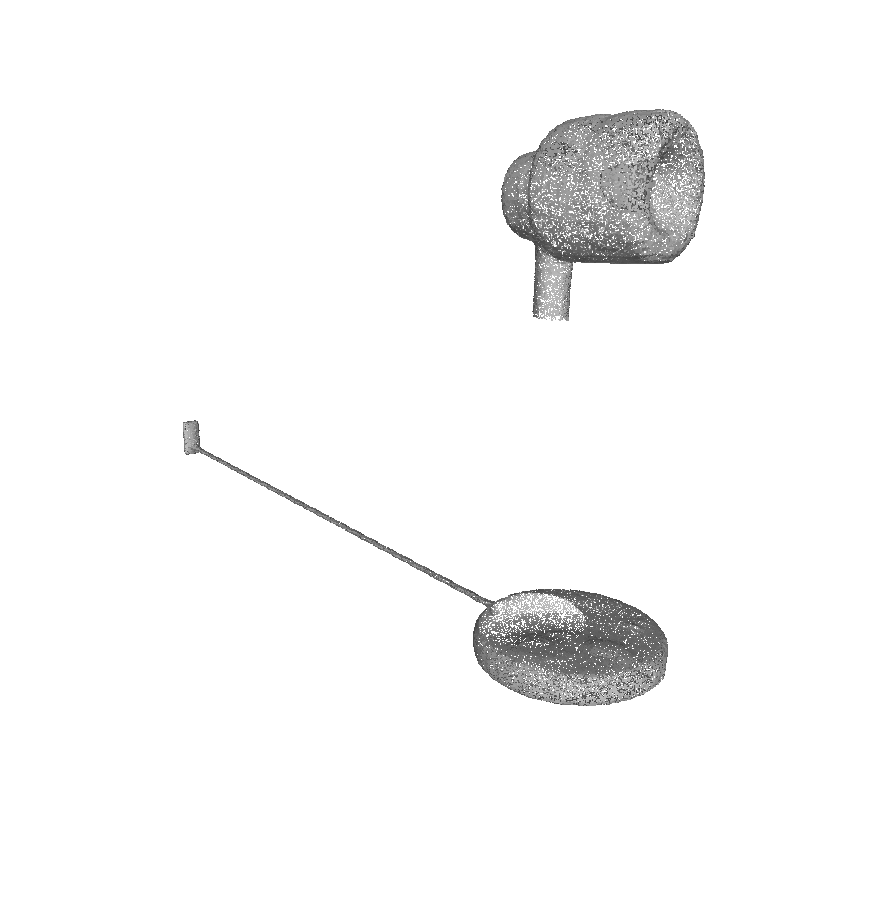} 
		& \includegraphics[width=\sz\textwidth]{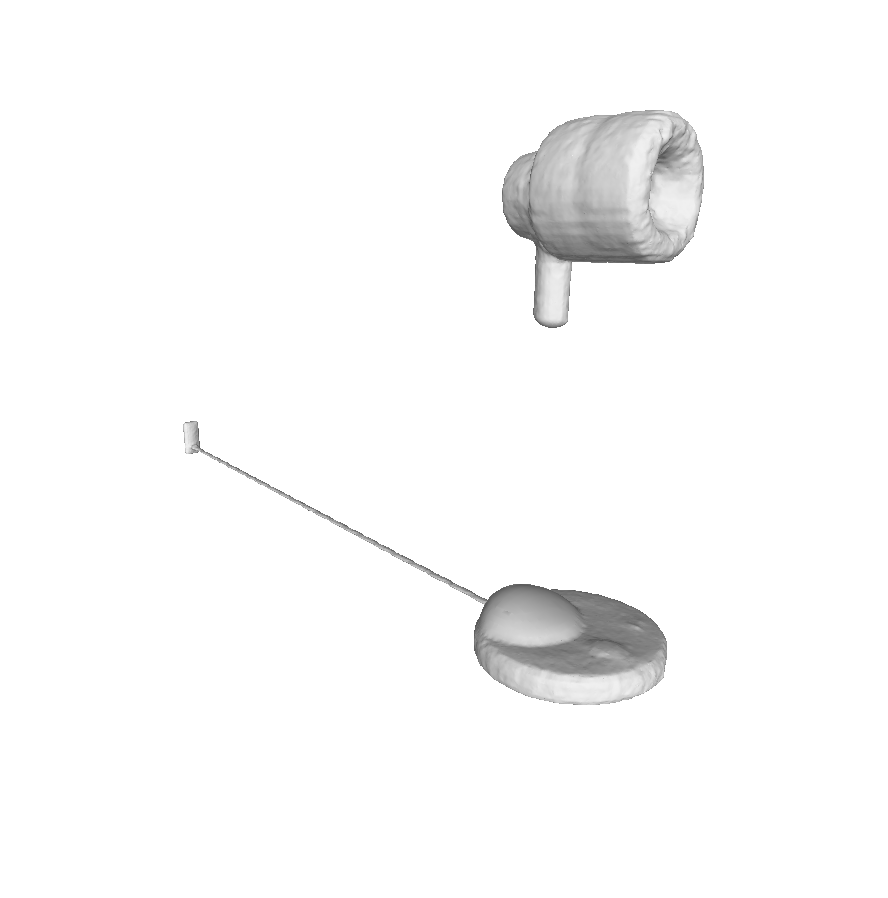} 
		& \includegraphics[width=\sz\textwidth]{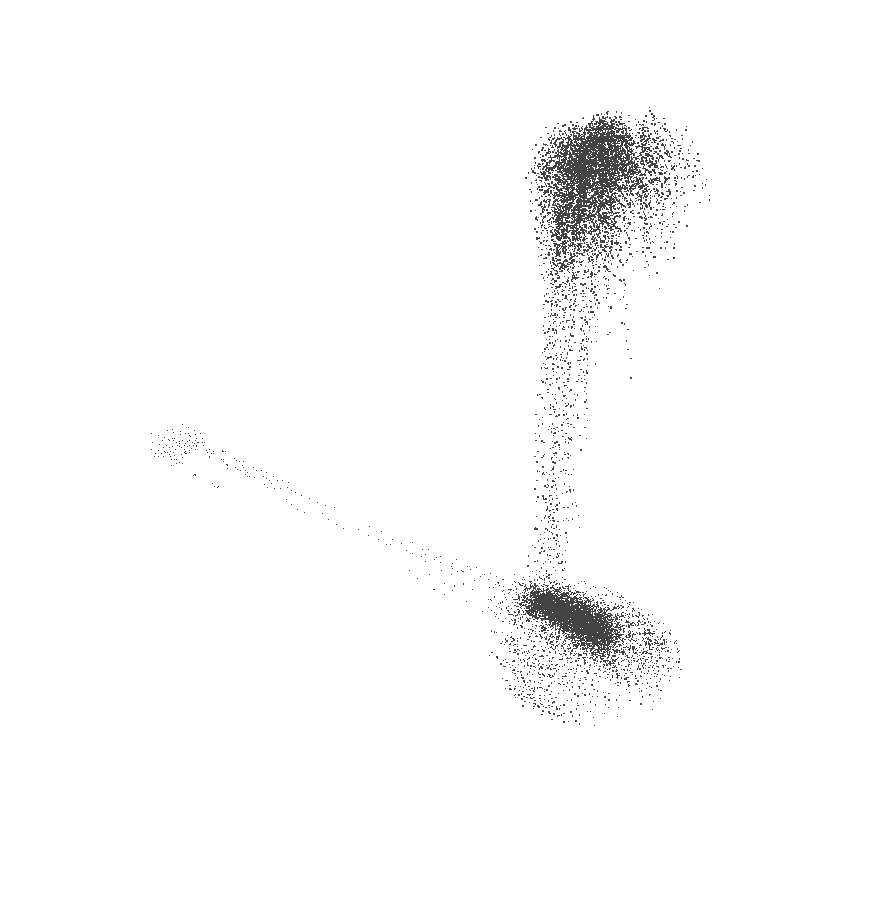} 
		& \includegraphics[width=\sz\textwidth]{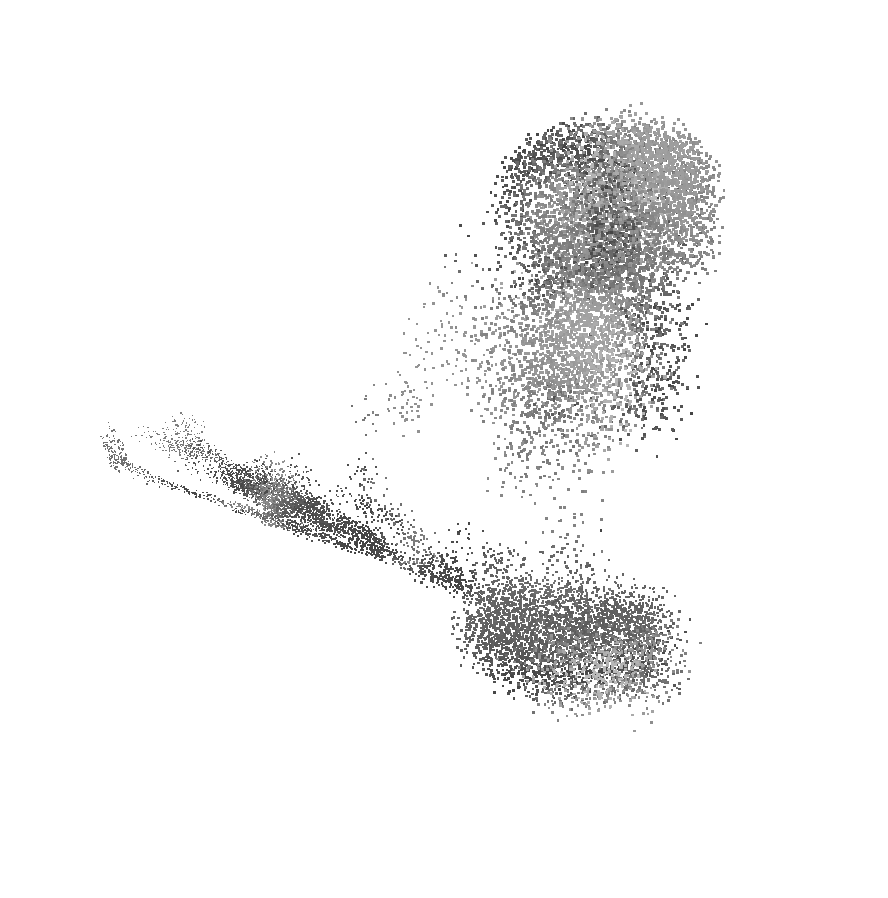}
		& \includegraphics[width=\sz\textwidth]{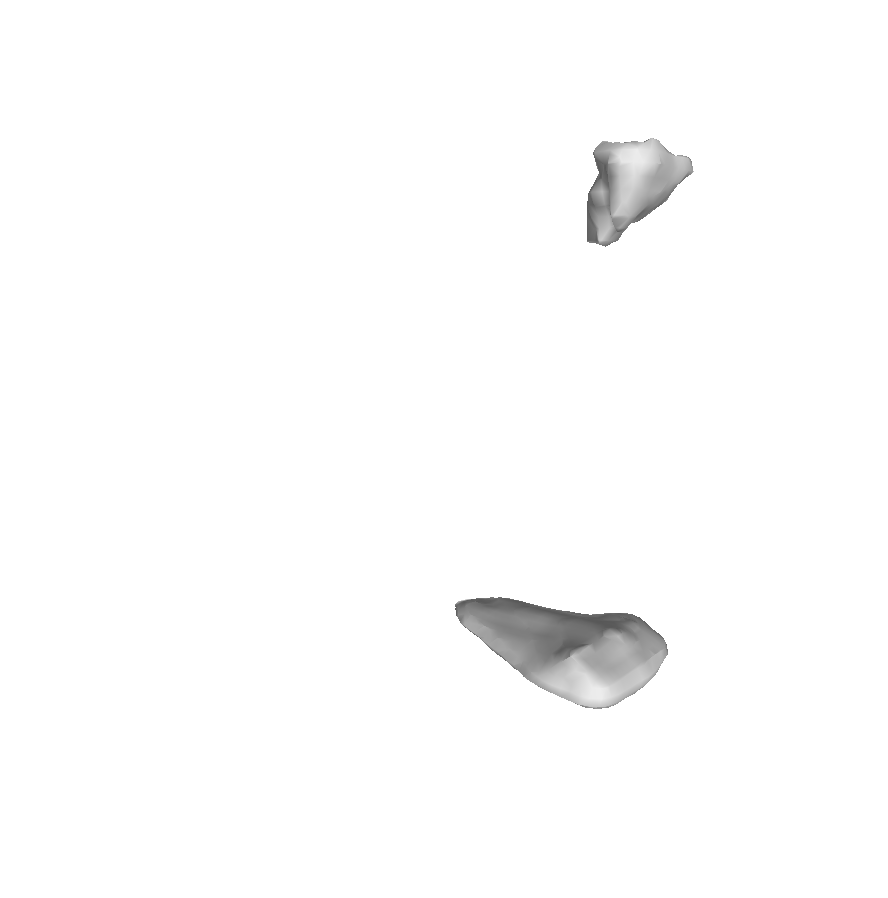} 
		& \includegraphics[width=\sz\textwidth]{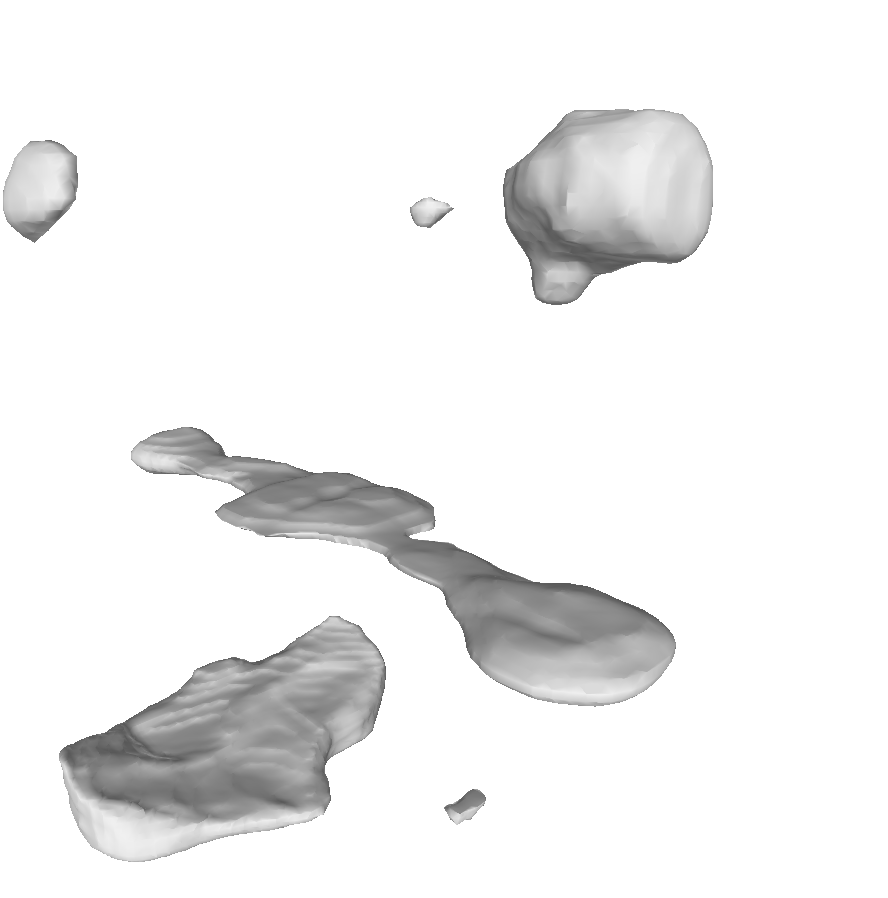}
		& \includegraphics[width=\sz\textwidth]{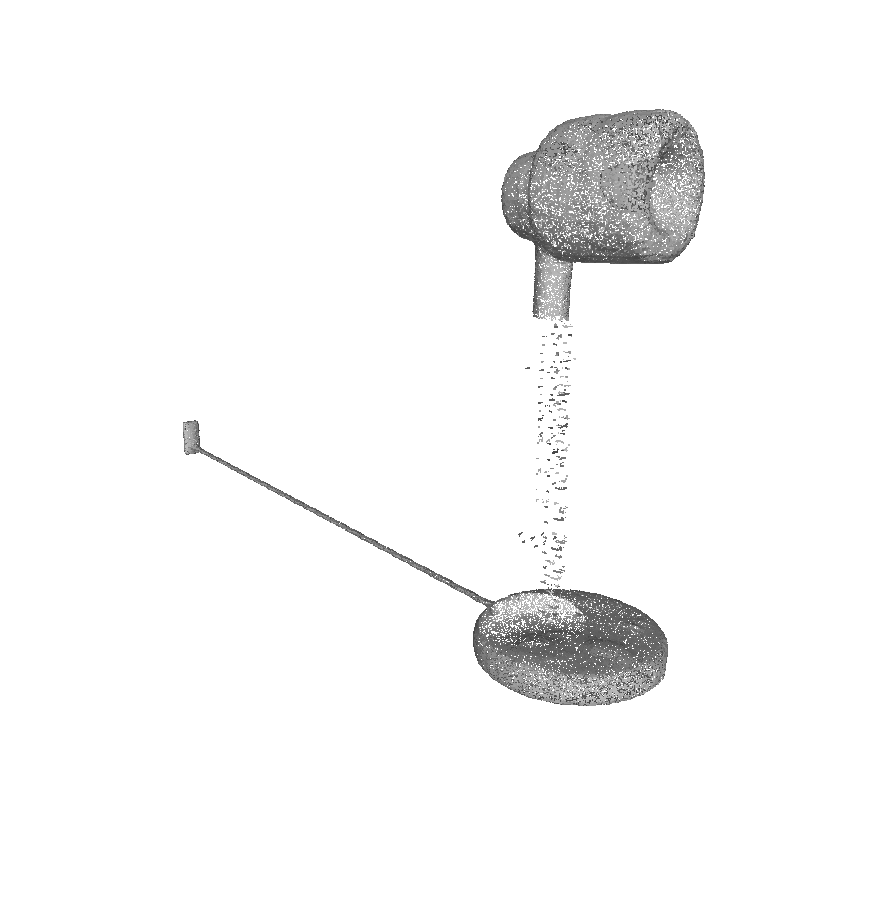} 
		& \includegraphics[width=\sz\textwidth]{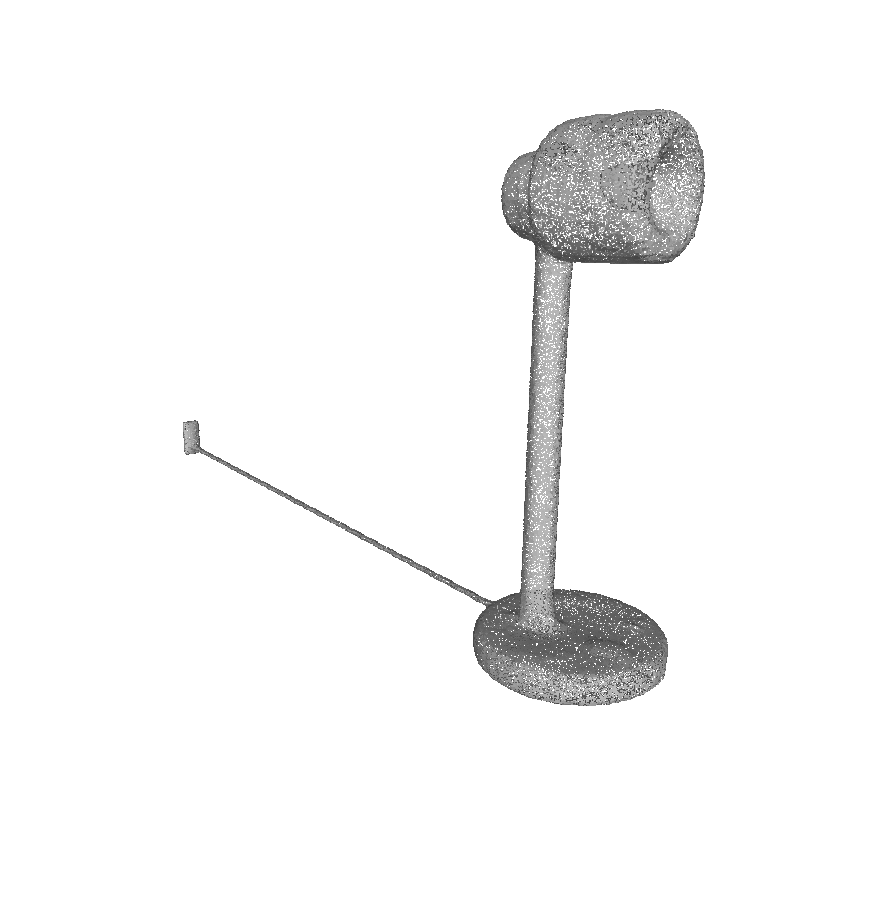}  
		\\[-1pt]
		&  
		& $2.57  \;\vert\; {\bf 94.53}$ 
		& $2.98  \;\vert\; 56.68$ 
		& ${\bf 1.91}  \;\vert\; 58.61$ 
		& $3.51  \;\vert\; 58.51$ 
		& $47.3  \;\vert\; 47.37$ 
		& ${\color{highlight2nd}2.40}  \;\vert\; {\color{highlight2nd}93.80}$ 
		\\[3pt]
		\hdashline\\[-7pt]
		\multirow{2}{*}[18pt]{\rotatebox{90}{\textbf{Sofa}}} 
		& \includegraphics[width=\sz\textwidth]{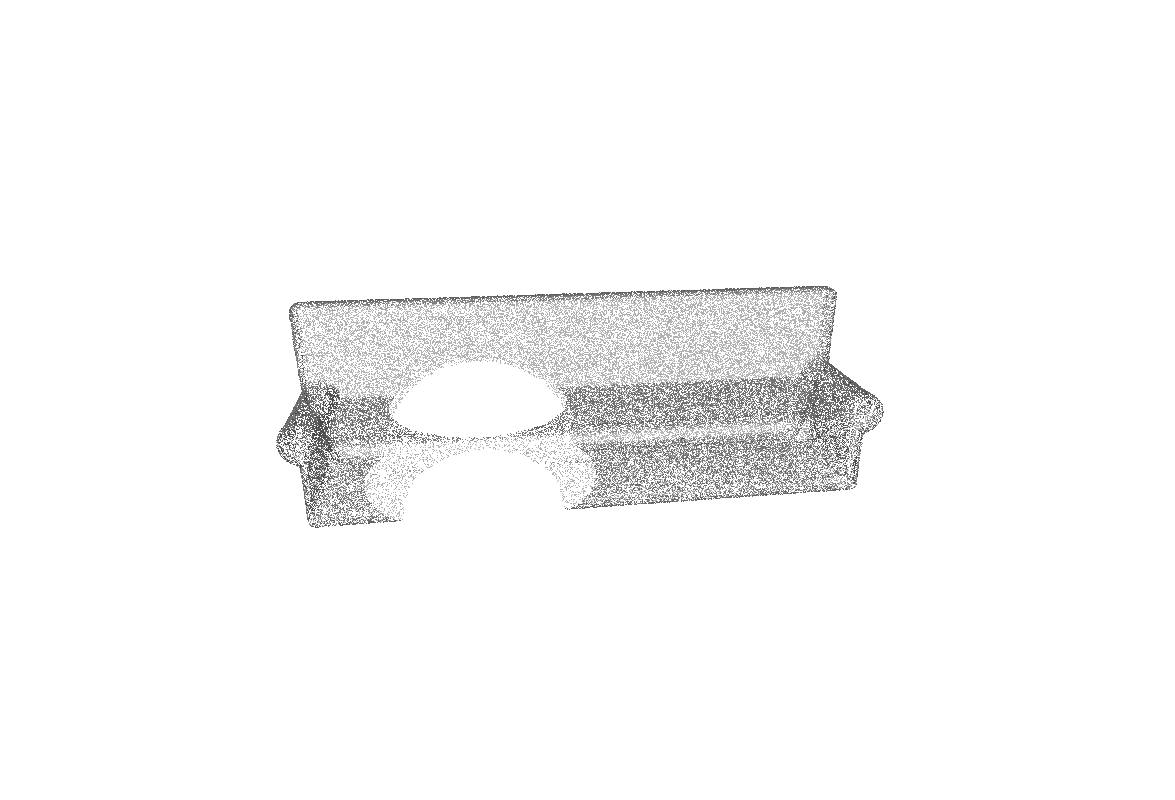} 
		& \includegraphics[width=\sz\textwidth]{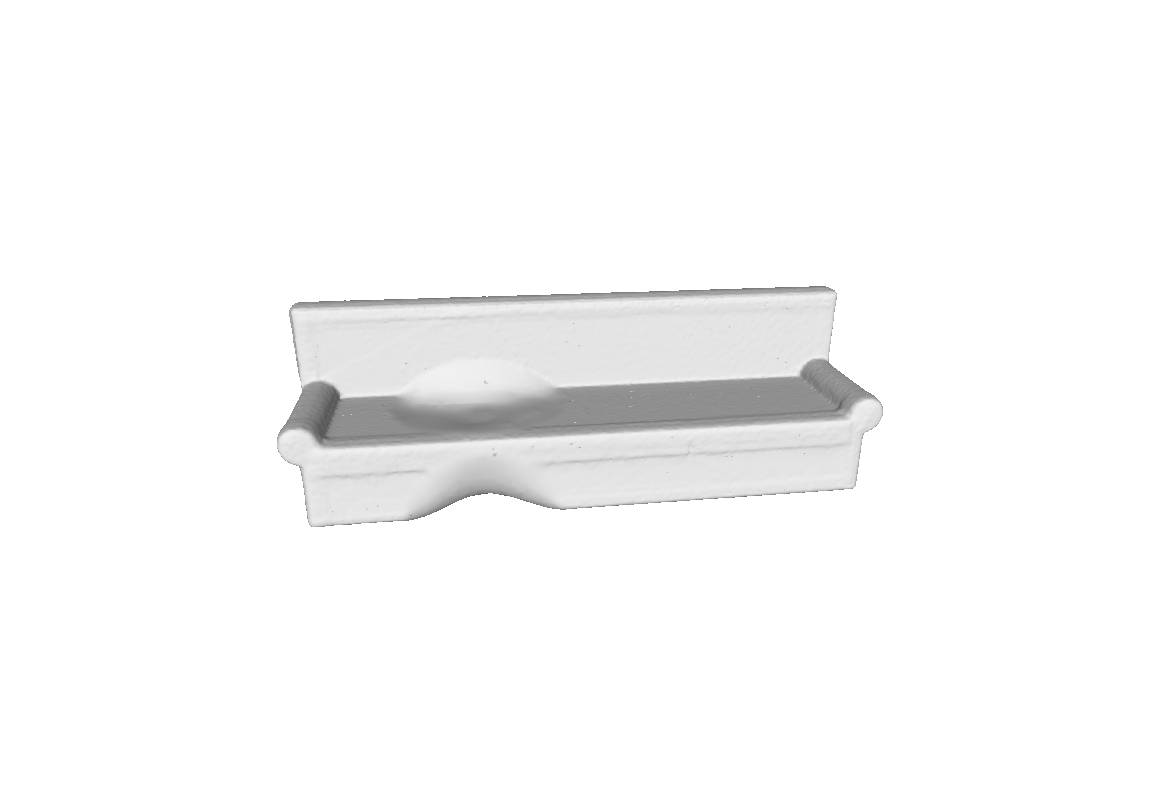} 
		& \includegraphics[width=\sz\textwidth]{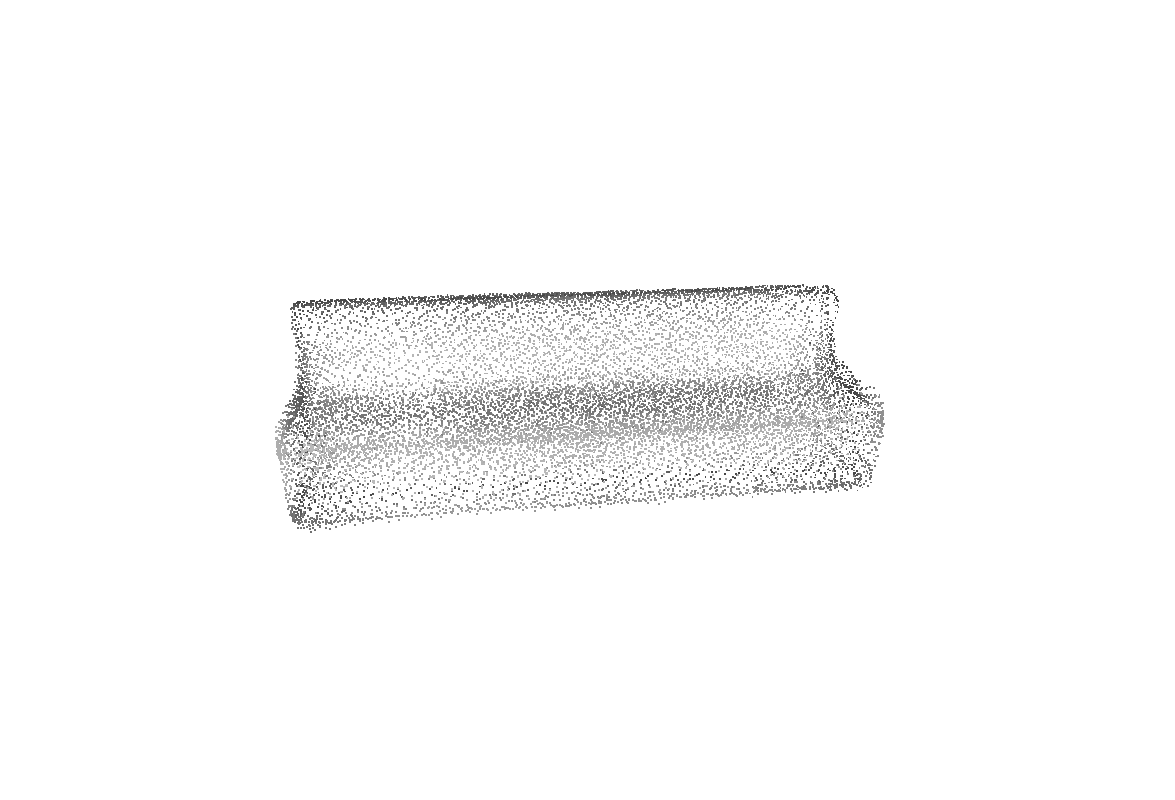} 
		& \includegraphics[width=\sz\textwidth]{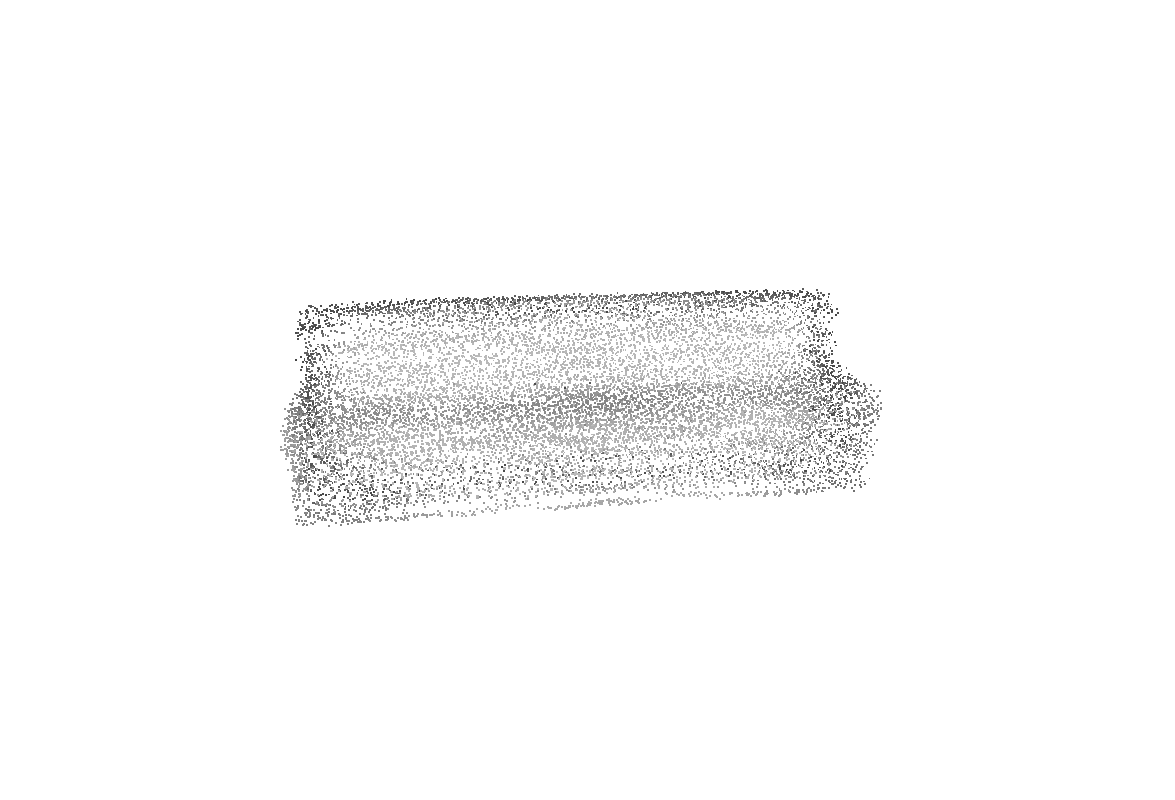} 
		& \includegraphics[width=\sz\textwidth]{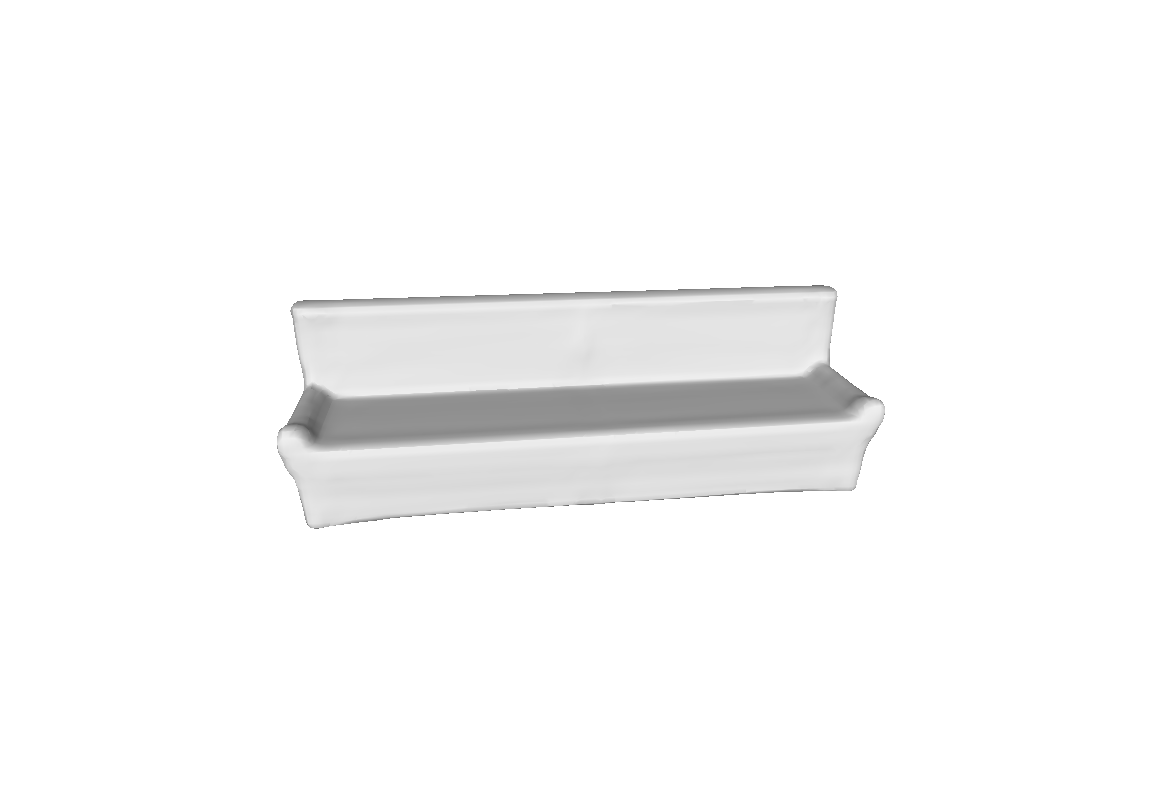} 
		& \includegraphics[width=\sz\textwidth]{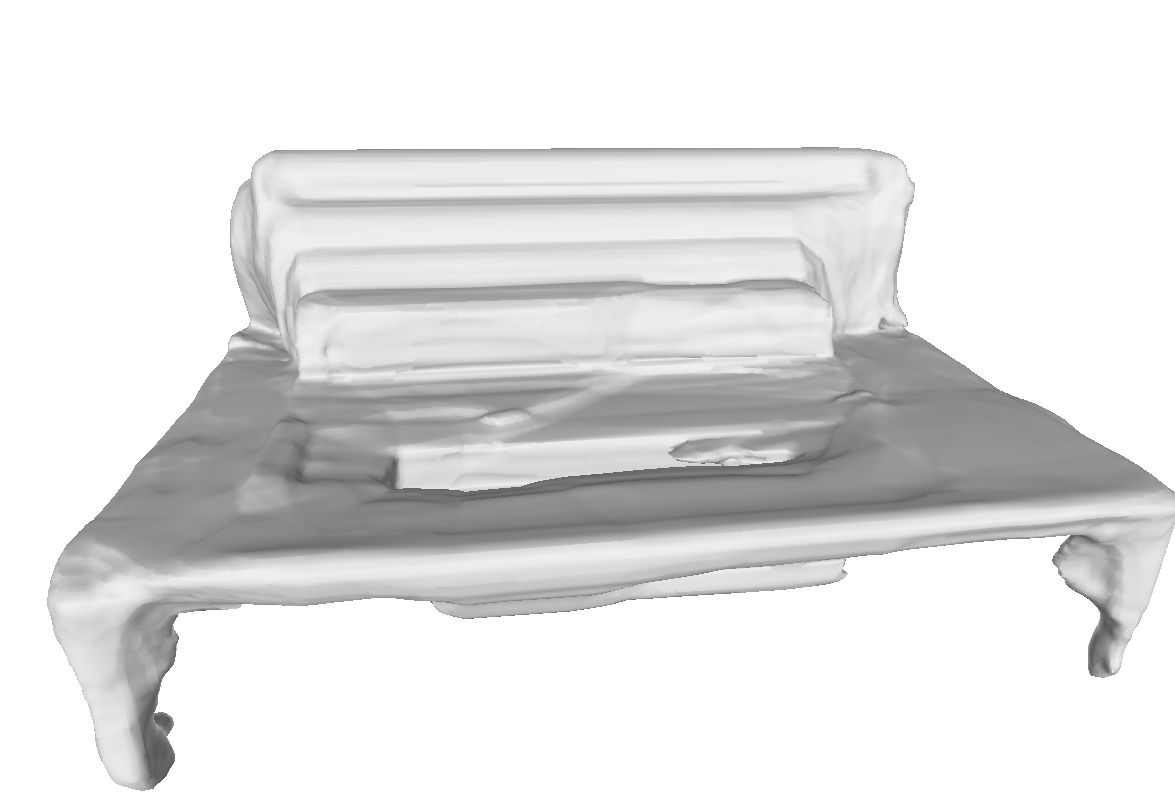} 
		& \includegraphics[width=\sz\textwidth]{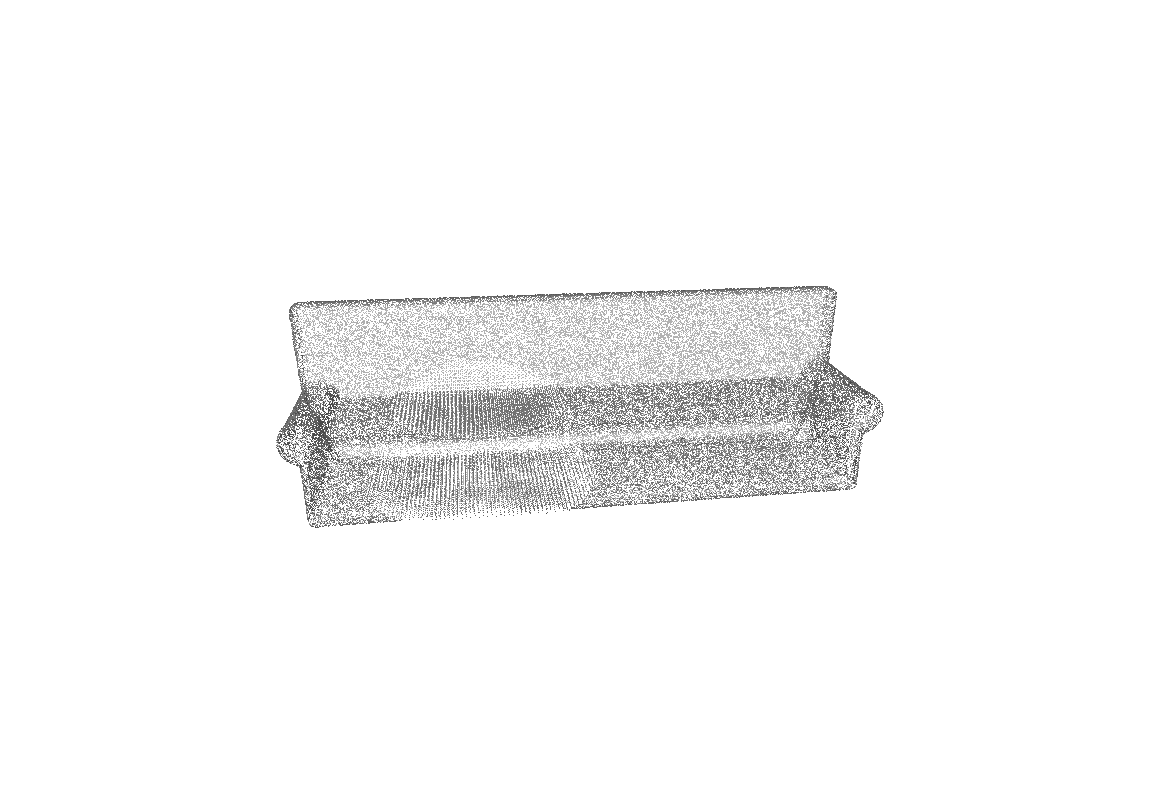} 
		& \includegraphics[width=\sz\textwidth]{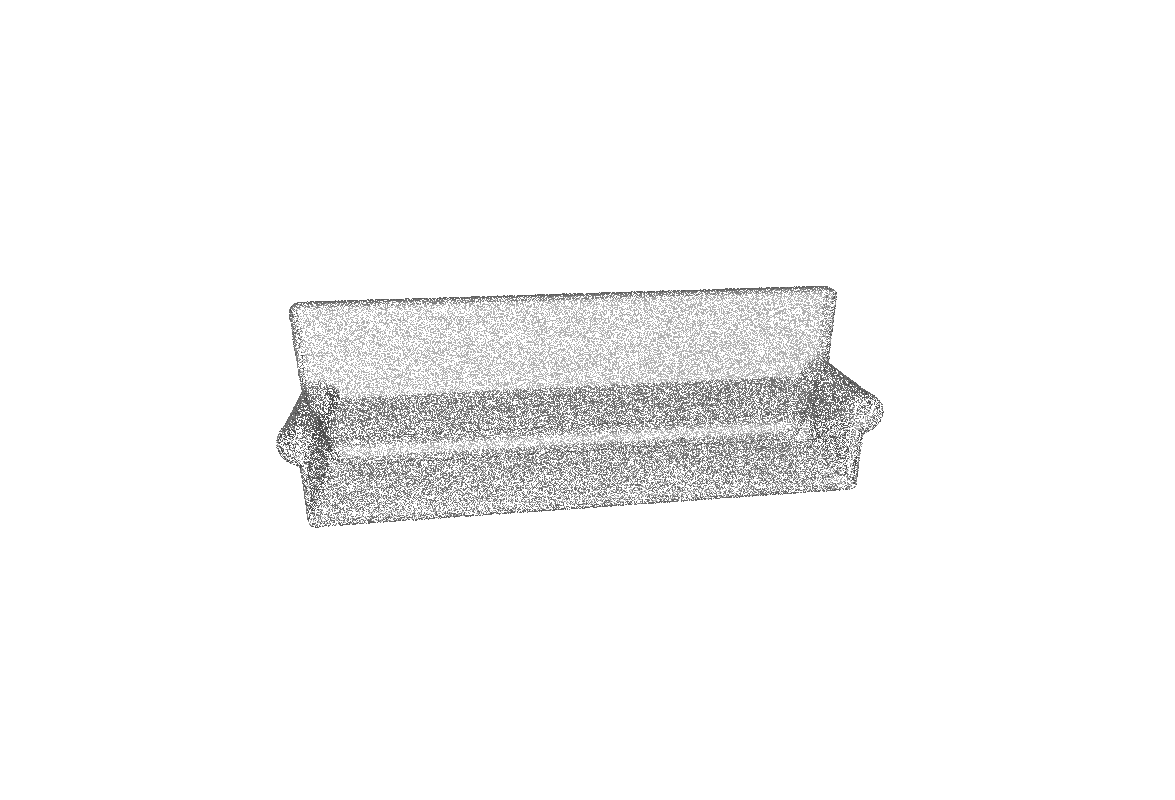} 
		\\[-6pt]
		&  
		& $1.37   \;\vert\; {\color{highlight2nd}94.19}$ 
		& ${\color{highlight2nd}0.503}  \;\vert\; 66.76$ 
		& $0.592   \;\vert\; 62.22$ 
		& $0.993  \;\vert\; 78.53$ 
		& $34.3   \;\vert\; 36.55$ 
		& ${\bf 0.274}  \;\vert\; {\bf 96.16}$
		\\[3pt]
		\hdashline\\[-7pt]
		\multirow{2}{*}[20pt]{\rotatebox{90}{\textbf{Table}}} 
		& \includegraphics[width=\sz\textwidth]{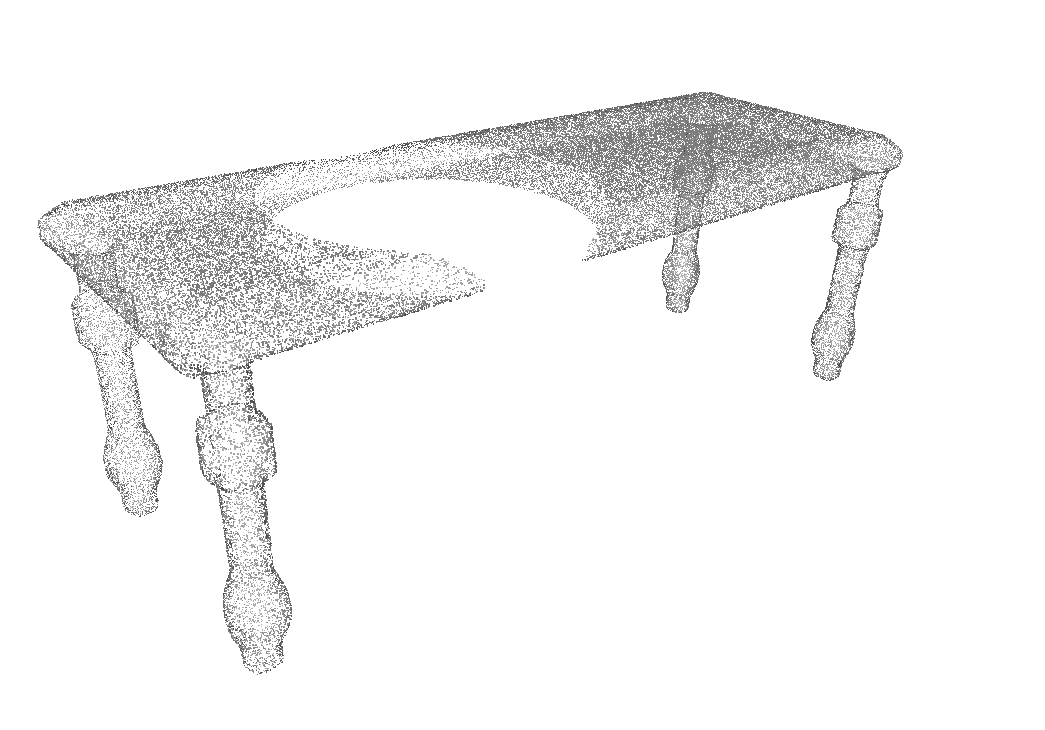} 
		& \includegraphics[width=\sz\textwidth]{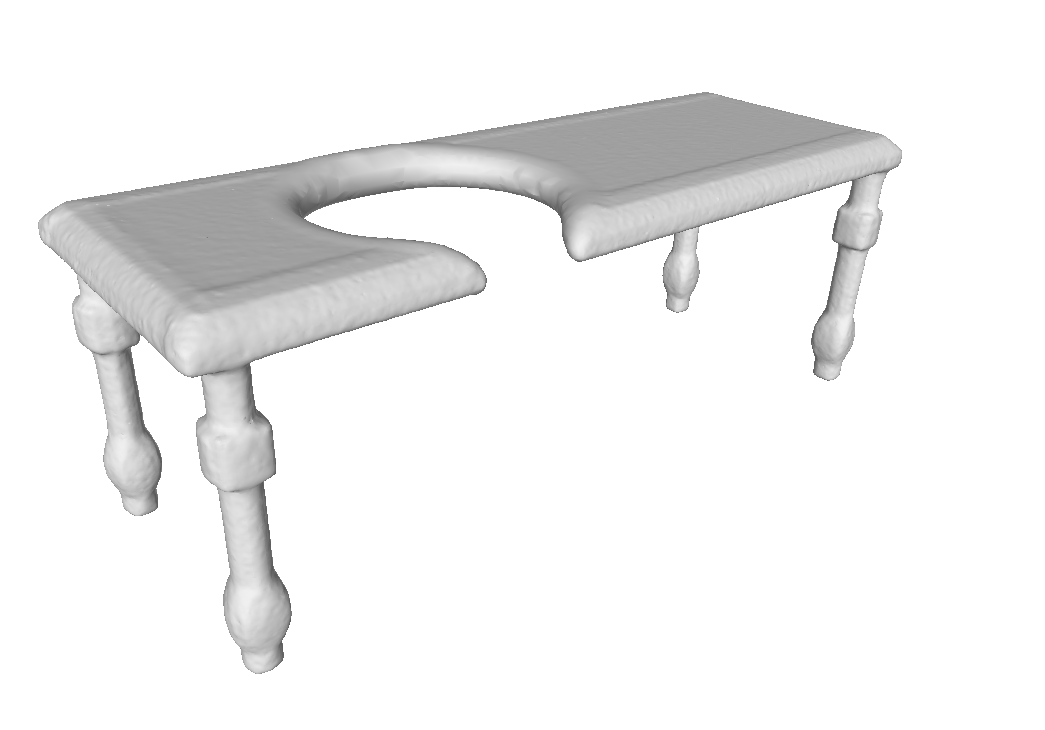} 
		& \includegraphics[width=\sz\textwidth]{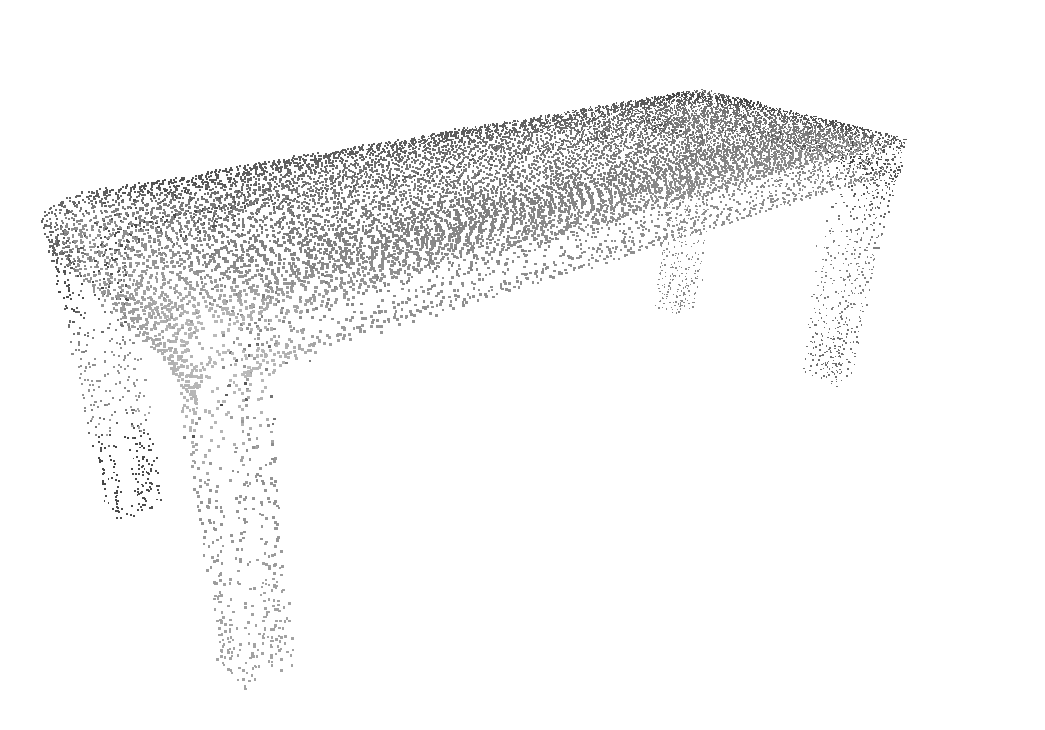} 
		& \includegraphics[width=\sz\textwidth]{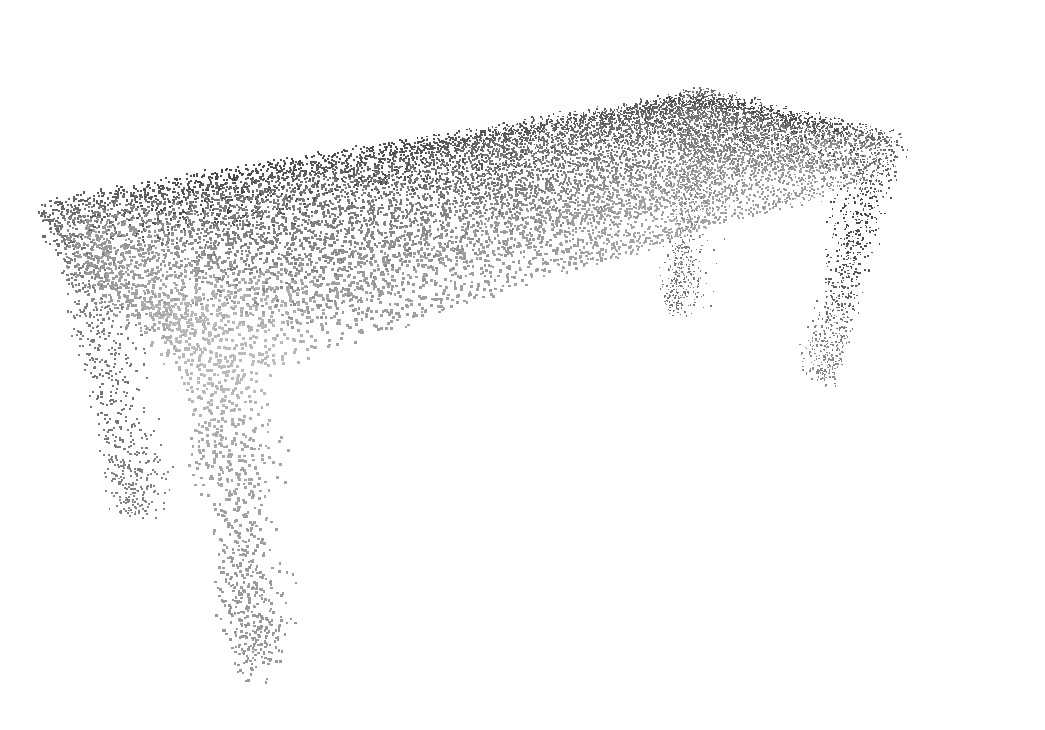} 
		& \includegraphics[width=\sz\textwidth]{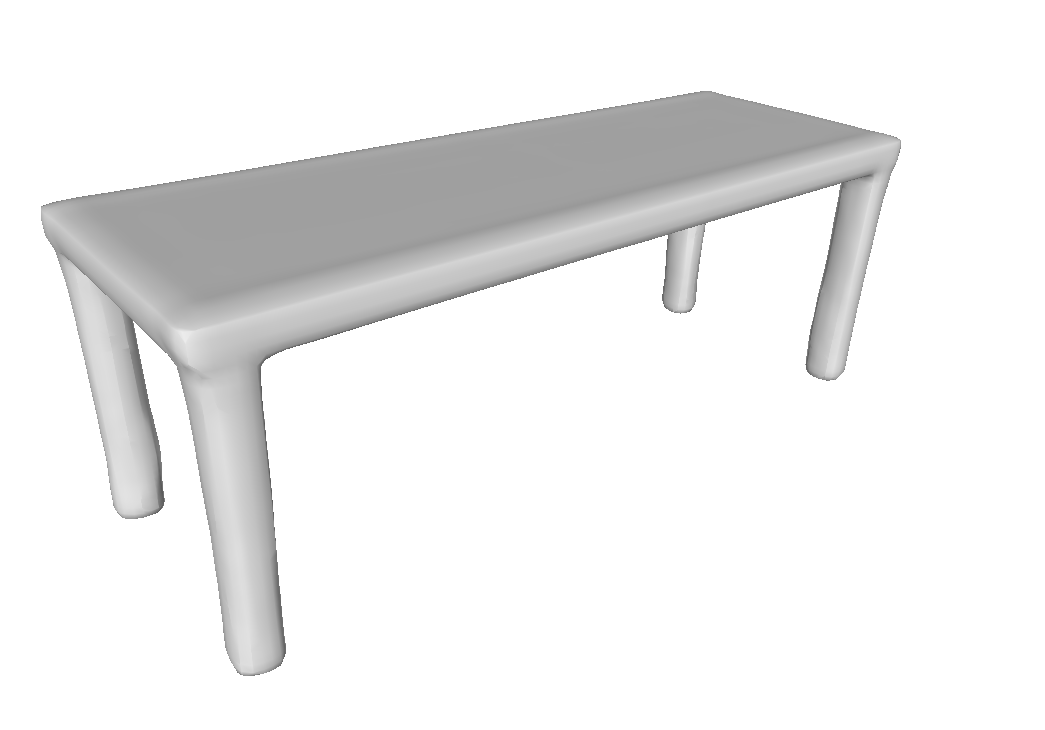} 
		& \includegraphics[width=\sz\textwidth]{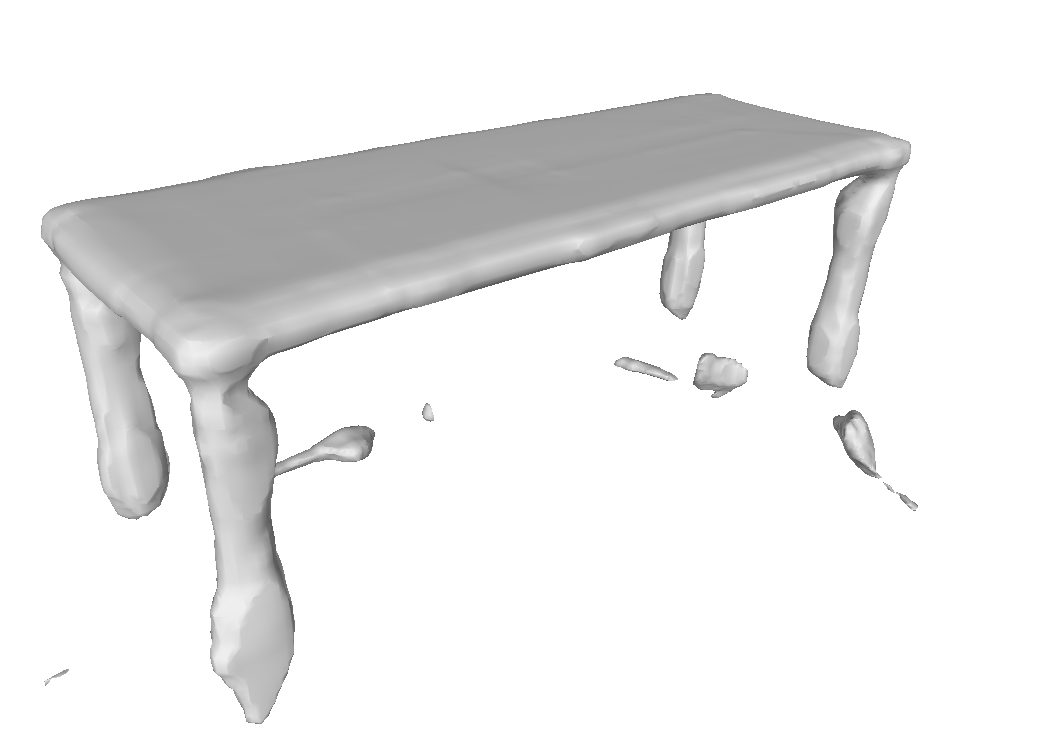} 
		& \includegraphics[width=\sz\textwidth]{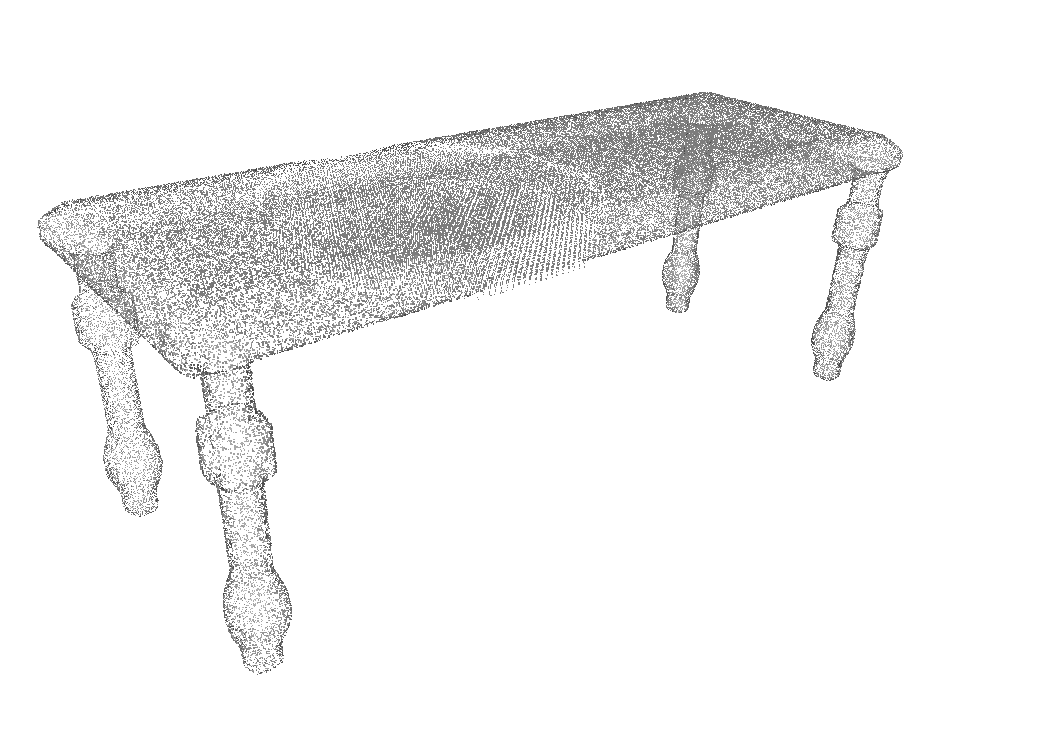} 
		& \includegraphics[width=\sz\textwidth]{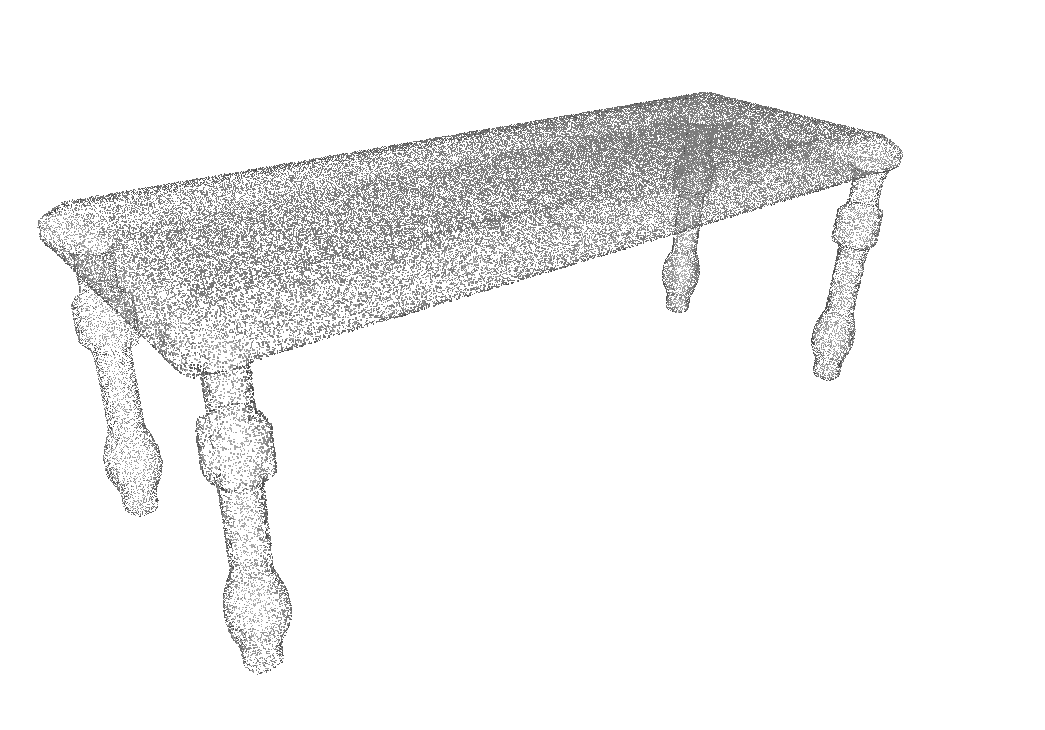} 
		\\[-1pt]
		&  
		& $3.16   \;\vert\; {\color{highlight2nd}93.64}$ 
		& ${\bf 0.763}  \;\vert\; 56.48$ 
		& ${\color{highlight2nd}0.851}  \;\vert\; 52.61$ 
		& $1.33   \;\vert\; 78.79$ 
		& $27.7   \;\vert\; 59.87$ 
		& $0.976  \;\vert\; {\bf 96.35}$
		\\[-6pt] 
	\end{tabular}
	\caption{\textbf{Qualitative and quantitative comparison to state-of-the-art shape completion methods.} The values below the pictures reproduce results from Fig.~\ref{fig:f1_chamfer_graphs} shown as $10^3\cdot CD \;\vert\; F1$. Best values are \textbf{bold}, 2$^\text{nd}$ best ones are \textcolor{Cerulean}{blue}.
	As explained, the DeepSDF results differ from the original paper, since we did not have access to the original code for shape completion.}
	\label{fig:qualitative_comparison}
	\vspace{1.5em}
\end{figure*}

\boldparagraph{Quantitative evaluation.}
Fig.~\ref{fig:f1_chamfer_graphs} shows the average $F1$ and CD values for all object categories, values for individual object instances are given in Fig.~\ref{fig:qualitative_comparison}.
For our method, we provide the values separately for each hierarchy level.
We recall that for the $F1$-score, larger values (corresponding to higher precision and recall) are better, whereas for CD, lower distance is better.

We observe that in term of $F1$-score our method and PSR perform clearly best.
This is due to the fact that the other baselines regenerate the complete 3D point cloud from the latent encoding, thus discarding the original input points and replacing them by necessarily imperfect predictions.
On the contrary, PSR and ours only fill in new points where needed, thus avoiding approximation errors on well-sampled surfaces.
Our full coarse-to-fine method consistently outperforms PSR by a small margin of 1.15--2.70 percent points, across all classes.

The Chamfer distance (CD) paints a more complete, slightly different picture.
For instance, we can see that for some categories the coarse-to-fine approach is more beneficial, in particular categories with large planar structures like airplanes and sofas.
Also in terms of CD, \kaplan{} performs on par with the strongest baselines, PCN and Cascaded, and consistently improves over all other methods, with the exception of OccNet for \emph{chairs}.

\boldparagraph{Qualitative evaluation.}
The qualitative results presented in Fig.~\ref{fig:qualitative_comparison} illustrate the main benefits of our method.
When large, smooth surfaces are missing, such as an airplane wing or a tabletop, the task reduces to smooth interpolation, and all methods perform reasonably well (except for PSR, which cannot capture long-range context).
However, unlike learning-based approaches which regenerate the complete point cloud, \kaplan{} preserves the input data. This is an important advantage in the presence of intricate geometry, as details tend to be washed out when decoding from a generalised encoding, such as on the legs of the table.
Moreover, \kaplan{} appears to strike a good compromise between global object shape and local surface properties, for instance it is able to recover thin structures, such as the legs of a foldable chair or the pole of the lamp, which are not captured correctly by the baselines. 
%

\boldparagraph{Ablation and parameter study.}
\begin{table}[t!]
	\ifthreedvfinal \vspace{-0.5em} \fi
	\centering
	\resizebox{.48\textwidth}{!}{
		\begin{tabular}{lcc}
			\toprule
			$\nK$ ($\nRes=35$) & $CD [\times 10^3]\downarrow$ & $F1 \uparrow$ \\
			\midrule
			{\phantom{1}1 tangential}	& 1.56 	& 	93.15 \\
			{\phantom{1}2 random}	    & 1.91 	& 	93.04 \\
			{\phantom{1}3 canonical}  	& 0.31  &	94.29 \\
			{\phantom{1}5 random}  		& 1.89	& 	93.08 \\
			{\phantom{1}9 canonical}	& 0.28	& 	94.39 \\
			{12 random}  	& 1.90	& 	93.08 \\
			{27 canonical}	& 0.27	& 	94.50 \\ 
			\bottomrule
		\end{tabular}
		\qquad
		\begin{tabular}{ccc} 
			\toprule
			$\nRes$ ($\nK=3$) & $CD [\times 10^3]\downarrow$ & $F1 \uparrow$ \\
			\midrule
			&		&		  \\      	
			{$15\times15$}		& 0.46	& 	93.50 \\
			{$35\times35$}  	& 0.31  &	94.29 \\
			{$49\times49$}  	& 0.26	& 	94.41 \\
			{$65\times65$}		& 0.17	&   95.33 \\
			{$85\times85$}  	& 0.13	&	95.51 \\
			{$105\times105$}  	& 0.12	&	95.60 \\
			\bottomrule
		\end{tabular}
	}
	\vspace{-2mm}
	\caption{\textbf{Parameter study at coarse level $\nLevelCoarse$.} The metrics CD and $F1$ score are reported for the category \textit{Plane}.} 
	\label{tab:parameters_study}
	\ifthreedvfinal \vspace{-0.6em} \else \vspace{0.3em} \fi
\end{table}
We conducted studies for the \emph{airplane} category, at level $\nLevelCoarse$ with $10$ query points.

\textit{Ablation -- normals. }
To assess here the contribution of the normals in our method, we simply remove them from the input, as well as from the decoder.
With normals, we obtain $F1=94.29$, respectively $10^3\cdot CD=0.31$. Without normals, these values drop to $F1=93.72$, respectively $10^3\cdot CD=0.35$.
We conclude that \kaplan{} still performs very well for unoriented point clouds (with no normal information).
Still, although normal estimation could in principle be learned implicitly, feeding in explicit normals does help the network and leads to more accurate predictions (as seen from the significant decrease of the CD).
A possible interpretation is that
depth and normals are complementary when it comes to discontinuities and smooth surfaces.
%

\textit{KAPLAN parameters. }
Two experiments are carried out to study the impact of the number of planes $\nK$ and the resolution $\nRes$.
In each experiment, we fix one parameter and vary the other.
For $\nK$, we also vary the plane orientation.
We start with the simplest setup, a single tangential plane aligned with the local surface normal. We then test randomly oriented planes, constrained to pairwise angles of at least 30$^\circ$ to avoid degenerate configurations with multiple very similar planes.
Lastly, we test ``canonical'' planes with uniformly distributed orientations.

It turns out that good resolution within a plane is beneficial: from Tab.~\ref{tab:parameters_study}, we observe a significant gain by increasing $\nRes$ from $15$ to $35$.
On the other hand, at most a small improvement is possible when increasing the number of planes beyond 3.
Interestingly, randomly oriented planes perform a lot worse than canonical ones, as the network has difficulties to learn the prediction of the valid flag.
This could be due to the fact that objects in ShapeNet have main directions aligned to the canonical coordinates.
Please see the \textit{supplementary material} for further analysis. 

\ifthreedvfinal
\noindent
\begin{minipage}{\columnwidth}
	\vspace{2mm}
	\footnotesize
	\noindent
	\textbf{Acknowledgments.}~
	We thank Songyou Peng for the help with DeepSDF.
	This work was partially supported by Innosuisse funding (Grant No. 34475.1 IP-ICT).
\end{minipage}
\fi

\section{Conclusion}
We presented a new approach to shape completion for 3D point clouds.
It is based on \kaplan{}, a novel 3D point descriptor that locally aggregates features in multiple 2D projections, making the 3D data amenable to standard 2D convolutional encoding and decoding.
Importantly, the encoding includes a \emph{valid flag} to mark unobserved object regions, thus making it possible to fill in new data only where needed, while keeping original samples if possible.
We also embed \kaplan{} in a coarse-to-fine scheme to reconcile the use of global context with the need for local geometric detail.
Empirically, the proposed approach is able to complete missing areas effectively, reaching high accuracy in terms of the predicted point (respectively surface) locations, while keeping existing geometry intact.
Future work will assess the potential of \kaplan{} as a generic descriptor for other tasks like semantic labelling, denoising or scene analysis.
%

\clearpage

\appendix
\section*{Supplementary}


In this document, we provide further technical details about the implementation and experiments, as well as additional visualizations.

\section{Network Details}
\label{sec:network}
We begin by describing the detailed network architecture, as well as design considerations regarding the batch size in order to improve training.

\subsection{Architecture of our U-net encoder-decoder}
\begin{figure}[h]
	\centerline{\includegraphics[scale=0.16]{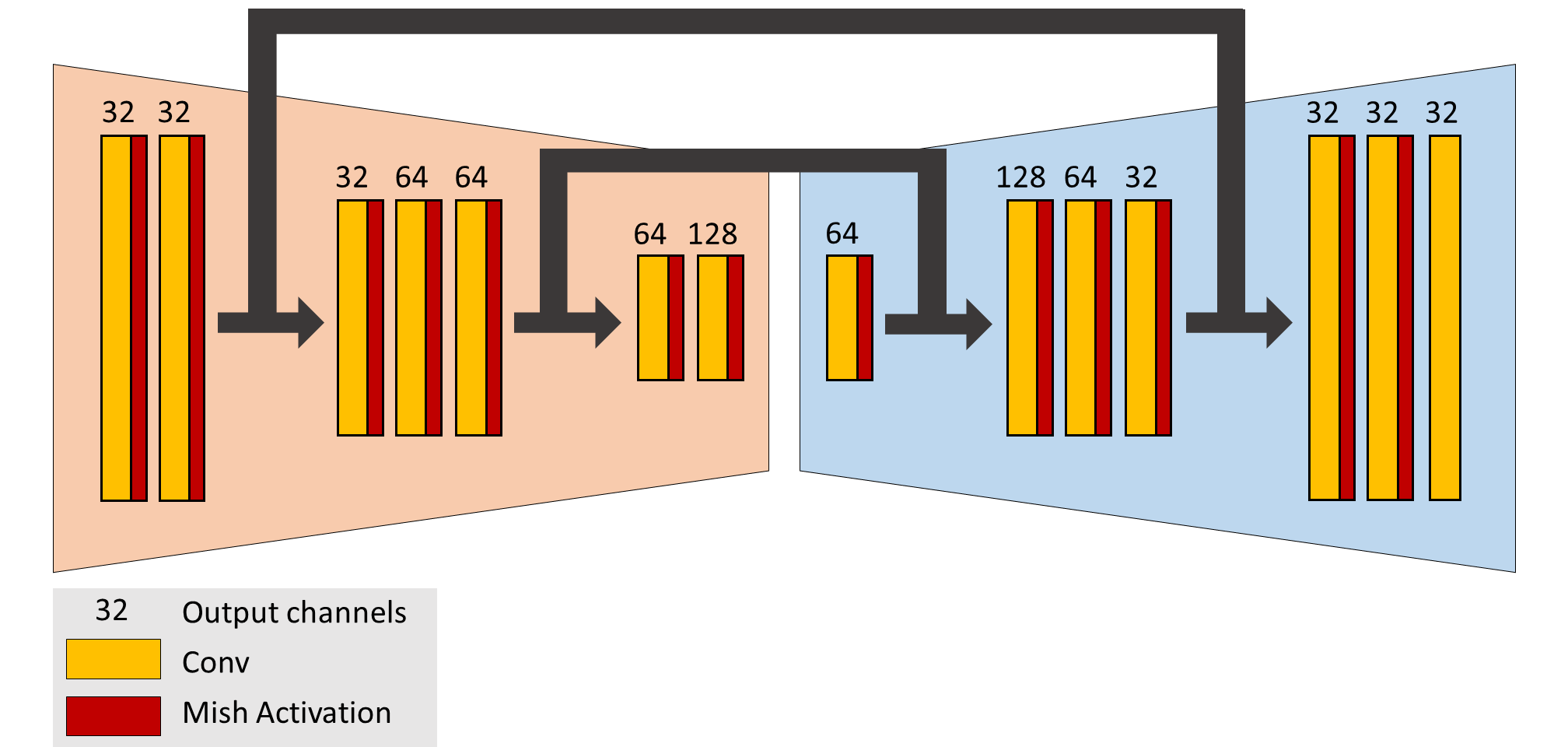}}
	\caption{\label{fig:ushaped_enc_dec} \textbf{Description of the U-shaped encoder-decoder}.}
\end{figure}

The network design of \kaplan{} follows the ideas of U-net \cite{Ronneberget_CoRR_2015}, and is represented in Fig.~\ref{fig:ushaped_enc_dec}. Both the encoder and the decoder are fully convolutional.
We experimented with different sizes of convolutions, and observed that larger convolutions improve the performance of the network, especially regarding the prediction of valid flags.
Our hypothesis is that a larger spatial context is needed to detect the missing regions and discriminate them from empty background.
In practice, we set the filter size to $35$ at the coarsest level, and divide it by two for every finer level.
Every convolution, except for the last one, is followed by the MISH activation function \cite{Misra-Arxiv-2019}, as we found that it improved our training.
On the one hand, MISH provides a smoother energy landscape, resulting in less peaked training losses than the ones obtained with ReLU. 
On the other hand, we also observed a significant improvement of the training performance for the valid flag and depth, $15\%$, respectively $28\%$.
We used max-pooling to downsample the data in the encoder, and nearest neighbour upsampling in the decoder. As ususal, skip connections propagate high-frequency encoder information to the corresponding decoder blocks.

\begin{figure*}[ht!]
	\centerline{\includegraphics[scale=0.3]{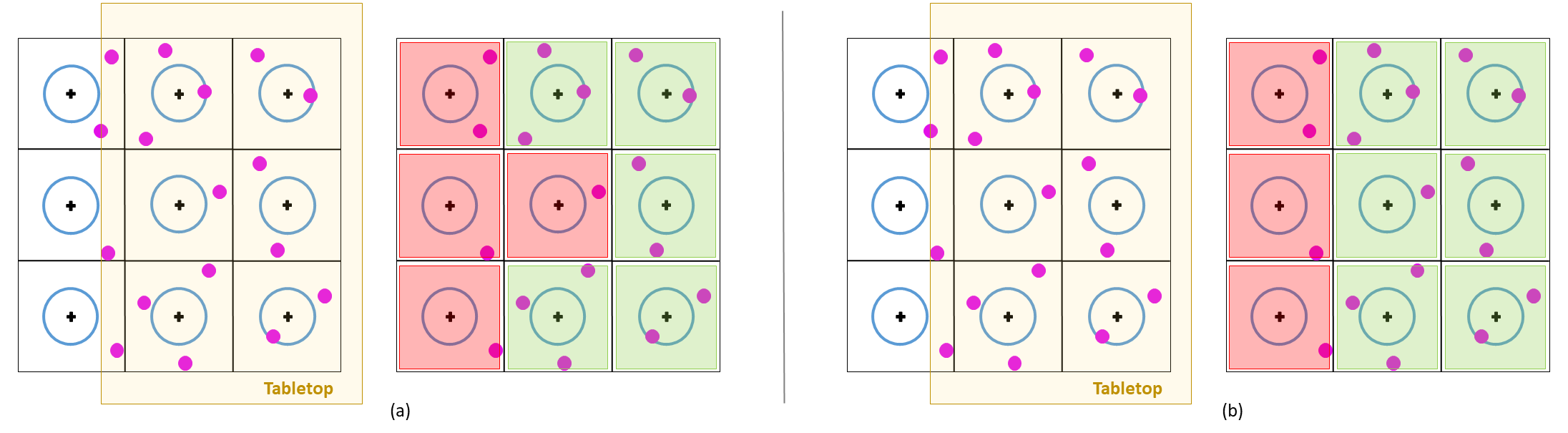}}
	\caption{\label{fig:ray_tracing} \textbf{Valid flag attribution}. (a) Points sampled on the tabletop are projected into the \kaplan{} plane. In each cell, we check if the barycenter of the projections is in the neighbourhood of the cell center. If so, the cell receives a valid flag. (b) In a second step, we verify for non-valid cells if they receive at least one projection, and have at least 3 neighbouring cells with a valid flag. If this is the case, as is for the center cell in our example, it also receives a valid flag.} 
\end{figure*}

\subsection{Batch Size}
During experimentation, we found that the batch size $bs$ is of great importance for the learning, especially at the coarse level where the network has to locate the holes in the object.
The initial query points for the coarse level are uniformly distributed over the object, and although the planes of each \kaplan{} are of size $1$ at that level (meaning they span the full object), there are several qualitatively different scenarios.
In some cases, the planes will be seeded and oriented such that they are near the hole, in that situation the network should learn to predict small depths.
In other cases, the plane may be located far from the hole, which means the network should predict large depths.
Finally, planes can be uninformative for the completion task, typically when the hole is occluded w.r.t.\ the plane, meaning that points from other object parts are projected onto it -- when this happens, the network should learn to do nothing.

It is thus important to use large batches, such most batches include examples from different scenarios (do nothing, small correction, big correction). As a positive side effect, this also considerably speeds up the training, by reducing the time per epoch. \Eg, one epoch takes $16$ minutes with $bs = 32$, against $11$ minutes with $bs = 128$ for the category \textit{Tables}.

%
\subsection{Runtime and Performance}\label{sec:runtime}
We compare the performances of our method to a full 3D approach on a voxel grid in terms of runtime.
To do so, we compared the runtimes for predicting the occupancy of $1000$ voxels and the location of $1000$ points.
We compared to ScanComplete \cite{Dai-et-al-CVPR-2018}, a state-of-the-art voxel-grid completion approach.
Both approaches operate in a coarse-to-fine scheme, we evaluated the times necessary to predict $1000$ voxels or points at each level, and report the total.
In both cases, we only consider the inference time, without pre-computations.
For \cite{Dai-et-al-CVPR-2018}, we used the numbers reported in the paper at the coarsest resolution.
We found that our method takes $0.078$ seconds, while ScanComplete takes $0.104$ seconds.
Besides being slower, the quality of a voxel-grid approaches is tied to the voxel resolution, which is inevitably limited. \kaplan{} reconstructions do not suffer from the associated artefacts like inflated objects and loss of thin structures.

\section{Precomputation and Query Points Selection}\label{sec:precomputation}
Before computing \kaplan{}, one must detect empty cells of the input to set the valid flag. Moreover, one needs a scheme to aggregate the depth information in a meaningful way to avoid artefacts.
The following sections describe these two steps.
We also specify how query points are selected at each level.

\subsection{Valid Flag Attribution}
We recall that our method only predicts points near cell centers.
The input to \kaplan{} should reflect this fact and only mark cells as \emph{valid} if they contain enough projections near the center.
Otherwise, the object could be shifted or inflated, especially at the coarsest level.
Fig.~\ref{fig:ray_tracing}a illustrates this with the example of a \kaplan{} seeded close to the edge of a table.
One can see in the example that, if the leftmost cells of the plane were marked as \emph{valid}, the table would be enlarged.

The most straightforward solution would be to consider valid only the cells which receive projections near their centers (the blue circles in Fig.~\ref{fig:ray_tracing}).
However, this solution proved to be too conservative, as too few cells remain \emph{valid}. With such limited input, the task for the network becomes very difficult.

\begin{figure*}
	\centerline{\includegraphics[scale=0.23]{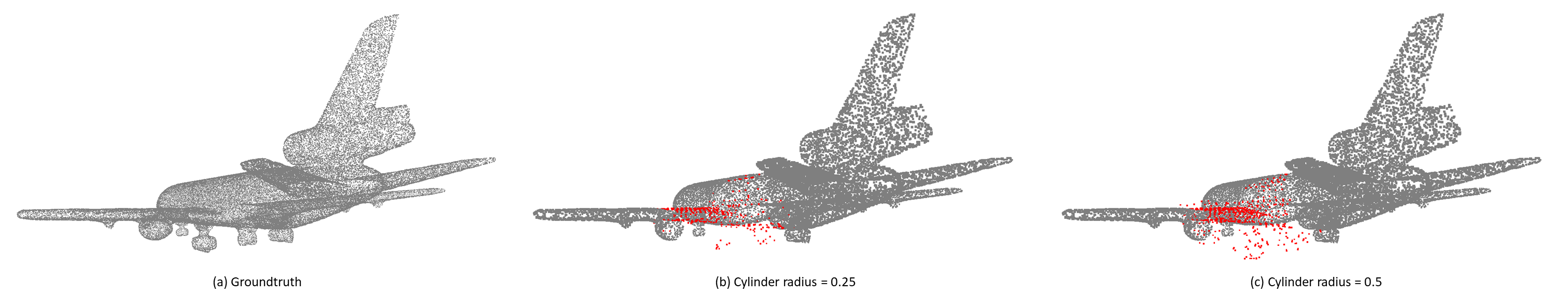}}
	\caption{\label{fig:radius_center} 
	\textbf{Effect of the average-to-center distance constraint for \emph{valid} cells.} (a) To illustrate the influence of the initial valid flags on completion, we use a complete point cloud of an airplane. We seed and precompute \kaplan{} on the the left wing, and use it to regenerate the corresponding part of the point cloud. If the neighbourhood radius is too large (c), then too many cells are \emph{valid} and give rise to spurious points.} 
\end{figure*}

Instead, we consider the \emph{average} location of all projections within the cell.
If that average lies near the cell center, then the cell is \emph{valid}.
Fig.~\ref{fig:radius_center} shows how this constraint mitigates artefacts during completion.

To further avoid a too strong pruning of the \emph{valid} cells due to aliasing, we finally switch the flag to \emph{valid} for all cells that contain at least one projected point and have $\geq$3 neighbours also marked \emph{valid}. See Fig.~\ref{fig:ray_tracing}b.

\subsection{Depth Aggregation}\label{sec:depth_aggreg}
\begin{figure*}
  \centering
  \begin{scriptsize}
  \centering
  \newcommand{\sz}{3.5cm}
  \newcommand{\insz}{1.25cm}
  \newcommand{\rd}{0pt}
  \newcommand{\rulesep}{\unskip\ \vrule\ }
  \setlength{\tabcolsep}{0.1pt}
  \vspace{-1em}
  \begin{tabular}{cccccc}
  &
  \textit{Plane View} & 
  \textit{Thresh 0.1} & 
  \textit{Thresh 0.01} &
  \textit{Thresh 0.005} &
  \textit{Thresh 0.001} \\[5pt]
      \rotatebox{90}{\hspace{0pt}\textit{Front View}} & 
  \includegraphics[height=\insz]{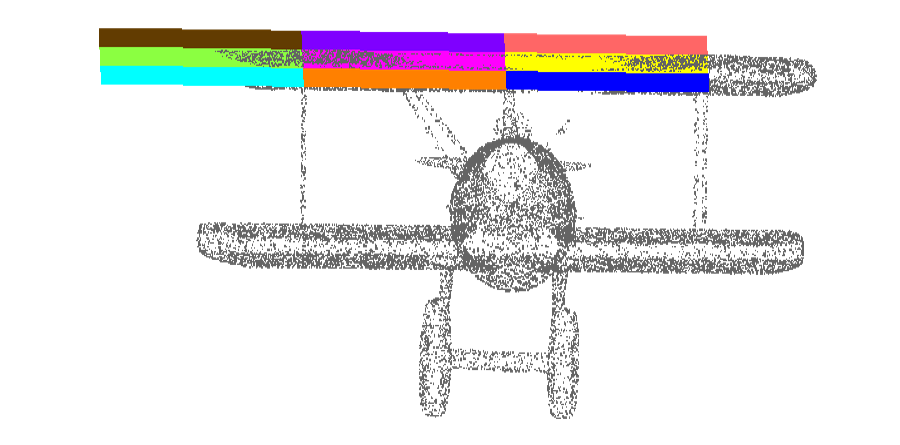} &
  \includegraphics[height=\insz]{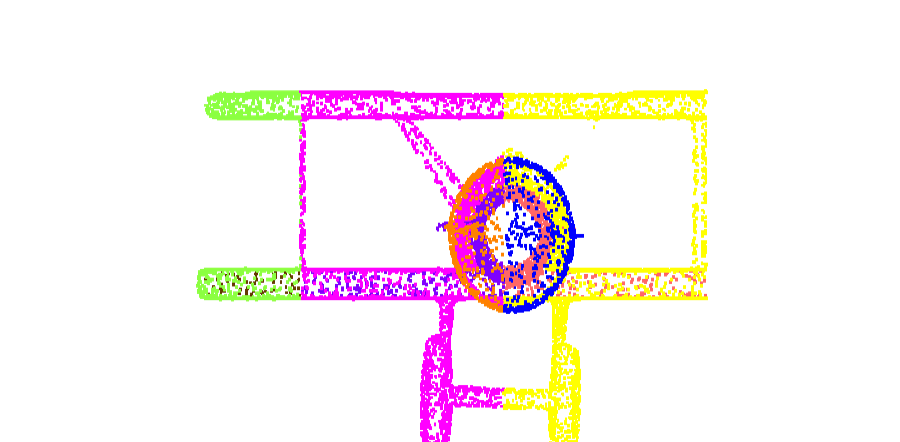} &
  \includegraphics[height=\insz]{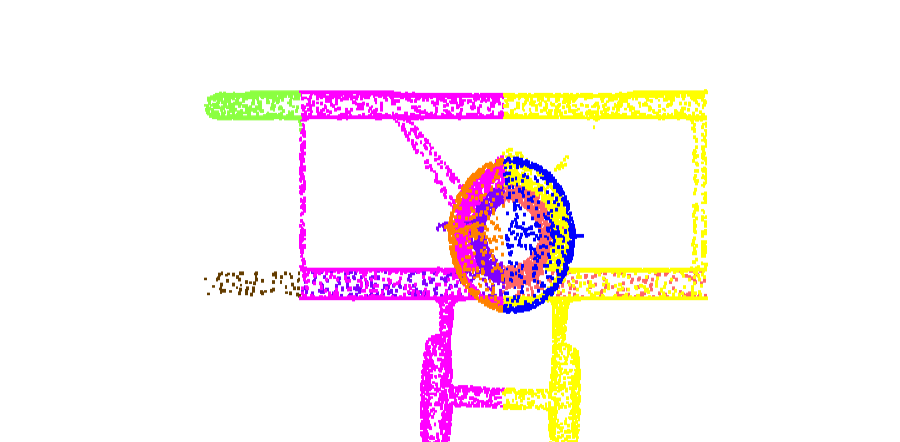} &
  \includegraphics[height=\insz]{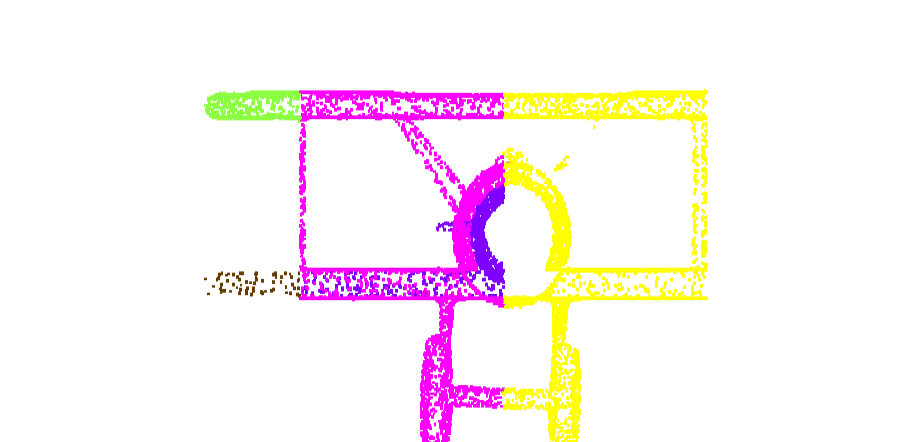} &
  \includegraphics[height=\insz]{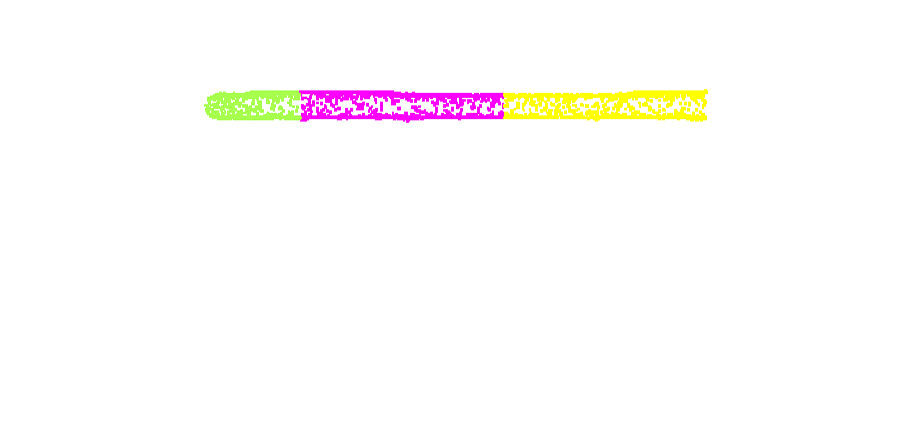} 
  \\[15pt]
  \rotatebox{90}{\hspace{0pt}\textit{Right View}} &
  \includegraphics[height=\insz]{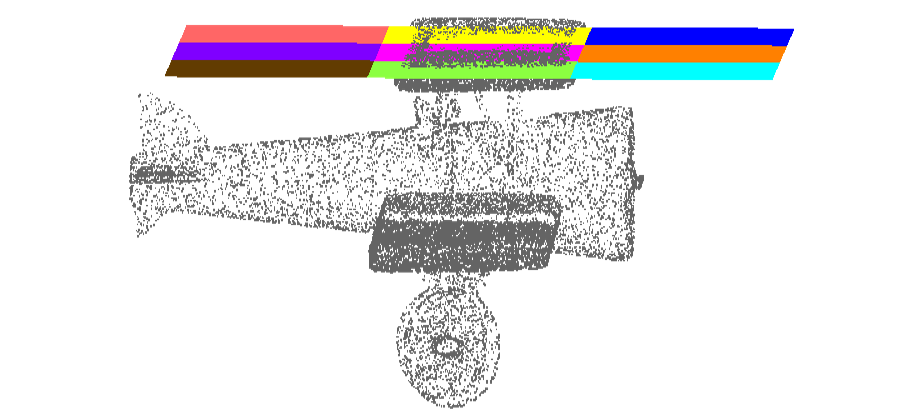} &
  \includegraphics[height=\insz]{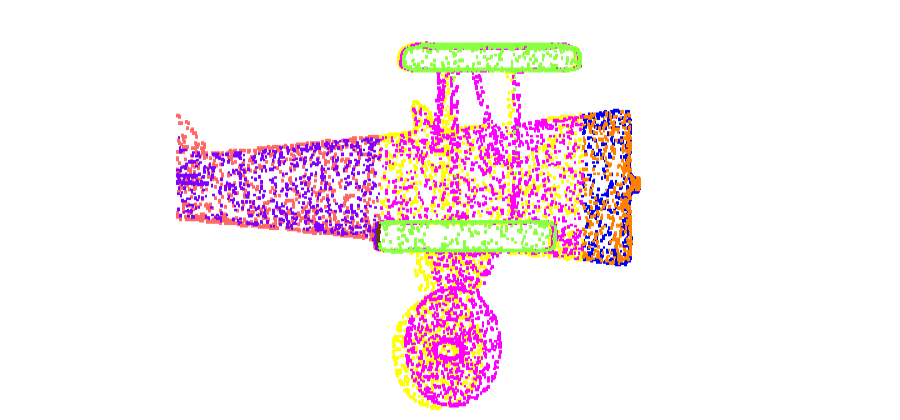} &
  \includegraphics[height=\insz]{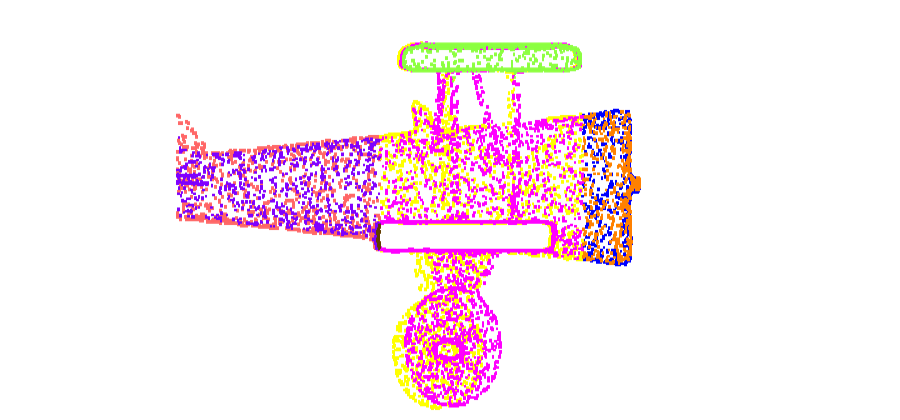} &
  \includegraphics[height=\insz]{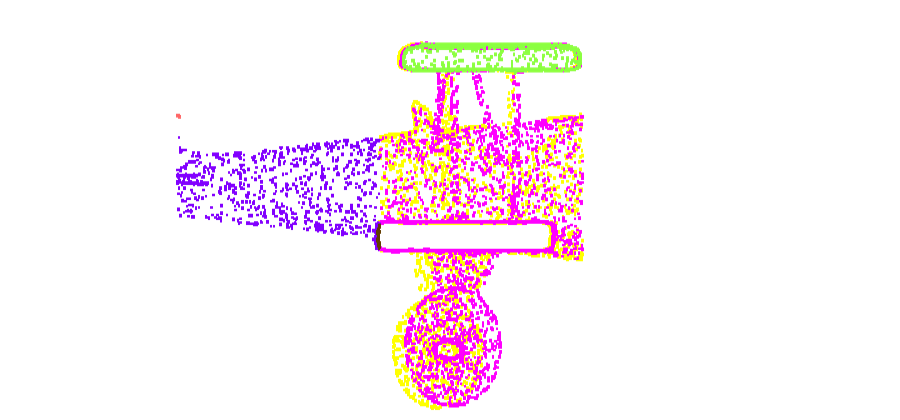} &
  \includegraphics[height=\insz]{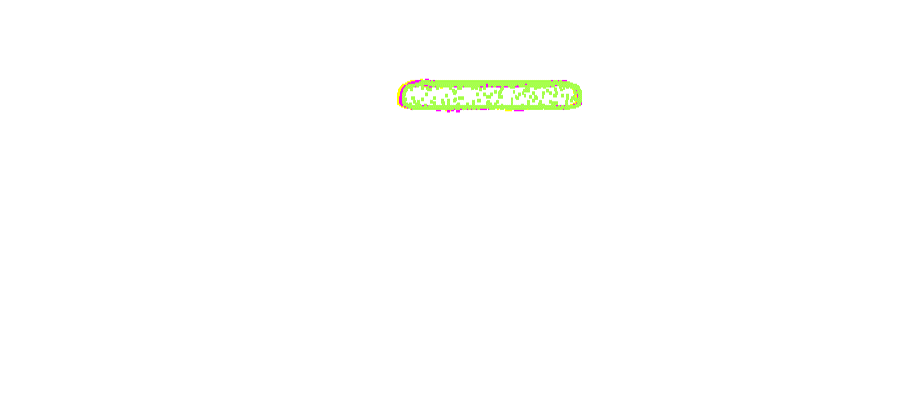} 
  \\
  \rotatebox{90}{\hspace{40pt}\textit{Top View}} &
  \includegraphics[height=\sz]{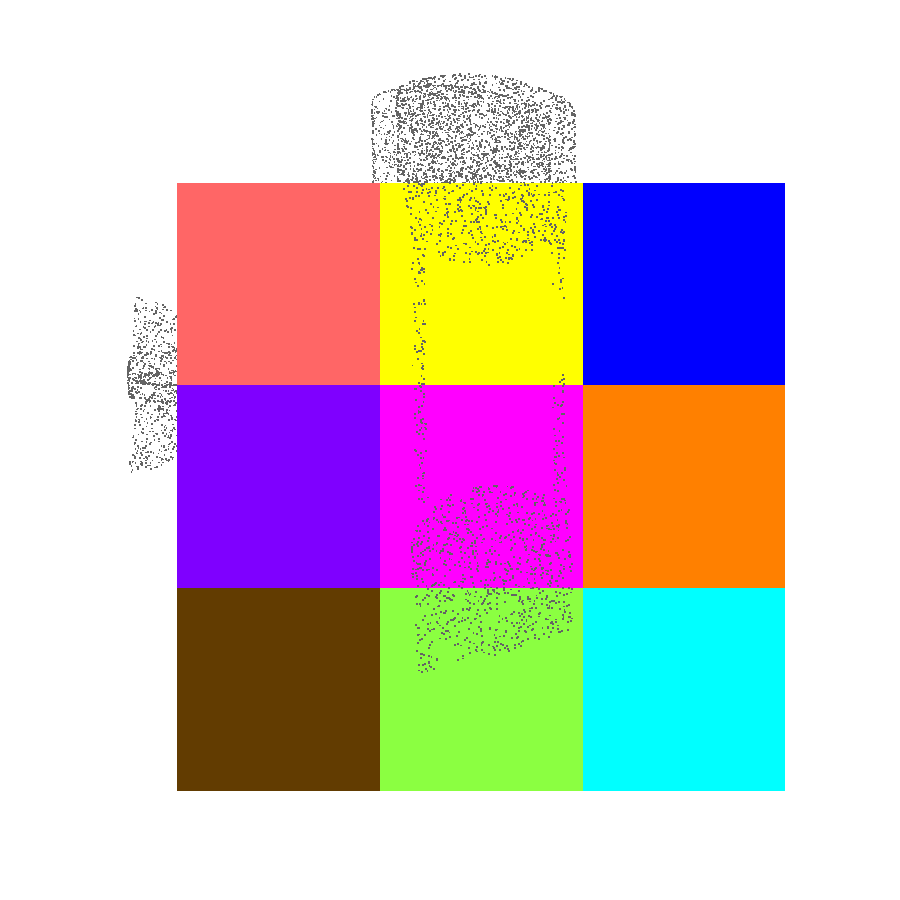} &
  \includegraphics[height=\sz]{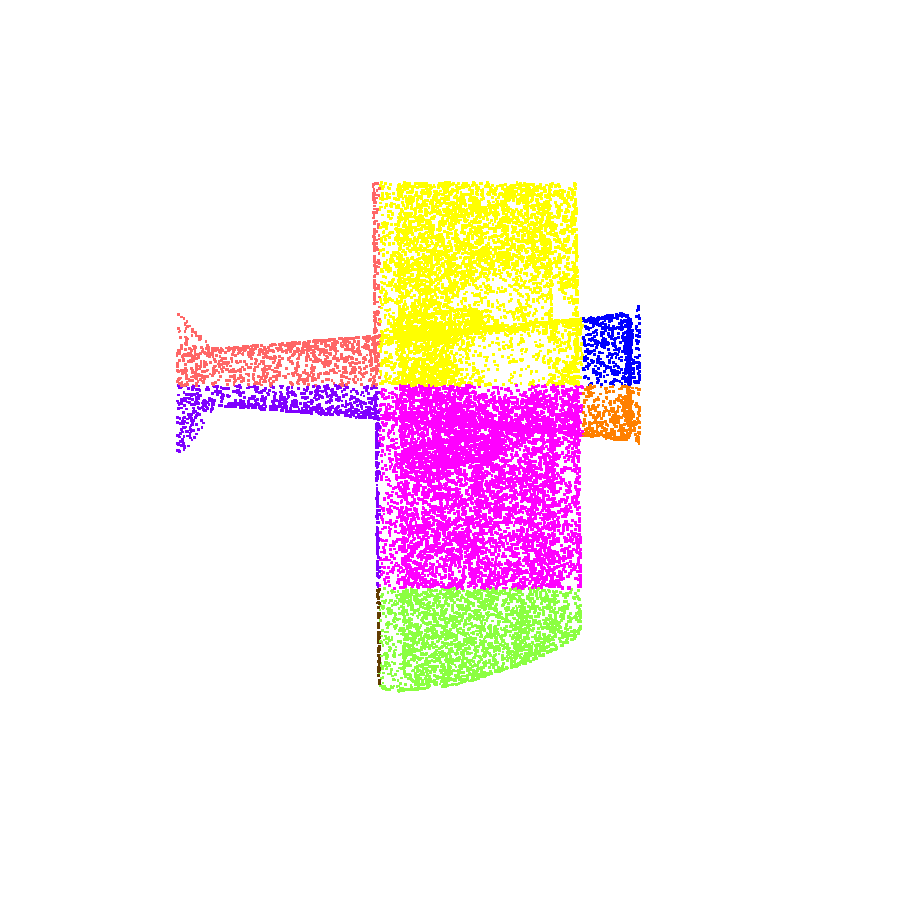} &
  \includegraphics[height=\sz]{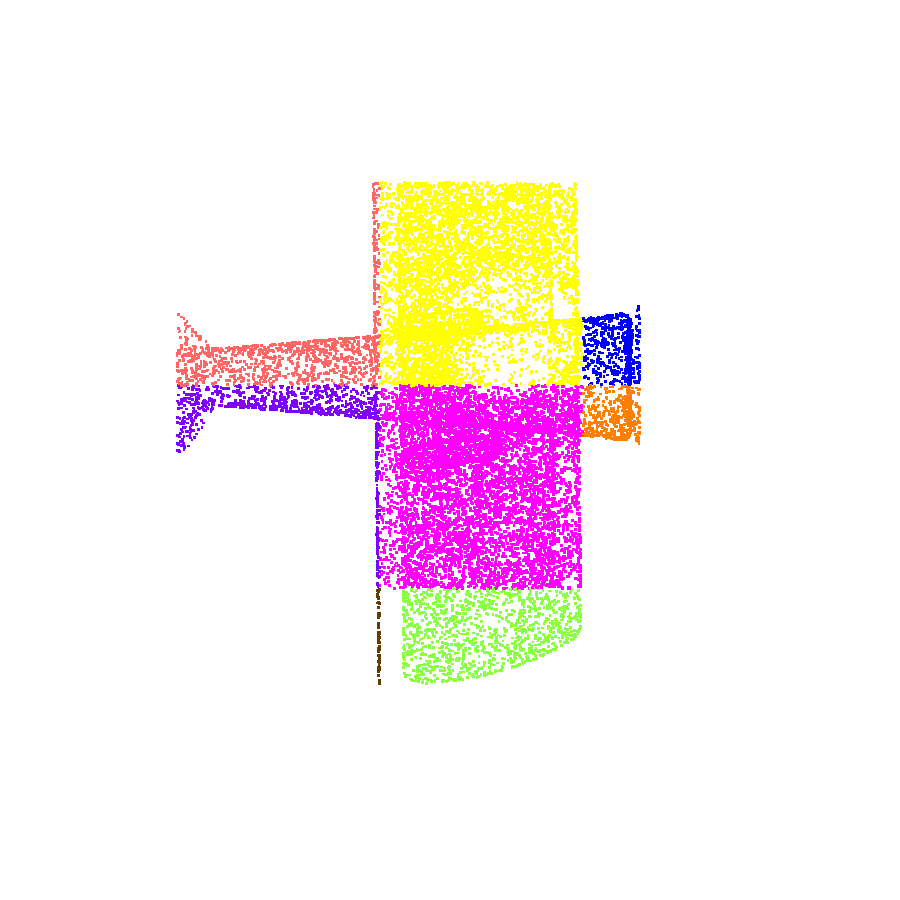} &
  \includegraphics[height=\sz]{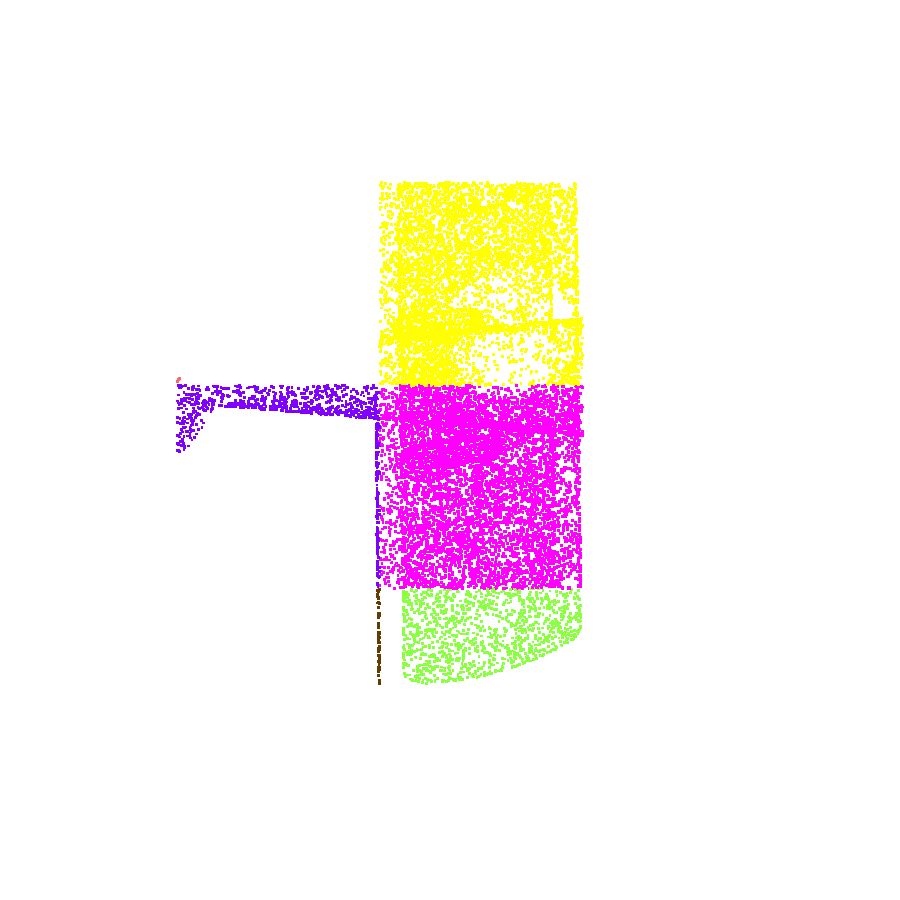} &
  \includegraphics[height=\sz]{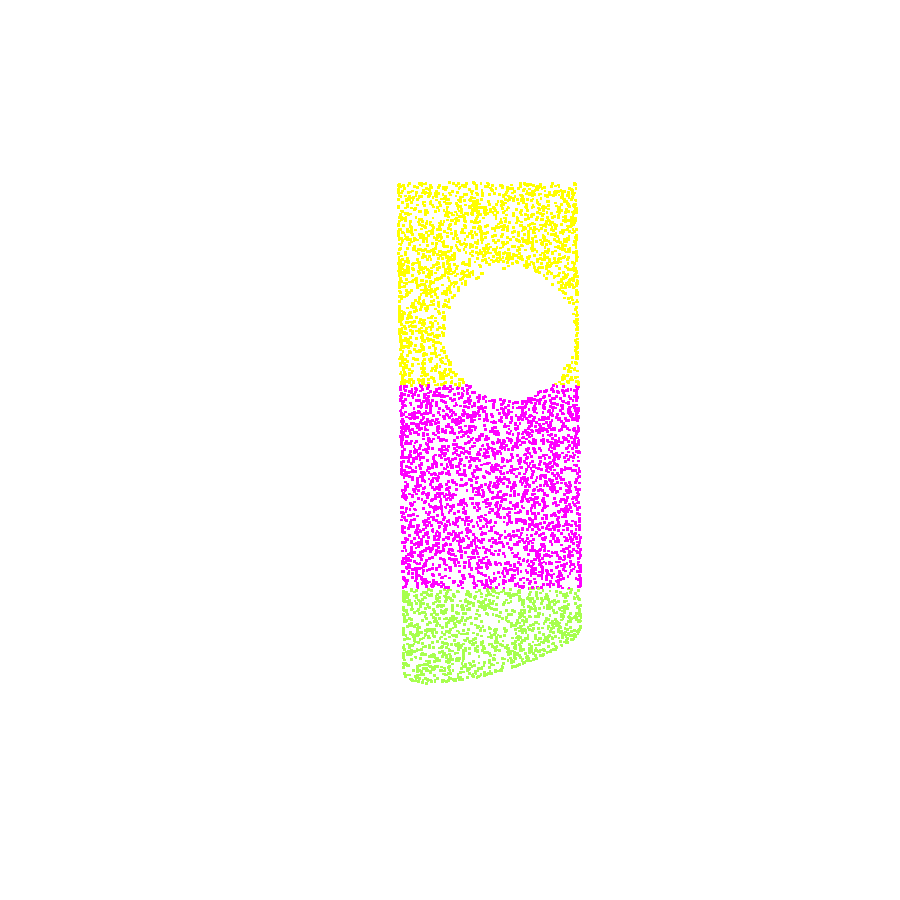} 
  \end{tabular}
  \end{scriptsize}
  \caption{\textbf{Effect of $\nThreshHeurist$ for depth aggregation in a cell.} We seed a \kaplan{} on the top wing of the airplane, consisting of $3\times3$ planes for the purpose of illustration. Each cell is represented in a different color. In all the figures, we only plot points that contribute to the average cell depth. Too large settings  conflate meaningless information from distant object parts, lowering the threshold leads to increasingly local descriptors.}
  \label{fig:aggregation_heuristic}
\end{figure*}

Once the valid flags have been attributed, we need to aggregate the depths in the corresponding cells to complete the \kaplan{} input.
Simply using the average depth over all points that project into the cell carries the risk of averaging together distinct parts of the object, as illustrated in Fig.~\ref{fig:aggregation_heuristic}.
In this figure, we represented an example a double winged airplane with a \kaplan{} seeded on the top wing, and show a \kaplan{} plane parallel to the wings.
Simply taking the average depth will cause averaging of the two wings to a meaningless intermediate depth.
Another solution could be to only project points within a limited depth range.
However this would again carry the risk to be too conservative and unnecessarily lose descriptors (and valid cells) on more curved surfaces.

As a practical compromise, we use a threshold on the \emph{moving average} of the depth inside a cell. 
We first sort all points that project into a cell by their absolute depth values.
We pick the lowest absolute depth as initial estimate. Then we find the next-lowest depth, and if its difference to the current estimate is below a threshold $\nThreshHeurist$, we add it to the moving average, and iterate until the depth gap to the next point exceeds  $\nThreshHeurist$.
This corresponds in practice to an instance of Kernel Density Estimation with a uniform kernel.
Other kernels could be considered, however we are confident that they would not greatly change the results.
Fig.~\ref{fig:aggregation_heuristic} illustrates the effects of different choices for $\nThreshHeurist$.
Too larger values aggregate depths from independent object parts, lower thresholds lead to a more local descriptor that captures meaningful surface information.
Following Fig.~\ref{fig:aggregation_heuristic}, we used a value of $0.001$ in all our experiments.

\subsection{Difference between Valid Flag and Depth Value}
Instead of storing the valid flag and depth information in different variables, an idea could be to merge them into a single depth value.
A threshold could be then used in order to determine if a point should be introduced or not.
The reason why we did not opt for this approach is because it makes the learning task very difficult.
Depth estimation is a regression task, while estimating if a point should be introduced or not is a classification task.
Decoupling both tasks makes the training more stable.

\subsection{Query Points Selection}\label{query_points}
In general, nearby points on densely sampled surfaces have similar local geometry and thus similar descriptors.
To avoid unnecessary redundancy (and to ensure diversity of the training samples), we adopt the following strategies to select query points:

\boldparagraph{Coarse level.}
The goal at this level is to detect the holes and introduce some points in them, to guide the completion at the finer levels.
Since coarse \kaplan{} are large and cover a large part of the object, only a few query points are necessary to achieve this.
We therefore sample $10$ query points uniformly, \ie $10$ descriptors are computed to represent the object's shape. 

\boldparagraph{Finer levels.}
For the next two levels, query points are selected among the newly predicted points (after filtering, see Sec.~\ref{sec:impl_filter}).
However, at these levels the cell size of \kaplan{} is decreased to achieve a finer resolution.
Unlike the situation at $\nLevelCoarse$, this means that a single plane within a \kaplan{} does not necessarily covers the entire missing region. Hence, to ensure complete coverage of the object we increase the number of query points to $\nLevelMedium=20$, respectively $\nLevelFine=30$.

\section{Additional Results}
\label{sec:add_results}

\subsection{Choice of Data}
We chose to evaluate our method on ShapeNet~\cite{shapenet2015} which is a standard dataset widely used in the community for benchmark.
ShapeNet is a synthetic human-made dataset, where all the data is gravity aligned, with a known orientation.
The main motivation for this choice is to compare against competing baselines which assume known gravity direction~\cite{Yuan-et-al-3DV-2018, Wang-et-al-CVPR-2020} or object orientation~\cite{Mescheder-et-al-CVPR-2019, Park-et-al-CVPR-2019}.

However, our method is not limited to such a particular alignment, although fixing it simplifies the learning task.
In order to run our method on real data that is not necessarily gravity aligned, we would need to retrain the network using data augmentation.
Instead of systematically using the canonical planes to initiate the generation of KAPLAN, we expect random orientation to perform better.

\subsection{Choice of Metrics}
We use two popular error metrics to assess the difference between the predicted point cloud and the ground truth.
The Chamfer distance (CD) is used to quantify the global (dis)similarity between the two 3D shapes.
It is computed as the mean, symmetric (forward-backward and backward-forward) nearest-neighbour distance between the two point sets. We refrain from sampling-based approximations and compute the CD over all points, using efficient kd-Tree search. The reported CD values are normalised by the number of points.

Since point cloud completion effectively tries to restore discrete points where a point ``should have been'' under ideal conditions, another approach is to measure completeness and accuracy of the predicted points \wrt the ground truth.
We follow~\cite{Schoeps-et-al-CVPR17} and define accuracy as the fraction of reconstructed points that coincide with a ground truth point up to the evaluation threshold, and completeness as the fraction of ground truth points that are covered by a predicted one up to the same threshold.
The two numbers are then combined in the ususal way via the harmonic mean to obtain the $F_1$-score.
The evaluation threshold in our experiments is set to $0.01$, according to the average spacing between nearest-neighbour ground truth points in the dataset.


\subsection{Design Choices using Ground Truth KAPLAN}
\begin{figure*}
	\centerline{\includegraphics[scale=0.35]{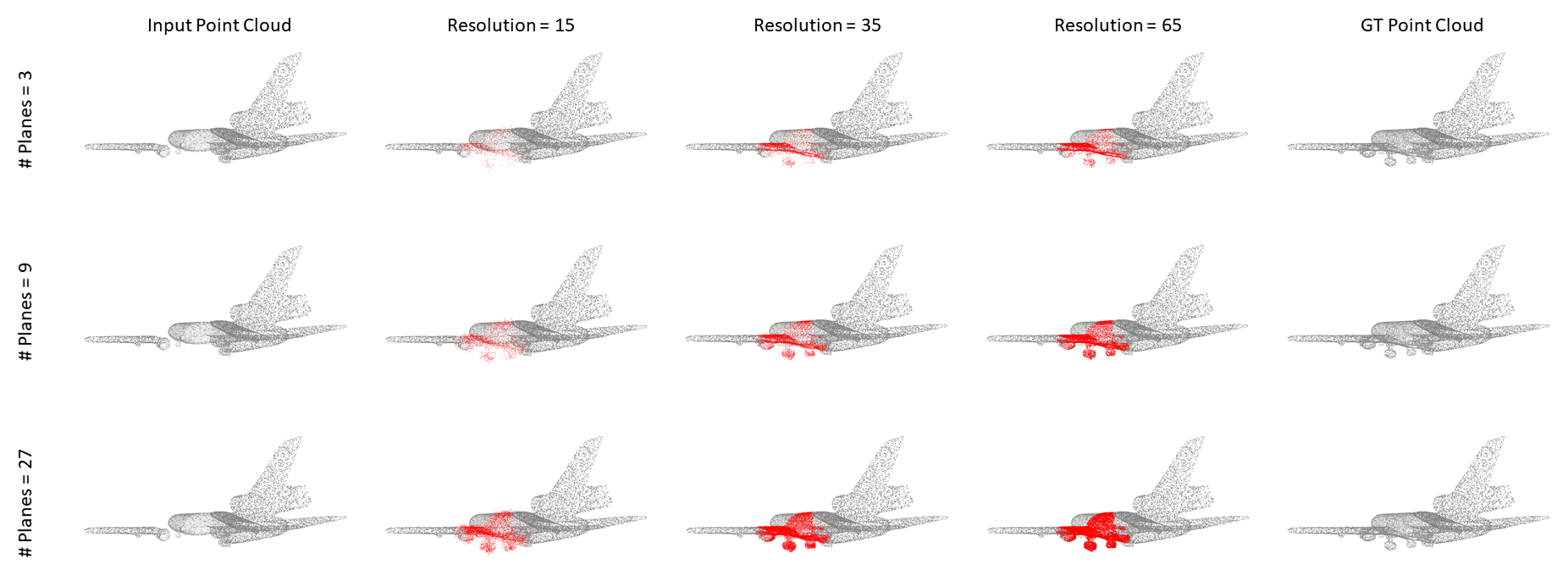}}
	\caption{
	\label{fig:kaplan_design}
	\textbf{Reconstruction using ground truth \kaplan{} for different configurations (only coarse level).} 
	The \kaplan{} configuration is defined by the number of planes and their resolution. 
	Coarse resolution leads to less accurate points. 
	More planes obviously lead to more points, but due to the coarse-to-fine scheme these are actually not required for the full scene completion pipeline.} 
\end{figure*}

The number of planes within a \kaplan{}, as well as their resolution, are hyper-parameters that need to be set in advance, before precomputing the input to the CNN.
Instead of grid-searching the network by training with many different combinations of planes number and resolution, we perform the empirical study directly on the ground truth.
For a given plane number and resolution, we picked \kaplan{} seed points in the complete ground truth point cloud and computed the corresponding valid flags and depths, to obtain ground truth \kaplan{} descriptors. We then synthetically generated a hole and used the ground truth descriptors to fill it.
The reconstruction quality of this procedure can be seen as an upper bound, corresponding to the case where the neural network would manage to recover the ideal descriptor from incomplete data.
The parameter search was run at the coarsest level, which both intuitively and empirically is the most crucial one for the overall scheme.
Fig.~\ref{fig:kaplan_design} and Tab.~\ref{tab:kaplan_config} show reconstruction results for different \kaplan{} configurations.
Note that the $F1$-score in the table are computed \emph{only for the hole}, not for the complete object as in Sec.~\ref{sec:experiments}. This accentuates the differences between different parameter settings. Visually, the improvement within the hole from resolution $35\times35$ to $65\times65$ only corresponds to a minor improvement, hence we opted for the slightly faster version.
Depending on the appliction and data characteristics, it may in some cases be beneficial to use \kaplan{} with higher resolution.

Beside the quality of the reconstruction, we also take into account the runtimes for pre-computation and reconstruction, as shown in Tab.~\ref{tab:kaplan_config_runtime}.
From the two tables, we see that the most important parameter setting is high resolution in the descriptor plane, whereas adding additional planes have comparatively little influence and can sometimes even hurt the quality.
Recall that \kaplan{} predicts points based on the cell centers in each plane, such that additional planes can lead to more artefacts, as discussed in Sec.~\ref{sec:precomputation}.
Moreover, fewer planes speed up the computation, whereas resolution has little influence on runtime.
Consequently, we opt for a recommended default configuration $3$ planes per \kaplan{}.

\begin{table}[h!]
    \centering
    \resizebox{.45\textwidth}{!}{%
    	\begin{tabular}{c||c|c|c} 
    	\backslashbox{\textbf{\# Planes}}{\textbf{Res}} &
      	\textbf{$15\times15$}  & 	
      	\textbf{$35\times35$}  &	
      	\textbf{$65\times65$} \\
      	\hline\hline
     	{3}  & 57.63 & 69.18 & 75.26 \\
      	{9}  & 53.68 & 68.47 & 76.99 \\
      	{27} & 53.36 & 69.56 & 79.60
    	\end{tabular}
    }
    \vspace{1mm}
    \caption{\textbf{F1-score for different ground truth \kaplan{} configurations.} Obtained for 10 query points at coarse level. The $F1$-scores are computed only on the missing regions instead of the full model.}
    \label{tab:kaplan_config}
\end{table}

\begin{table}[h!]
    \centering
    \resizebox{.45\textwidth}{!}{%
    	\begin{tabular}{c||c|c|c} 
    	\backslashbox{\textbf{\# Planes}}{\textbf{Res}} &
      	\textbf{$15\times15$}  & 	
      	\textbf{$35\times35$}  &	
      	\textbf{$65\times65$} \\
      	\hline\hline
     	{3}  & 0.84 & 0.72 & 0.77 \\
      	{9}  & 1.37 & 1.25 & 1.70 \\
      	{27} & 3.18 & 3.41 & 4.26
    	\end{tabular}
    }
    \vspace{1mm}
    \caption{\textbf{Runtime for different ground truth \kaplan{} configurations, in seconds}.
    The runtime takes into account the \kaplan{} precomputations for all query points and the consecutive reconstruction, using 10 query points.}
    \label{tab:kaplan_config_runtime}
\end{table}

\begin{figure*}[p]
	\centerline{\includegraphics[scale=0.4]{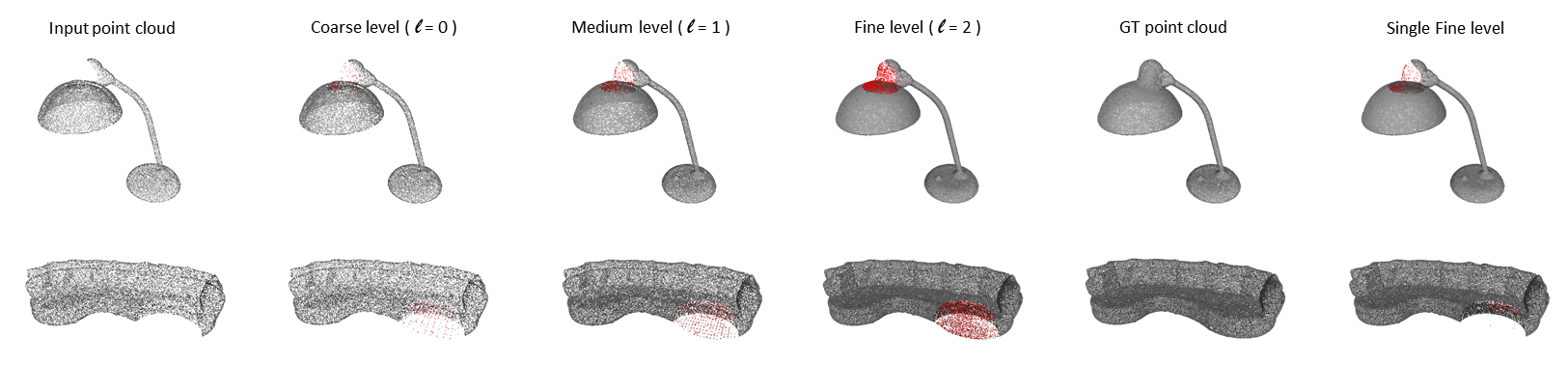}}
	\caption{
	\label{fig:coarse_to_fine} 
	\textbf{Reconstruction steps of the coarse-to-fine scheme.}
	Every level improves the point density of the reconstruction, with the coarsest level being in charge of placing the first seeds in the missing regions.
        Using only the finest level can give acceptable (but sparse) results for some geometries like the lamp, but fails to complete larger holes like on the sofa.
	} 
\end{figure*}

\boldparagraph{Note on the coarse-to-fine scheme.}
Still using our ground truth, we also present qualitative examples to illustrate the coarse-to-fine scheme. In Fig.~\ref{fig:coarse_to_fine}, we show a lamp and a sofa being reconstructed using ground truth \kaplan{} descriptors. 
Intermediate results are shown at each hierarchy level. One can nicely see the complementary tasks of the different scale levels: the coarse initial step detects holes and fills them with sparse ``anchor points''. The subsequent levels densify and refine the completion. The gradual densification through the coarse-to-fine scheme successfully fills in large holes and achieves sufficient point density.

\subsection{Qualitative Visualizations}

\begin{figure*}
	\centering
	\scriptsize
	\setlength{\tabcolsep}{0.8mm}
	\newcommand{\sz}{0.12}
	\newcommand{\insz}{0.065}
	\begin{tabular}{ccccccccc}
		& \textbf{Input} & \textbf{PSR~\cite{Kazhdan-et-al-SGP-2006}} &  \textbf{PCN~\cite{Yuan-et-al-3DV-2018}}  &  \textbf{Cascaded~\cite{Wang-et-al-CVPR-2020}}  & \textbf{OccNet~\cite{Mescheder-et-al-CVPR-2019}} & \textbf{DeepSDF~\cite{Park-et-al-CVPR-2019}} & \textbf{Ours} & \textbf{GT} \\[12pt]
		\multirow{2}{*}[35pt]{\rotatebox{90}{\textbf{Plane}}} &
		\includegraphics[width=\sz\textwidth]{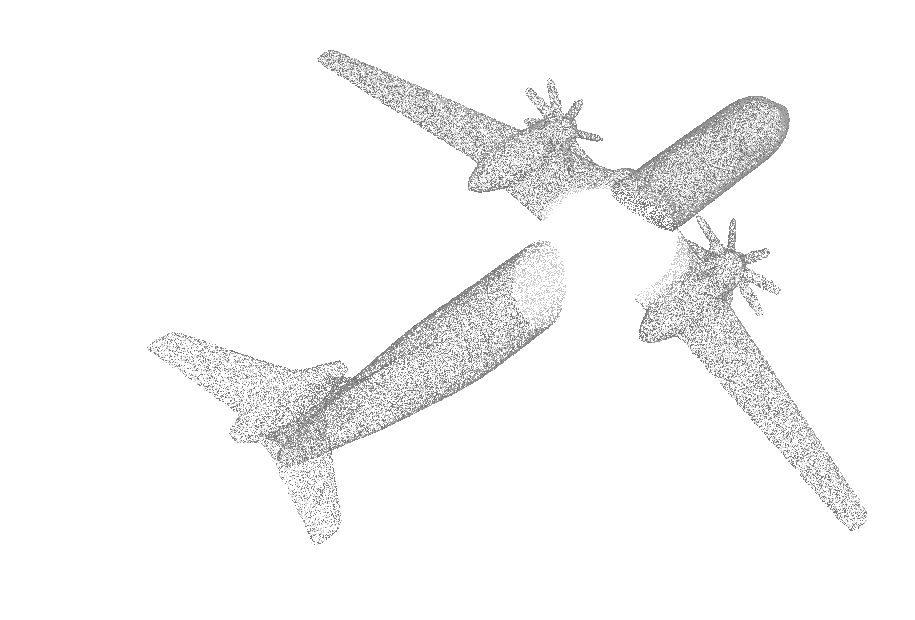} &
		\includegraphics[width=\sz\textwidth]{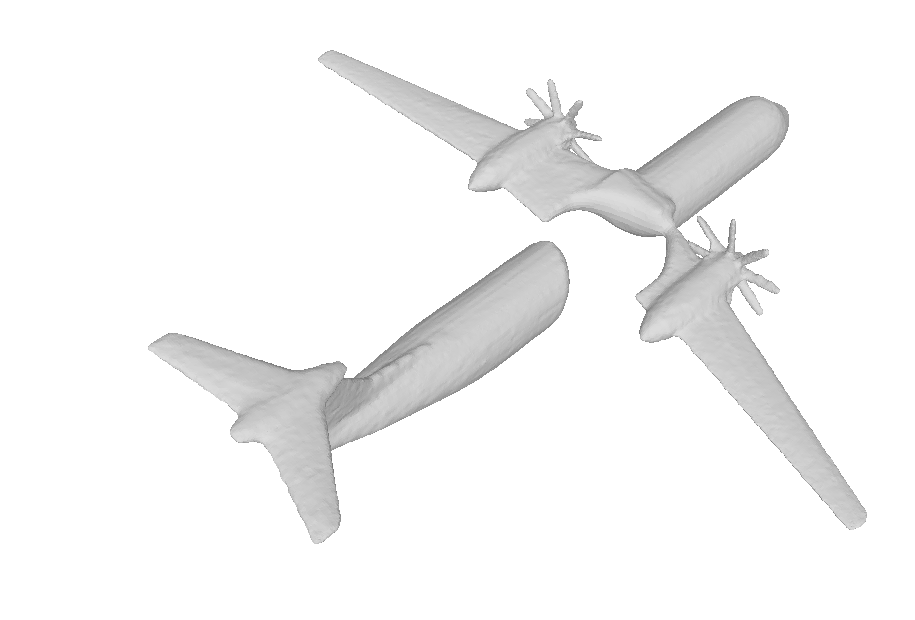} &
		\includegraphics[width=\sz\textwidth]{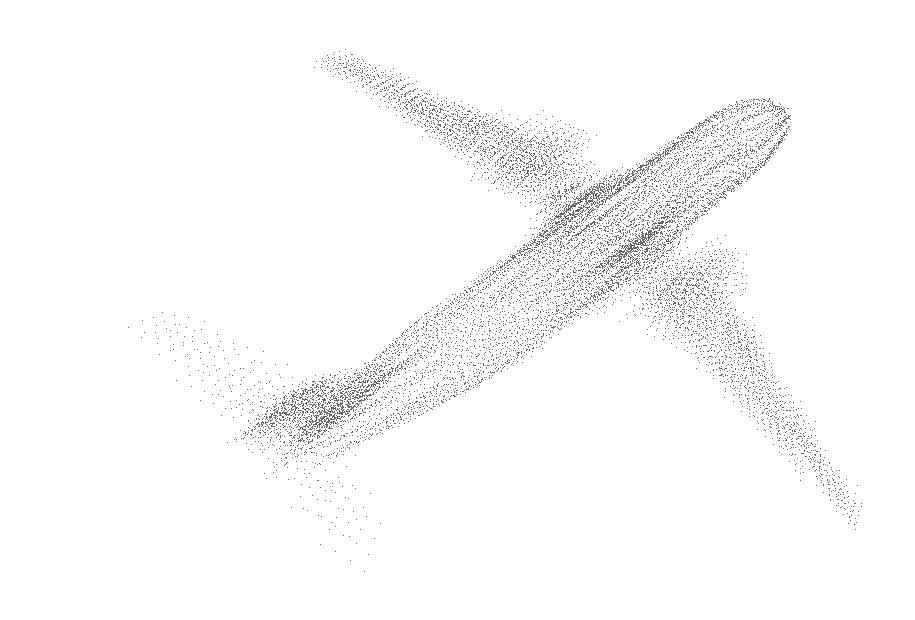} &
		\includegraphics[width=\sz\textwidth]{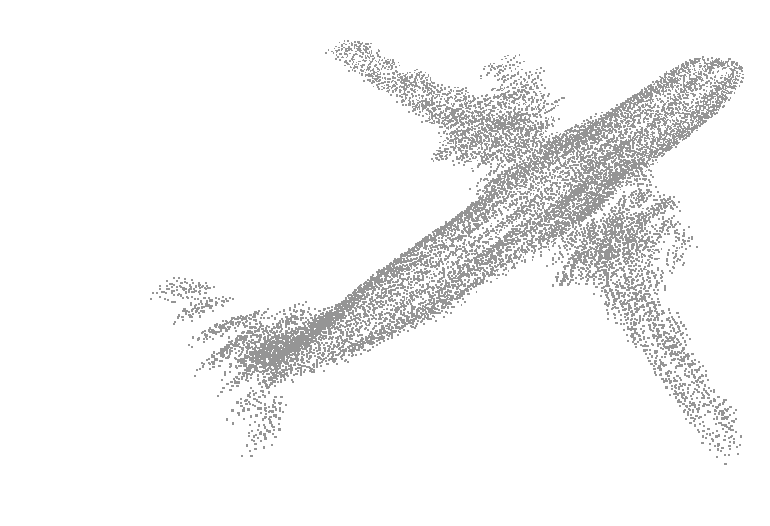} &
		\includegraphics[width=\sz\textwidth]{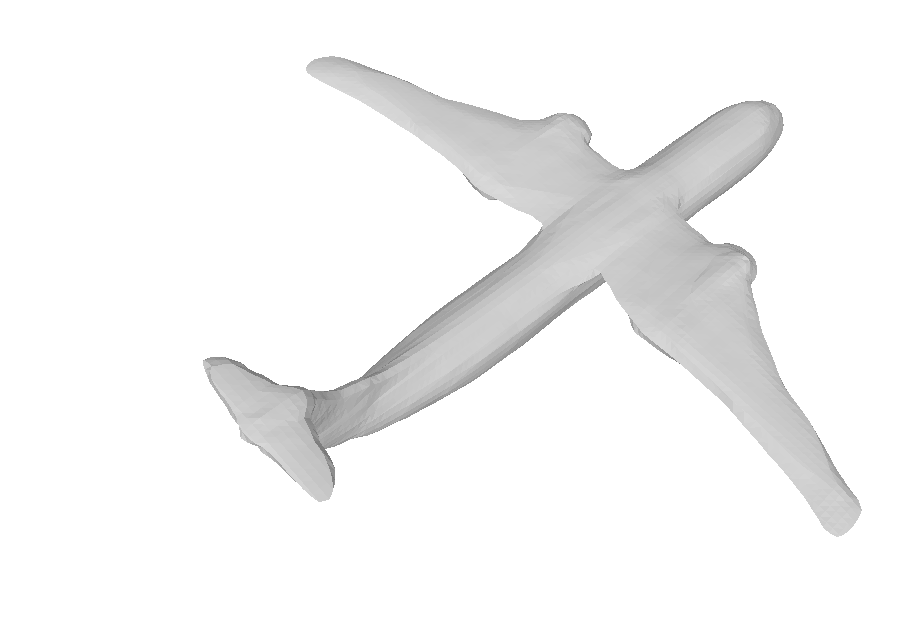} &
		\includegraphics[width=\sz\textwidth]{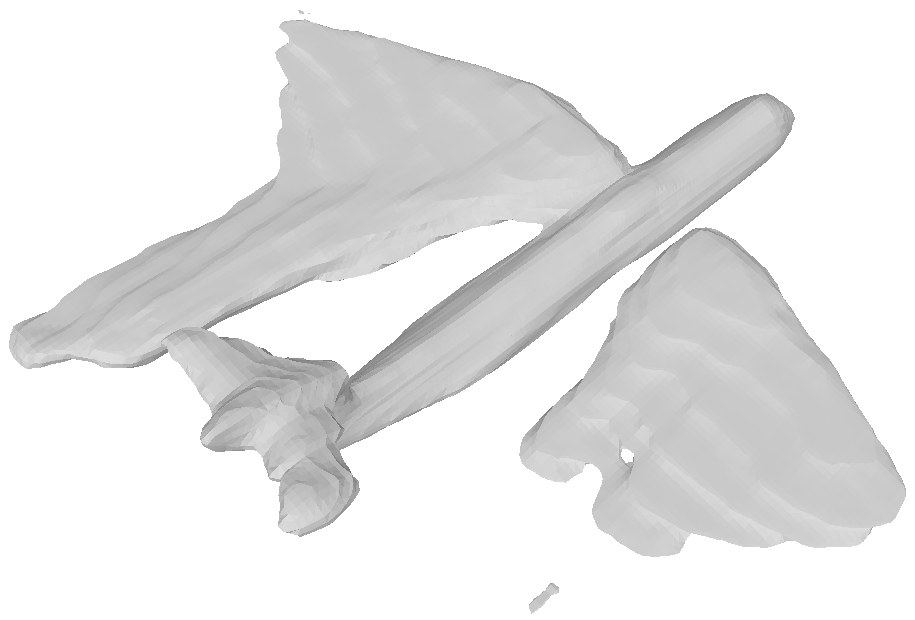} &
		\includegraphics[width=\sz\textwidth]{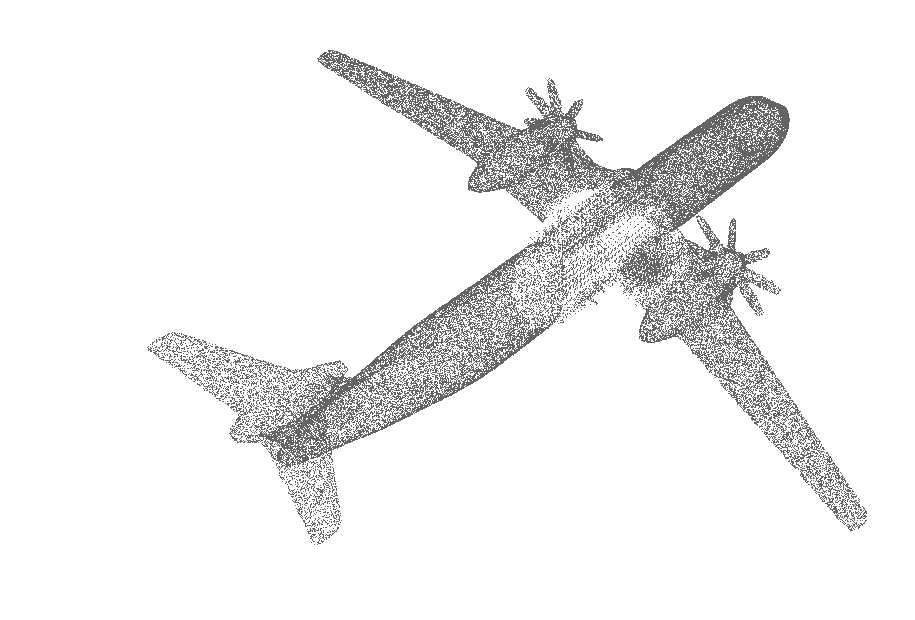} &
		\includegraphics[width=\sz\textwidth]{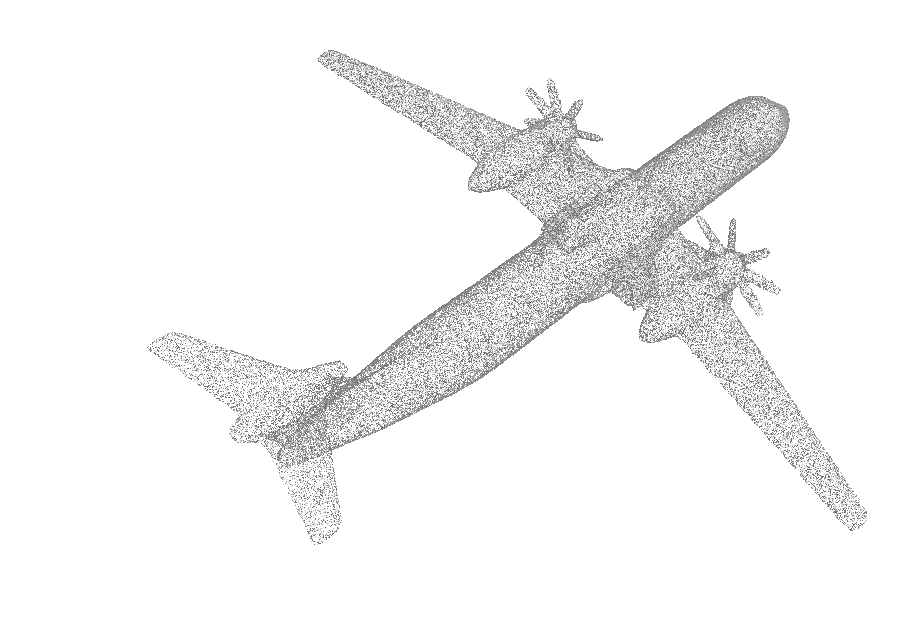} \\[12pt]
		\multirow{2}{*}[40pt]{\rotatebox{90}{\textbf{Plane}}} &
		\includegraphics[width=\sz\textwidth]{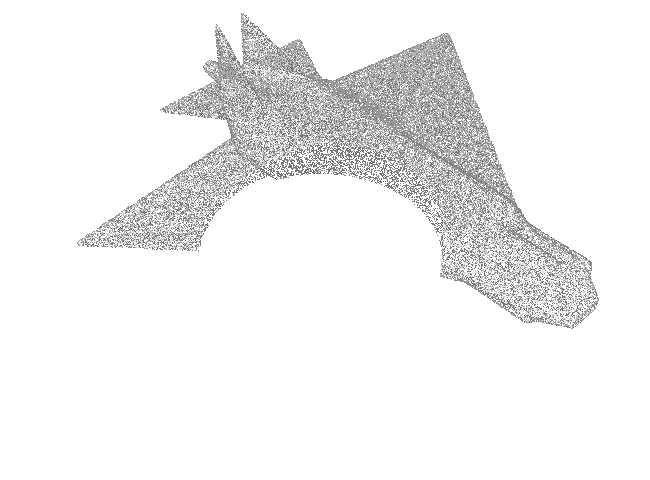} &
		\includegraphics[width=\sz\textwidth]{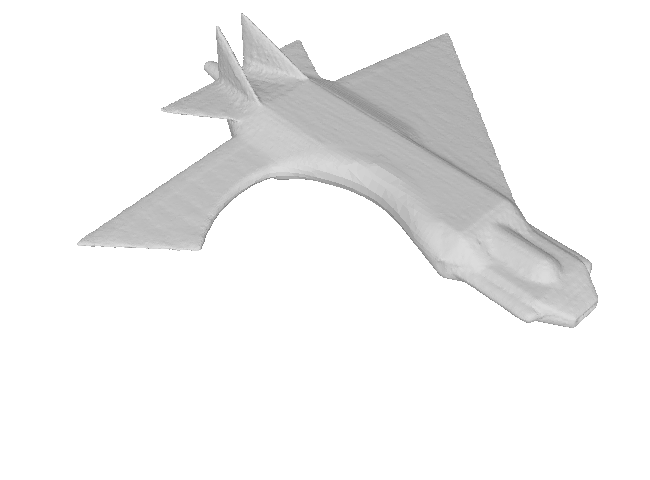} &
		\includegraphics[width=\sz\textwidth]{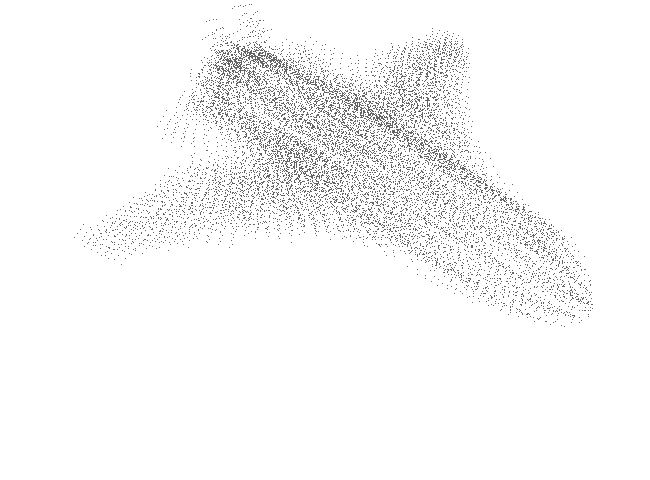} &
		\includegraphics[width=\sz\textwidth]{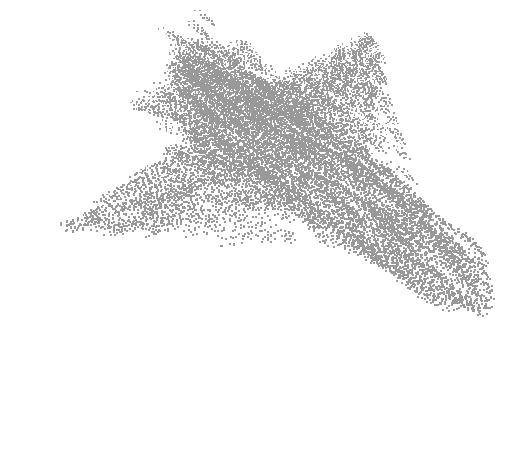} &
		\includegraphics[width=\sz\textwidth]{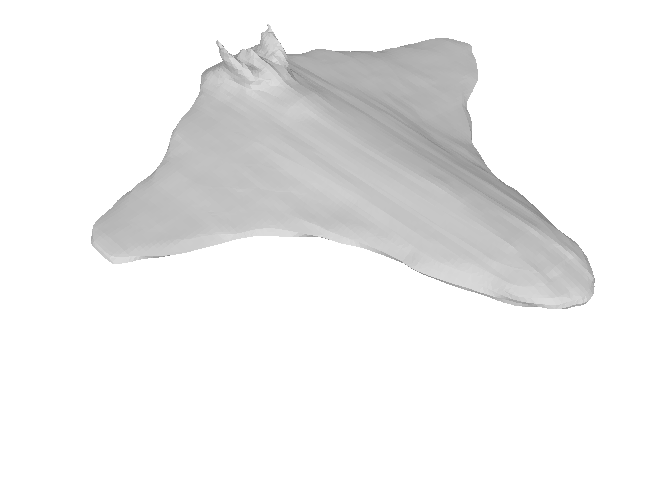} &
		\includegraphics[width=\sz\textwidth]{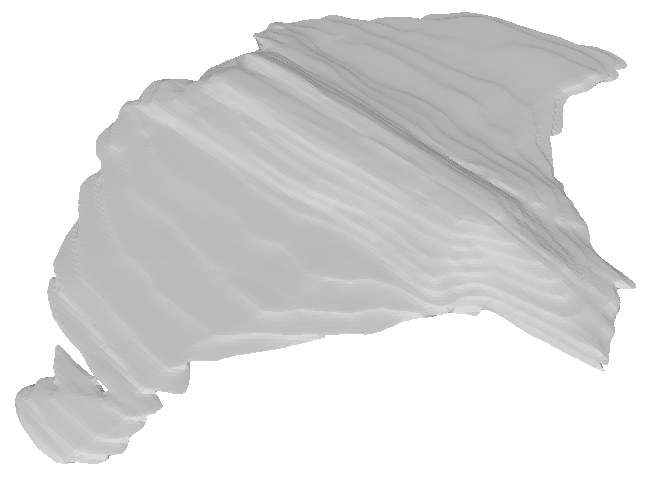} &
		\includegraphics[width=\sz\textwidth]{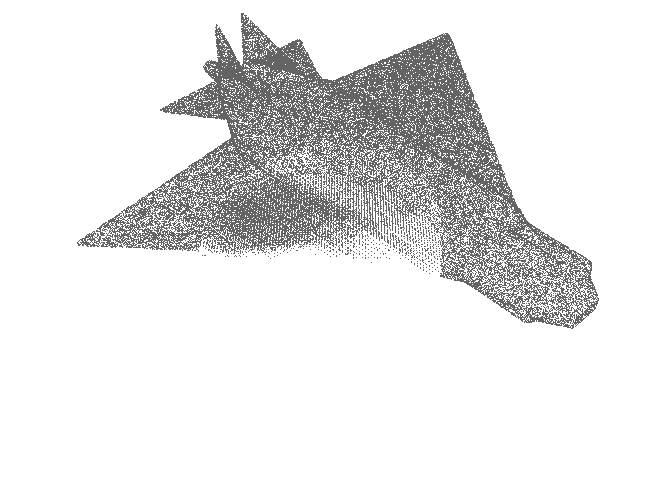} &
		\includegraphics[width=\sz\textwidth]{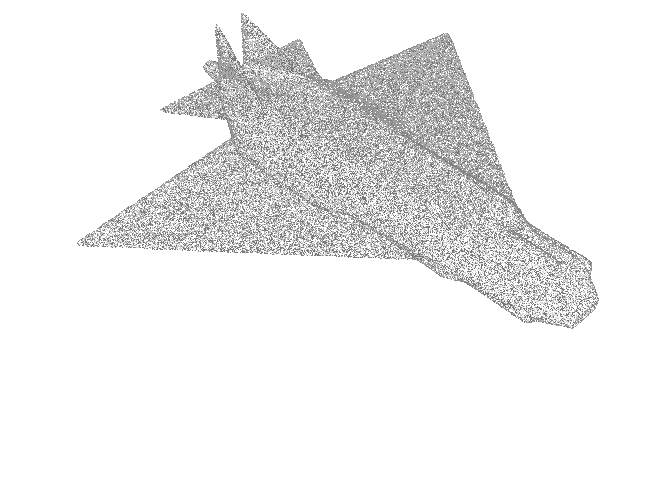} \\[12pt]
		\multirow{2}{*}[60pt]{\rotatebox{90}{\textbf{Chair}}} &
		\includegraphics[width=\sz\textwidth]{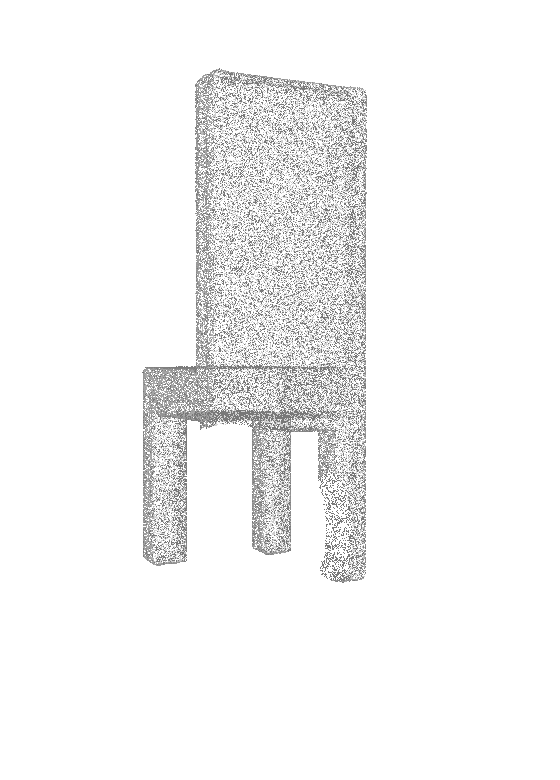} &
		\includegraphics[width=\sz\textwidth]{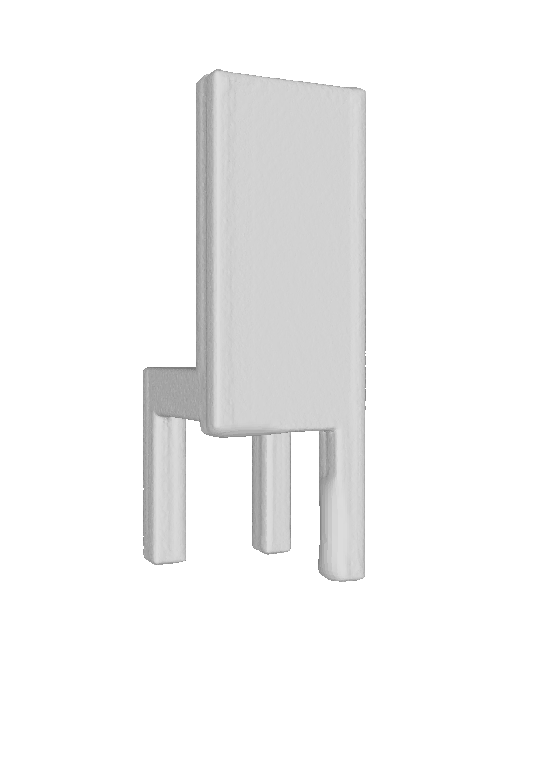} &
		\includegraphics[width=\sz\textwidth]{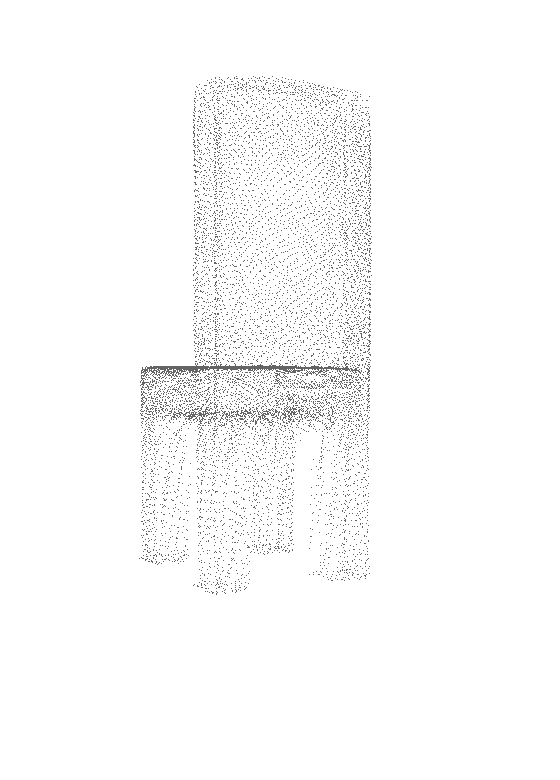} &
		\includegraphics[width=\sz\textwidth]{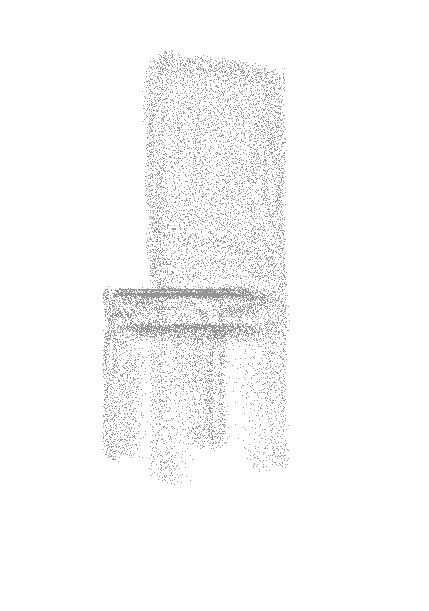} &
		\includegraphics[width=\sz\textwidth]{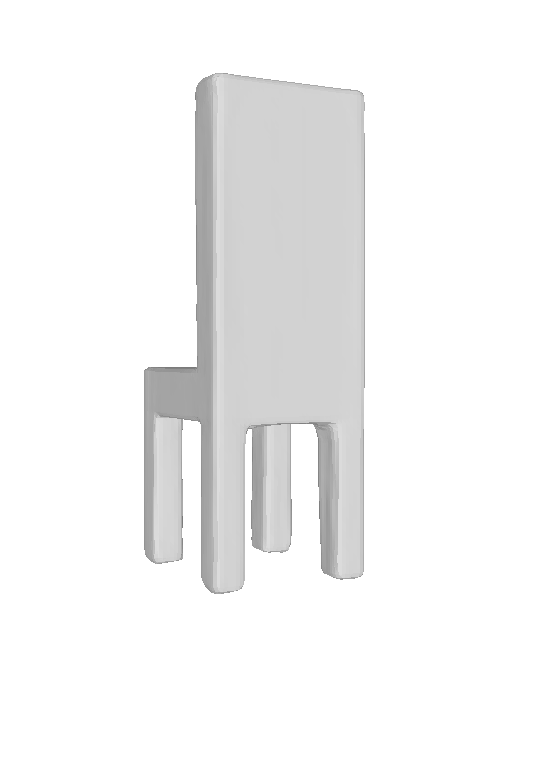} &
		\includegraphics[width=\sz\textwidth]{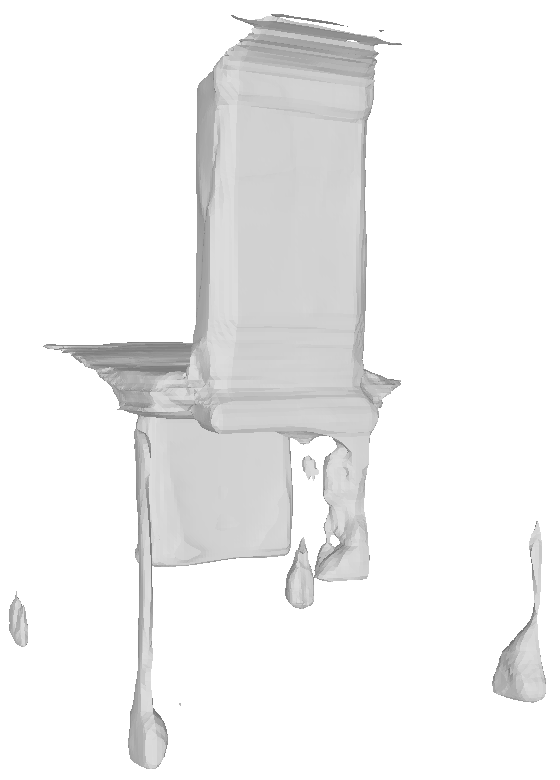} &
		\includegraphics[width=\sz\textwidth]{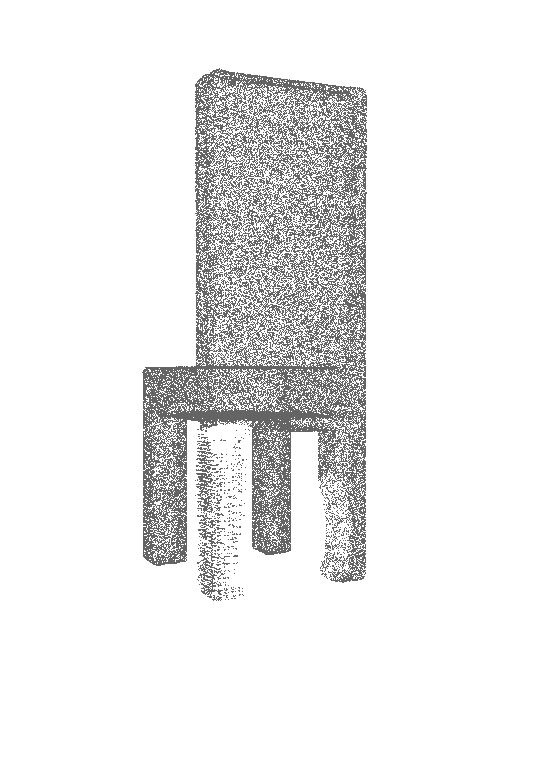} &
		\includegraphics[width=\sz\textwidth]{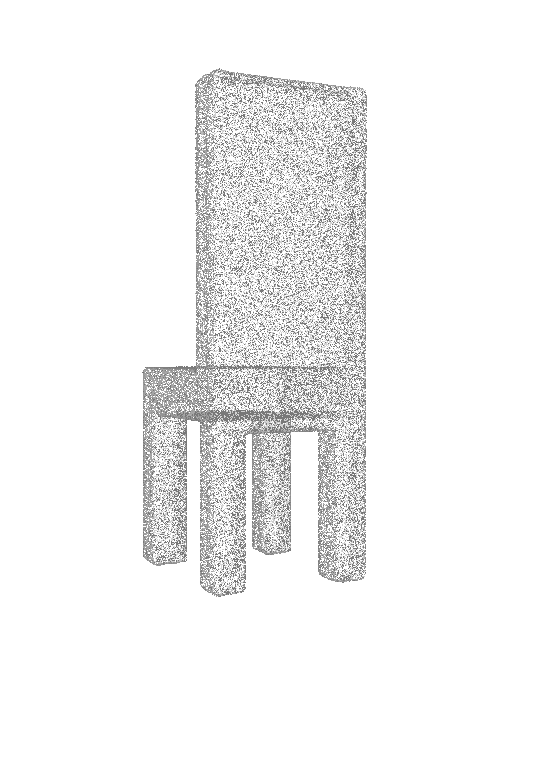}  \\[12pt]
		\multirow{2}{*}[65pt]{\rotatebox{90}{\textbf{Lamp}}} &
		\includegraphics[width=\sz\textwidth]{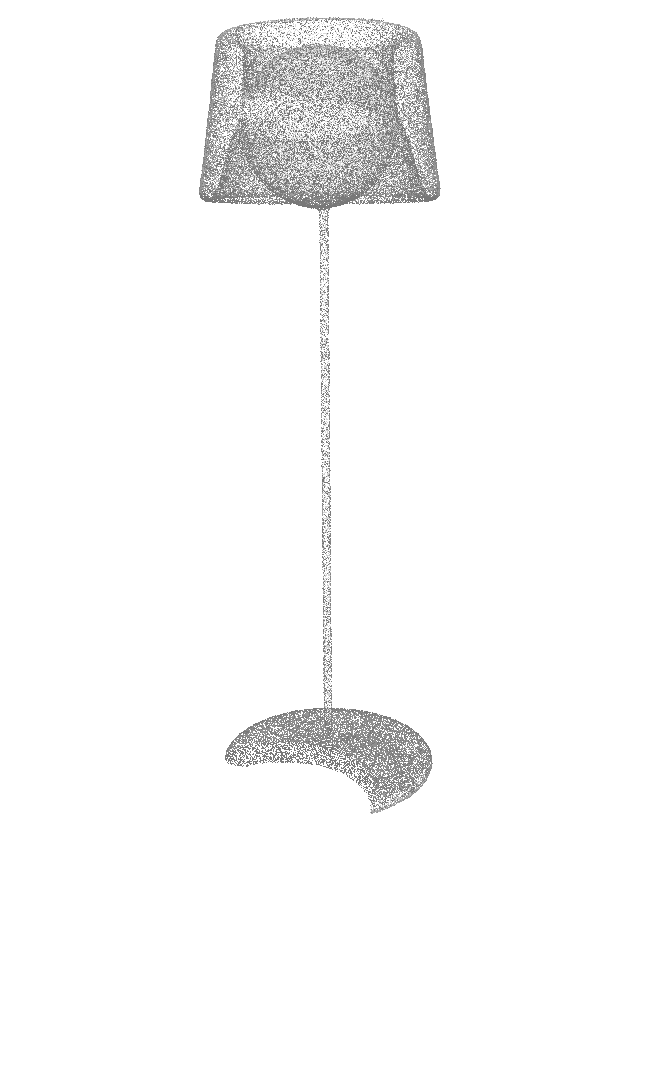} &
		\includegraphics[width=\sz\textwidth]{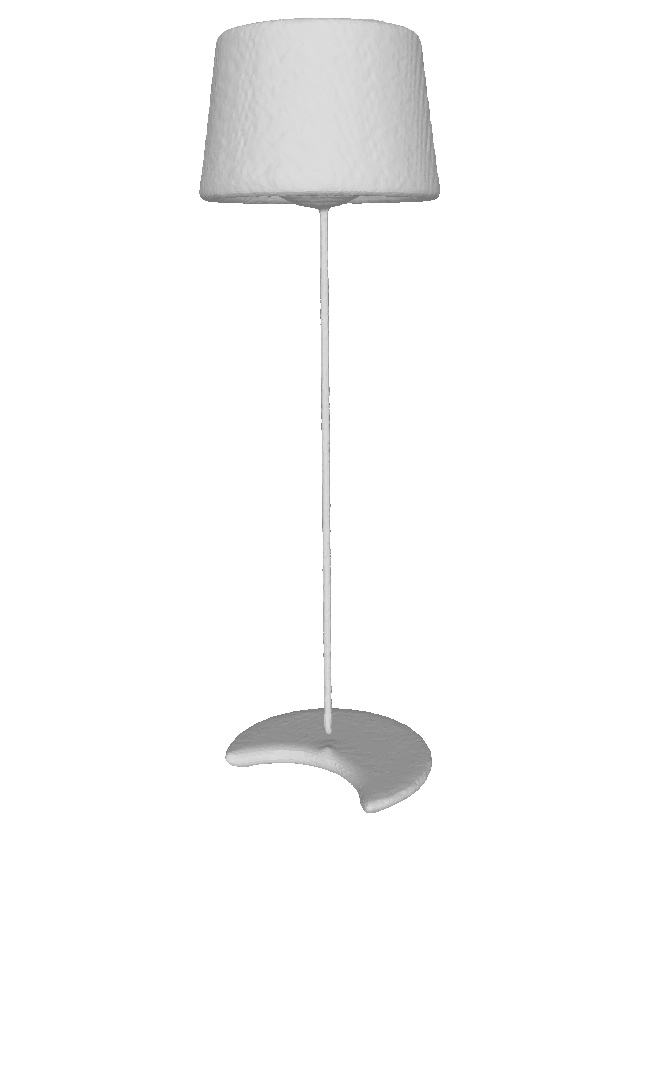} &
		\includegraphics[width=\sz\textwidth]{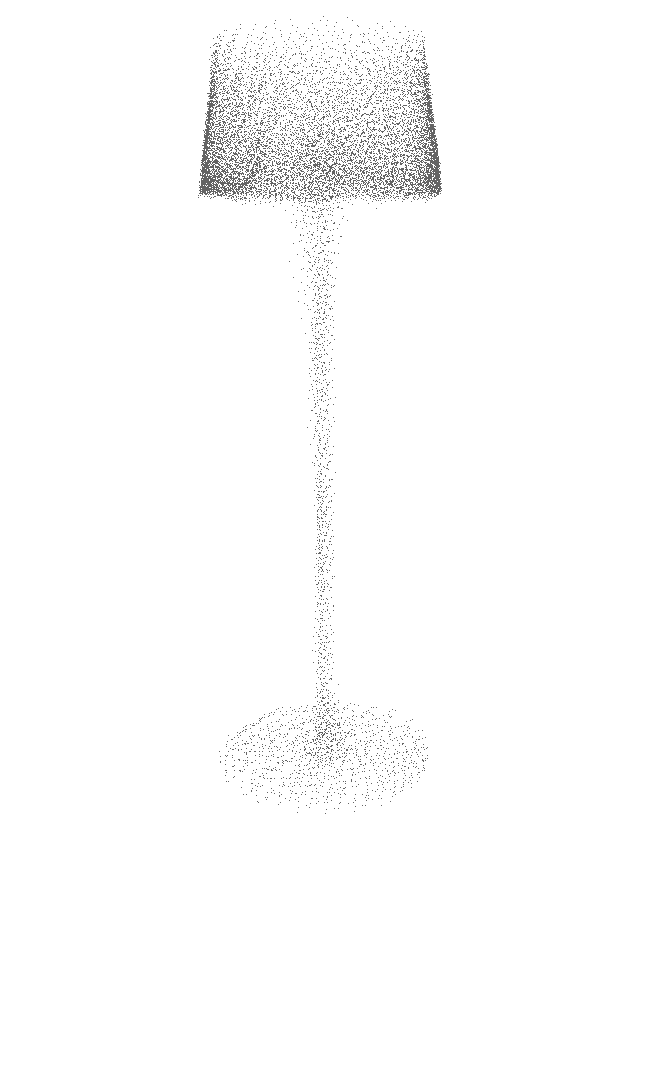} &
		\includegraphics[width=0.0985\textwidth]{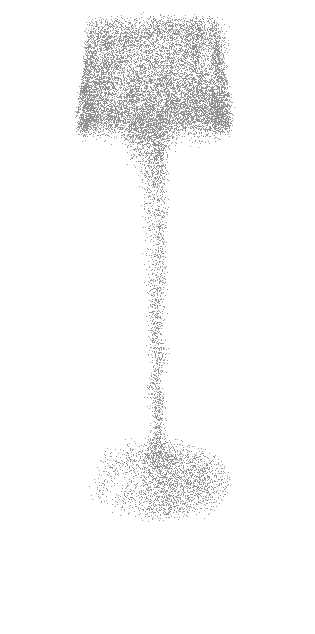} &
		\includegraphics[width=\sz\textwidth]{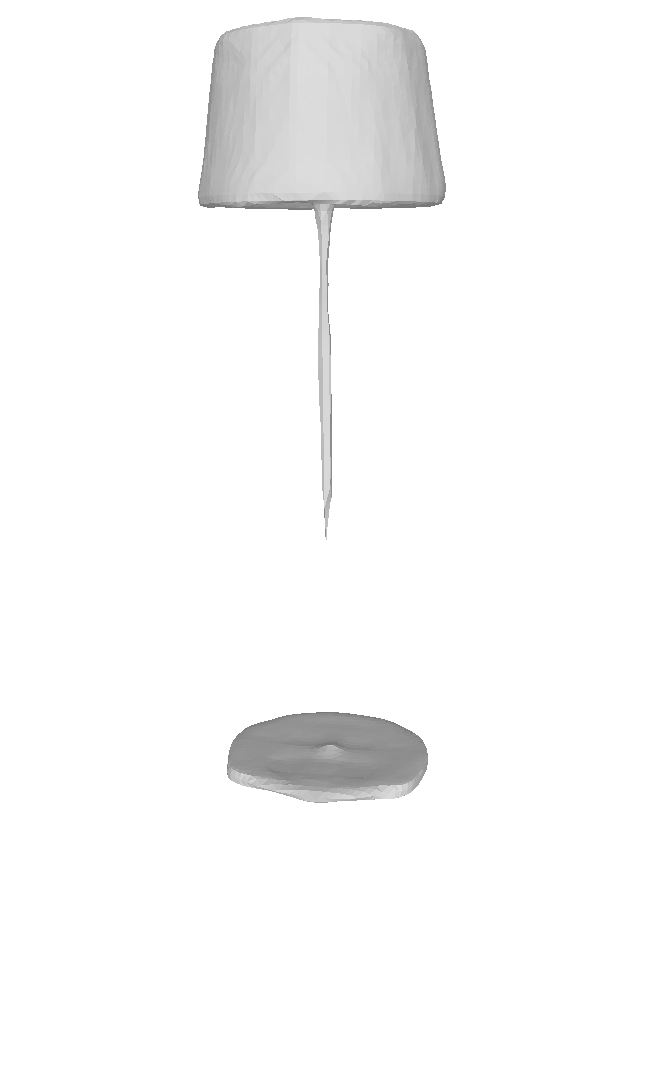} &
		\includegraphics[width=\sz\textwidth]{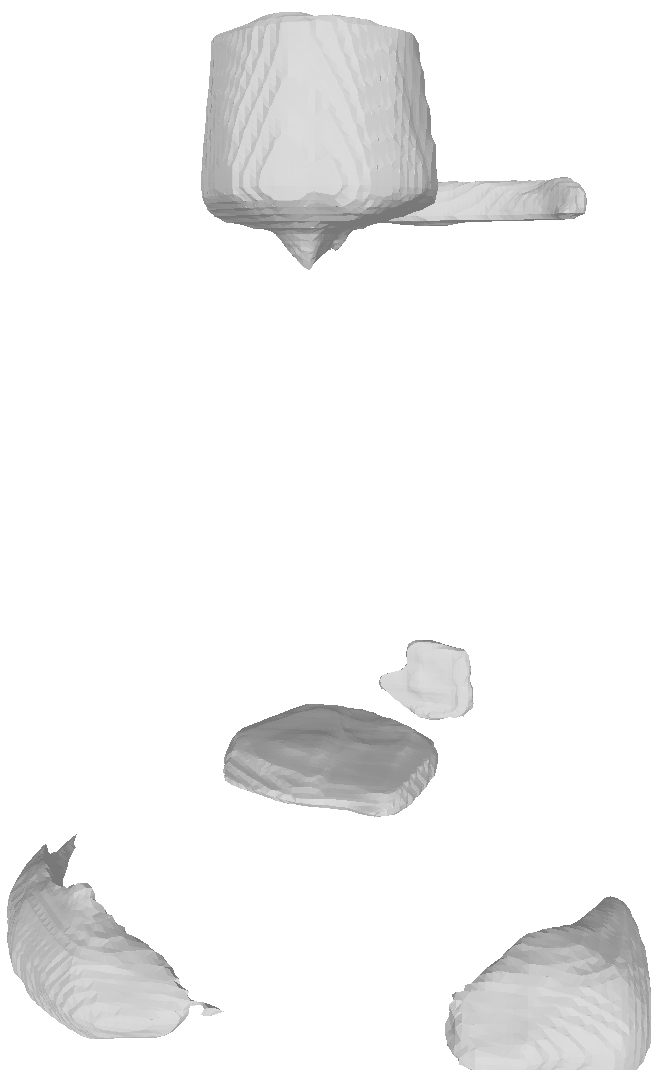} &
		\includegraphics[width=\sz\textwidth]{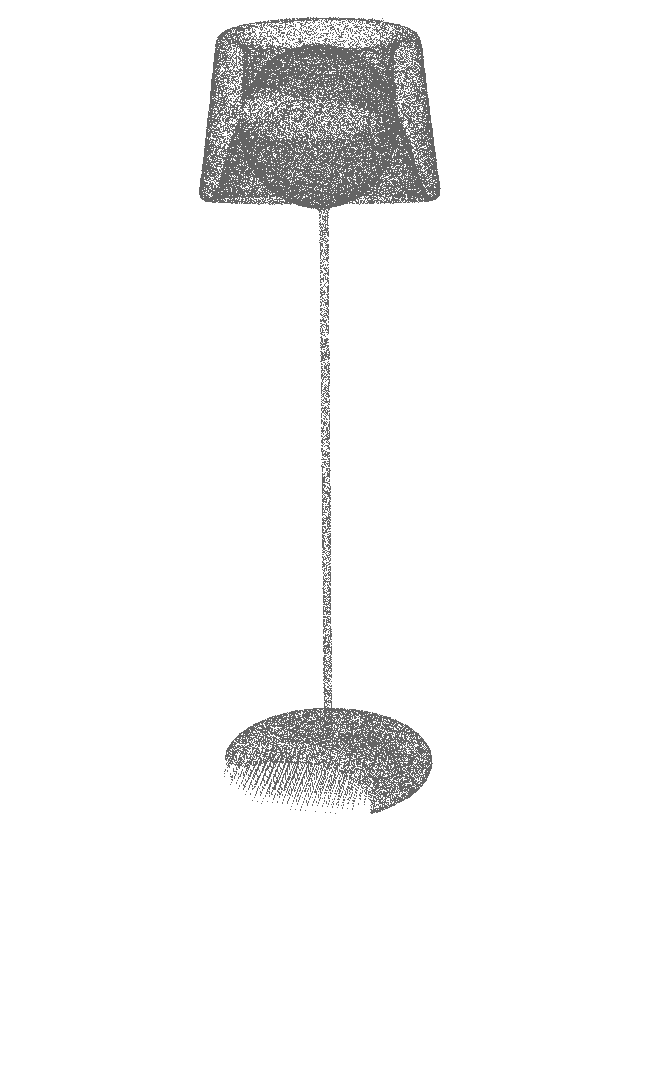} &
		\includegraphics[width=\sz\textwidth]{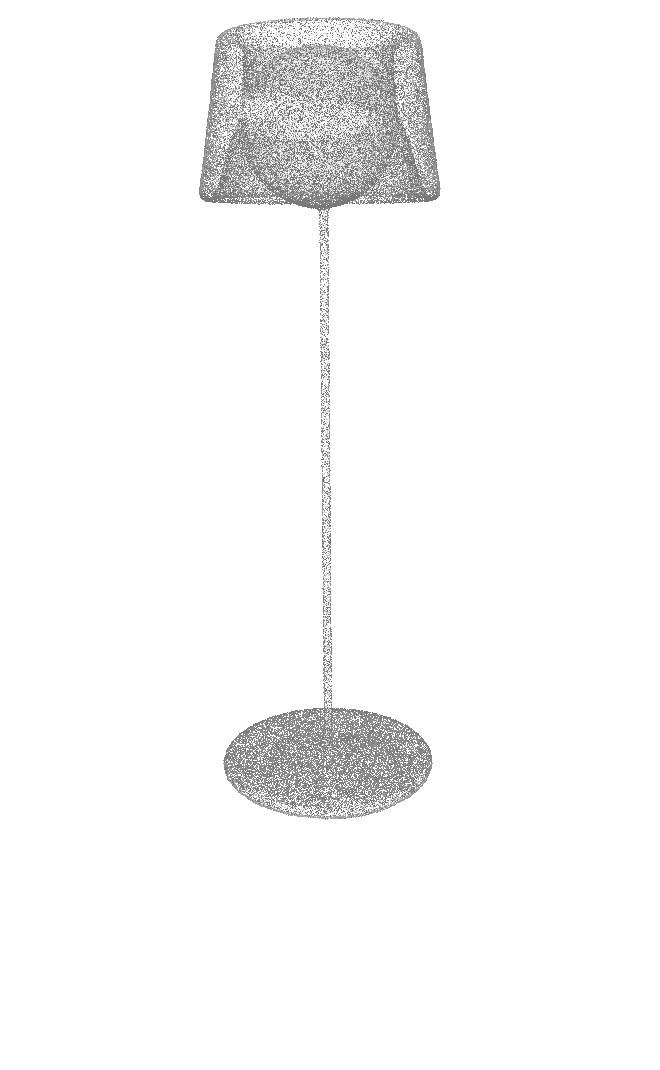} \\[12pt]
		\multirow{2}{*}[35pt]{\rotatebox{90}{\textbf{Sofa}}} &
		\includegraphics[width=\sz\textwidth]{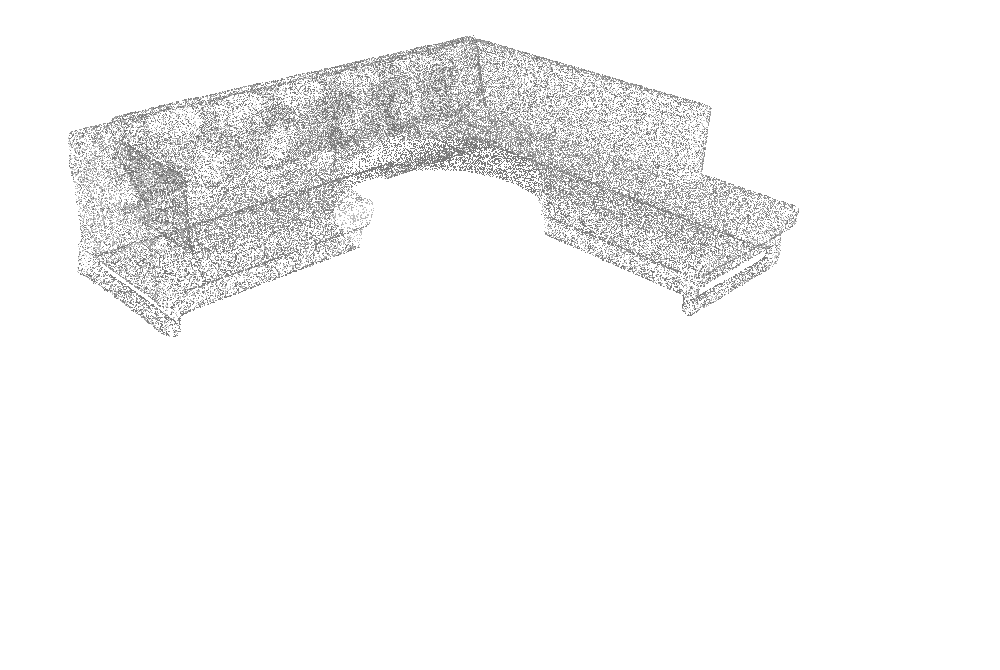} &
		\includegraphics[width=\sz\textwidth]{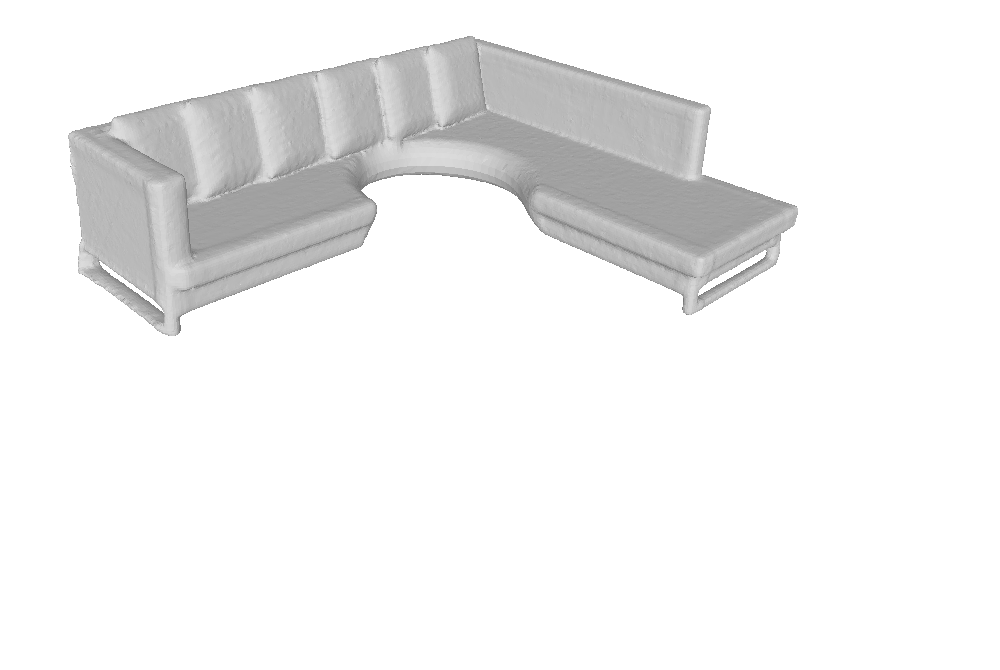} &
		\includegraphics[width=\sz\textwidth]{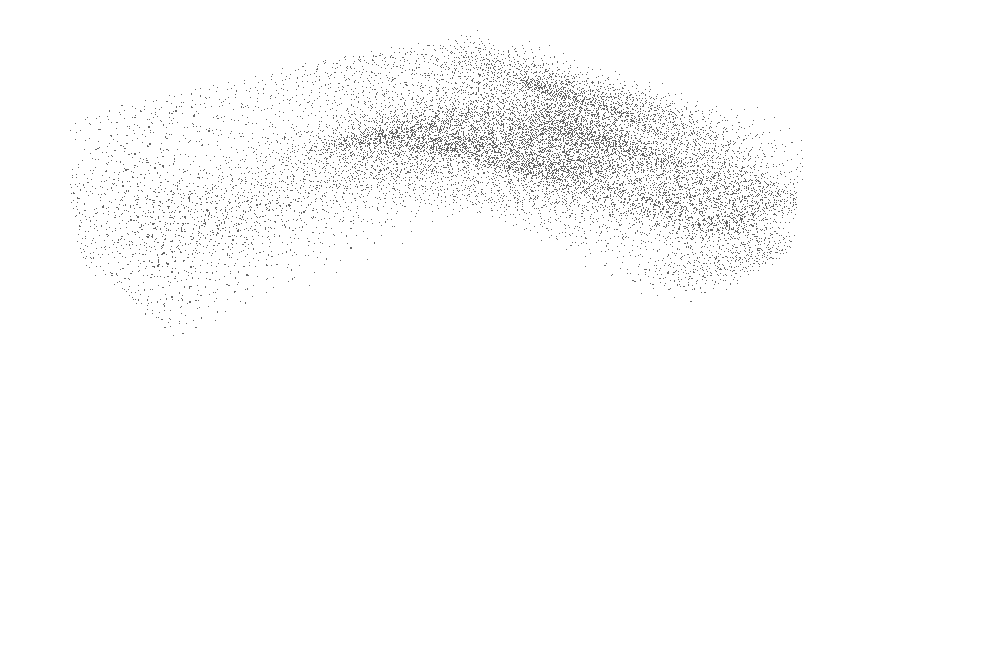} &
		\includegraphics[width=\sz\textwidth]{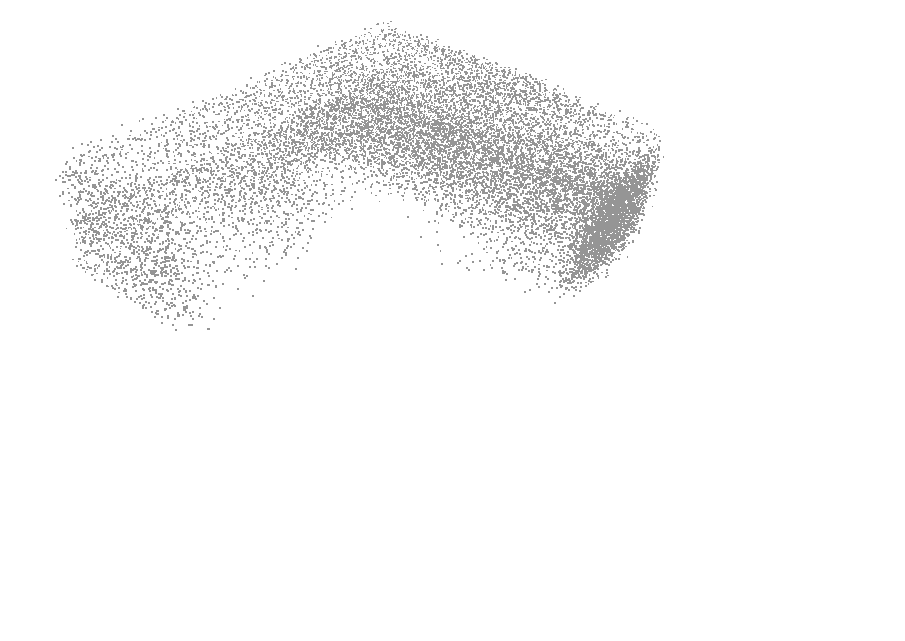} &
		\includegraphics[width=\sz\textwidth]{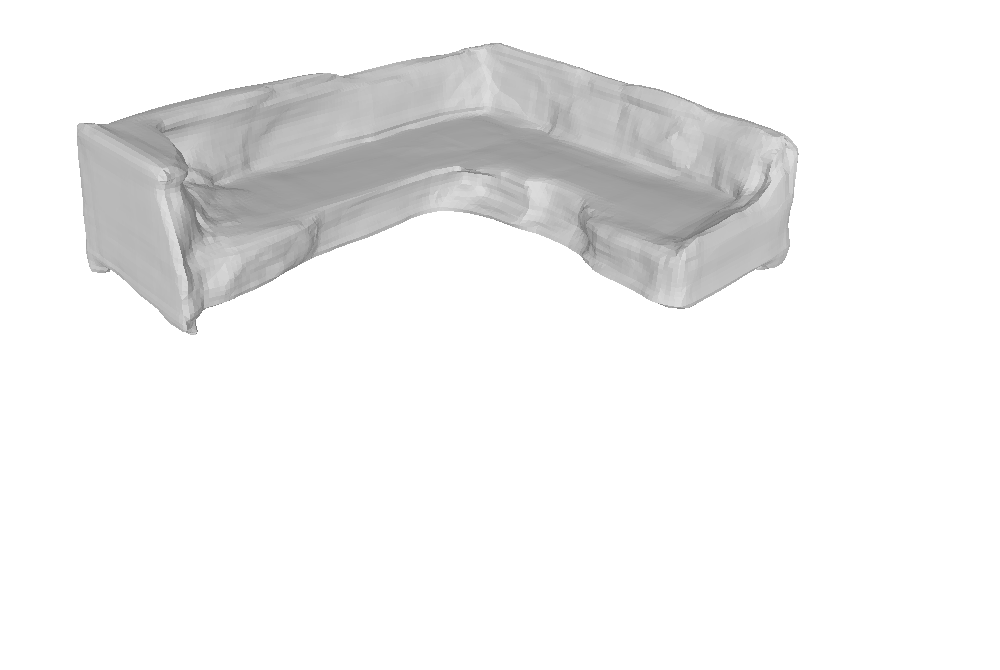} &
		\includegraphics[width=\sz\textwidth]{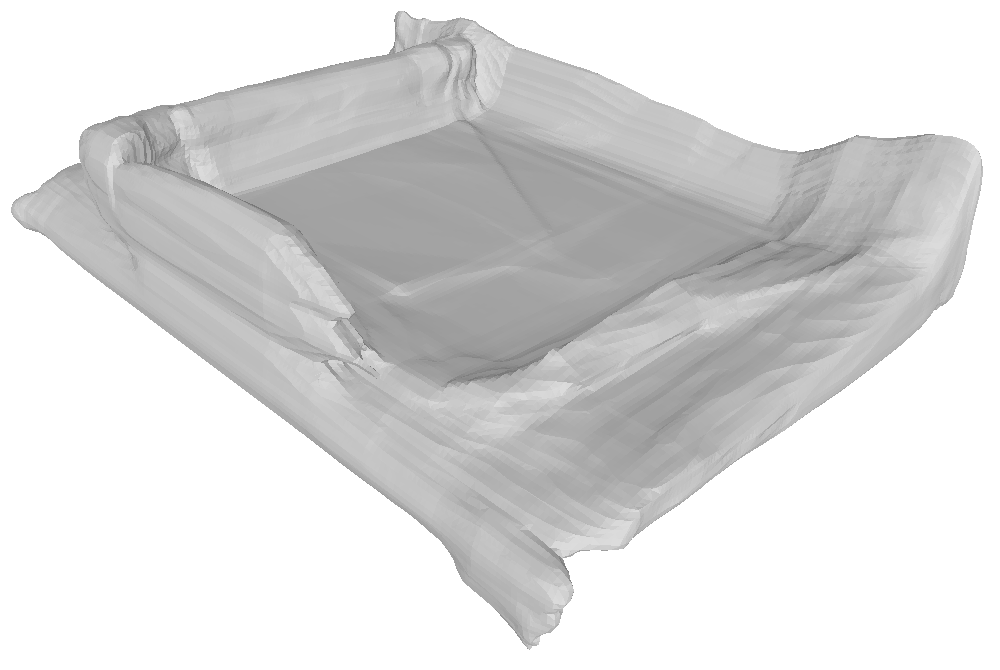} &
		\includegraphics[width=\sz\textwidth]{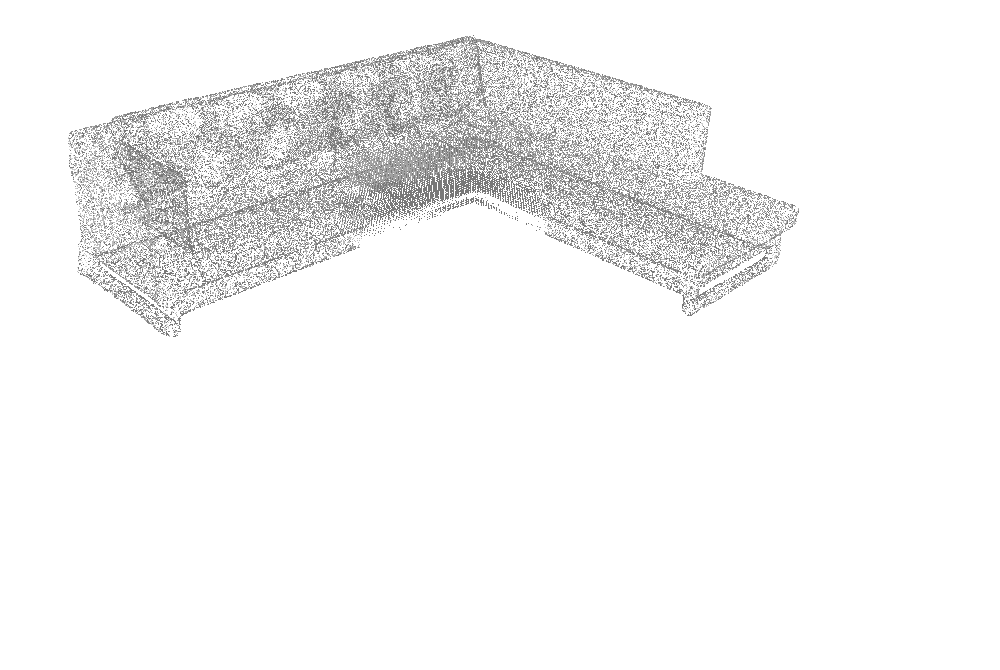} &
		\includegraphics[width=\sz\textwidth]{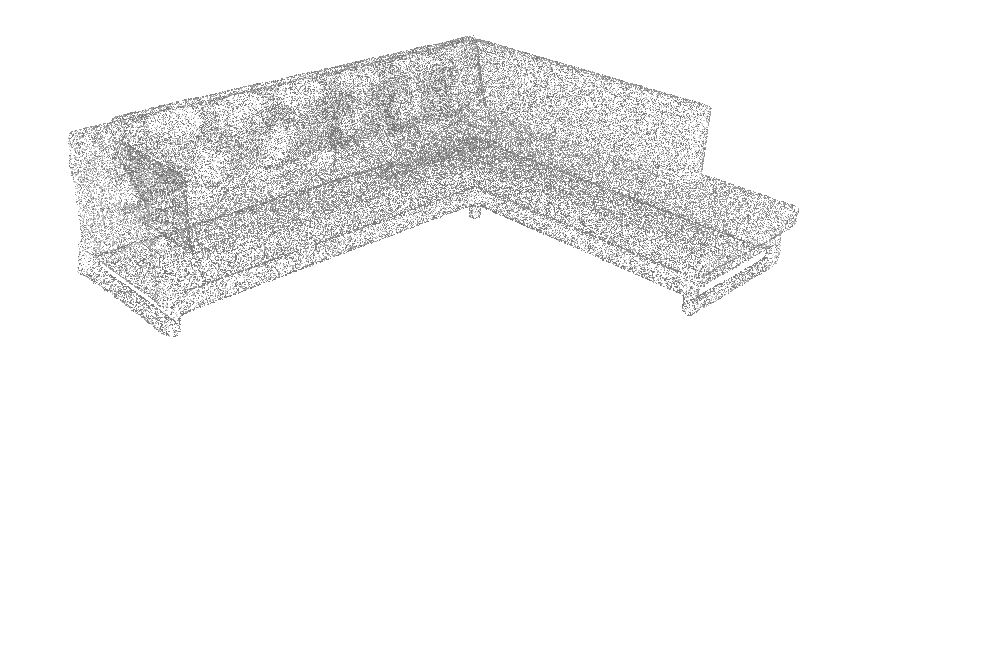} \\[12pt]
		\multirow{2}{*}[40pt]{\rotatebox{90}{\textbf{Table}}} &
		\includegraphics[width=\sz\textwidth]{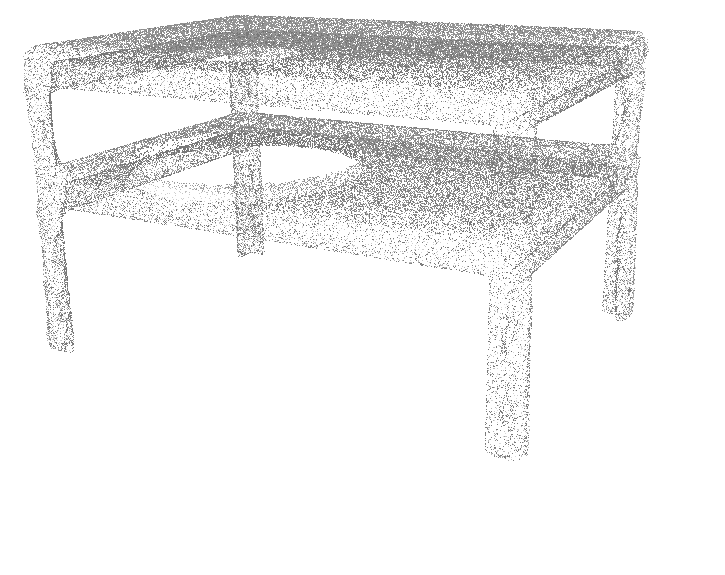} &
		\includegraphics[width=\sz\textwidth]{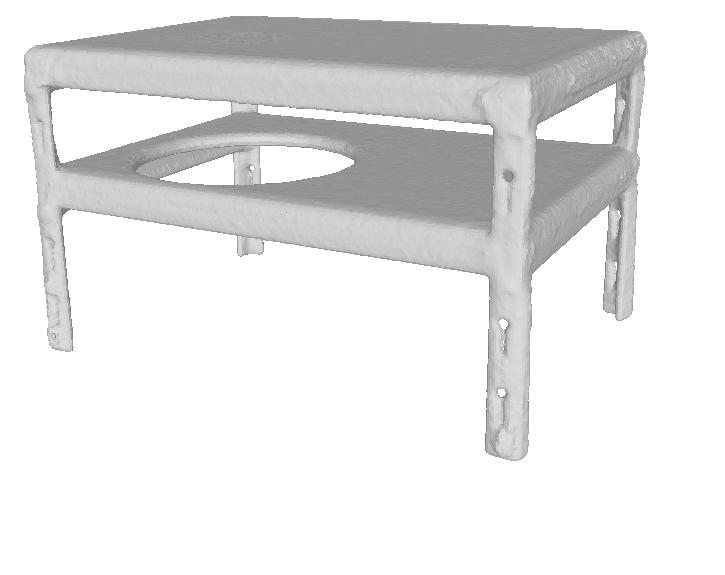} &
		\includegraphics[width=\sz\textwidth]{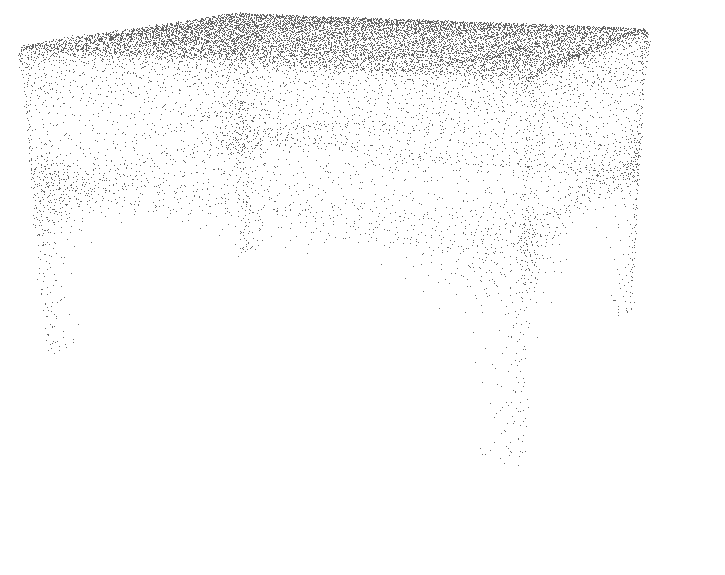} &
		\includegraphics[width=0.125\textwidth]{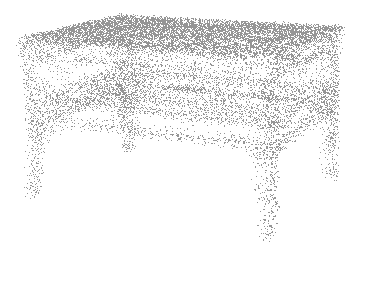} &
		\includegraphics[width=\sz\textwidth]{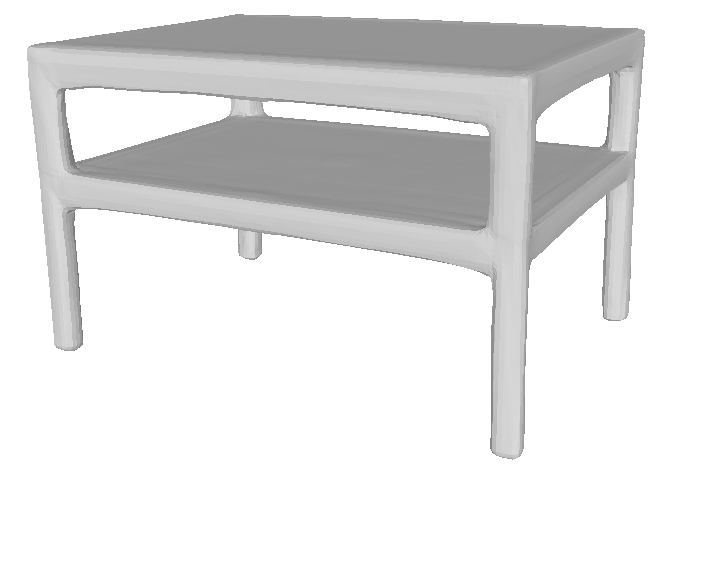} &
		\includegraphics[width=\sz\textwidth]{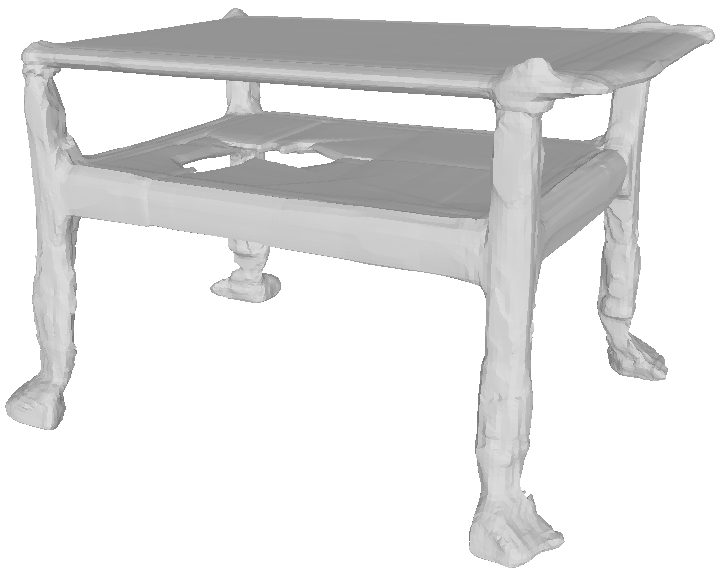} &
		\includegraphics[width=\sz\textwidth]{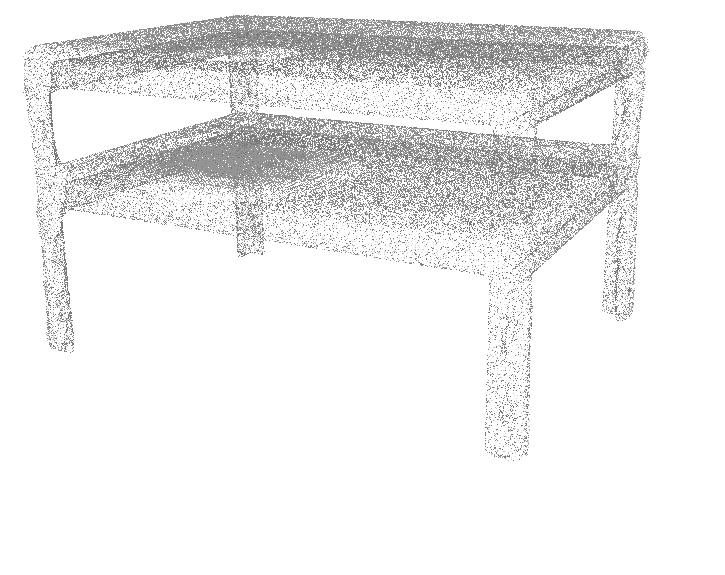} &
		\includegraphics[width=\sz\textwidth]{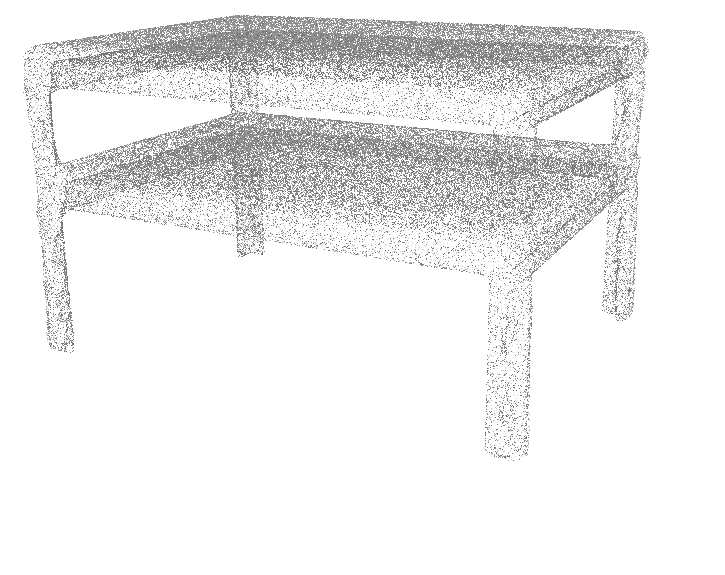} \\[12pt]
		\multirow{2}{*}[25pt]{\rotatebox{90}{\textbf{Table}}} &
		\includegraphics[width=\sz\textwidth]{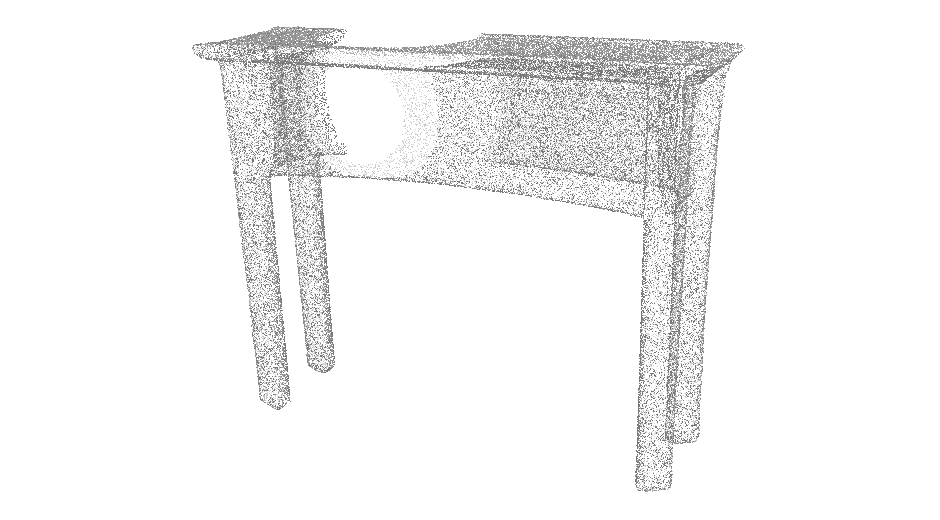} &
		\includegraphics[width=\sz\textwidth]{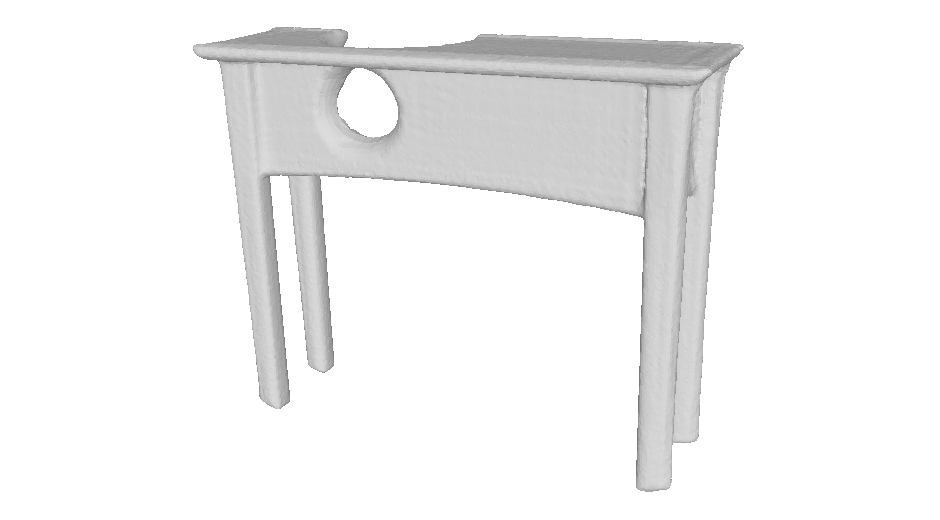} &
		\includegraphics[width=\sz\textwidth]{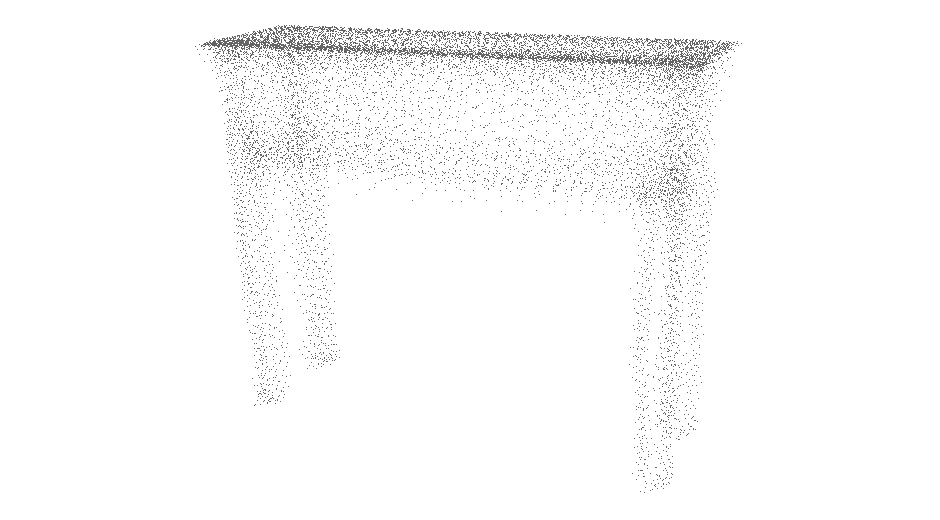} &
		\includegraphics[width=0.10\textwidth]{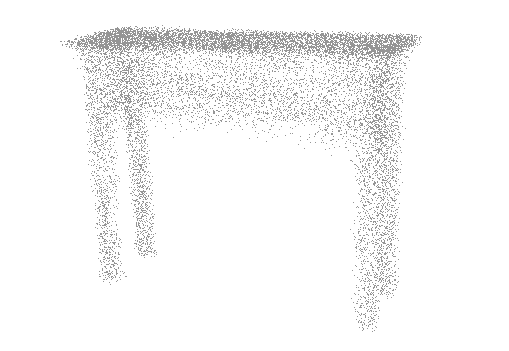} &
		\includegraphics[width=\sz\textwidth]{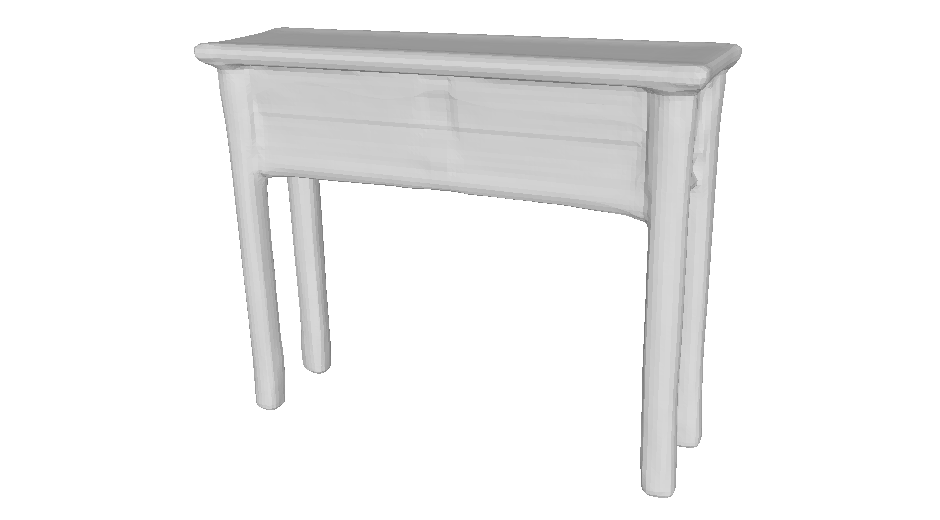} &
		\includegraphics[width=\sz\textwidth]{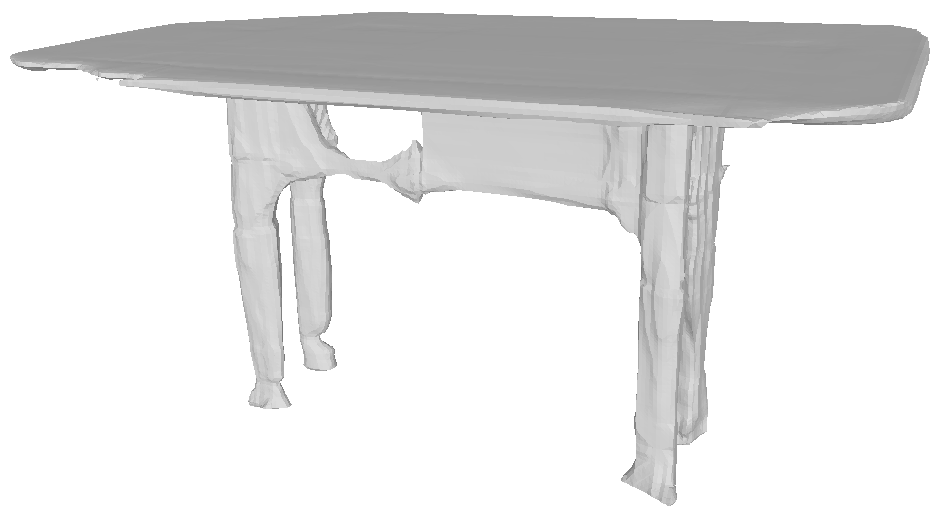} &
		\includegraphics[width=\sz\textwidth]{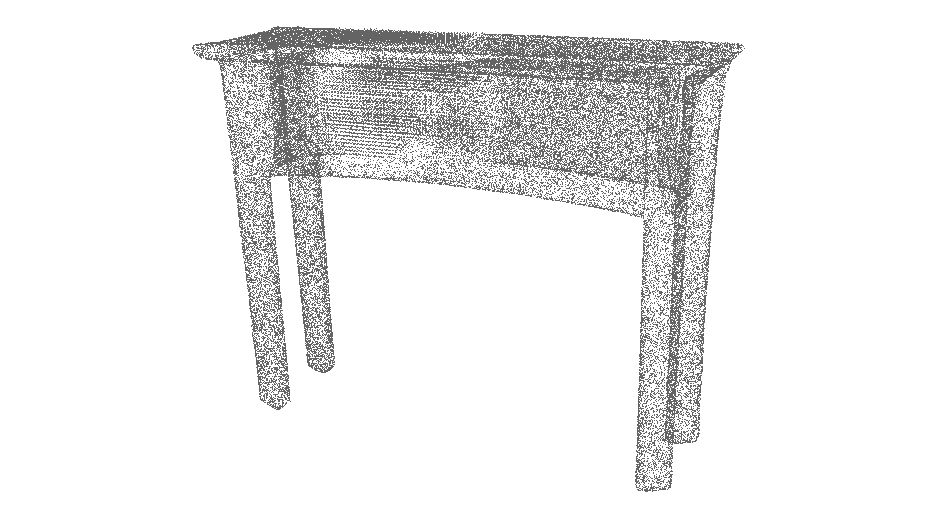} &
		\includegraphics[width=\sz\textwidth]{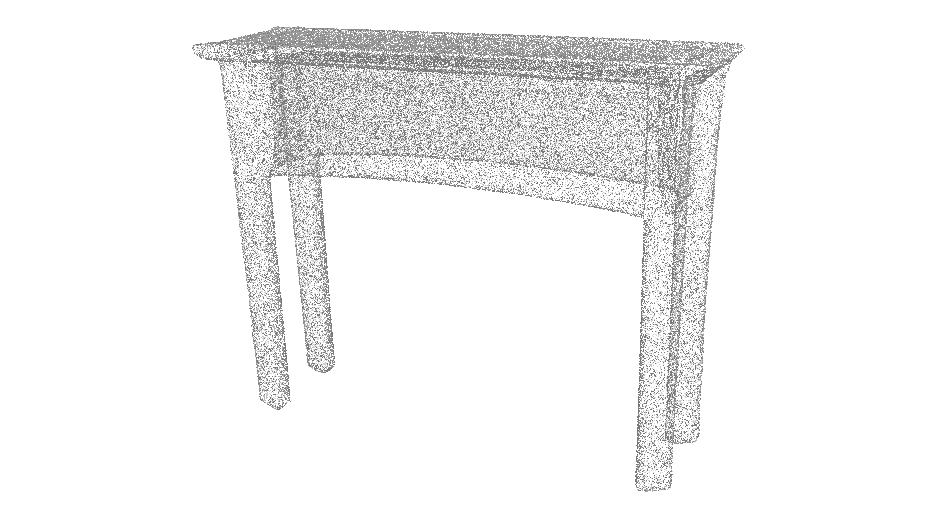} \\[12pt]
		\multirow{2}{*}[35pt]{\rotatebox{90}{\textbf{Table}}} &
		\includegraphics[width=\insz\textwidth]{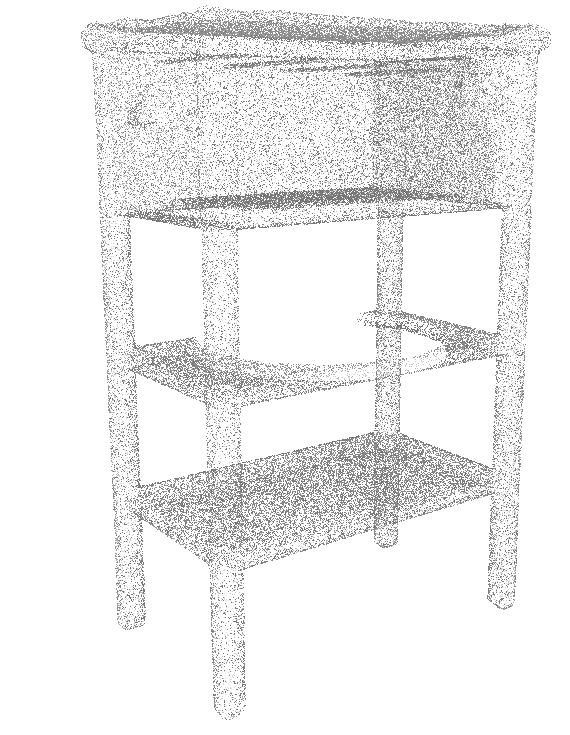} &
		\includegraphics[width=\insz\textwidth]{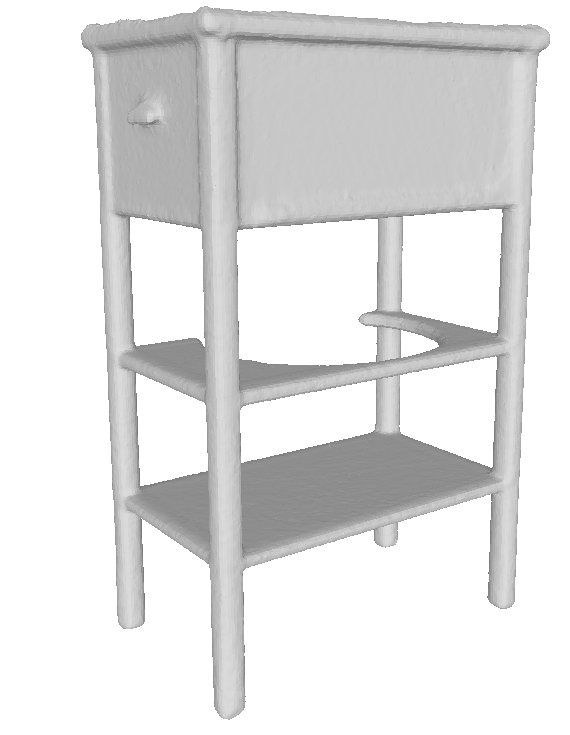} &
		\includegraphics[width=\insz\textwidth]{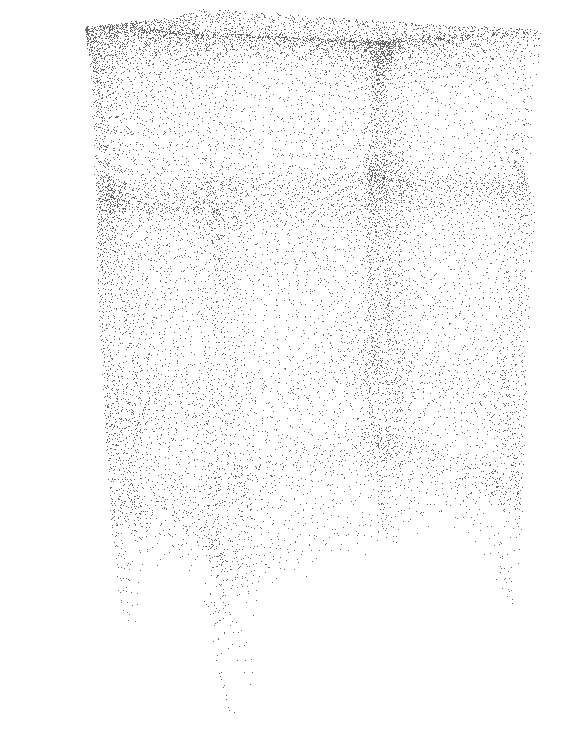} &
		\includegraphics[width=\insz\textwidth]{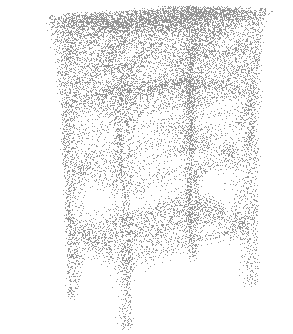} &
		\includegraphics[width=\insz\textwidth]{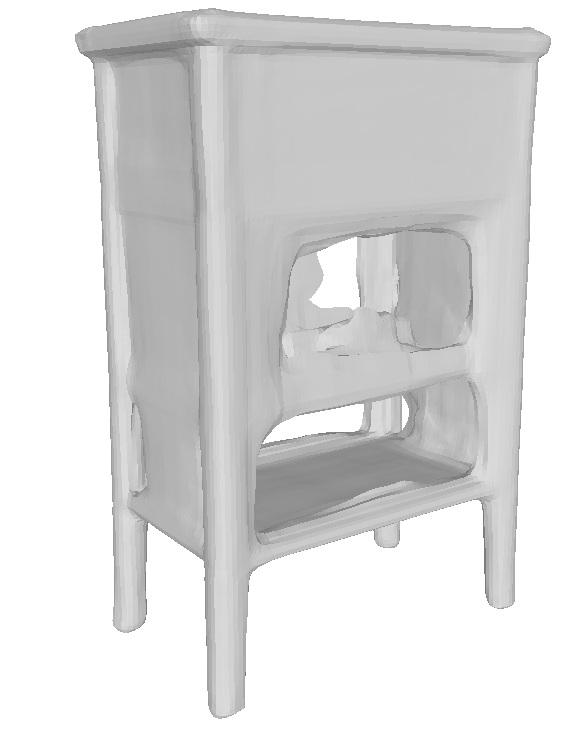} &
		\includegraphics[width=\insz\textwidth]{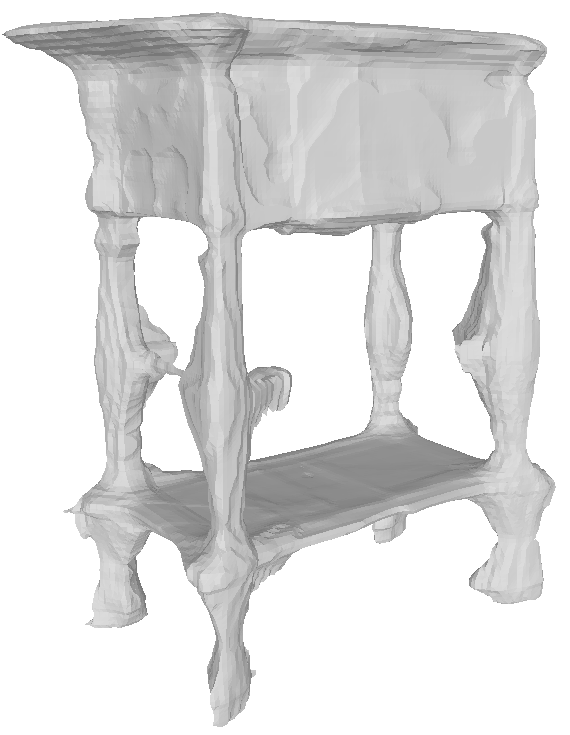} &
		\includegraphics[width=\insz\textwidth]{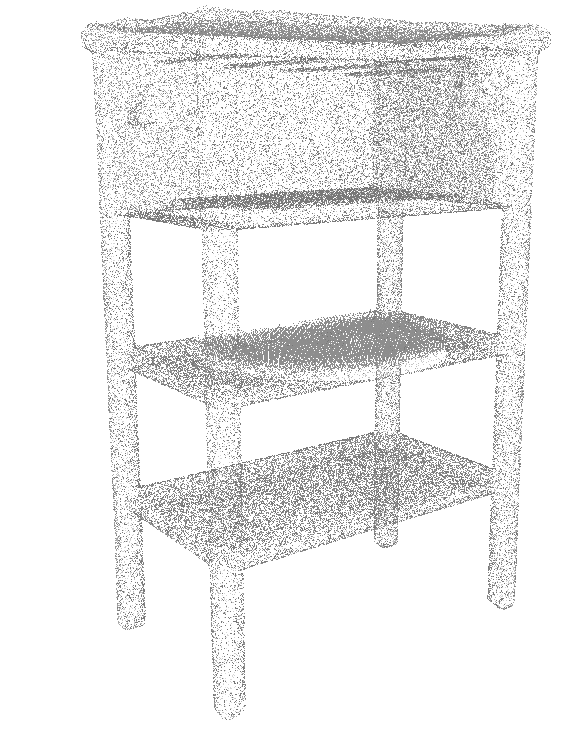} &
		\includegraphics[width=\insz\textwidth]{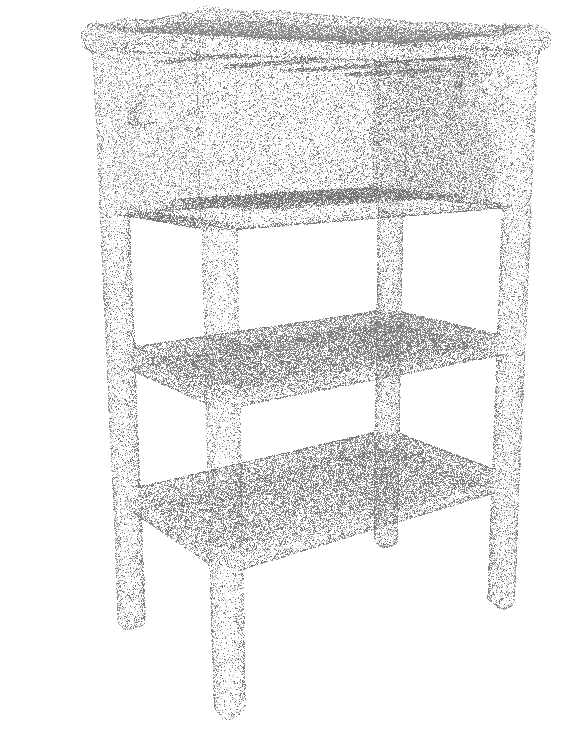} \\[12pt]
	\end{tabular}
	\caption{\textbf{Additional qualitative results on ShapeNet.}}
	\label{fig:supp_qualitative_comparison}
\end{figure*}

We present additional qualitative results in Fig.~\ref{fig:supp_qualitative_comparison} to illustrate the performance of \kaplan{}.
The first row nicely illustrates the benefits of preserving original geometry where available, as methods that recreate the complete shape miss details like the propellers.
Moreover, as illustrated by the second row, it is also preferrable for untypical examples to condition on the learned prior only where necessary, so as not to overwrite rare but valid geometry with prior expectations.
As a general observation, \kaplan{} strikes a good compromise between global shape context and local surface cues. We find that our baselines often are either good at recovering missing geometry from global high-level cues, but deviate a lot from the existing surface geometry (e.g., OccNet, PCN); or they perform accurate local surface fitting, but disregard global cues like symmetry or the number of legs (e.g., PSR, DeepSDF).

\begin{figure*}[p]
	\centering
	\scriptsize
	\setlength{\tabcolsep}{0.3mm}
	\newcommand{\sz}{0.12}
	\newcommand{\insz}{0.08}
	\newcommand{\inszz}{0.12}
	\begin{tabular}{ccccccccc}
		& \textbf{Input} & \textbf{PSR~\cite{Kazhdan-et-al-SGP-2006}} &  \textbf{PCN~\cite{Yuan-et-al-3DV-2018}}  & \textbf{Cascaded~\cite{Wang-et-al-CVPR-2020}}  & \textbf{OccNet~\cite{Mescheder-et-al-CVPR-2019}} & \textbf{DeepSDF~\cite{Park-et-al-CVPR-2019}} & \textbf{Ours} & \textbf{GT} \\[12pt]
		\multirow{2}{*}[30pt]{\rotatebox{90}{\textbf{Plane}}} &
		\includegraphics[width=\sz\textwidth]{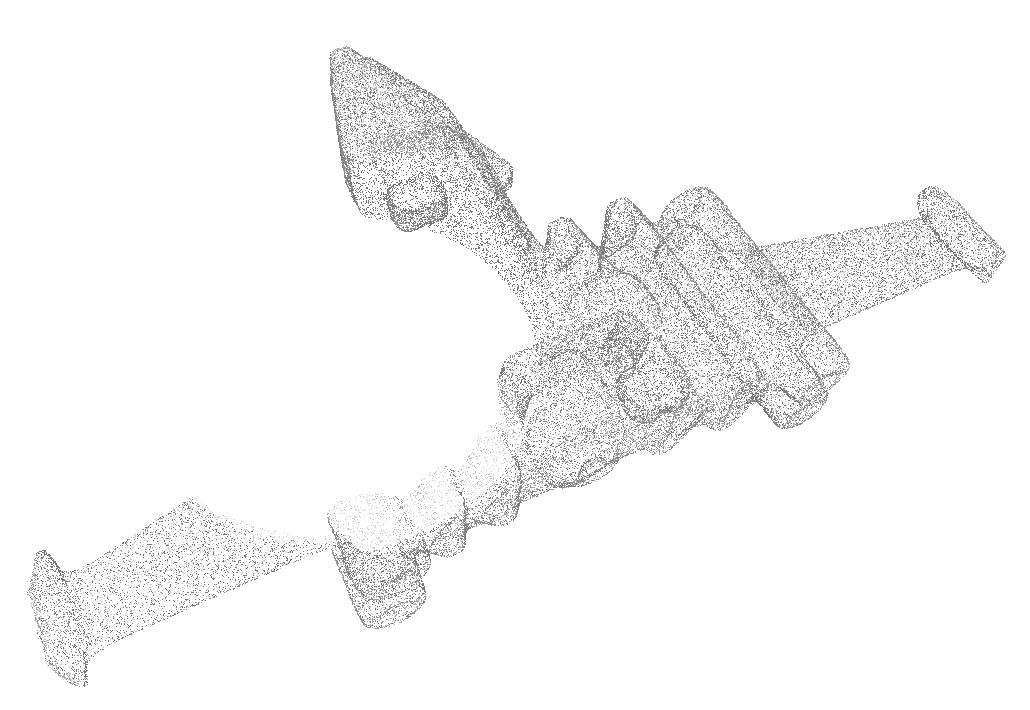} &
		\includegraphics[width=\sz\textwidth]{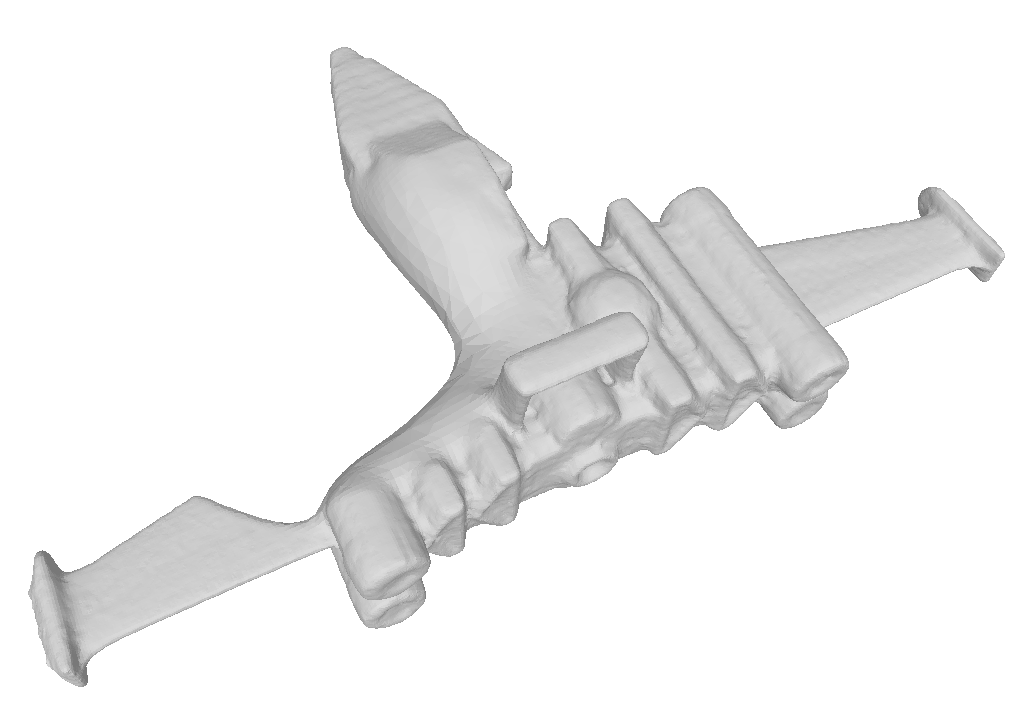} &
		\includegraphics[width=\sz\textwidth]{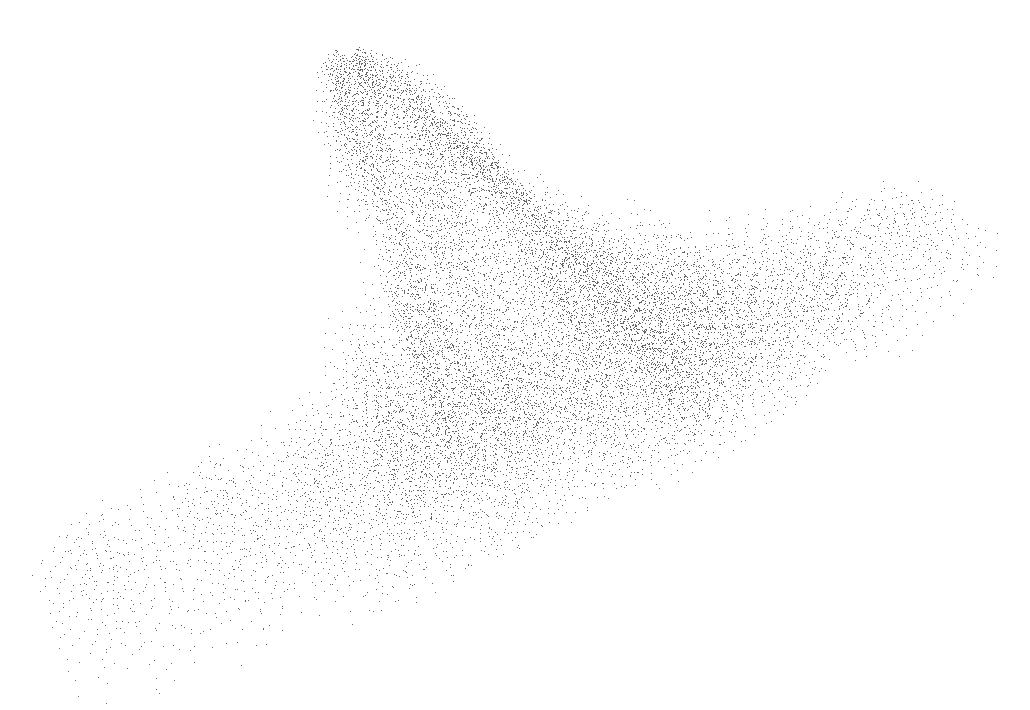} &
		\includegraphics[width=\sz\textwidth]{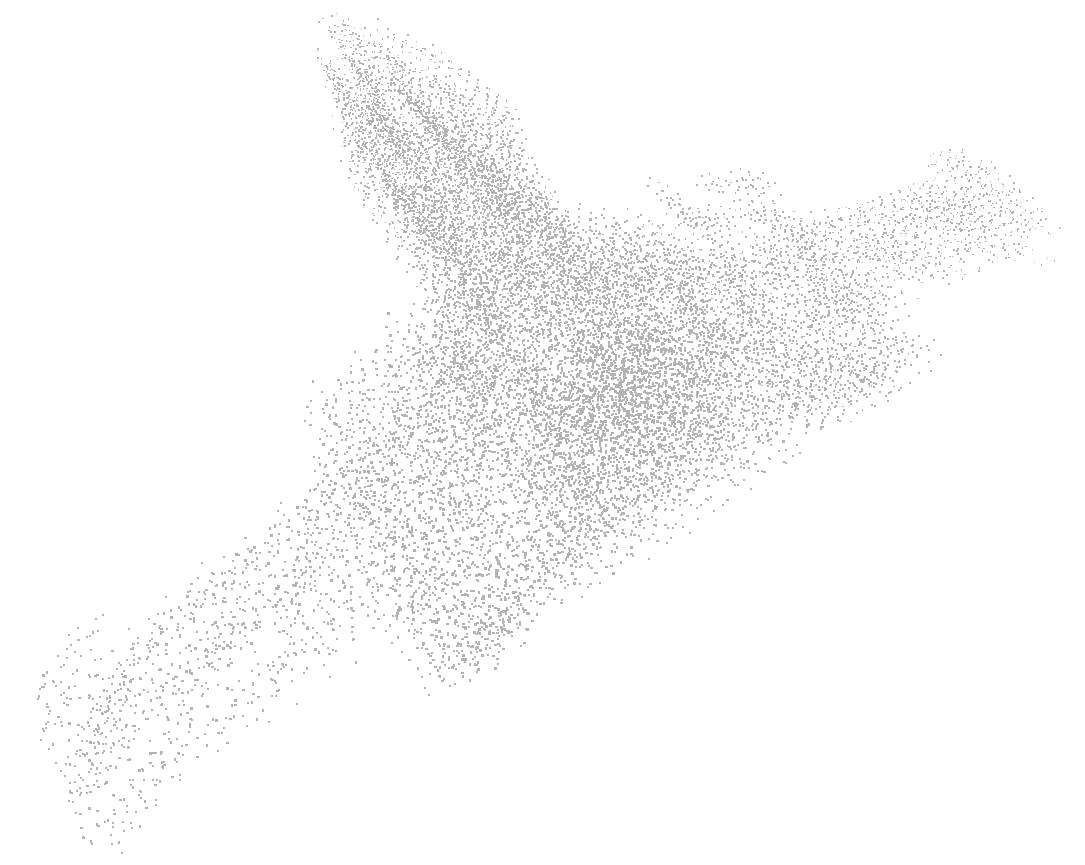} &
		\includegraphics[width=\sz\textwidth]{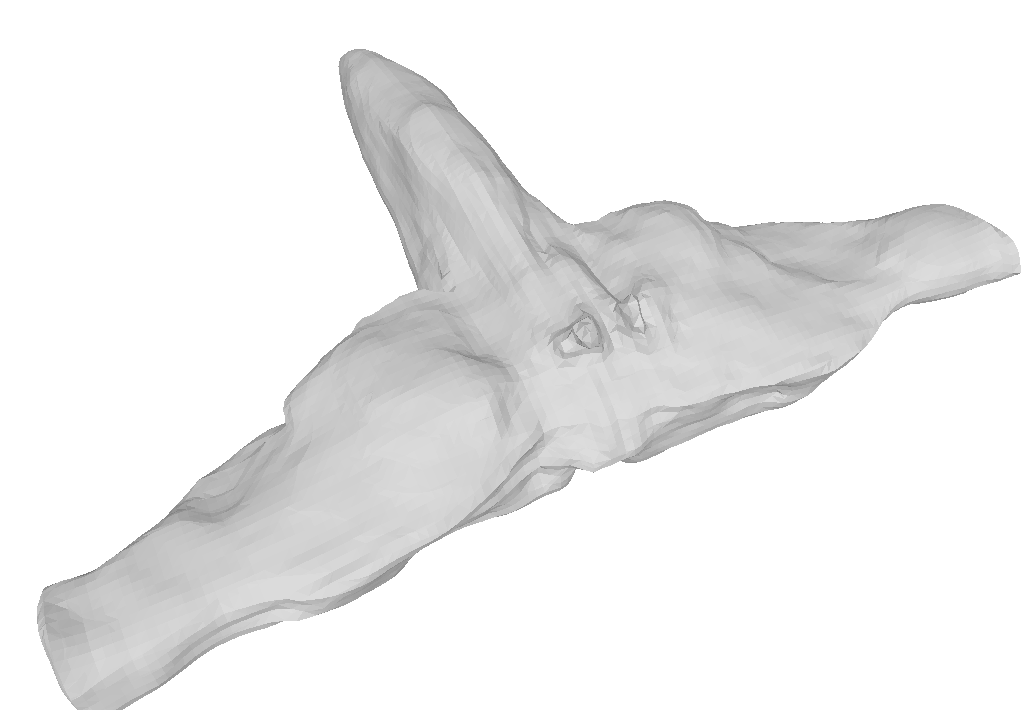} &
		\includegraphics[width=\sz\textwidth]{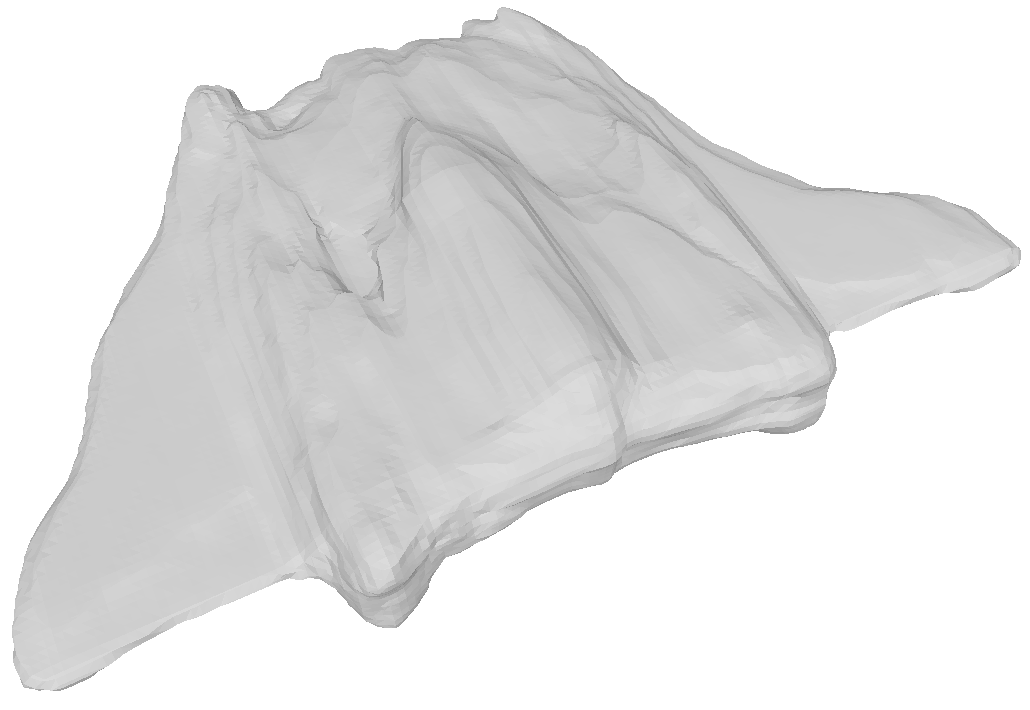} &
		\includegraphics[width=\sz\textwidth]{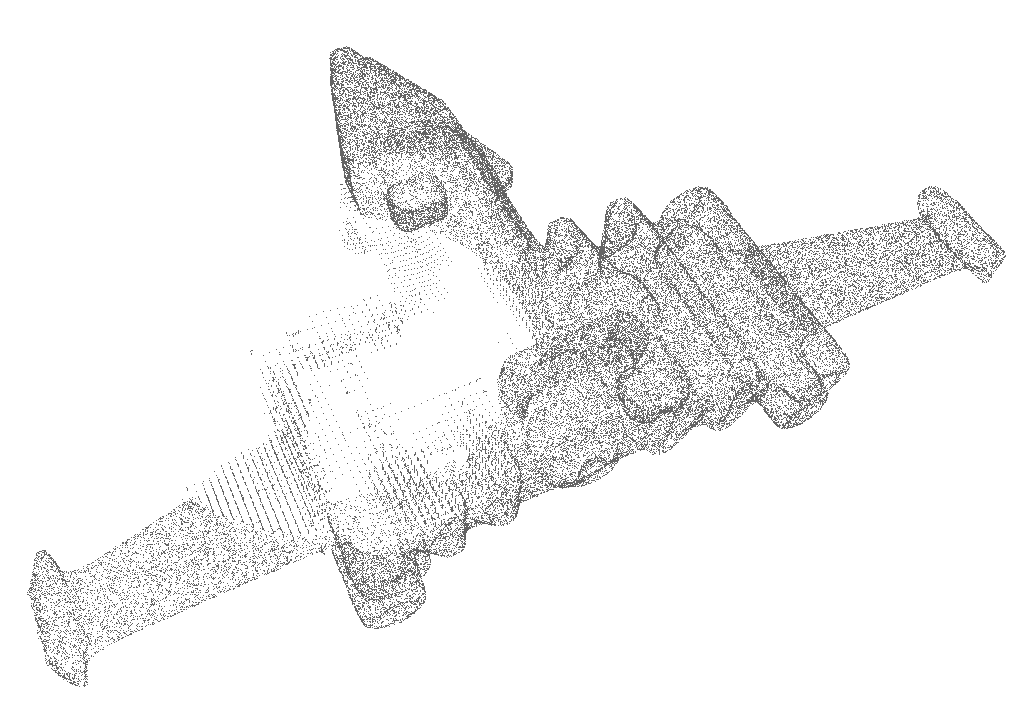} &
		\includegraphics[width=\sz\textwidth]{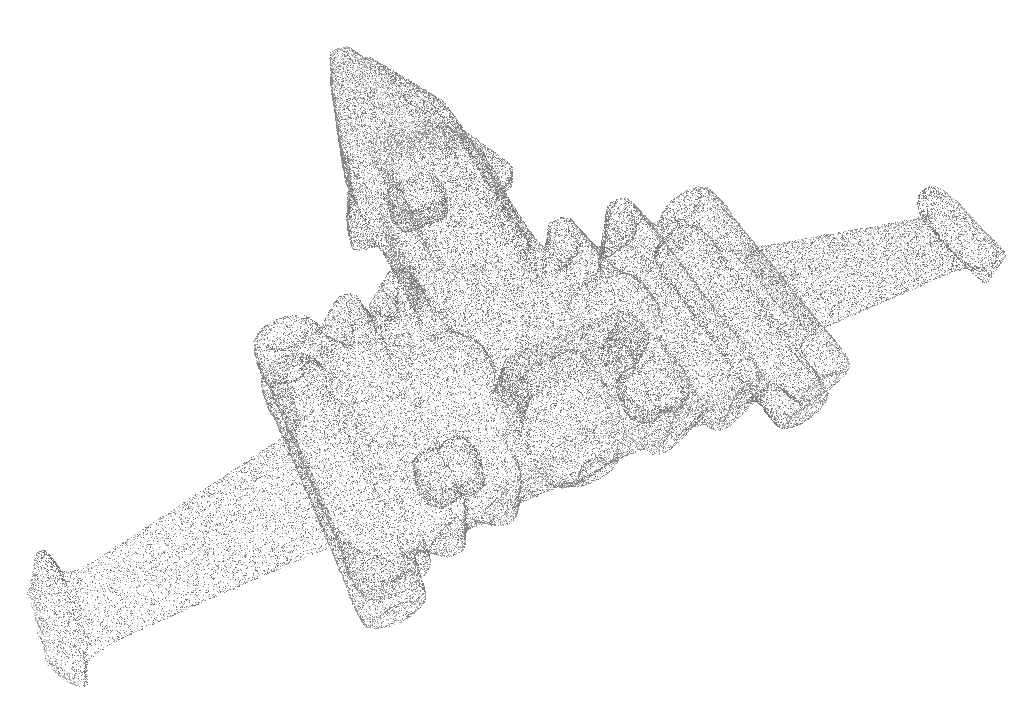} \\[12pt]
		\multirow{2}{*}[60pt]{\rotatebox{90}{\textbf{Chair}}} &
		\includegraphics[width=\sz\textwidth]{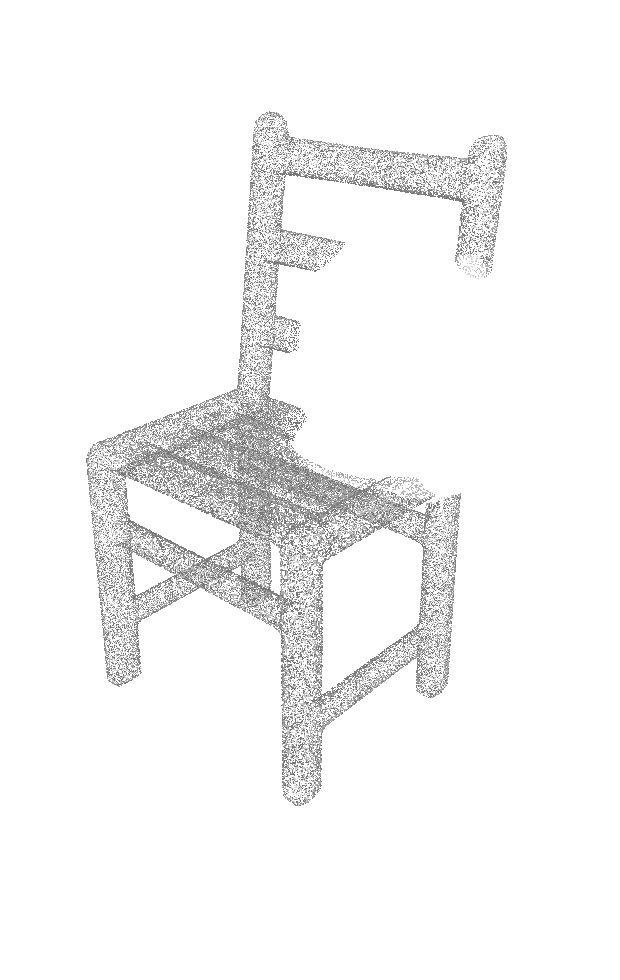} &
		\includegraphics[width=\sz\textwidth]{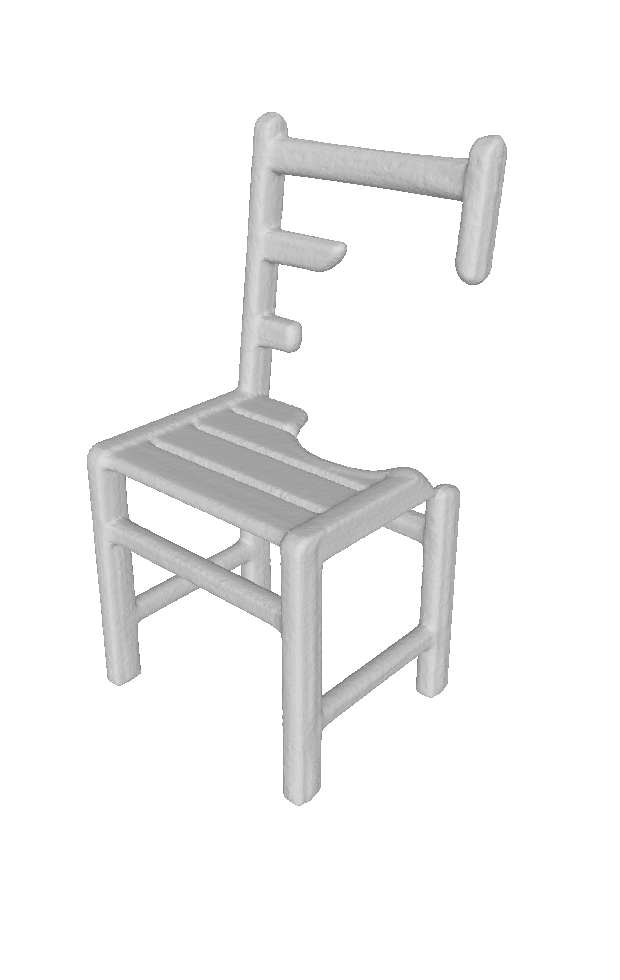} &
		\includegraphics[width=\sz\textwidth]{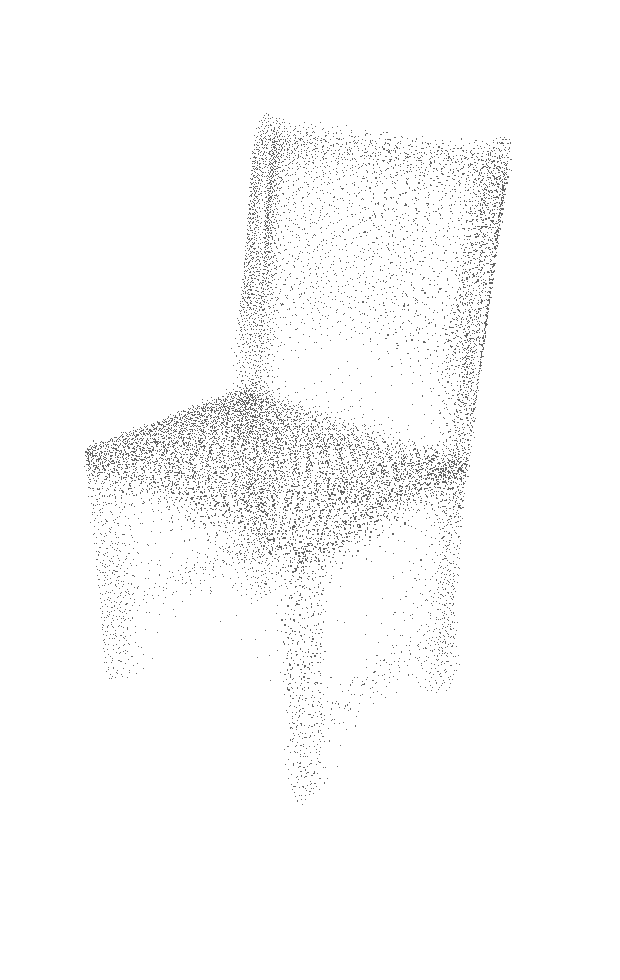} &
		\includegraphics[width=0.10\textwidth]{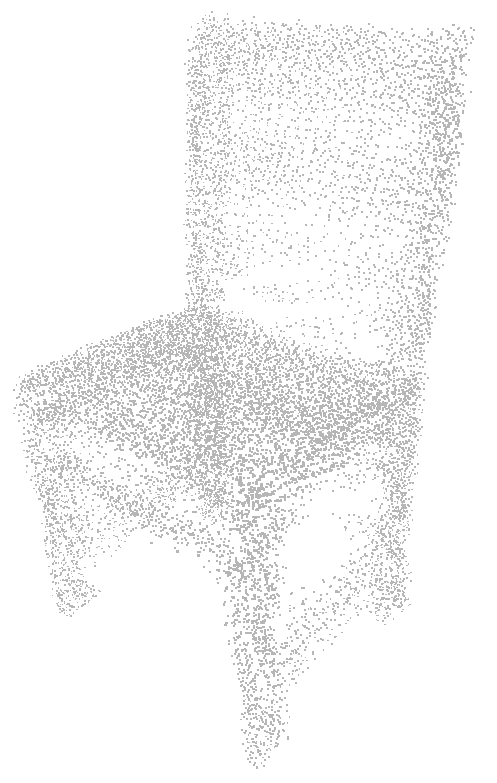} &
		\includegraphics[width=\sz\textwidth]{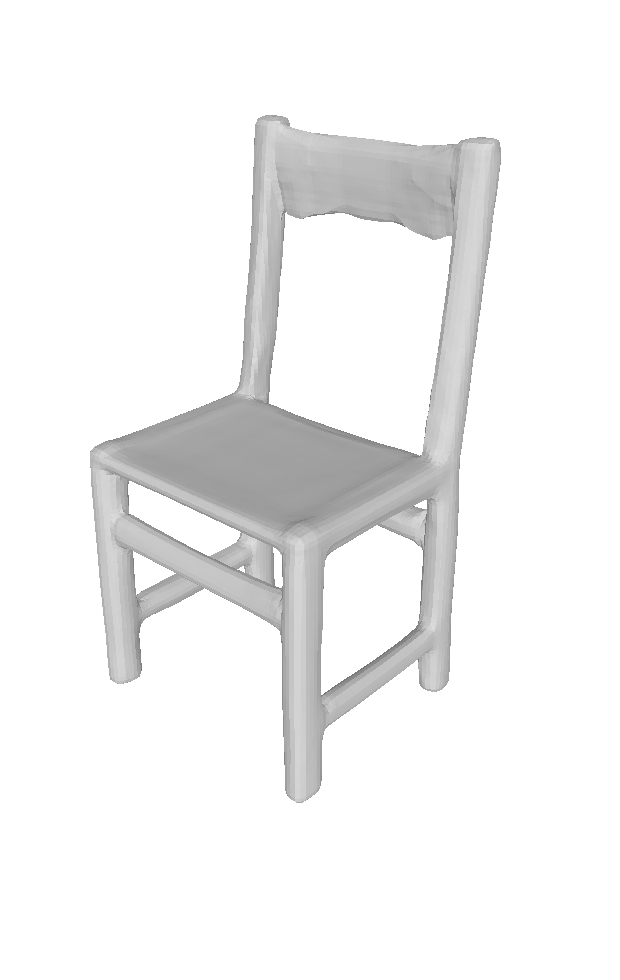} &
		\includegraphics[width=\sz\textwidth]{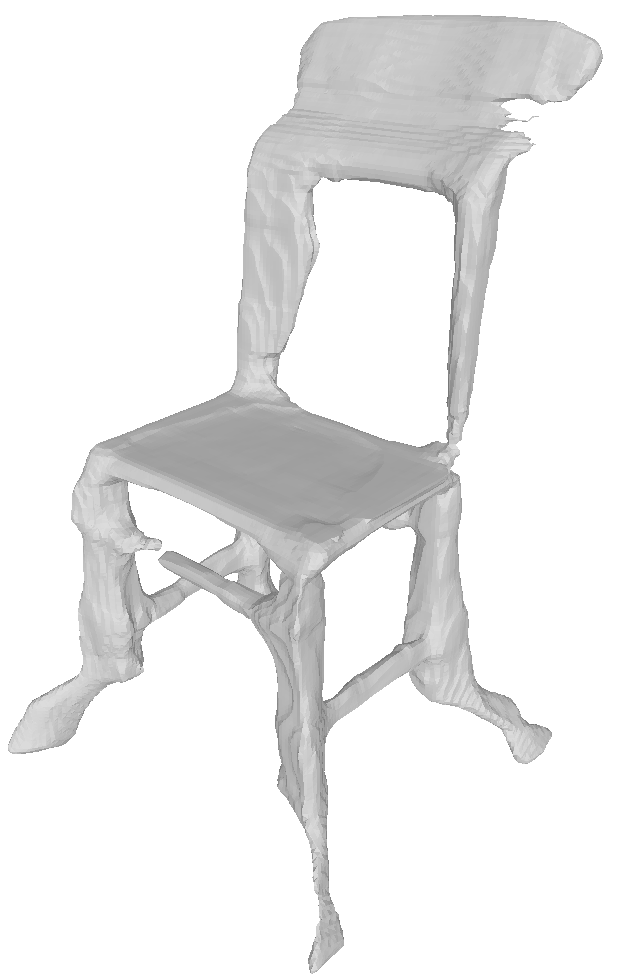} &
		\includegraphics[width=\sz\textwidth]{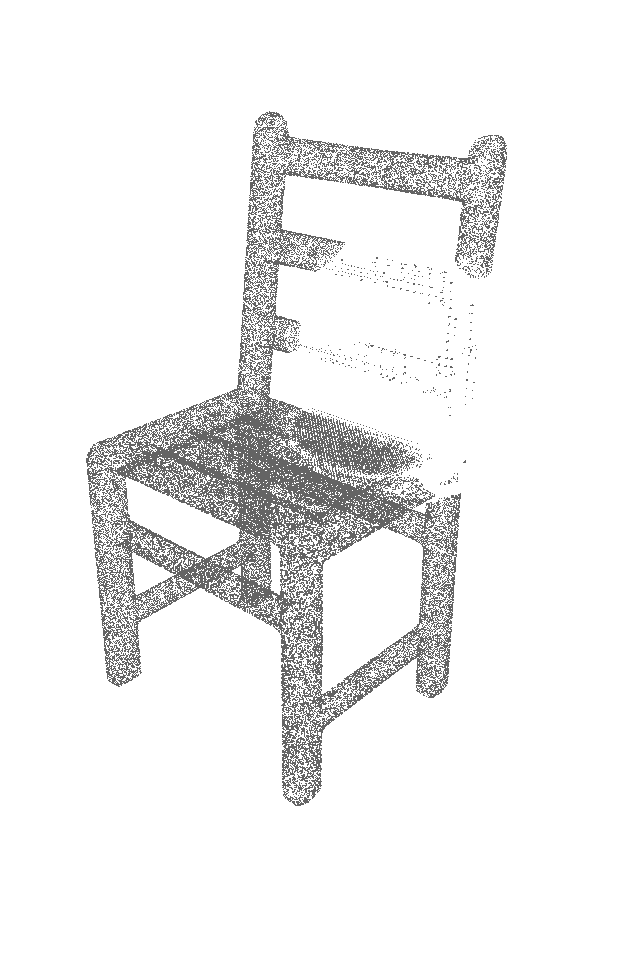} &
		\includegraphics[width=\sz\textwidth]{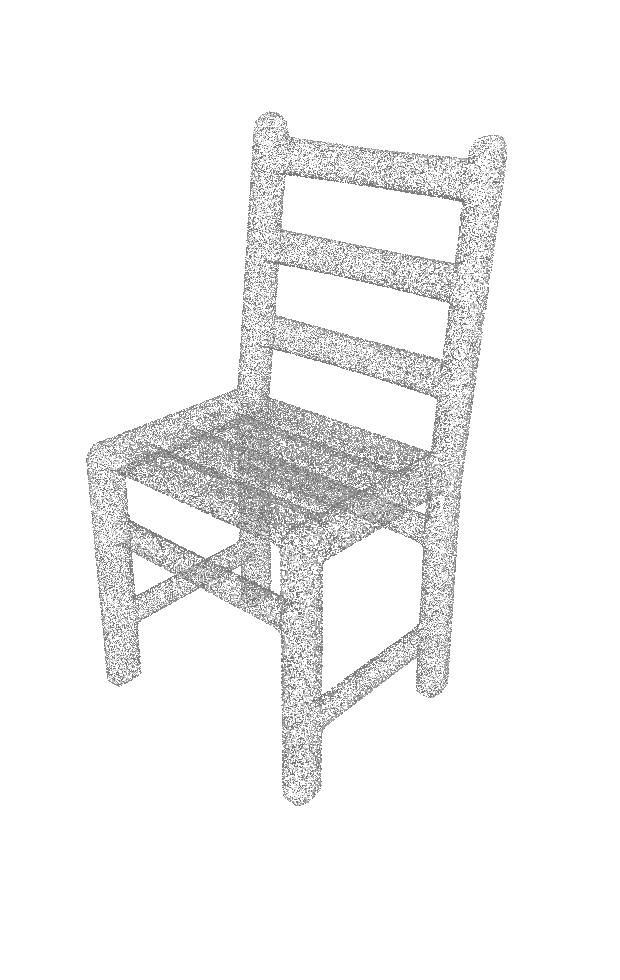}  \\[12pt]
		\multirow{2}{*}[40pt]{\rotatebox{90}{\textbf{Lamp}}} &
		\includegraphics[width=\insz\textwidth]{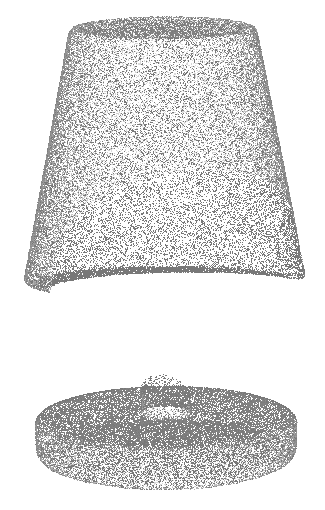} &
		\includegraphics[width=\insz\textwidth]{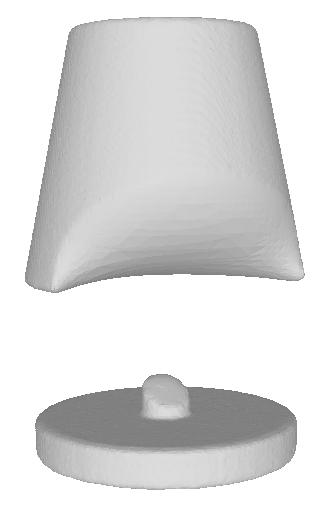} &
		\includegraphics[width=\insz\textwidth]{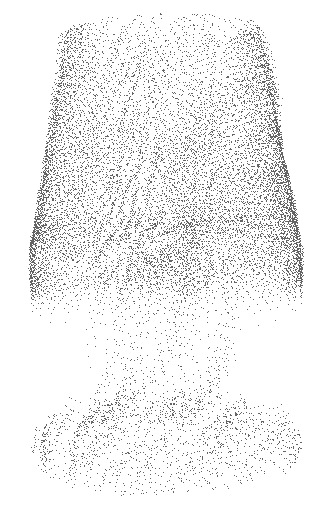} &
		\includegraphics[width=0.07\textwidth]{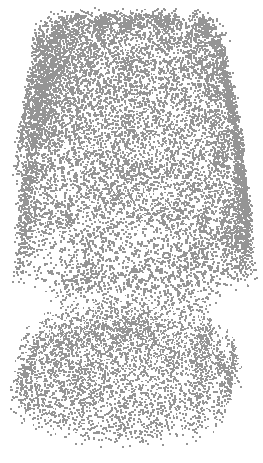} &
		\includegraphics[width=\insz\textwidth]{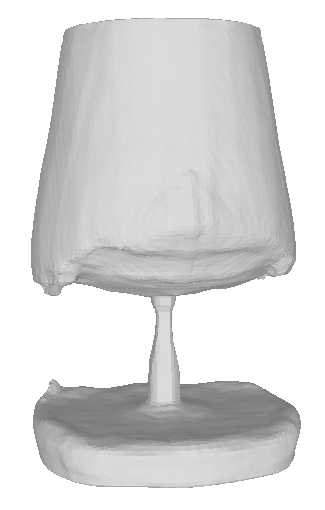} &
		\includegraphics[width=\insz\textwidth]{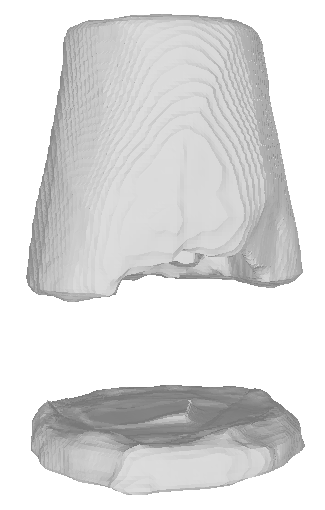} &
		\includegraphics[width=\insz\textwidth]{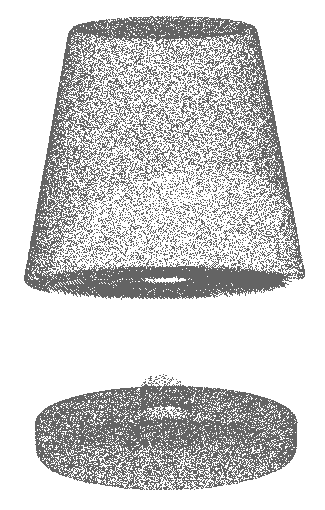} &
		\includegraphics[width=\insz\textwidth]{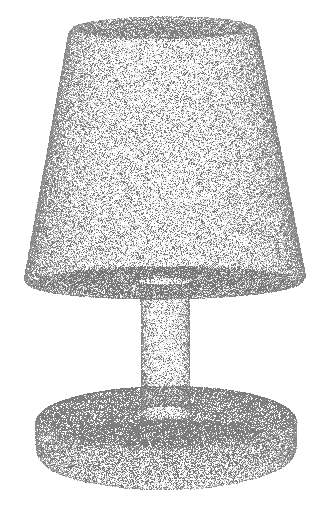} \\[23pt]
		\multirow{2}{*}[40pt]{\rotatebox{90}{\textbf{Table}}} &
		\includegraphics[width=\inszz\textwidth]{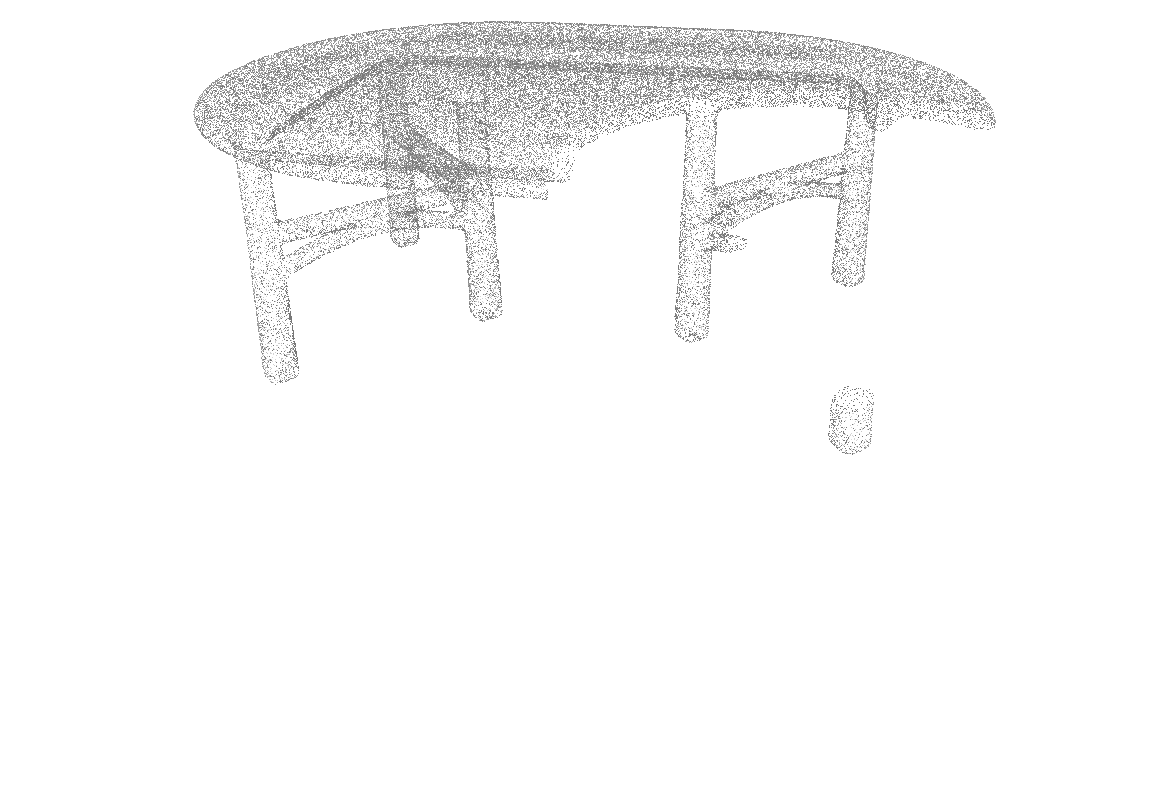} &
		\includegraphics[width=\inszz\textwidth]{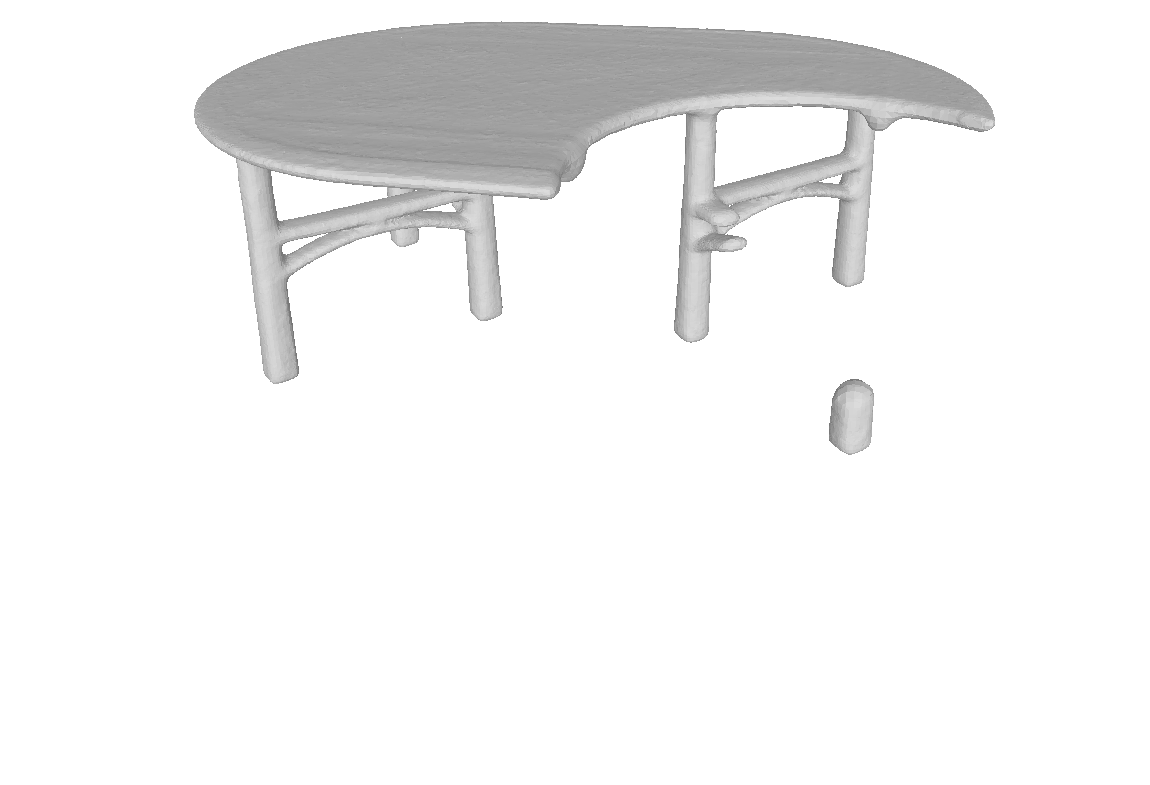} &
		\includegraphics[width=\inszz\textwidth]{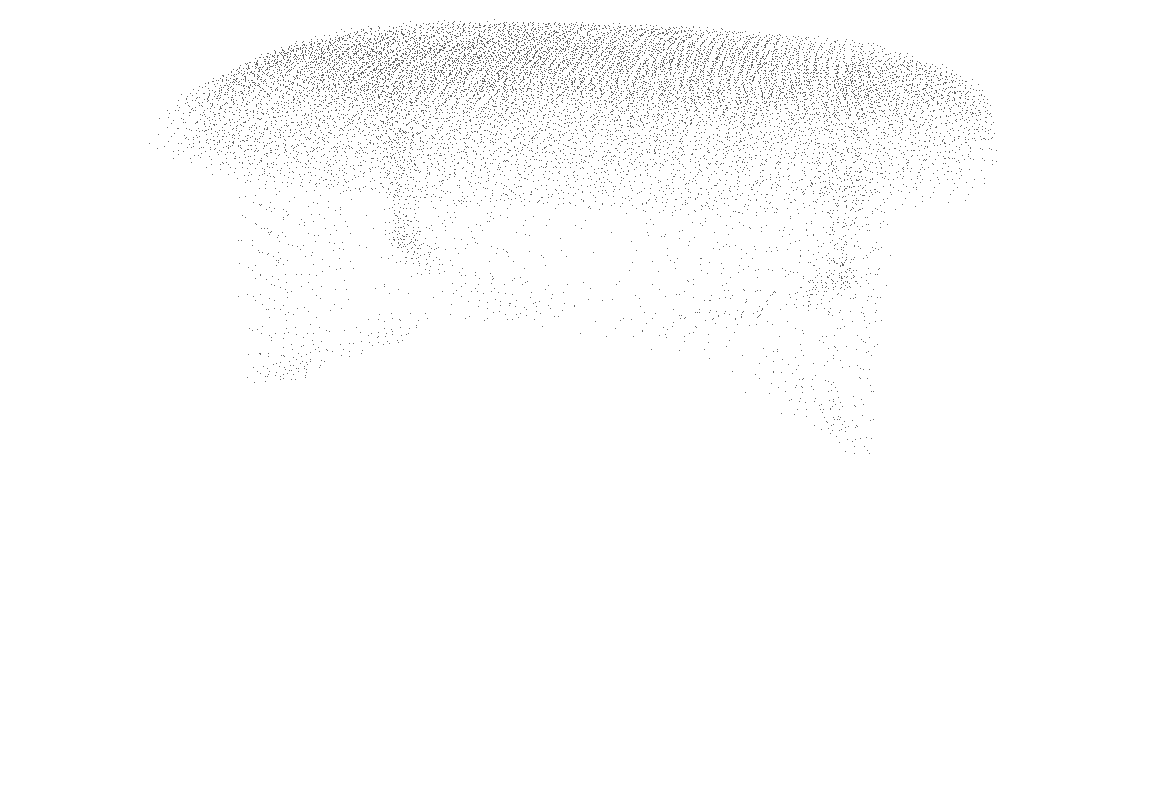} &
		\includegraphics[width=\inszz\textwidth]{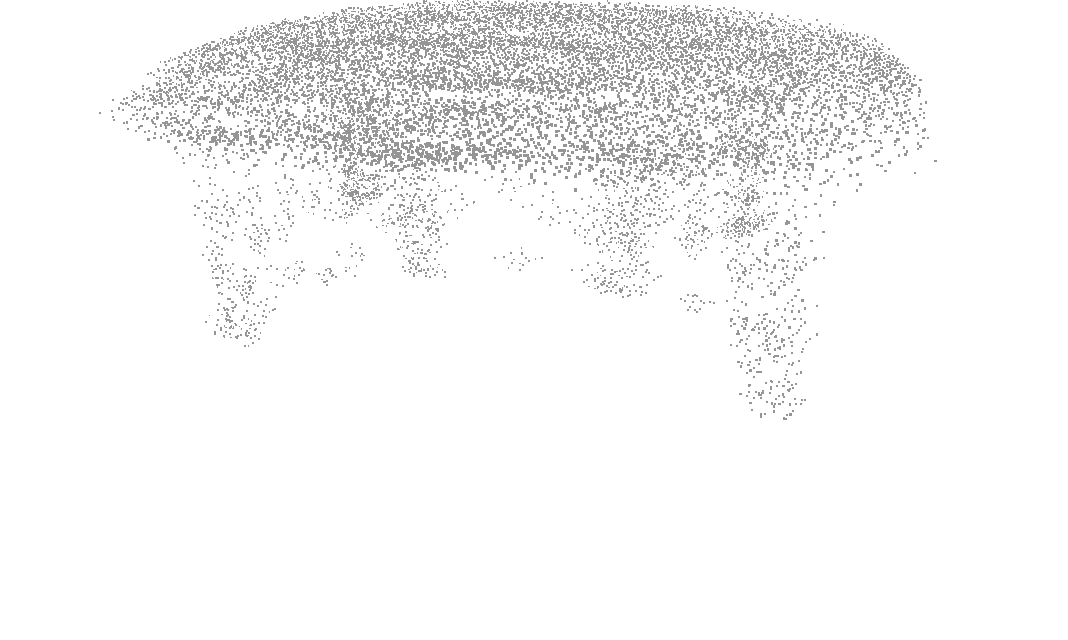} &
		\includegraphics[width=\inszz\textwidth]{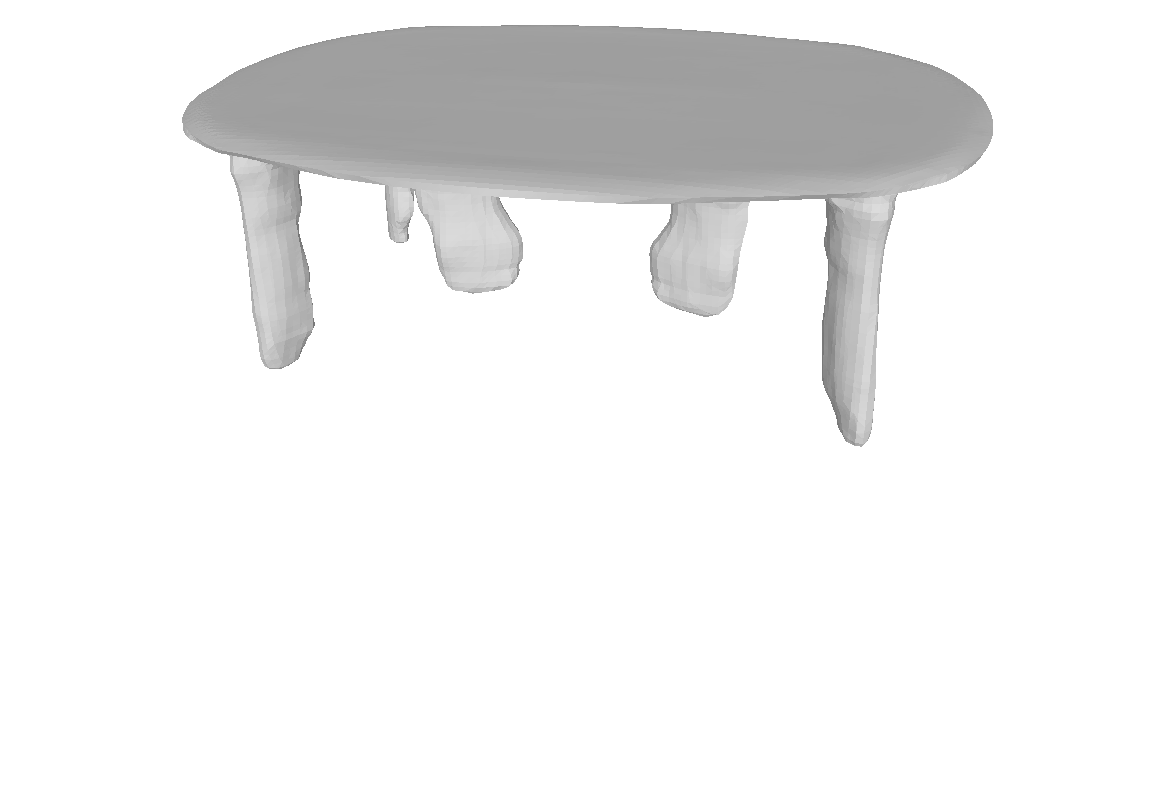} &
		\includegraphics[width=\inszz\textwidth]{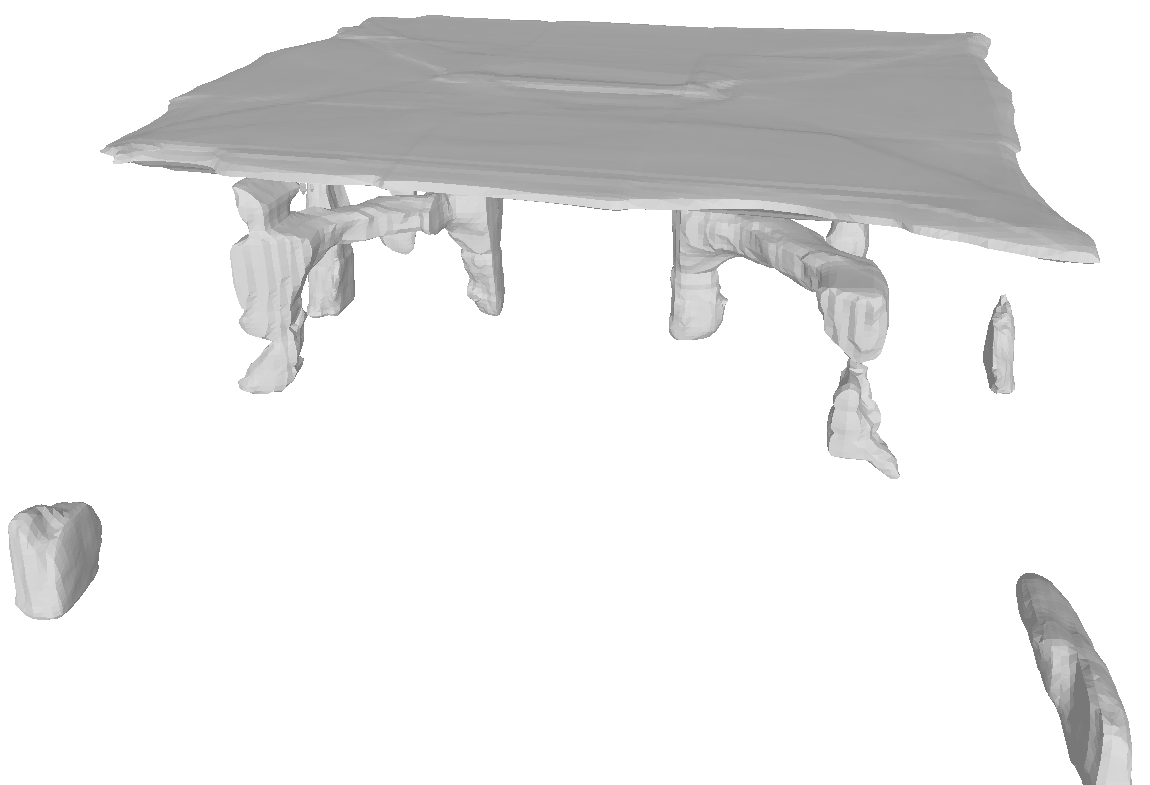} &
		\includegraphics[width=\inszz\textwidth]{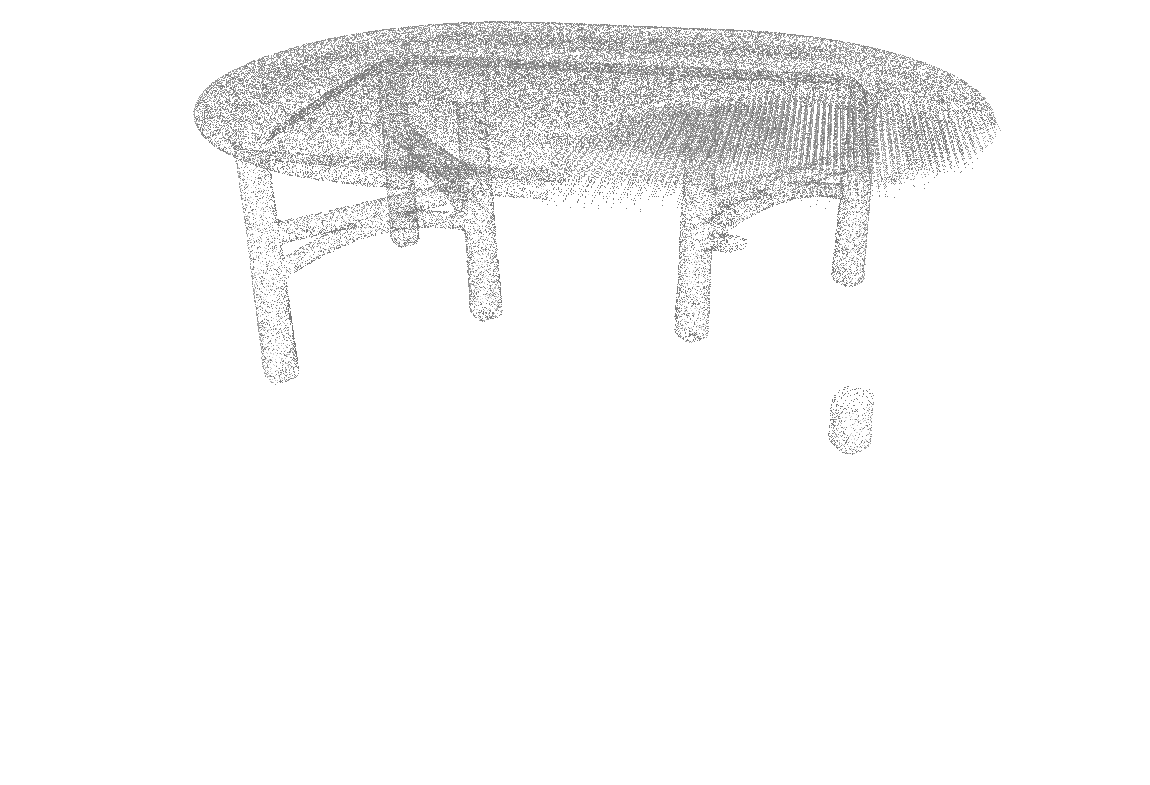} &
		\includegraphics[width=\inszz\textwidth]{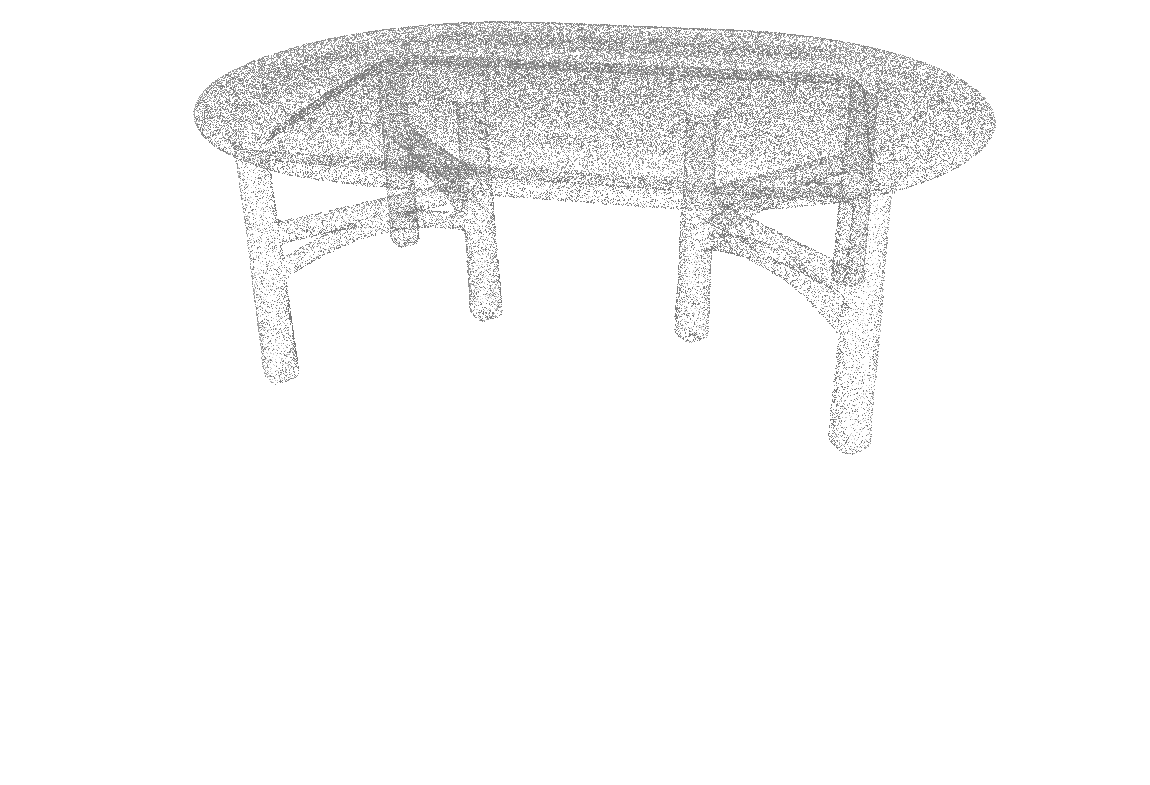} \\[12pt]
	\end{tabular}
	\caption{\textbf{Failure cases.} Examples where the completion
          is incomplete or overly sparse. Note though that, despite
          recovering too few points, \kaplan{} was the only method
          able to reconstruct missing fine structure like the back
          rest of the chair in the second row.}
	\label{fig:supp_failure_modes}
\end{figure*}

\begin{figure*}[h]
	\centering
	\scriptsize
	\setlength{\tabcolsep}{0.1mm}
	\newcommand{\sz}{0.125}
	\newcommand{\insz}{0.09}
	\begin{tabular}{cccccccccc}
		& \textbf{Input} 
		& \textbf{PSR~\cite{Kazhdan-et-al-SGP-2006}} 
		& \textbf{PCN~\cite{Yuan-et-al-3DV-2018}}  
		& \textbf{Cascaded~\cite{Wang-et-al-CVPR-2020}}
		& \textbf{OccNet~\cite{Mescheder-et-al-CVPR-2019}} 
		& \textbf{DeepSDF~\cite{Park-et-al-CVPR-2019}} 
		& \textbf{Ours} 
		& \textbf{GT} 
		\\[-1pt]
		\multirow{2}{*}[40pt]{\rotatebox{90}{\textbf{Plane}}}
		& \includegraphics[width=\sz\textwidth]{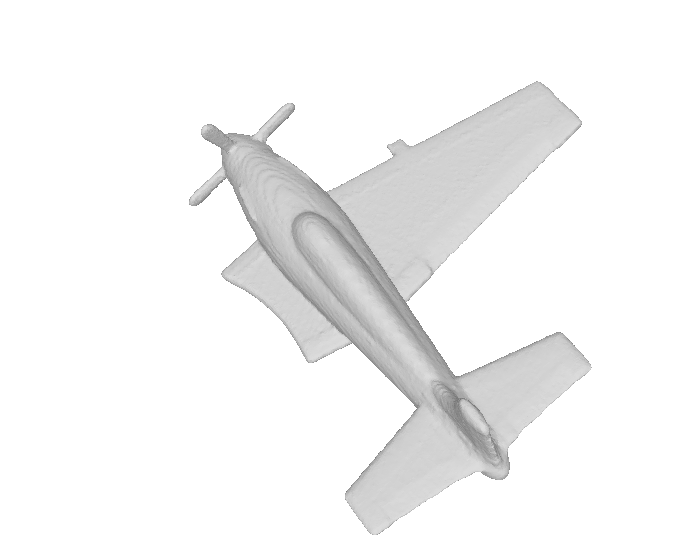} 
		& \includegraphics[width=\sz\textwidth]{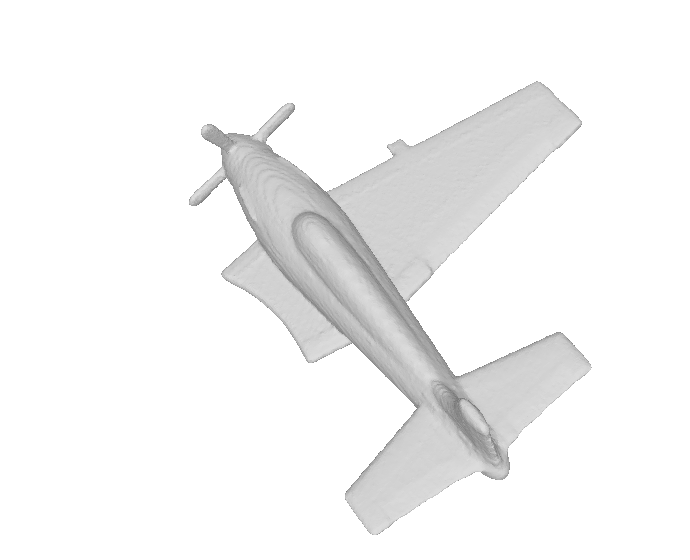} 
		& \includegraphics[width=\sz\textwidth]{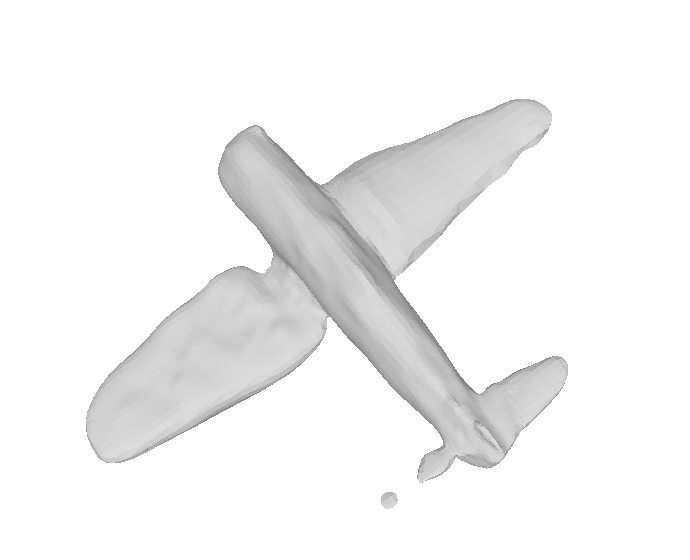} 
		& \includegraphics[width=\sz\textwidth]{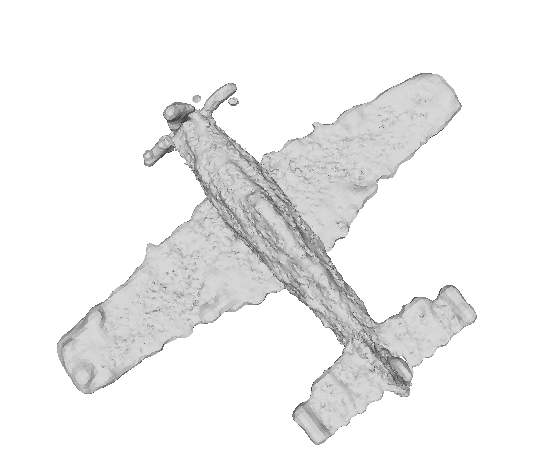} 
		& \includegraphics[width=\sz\textwidth]{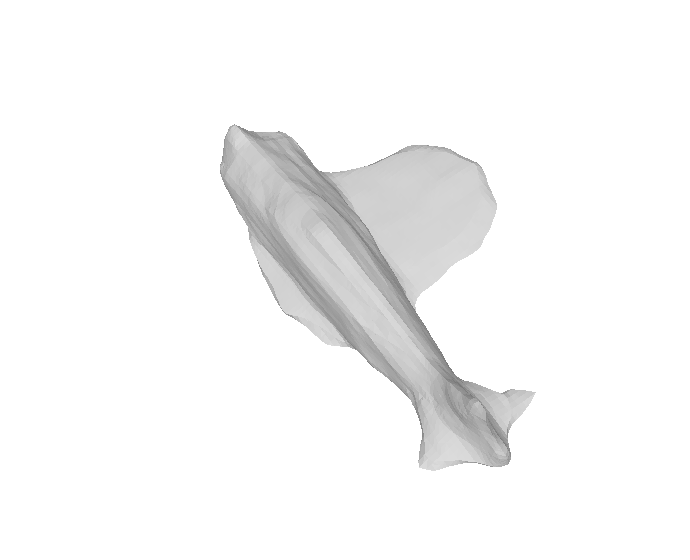} 
		& \includegraphics[width=\sz\textwidth]{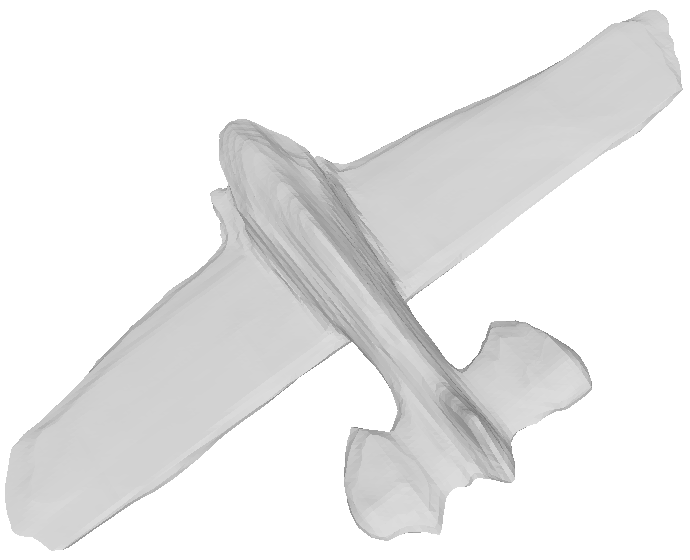} 
		& \includegraphics[width=\sz\textwidth]{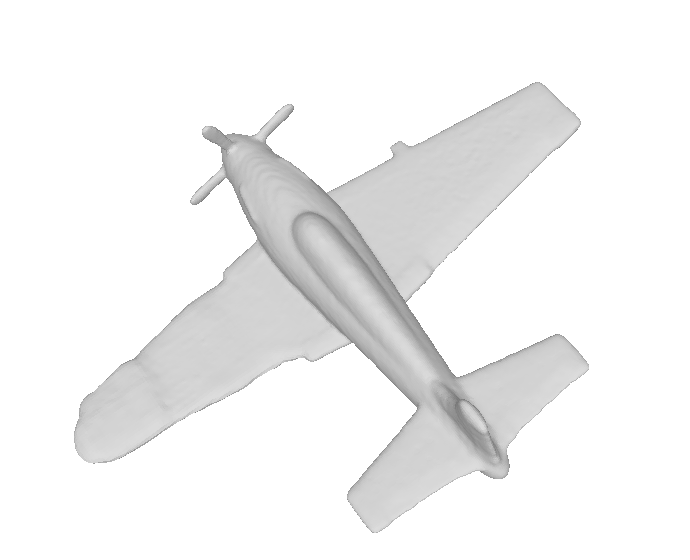} 
		& \includegraphics[width=\sz\textwidth]{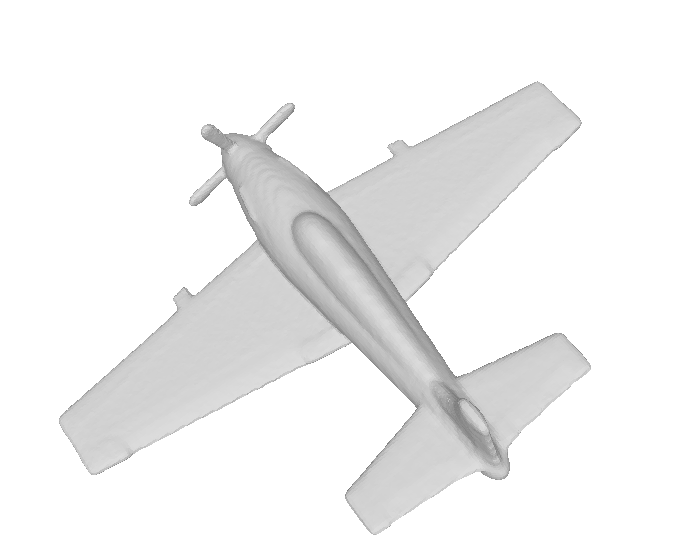} 
		\\[-3pt]
		&
		& $1.69  \;\vert\; 94.51$ 
		& ${\color{highlight2nd} 0.175} \;\vert\; {\color{highlight2nd}82.45}$ 
		& ${\bf 0.151}  \;\vert\; 85.45$ 
		& $1.46  \;\vert\; 78.87$ 
		& $24.5  \;\vert\; 51.41$ 
		& $ 0.219 \;\vert\; {\bf 96.40}$
		\\[3pt]
		\hdashline\\[-6pt]
		\multirow{2}{*}[40pt]{\rotatebox{90}{\textbf{Chair}}} 
		& \includegraphics[width=\insz\textwidth]{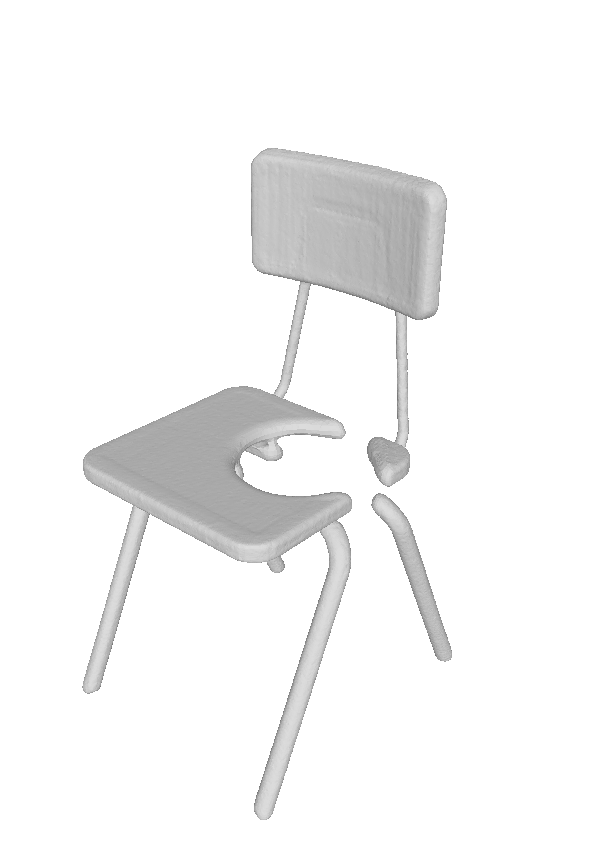} 
		& \includegraphics[width=\insz\textwidth]{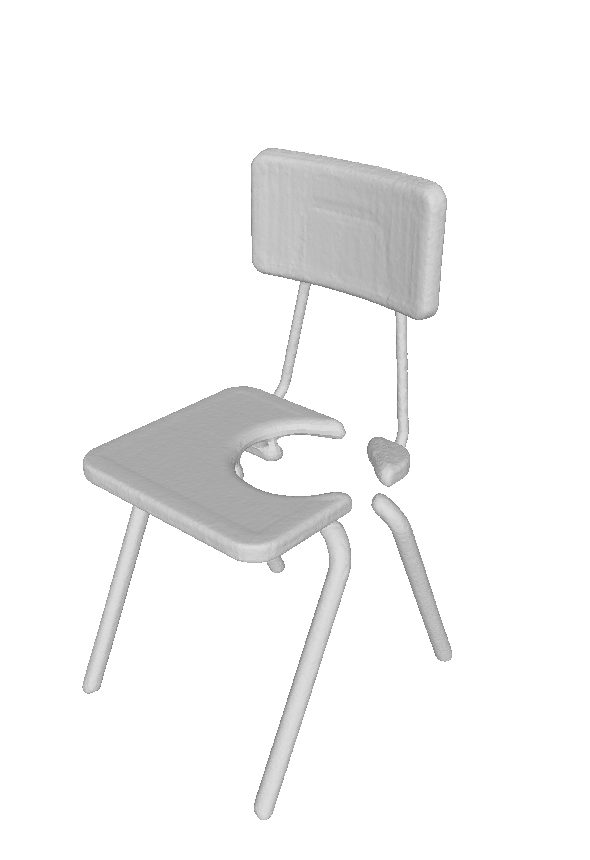} 
		& \includegraphics[width=\insz\textwidth]{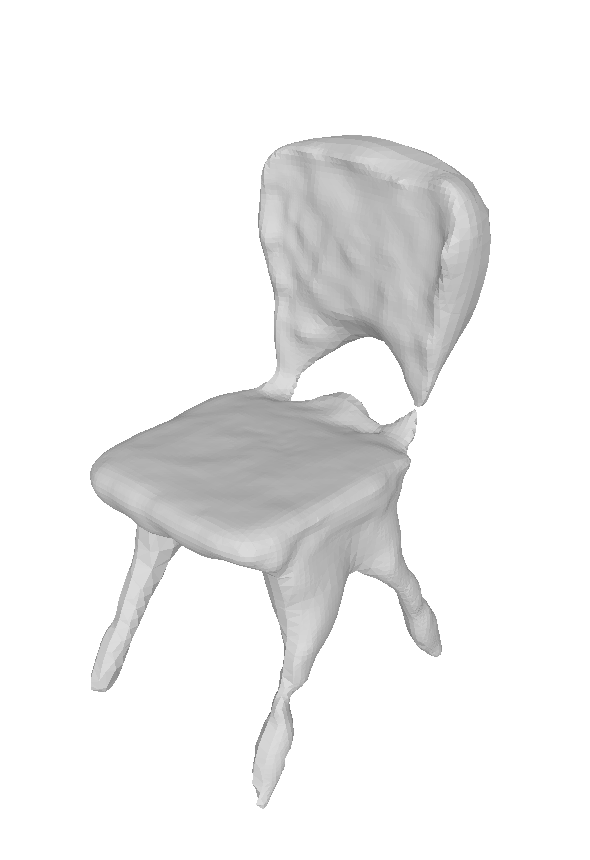} 
		& \includegraphics[width=\insz\textwidth]{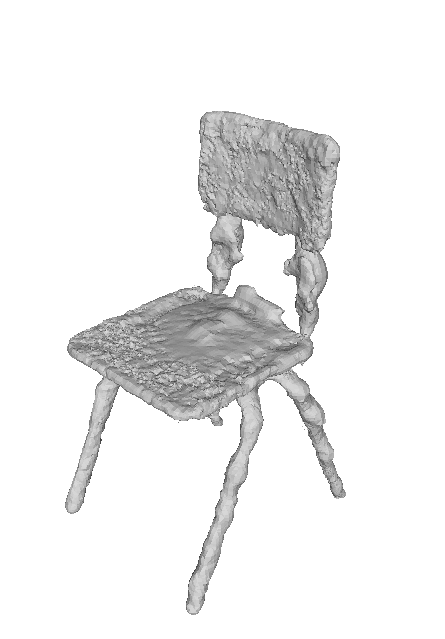} 
		& \includegraphics[width=\insz\textwidth]{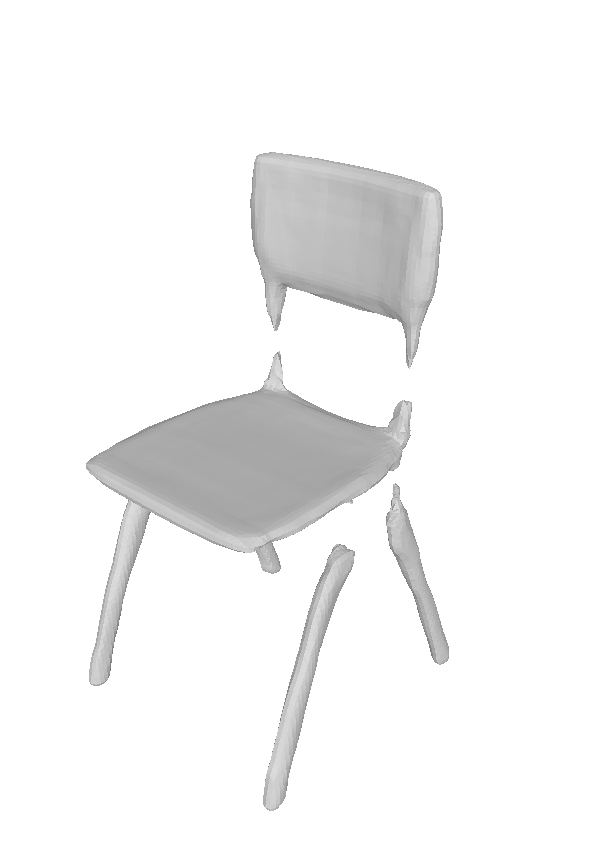} 
		& \includegraphics[width=\insz\textwidth]{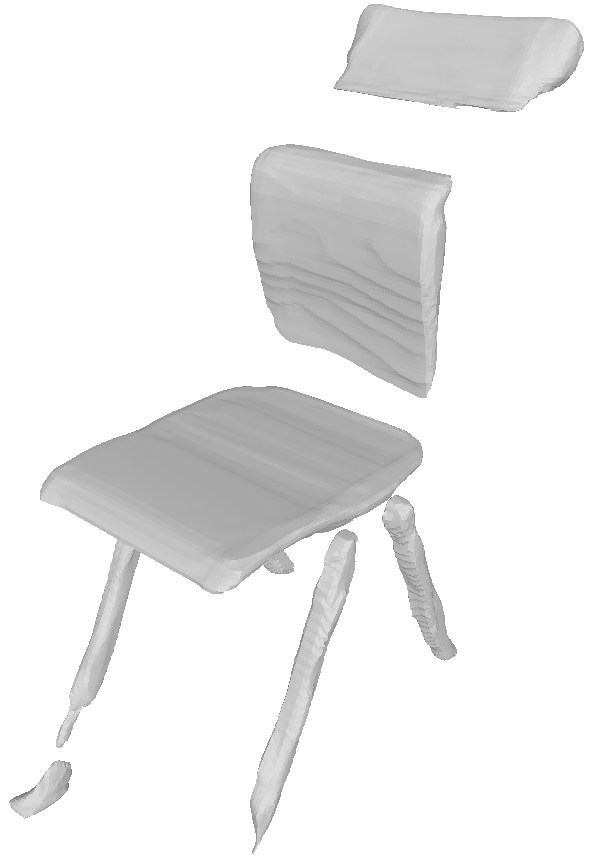} 
		& \includegraphics[width=\insz\textwidth]{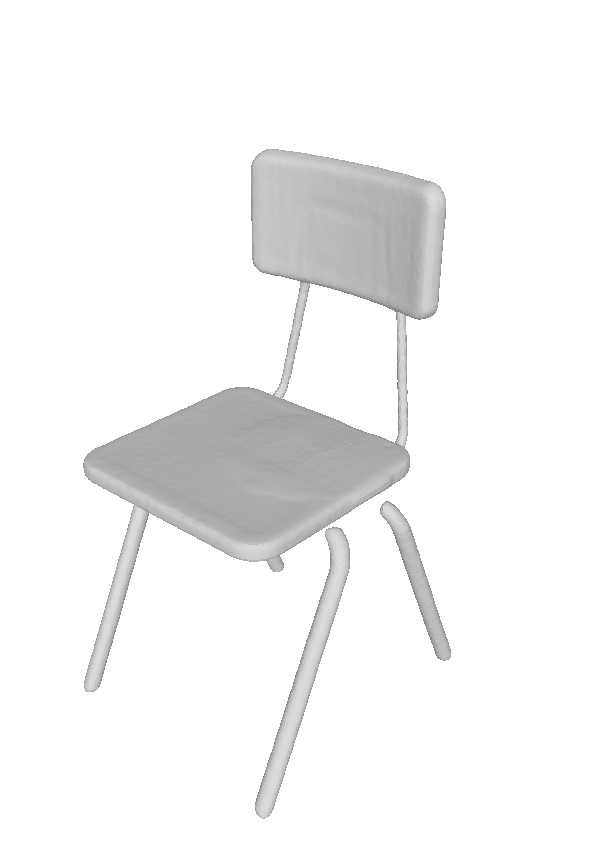} 
		& \includegraphics[width=\insz\textwidth]{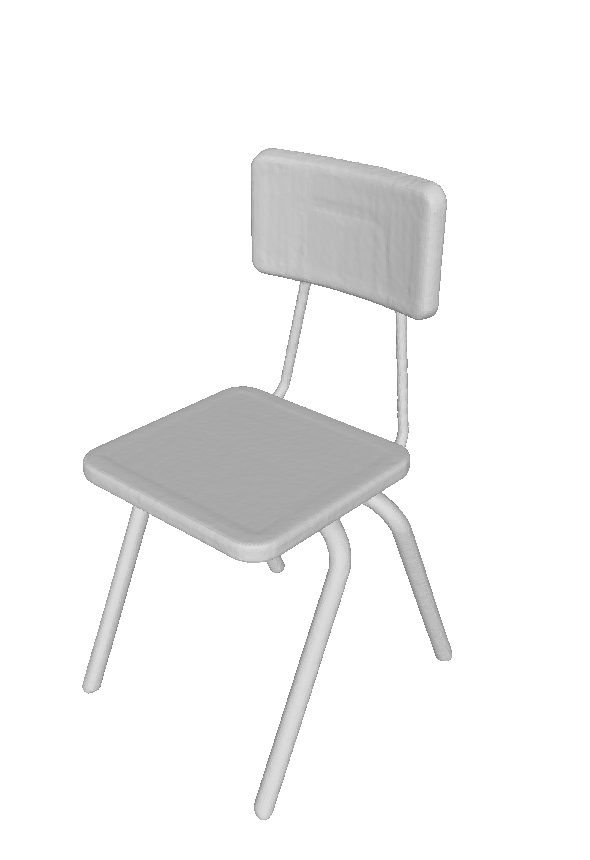} 
		\\[-3pt]
		&   
		& $2.45  \;\vert\; {\color{highlight2nd} 93.87}$ 
		& ${\bf 0.783} \;\vert\; 54.69$ 
		& ${\color{highlight2nd}0.815}  \;\vert\; 53.26$ 
		& $1.20  \;\vert\; 70.77$ 
		& $13.9  \;\vert\; 55.75$ 
		& $1.32  \;\vert\; {\bf 95.02}$
		\\[3pt]
		\hdashline\\[-6pt]
		\multirow{2}{*}[40pt]{\rotatebox{90}{\textbf{Lamp}}} 
		& \includegraphics[width=\sz\textwidth]{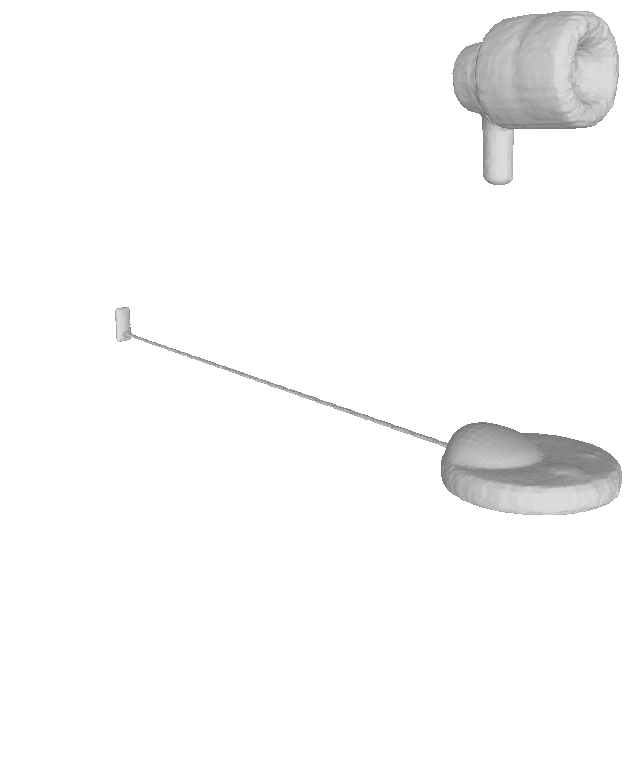} 
		& \includegraphics[width=\sz\textwidth]{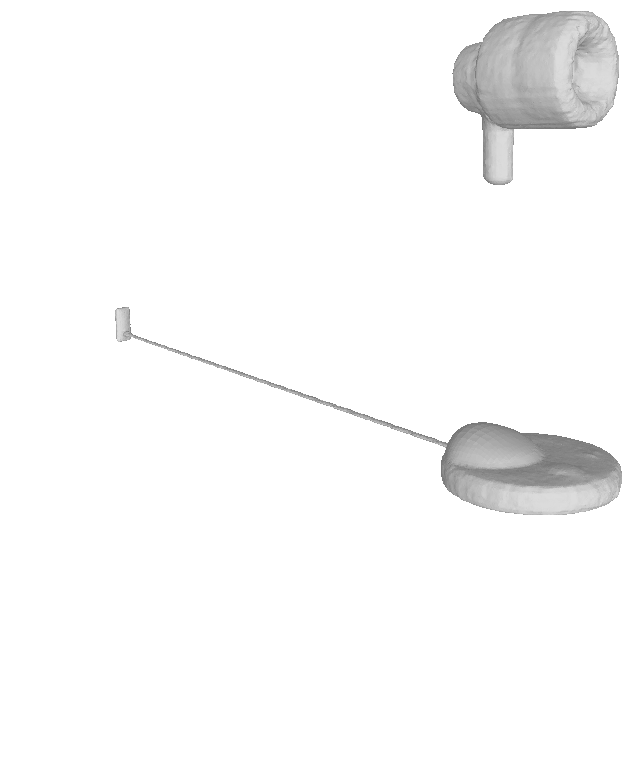} 
		& \includegraphics[width=\sz\textwidth]{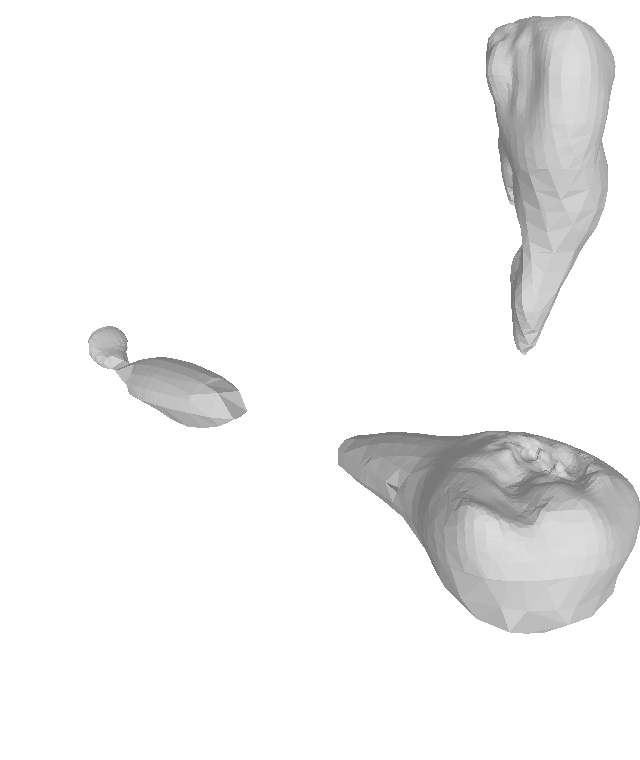} 
		& \includegraphics[width=0.135\textwidth]{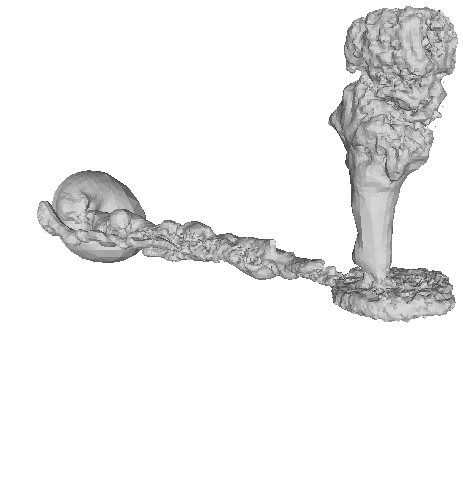}
		& \includegraphics[width=\sz\textwidth]{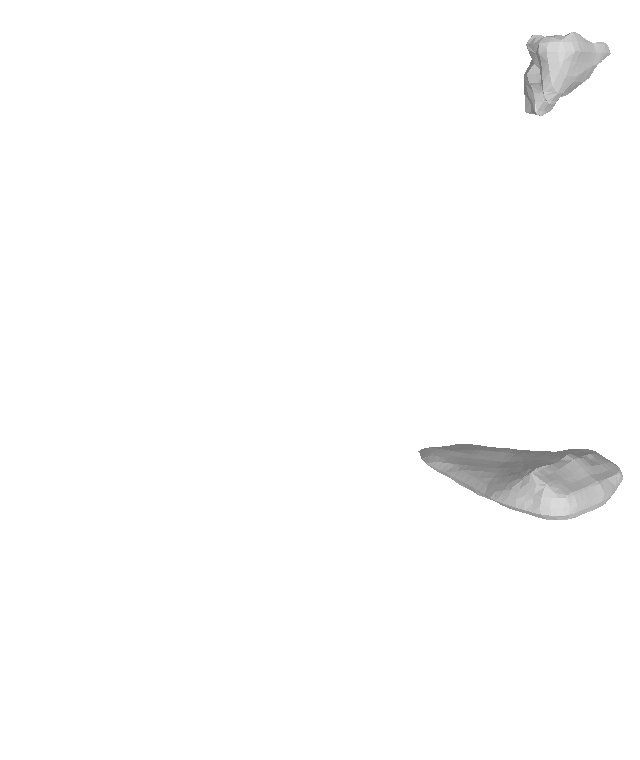} 
		& \includegraphics[width=\sz\textwidth]{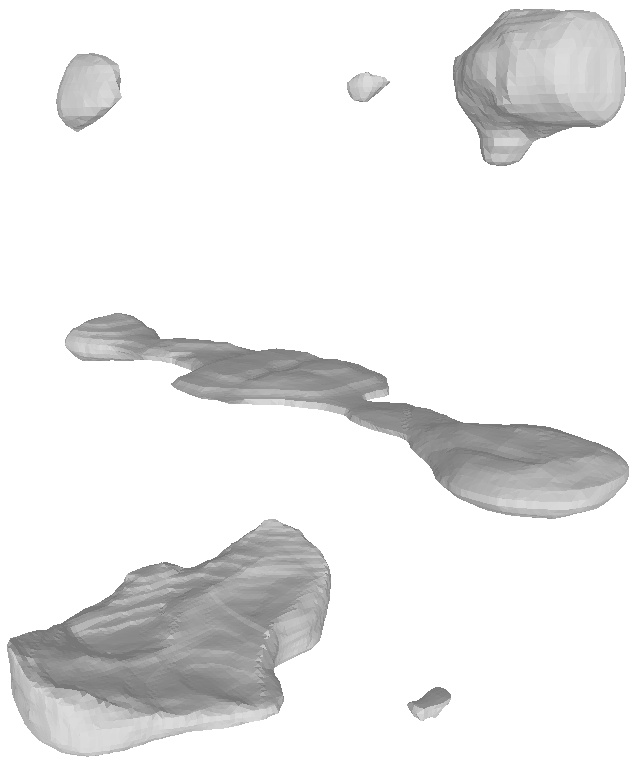}
		& \includegraphics[width=\sz\textwidth]{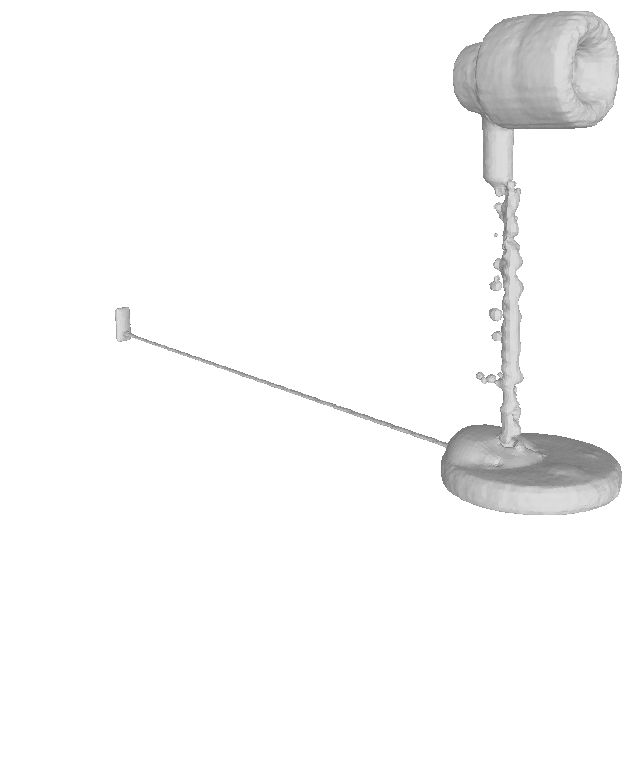} 
		& \includegraphics[width=\sz\textwidth]{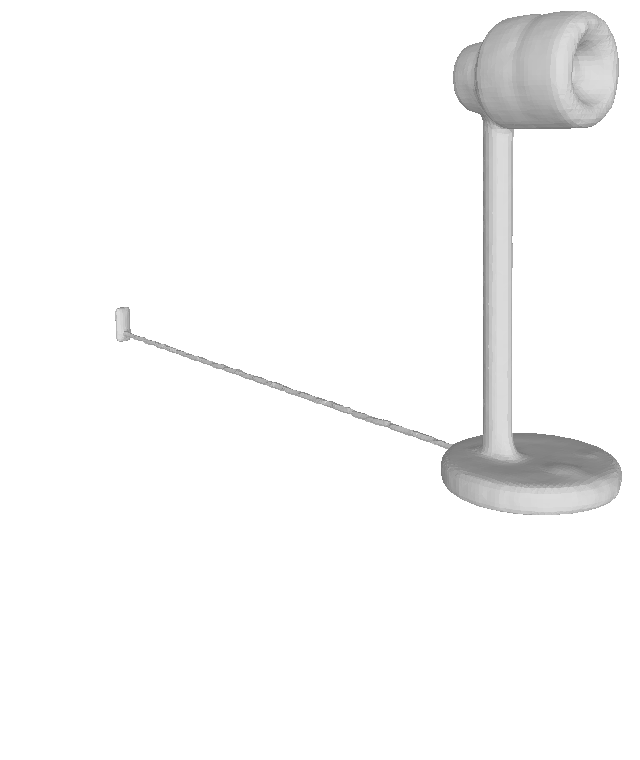}  
		\\[-1pt]
		&  
		& $2.57  \;\vert\; {\bf 94.53}$ 
		& $2.98  \;\vert\; 56.68$ 
		& ${\bf 1.91}  \;\vert\; 58.61$ 
		& $3.51  \;\vert\; 58.51$ 
		& $47.3  \;\vert\; 47.37$ 
		& ${\color{highlight2nd}2.40}  \;\vert\; {\color{highlight2nd}93.80}$ 
		\\[3pt]
		\hdashline\\[-4pt]
		\multirow{2}{*}[26pt]{\rotatebox{90}{\textbf{Sofa}}} 
		& \includegraphics[width=\sz\textwidth]{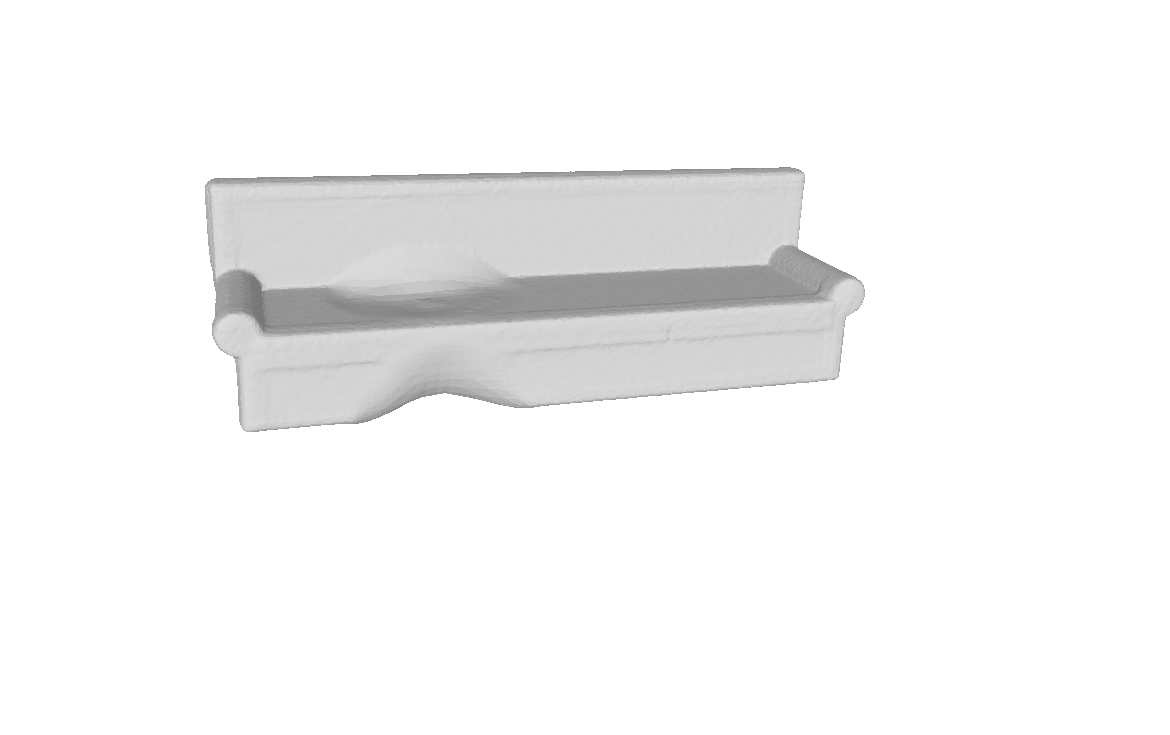} 
		& \includegraphics[width=\sz\textwidth]{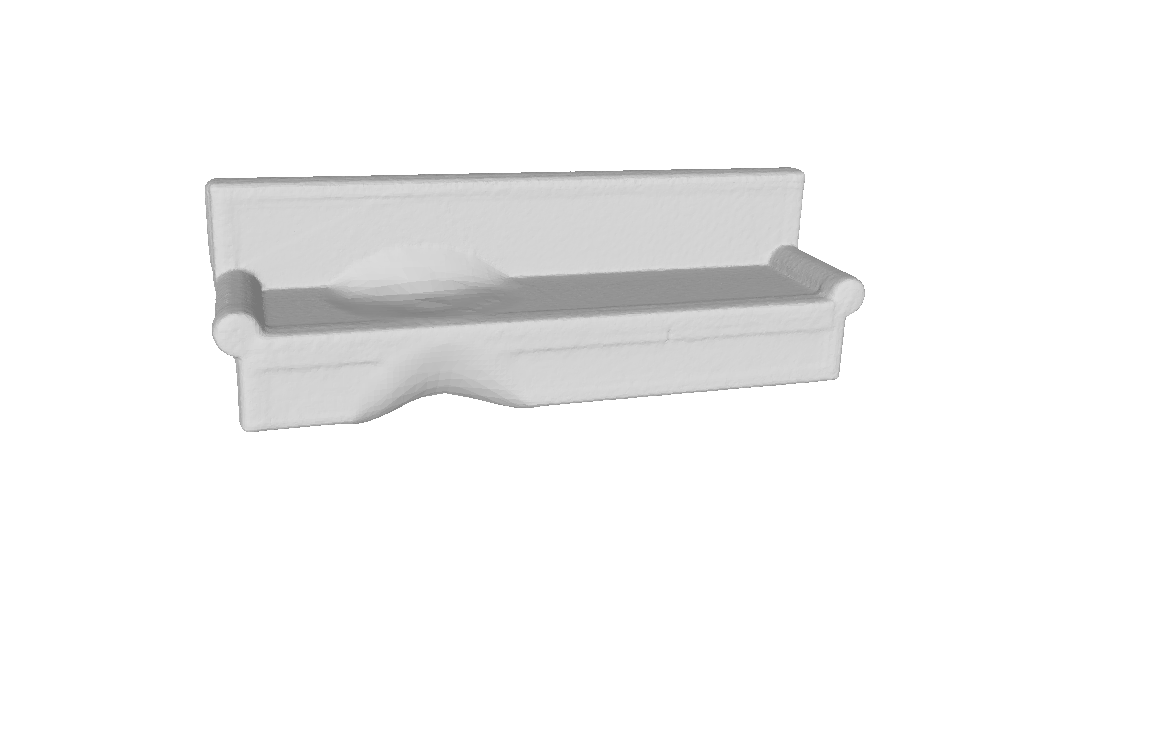} 
		& \includegraphics[width=\sz\textwidth]{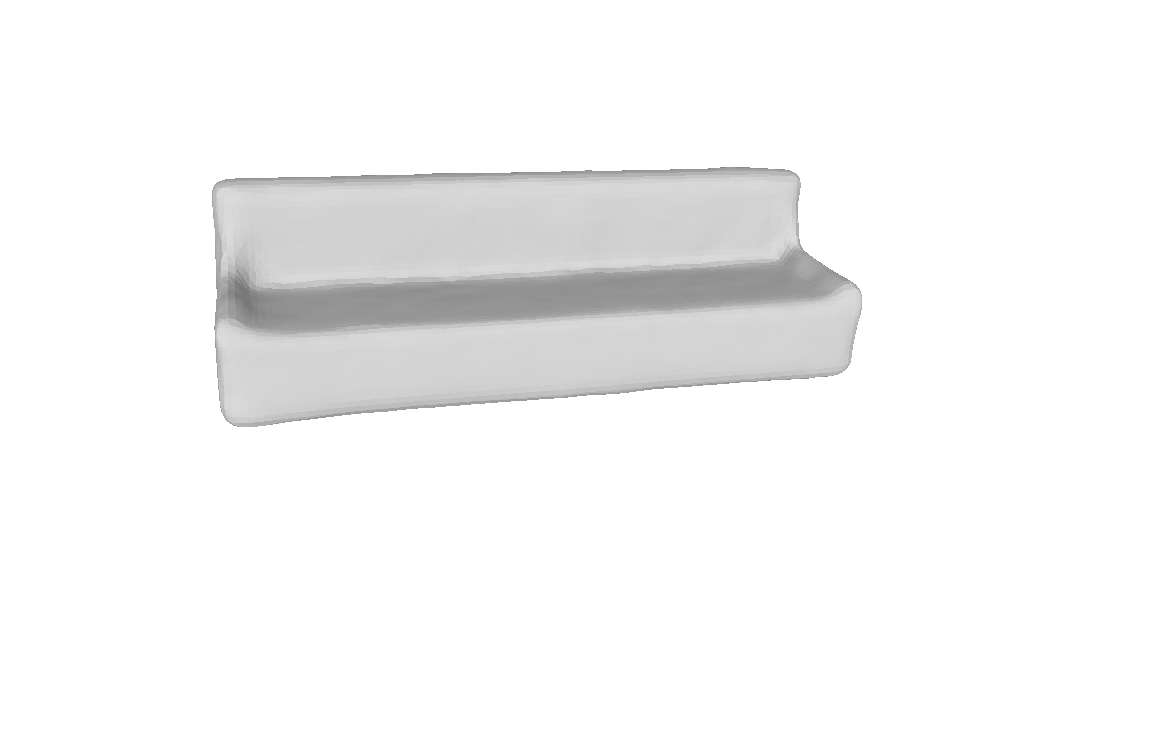} 
		& \includegraphics[width=\sz\textwidth]{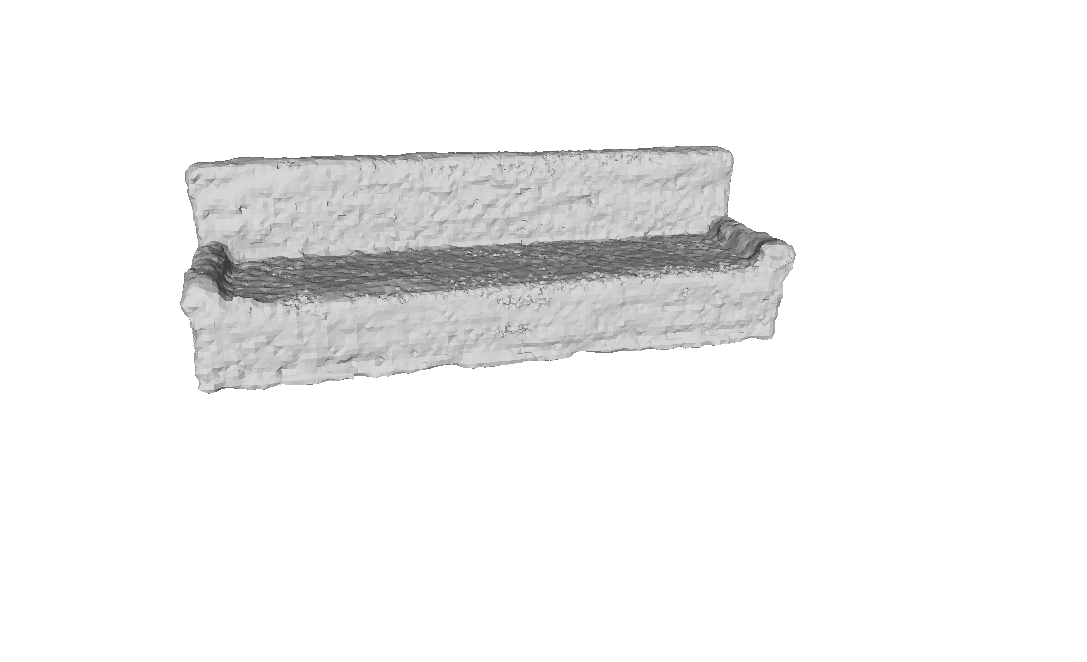} 
		& \includegraphics[width=\sz\textwidth]{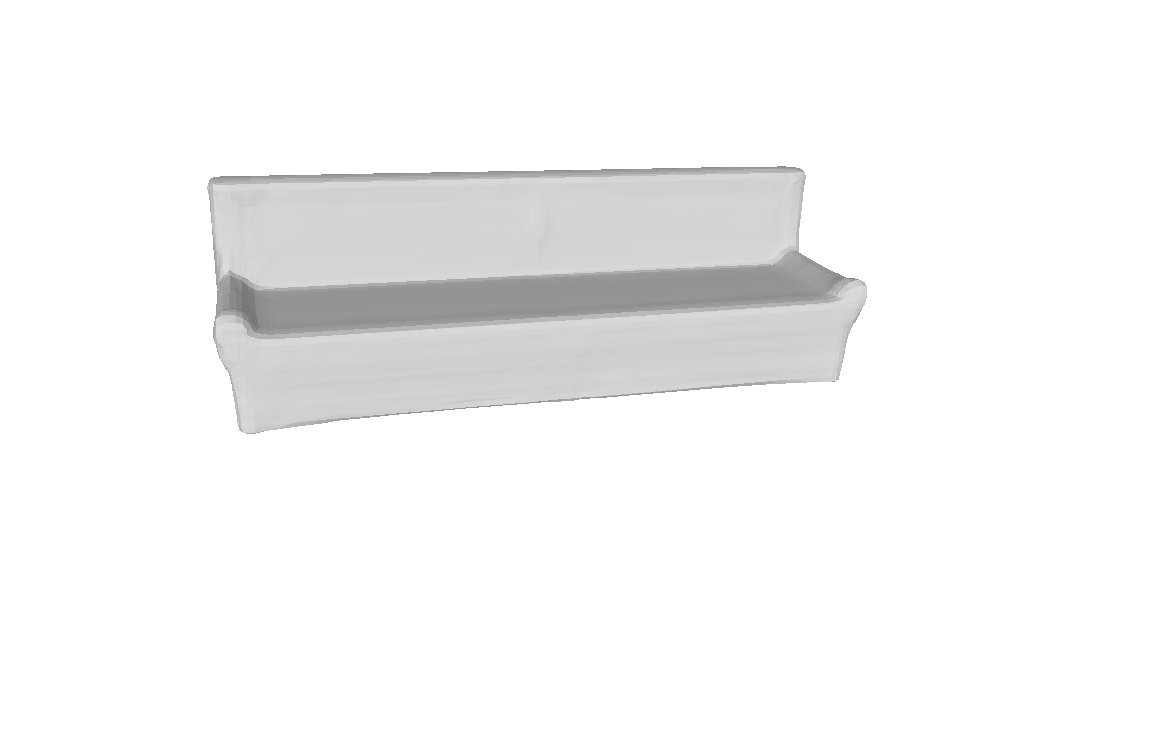} 
		& \includegraphics[width=\sz\textwidth]{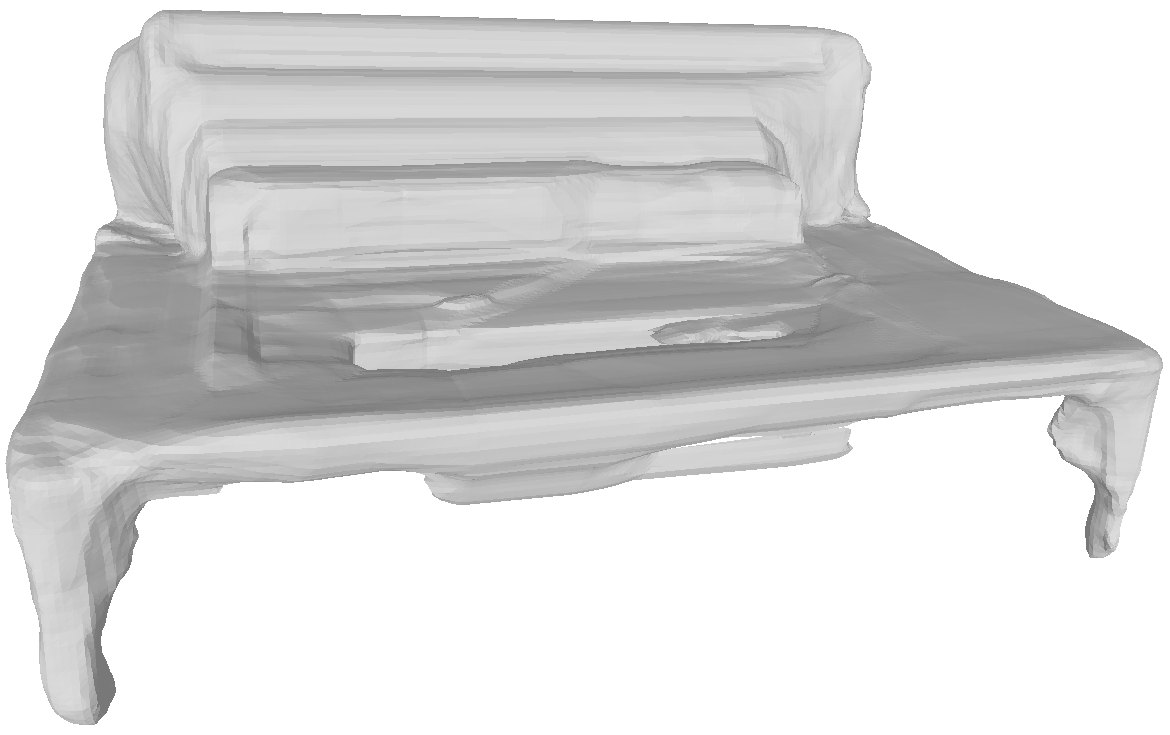} 
		& \includegraphics[width=\sz\textwidth]{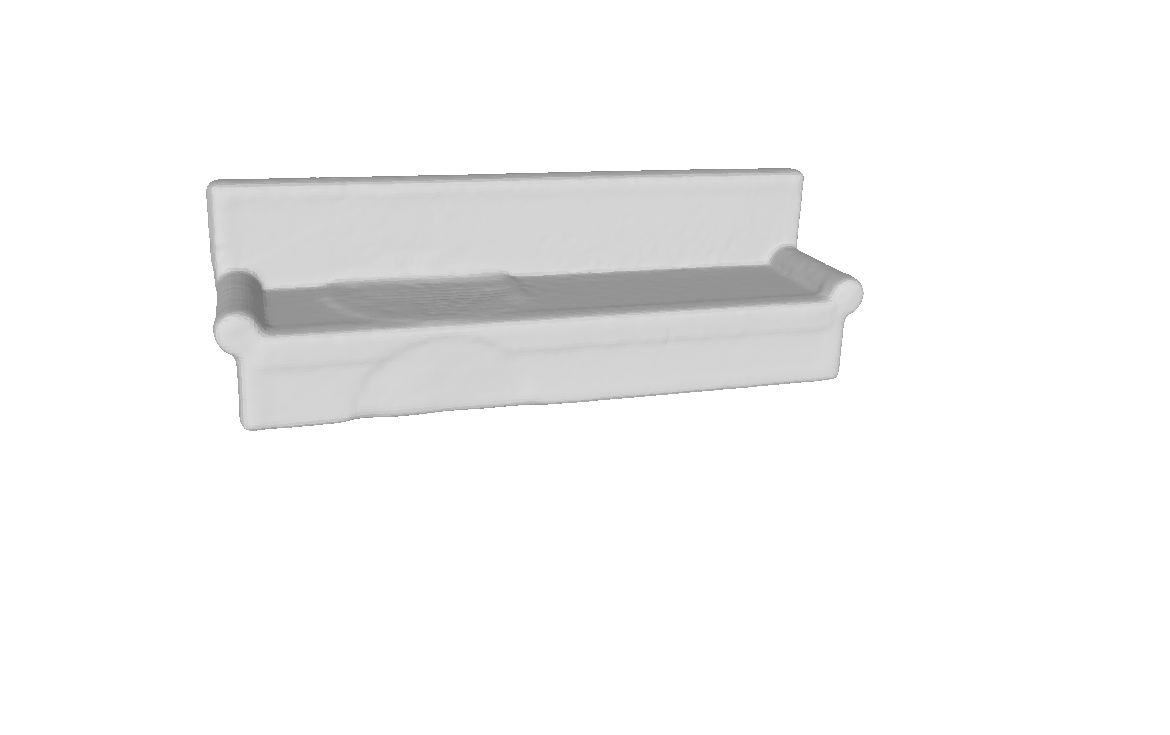} 
		& \includegraphics[width=\sz\textwidth]{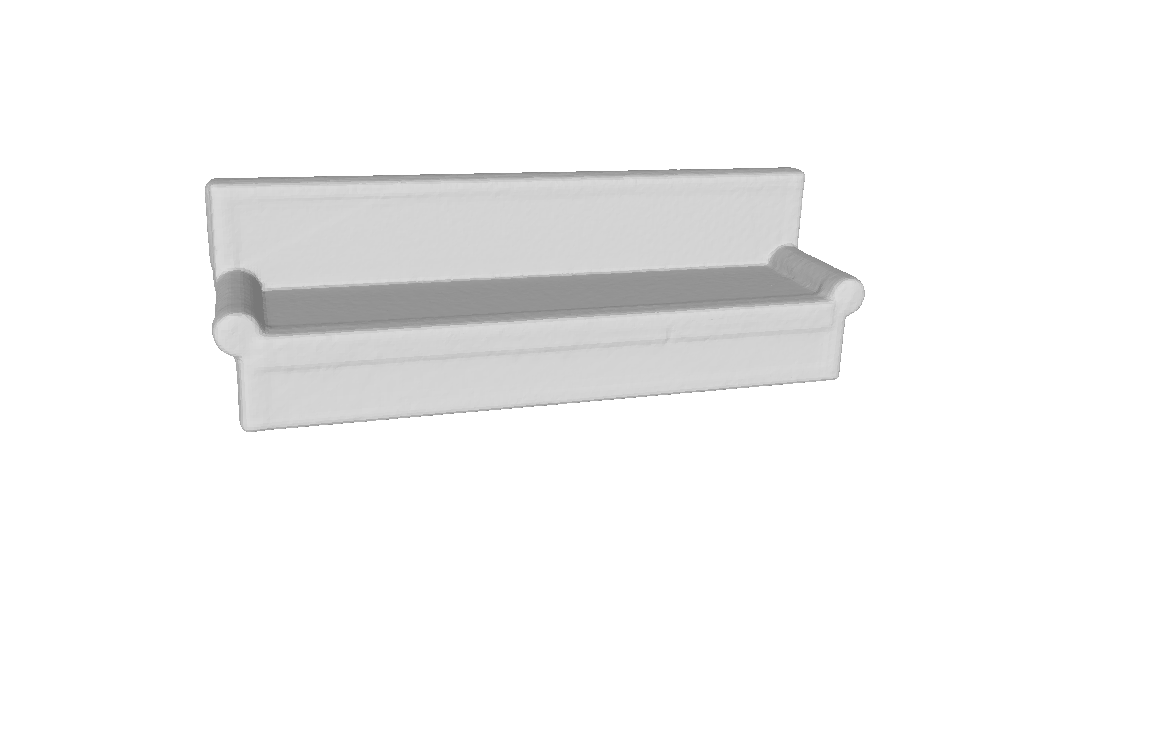} 
		\\[-6pt]
		&  
		& $1.37   \;\vert\; {\color{highlight2nd}94.19}$ 
		& ${\color{highlight2nd}0.503}  \;\vert\; 66.76$ 
		& $0.592   \;\vert\; 62.22$ 
		& $0.993  \;\vert\; 78.53$ 
		& $34.3   \;\vert\; 36.55$ 
		& ${\bf 0.274}  \;\vert\; {\bf 96.16}$
		\\[3pt]
		\hdashline\\[-4pt]
		\multirow{2}{*}[27pt]{\rotatebox{90}{\textbf{Table}}} 
		& \includegraphics[width=\sz\textwidth]{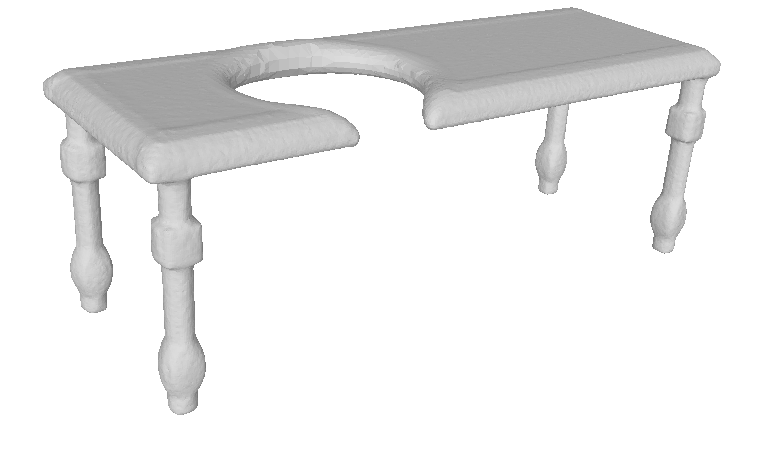} 
		& \includegraphics[width=\sz\textwidth]{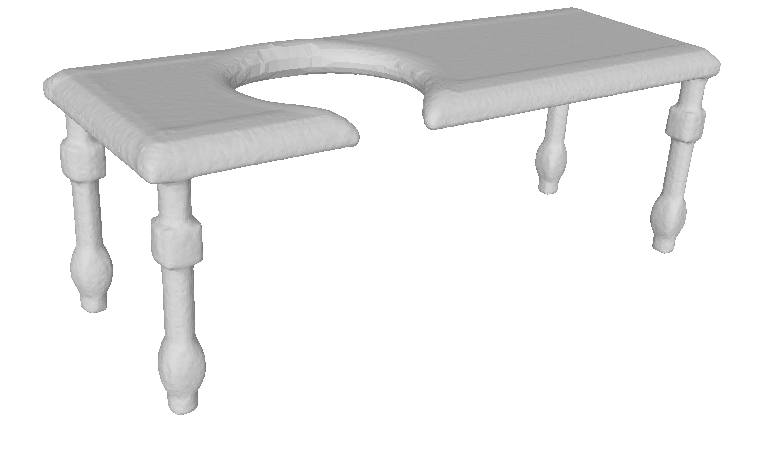} 
		& \includegraphics[width=\sz\textwidth]{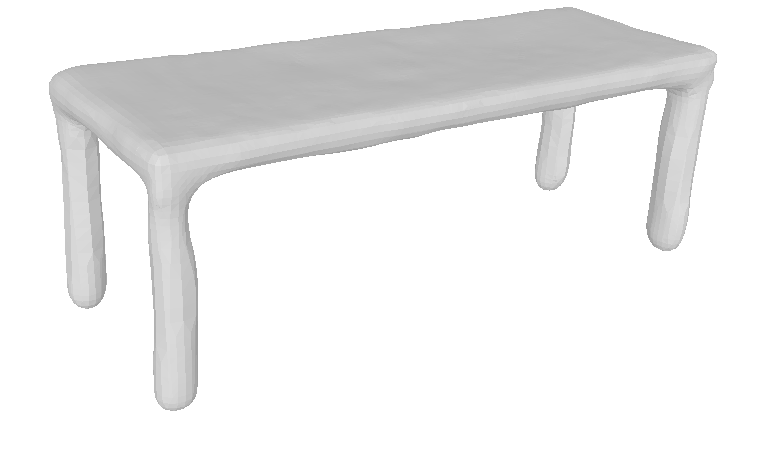} 
		& \includegraphics[width=\sz\textwidth]{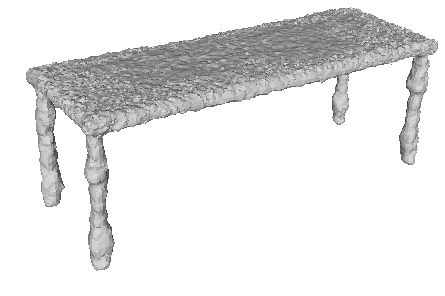} 
		& \includegraphics[width=\sz\textwidth]{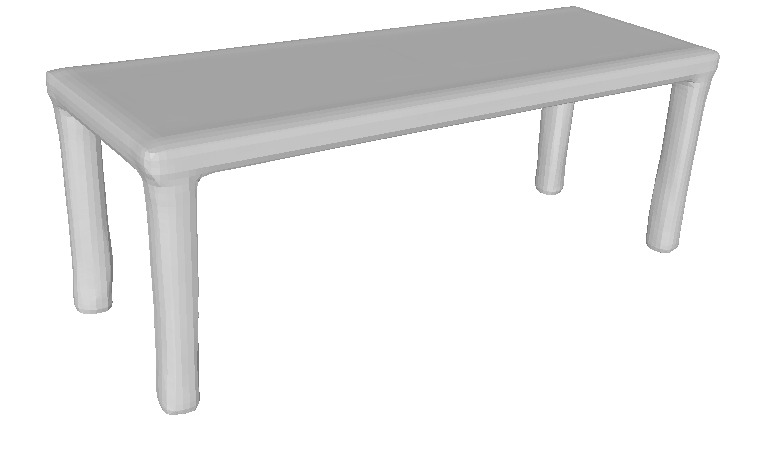} 
		& \includegraphics[width=\sz\textwidth]{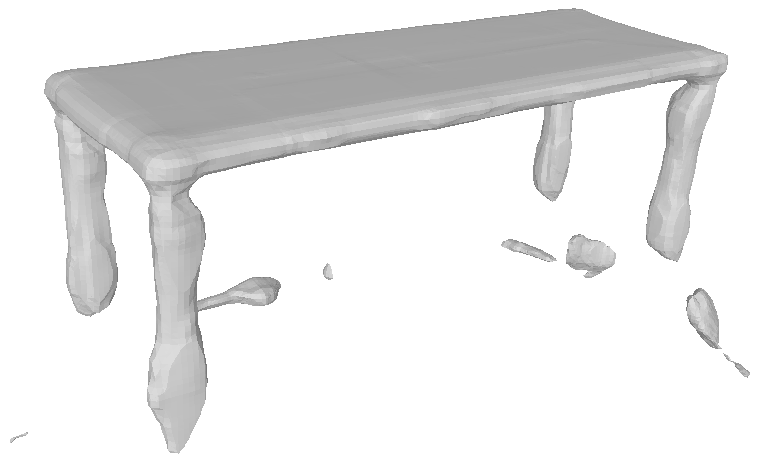} 
		& \includegraphics[width=\sz\textwidth]{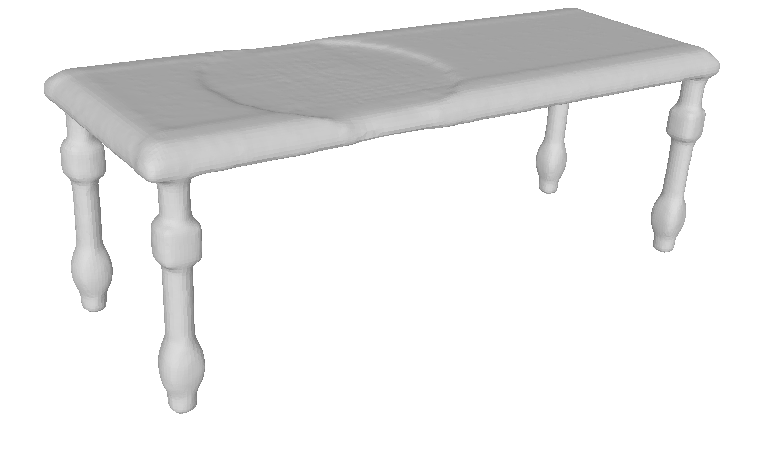} 
		& \includegraphics[width=\sz\textwidth]{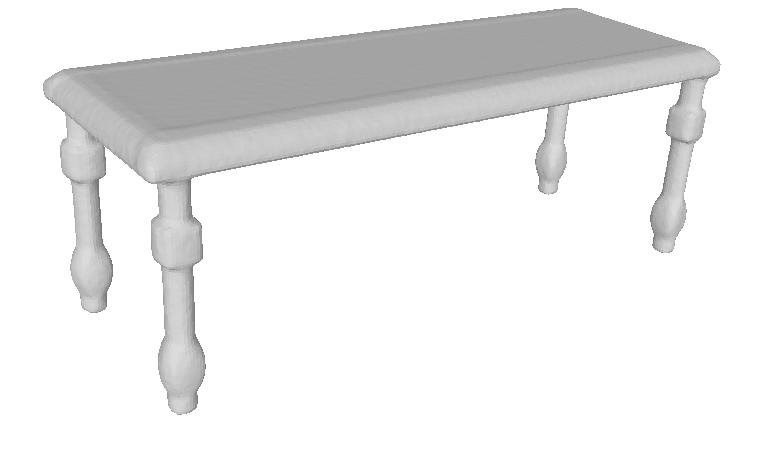} 
		\\[-1pt]
		&  
		& $3.16   \;\vert\; {\color{highlight2nd}93.64}$ 
		& ${\bf 0.763}  \;\vert\; 56.48$ 
		& ${\color{highlight2nd}0.851}  \;\vert\; 52.61$ 
		& $1.33   \;\vert\; 78.79$ 
		& $27.7   \;\vert\; 59.87$ 
		& $0.976  \;\vert\; {\bf 96.35}$
		\\[-6pt] 
	\end{tabular}
	\caption{\textbf{Qualitative and quantitative comparison to state-of-the-art scene completion methods (meshed version).} We repeat the Fig.~\ref{fig:qualitative_comparison} after meshing all the point clouds using Poisson Surface Reconstruction. The values below the pictures reproduce results from Fig.~\ref{fig:f1_chamfer_graphs} shown as $10^3\cdot CD \;\vert\; F1$. Best values are \textbf{bold}, 2$^\text{nd}$ best ones are \textcolor{Cerulean}{blue}.
	As explained in Sec.~\ref{sec:exp_results}, the DeepSDF results differ from those presented in their paper, since we did not have access to the original code for shape completion.}
	\label{fig:qualitative_comparison_mesh}
\end{figure*}

\boldparagraph{Meshed results.}
In Fig.~\ref{fig:qualitative_comparison_mesh}, we reproduce the Fig.~\ref{fig:qualitative_comparison} where we present a meshed version obtained by applying Poisson Surface Reconstruction (PSR) on the point clouds. 
Our network is able to reliably predict the normals of the points, unlike PCN that requires to apply an additional normal estimator. 
This leads to a better final mesh in our case. 

\subsection{Failure Modes}
\kaplan{} exhibits two main types of failures, shown in Fig.~\ref{fig:supp_failure_modes}.
The first type corresponds to situations where the reconstructed geometry is undersampled and too sparse.
In the two top rows, the missing region was correctly detected, but not appropriately filled with enough points.
This situation could likely be solved by sampling more query points at finer levels, or by increasing the resolution of descriptor planes.

The second type of failures happens when two or more parts of the object are incomplete, and only one of them is completed correctly. For example, the lamp shade, respectively the tabletop are nicely recovered in the bottom two rows, whereas the lamp post and the leg of the table are not.
In fact, a weaker version of the same problem appears in the second row, where the seat of the chair is completed much better than the fine structures of the back rest.

We are still investigating this problem, but suspect that too few descriptors at the coarse level are placed such that they support the ``weaker'' part. Once a region is ignored during the initial completion, finer levels can obviously not repair it.

%


{\small
\bibliographystyle{ieee}
\bibliography{bibliography}

\begin{thebibliography}{10}\itemsep=-1pt

\bibitem{Achlioptas-et-al-ICML-2018}
P.~Achlioptas, O.~Diamanti, I.~Mitliagkas, and L.~J. Guibas.
\newblock Learning representations and generative models for 3d point clouds.
\newblock In {\em ICML}, 2018.

\bibitem{Bello-et-al-Arxiv-2020}
S.~A. Bello, S.~Yu, and C.~Wang.
\newblock Review: deep learning on 3d point clouds.
\newblock {\em arXiv preprint 2001.06280}, 2020.

\bibitem{boulch2018snapnet}
A.~Boulch, J.~Guerry, B.~Le~Saux, and N.~Audebert.
\newblock {SnapNet}: 3d point cloud semantic labeling with 2d deep segmentation
  networks.
\newblock {\em Computers \& Graphics}, 71:189--198, 2018.

\bibitem{shapenet2015}
A.~X. Chang, T.~Funkhouser, L.~Guibas, P.~Hanrahan, Q.~Huang, Z.~Li,
  S.~Savarese, M.~Savva, S.~Song, H.~Su, J.~Xiao, L.~Yi, and F.~Yu.
\newblock {ShapeNet}: An information-rich 3d model repository.
\newblock {\em arXiv preprint 1512.03012}, 2015.

\bibitem{Chen-et-al-ICLR-20}
X.~Chen, B.~Chen, and N.~J. Mitra.
\newblock Unpaired point cloud completion on real scans using adversarial
  training.
\newblock In {\em Proceedings of the International Conference on Learning
  Representations (ICLR)}, 2020.

\bibitem{Chen_CVPR_2019}
Z.~Chen and H.~Zhang.
\newblock Learning implicit fields for generative shape modeling.
\newblock In {\em CVPR}, 2019.

\bibitem{Choy-et-al-ECCV16}
C.~Choy, D.~Xu, J.~Gwak, K.~Chen, and S.~Savarese.
\newblock {3D-R2N2}: A unified approach for single and multi-view 3d object
  reconstruction.
\newblock In {\em ECCV}, 2016.

\bibitem{Chui-et-al-CAGD-2000}
C.~K. Chui and M.~Lai.
\newblock Filling polygonal holes using {C1} cubic triangular spline patches.
\newblock {\em Computer Aided Geometric Design}, 17(4):297--307, 2000.

\bibitem{cyr2001_3d}
C.~M. Cyr and B.~B. Kimia.
\newblock 3d object recognition using shape similiarity-based aspect graph.
\newblock In {\em ICCV}, 2001.

\bibitem{Dai-et-al-CVPR-2020}
A.~Dai, C.~Diller, and M.~Niessner.
\newblock Sg-nn: Sparse generative neural networks for self-supervised scene
  completion of rgb-d scans.
\newblock In {\em Proceedings of the IEEE/CVF Conference on Computer Vision and
  Pattern Recognition (CVPR)}, June 2020.

\bibitem{Dai-et-al-CVPR-2017-shape-completion}
A.~Dai, C.~R. Qi, and M.~Nie{\ss}ner.
\newblock Shape completion using 3d-encoder-predictor cnns and shape synthesis.
\newblock In {\em CVPR}, 2017.

\bibitem{Dai-et-al-CVPR-2018}
A.~Dai, D.~Ritchie, M.~Bokeloh, S.~Reed, J.~Sturm, and M.~Nie{\ss}ner.
\newblock {ScanComplete}: Large-scale scene completion and semantic
  segmentation for 3d scans.
\newblock In {\em CVPR}, 2018.

\bibitem{Davis-et-al-3DPVT-2002}
J.~Davis, S.~R. Marschner, M.~Garr, and M.~Levoy.
\newblock Filling holes in complex surfaces using volumetric diffusion.
\newblock In {\em 3DPVT}, 2002.

\bibitem{Graham-et-al-CVPR-2018}
B.~Graham, M.~Engelcke, and L.~van~der Maaten.
\newblock 3d semantic segmentation with submanifold sparse convolutional
  networks.
\newblock In {\em CVPR}, 2018.

\bibitem{Groueix-et-al-CVPR-2018}
T.~Groueix, M.~Fisher, V.~G. Kim, B.~C. Russell, and M.~Aubry.
\newblock {AtlasNet}: A papier-m{\^{a}}ch{\'{e}} approach to learning 3d
  surface generation.
\newblock In {\em CVPR}, 2018.

\bibitem{Guo-et-al-Arxiv-2019}
Y.~Guo, H.~Wang, Q.~Hu, H.~Liu, L.~Liu, and M.~Bennamoun.
\newblock Deep learning for 3d point clouds: A survey.
\newblock {\em arXiv preprint 1912.12033}, 2019.

\bibitem{Han-et-al-ICCV-2017}
X.~Han, Z.~Li, H.~Huang, E.~Kalogerakis, and Y.~Yu.
\newblock High-resolution shape completion using deep neural networks for
  global structure and local geometry inference.
\newblock In {\em ICCV}, 2017.

\bibitem{Han-et-al-CVPR-2019}
X.~Han, Z.~Zhang, D.~Du, M.~Yang, J.~Yu, P.~Pan, X.~Yang, L.~Liu, Z.~Xiong, and
  S.~Cui.
\newblock Deep reinforcement learning of volume-guided progressive view
  inpainting for 3d point scene completion from a single depth image.
\newblock In {\em CVPR}, 2019.

\bibitem{Hou-et-al-Arxiv-2019}
J.~Hou, A.~Dai, and M.~Nie{\ss}ner.
\newblock {3D-SIC}: 3d semantic instance completion for {RGB-D} scans.
\newblock {\em arXiv preprint 1904.12012}, 2019.

\bibitem{Hu-et-al-CVPR-2020}
Q.~Hu, B.~Yang, L.~Xie, S.~Rosa, Y.~Guo, Z.~Wang, N.~Trigoni, and A.~Markham.
\newblock Randla-net: Efficient semantic segmentation of large-scale point
  clouds.
\newblock In {\em CVPR}, 2020.

\bibitem{Hu-et-al-AAAI-2020}
T.~Hu, Z.~Han, and M.~Zwicker.
\newblock 3d shape completion with multi-view consistent inference.
\newblock In {\em AAAI}, 2020.

\bibitem{Huang-et-al-CVPR-2020}
Z.~Huang, Y.~Yu, J.~Xu, F.~Ni, and X.~Le.
\newblock Pf-net: Point fractal network for 3d point cloud completion.
\newblock In {\em CVPR}, 2020.

\bibitem{Kazhdan-et-al-SGP-2006}
M.~M. Kazhdan, M.~Bolitho, and H.~Hoppe.
\newblock Poisson surface reconstruction.
\newblock In A.~Sheffer and K.~Polthier, editors, {\em Eurographics Symposium
  on Geometry Processing}, 2006.

\bibitem{Kazhdan-Hoppe-SIGGRAPH-2013}
M.~M. Kazhdan and H.~Hoppe.
\newblock Screened poisson surface reconstruction.
\newblock {\em ACM TOG}, 32(3):29:1--29:13, 2013.

\bibitem{Kingma-Ba-ICLR-2015}
D.~P. Kingma and J.~Ba.
\newblock Adam: {A} method for stochastic optimization.
\newblock In Y.~Bengio and Y.~LeCun, editors, {\em ICLR}, 2015.

\bibitem{Ladicky-et-al-ICCV-2017}
L.~Ladicky, O.~Saurer, S.~Jeong, F.~Maninchedda, and M.~Pollefeys.
\newblock From point clouds to mesh using regression.
\newblock In {\em ICCV}, 2017.

\bibitem{Li-et-al-NIPS-2018}
Y.~Li, R.~Bu, M.~Sun, W.~Wu, X.~Di, and B.~Chen.
\newblock {PointCNN}: Convolution on x-transformed points.
\newblock In {\em NeurIPS}, 2018.

\bibitem{Lin-et-al-Arxiv-2020}
Y.~Lin, Z.~Yan, H.~Huang, D.~Du, L.~Liu, S.~Cui, and X.~Han.
\newblock {FPConv}: Learning local flattening for point convolution.
\newblock {\em arXiv preprint 2002.10701}, 2020.

\bibitem{Minghua-et-al-AAAI-20}
M.~Liu, L.~Sheng, S.~Yang, J.~Shao, and S.~Hu.
\newblock Morphing and sampling network for dense point cloud completion.
\newblock In {\em The Thirty-Fourth {AAAI} Conference on Artificial
  Intelligence, {AAAI} 2020, The Thirty-Second Innovative Applications of
  Artificial Intelligence Conference, {IAAI} 2020, The Tenth {AAAI} Symposium
  on Educational Advances in Artificial Intelligence, {EAAI} 2020, New York,
  NY, USA, February 7-12, 2020}, pages 11596--11603. {AAAI} Press, 2020.

\bibitem{Mescheder-et-al-CVPR-2019}
L.~M. Mescheder, M.~Oechsle, M.~Niemeyer, S.~Nowozin, and A.~Geiger.
\newblock Occupancy networks: Learning 3d reconstruction in function space.
\newblock In {\em CVPR}, 2019.

\bibitem{Misra-Arxiv-2019}
D.~Misra.
\newblock Mish: A self regularized non-monotonic neural activation function.
\newblock {\em arXiv preprint 1908.08681}, 2019.

\bibitem{Park-et-al-CVPR-2019}
J.~J. Park, P.~Florence, J.~Straub, R.~A. Newcombe, and S.~Lovegrove.
\newblock {DeepSDF}: Learning continuous signed distance functions for shape
  representation.
\newblock In {\em CVPR}, 2019.

\bibitem{Pauly-et-al-SGP-2005}
M.~Pauly, N.~J. Mitra, J.~Giesen, M.~H. Gross, and L.~J. Guibas.
\newblock Example-based 3d scan completion.
\newblock In M.~Desbrun and H.~Pottmann, editors, {\em Eurographics Symposium
  on Geometry Processing}, 2005.

\bibitem{Peyghambarzadeh-et-al-DSP-2020}
S.~M.~M. Peyghambarzadeh, F.~Azizmalayeri, H.~Khotanlou, and A.~Salarpour.
\newblock {Point-PlaneNet}: Plane kernel based convolutional neural network for
  point clouds analysis.
\newblock {\em Digit. Signal Process.}, 98:102633, 2020.

\bibitem{Qi-et-al-CVPR-2017}
C.~R. Qi, H.~Su, K.~Mo, and L.~J. Guibas.
\newblock {PointNet}: Deep learning on point sets for 3d classification and
  segmentation.
\newblock In {\em CVPR}, 2017.

\bibitem{Qi-et-al-NIPS-2017}
C.~R. Qi, L.~Yi, H.~Su, and L.~J. Guibas.
\newblock {PointNet++}: Deep hierarchical feature learning on point sets in a
  metric space.
\newblock In {\em NeurIPS}, 2017.

\bibitem{Ramamonjisoa-et-al-ICCV19}
M.~Ramamonjisoa and V.~Lepetit.
\newblock {SharpNet}: Fast and accurate recovery of occluding contours in
  monocular depth estimation.
\newblock In {\em ICCV}, 2019.

\bibitem{Ronneberget_CoRR_2015}
O.~Ronneberger, P.~Fischer, and T.~Brox.
\newblock {U-Net}: Convolutional networks for biomedical image segmentation.
\newblock {\em arXiv preprint 1505.04597}, 2015.

\bibitem{Kripasindhu-et-al-SGP-2018}
K.~Sarkar, F.~Bernard, K.~Varanasi, C.~Theobalt, and D.~Stricker.
\newblock {Denoising of Point-clouds Based on Structured Dictionary Learning}.
\newblock In {\em Eurographics Symposium on Geometry Processing}, 2018.

\bibitem{Sarmad-et-al-CVPR-2019}
M.~Sarmad, H.~J. Lee, and Y.~M. Kim.
\newblock Rl-gan-net: A reinforcement learning agent controlled gan network for
  real-time point cloud shape completion.
\newblock In {\em The IEEE Conference on Computer Vision and Pattern
  Recognition (CVPR)}, June 2019.

\bibitem{Schoeps-et-al-CVPR17}
T.~Schoeps, J.~Schönberger, S.~Galliani, T.~Sattler, M.~Pollefeys, and
  A.~Geiger.
\newblock A multi-view stereo benchmark with high-resolution images and
  multi-camera videos.
\newblock In {\em CVPR}, 2017.

\bibitem{Sitzmann-et-al-NIPS-2019}
V.~Sitzmann, M.~Zollh{\"{o}}fer, and G.~Wetzstein.
\newblock Scene representation networks: Continuous 3d-structure-aware neural
  scene representations.
\newblock In {\em NeurIPS}, 2019.

\bibitem{Song-et-al-CVPR-2017}
S.~Song, F.~Yu, A.~Zeng, A.~X. Chang, M.~Savva, and T.~A. Funkhouser.
\newblock Semantic scene completion from a single depth image.
\newblock In {\em CVPR}, 2017.

\bibitem{Speciale-et-al-ECCV-2016}
P.~Speciale, M.~R. Oswald, A.~Cohen, and M.~Pollefeys.
\newblock A symmetry prior for convex variational 3d reconstruction.
\newblock In {\em ECCV}, 2016.

\bibitem{Stutz-Geiger-CVPR-2018}
D.~Stutz and A.~Geiger.
\newblock Learning 3d shape completion from laser scan data with weak
  supervision.
\newblock In {\em CVPR}, 2018.

\bibitem{Su-et-al-CVPR-2018}
H.~Su, V.~Jampani, D.~Sun, S.~Maji, E.~Kalogerakis, M.~Yang, and J.~Kautz.
\newblock {SPLATNet}: Sparse lattice networks for point cloud processing.
\newblock In {\em CVPR}, 2018.

\bibitem{Tatarchenko-et-al-CVPR-2018}
M.~Tatarchenko, J.~Park, V.~Koltun, and Q.~Zhou.
\newblock Tangent convolutions for dense prediction in 3d.
\newblock In {\em CVPR}, 2018.

\bibitem{Tchapmi-et-al-CVPR-19}
L.~P. Tchapmi, V.~Kosaraju, H.~Rezatofighi, I.~Reid, and S.~Savarese.
\newblock Topnet: Structural point cloud decoder.
\newblock In {\em Proceedings of the IEEE/CVF Conference on Computer Vision and
  Pattern Recognition (CVPR)}, June 2019.

\bibitem{Terzopoulos-et-al-IJCV-1987}
D.~Terzopoulos, A.~Witkin, and M.~Kass.
\newblock Symmetry-seeking models and 3d object reconstruction.
\newblock {\em International Journal of Computer Vision}, 1:211--221, 1987.

\bibitem{Thomas-et-al-ICCV-2019}
H.~Thomas, C.~R. Qi, J.~Deschaud, B.~Marcotegui, F.~Goulette, and L.~J. Guibas.
\newblock {KPConv}: Flexible and deformable convolution for point clouds.
\newblock In {\em ICCV}, 2019.

\bibitem{Wang-et-al-CVPR-2020}
X.~Wang, M.~H. A.~J. ~, and G.~H. Lee.
\newblock Cascaded refinement network for point cloud completion.
\newblock In {\em CVPR}, 2020.

\bibitem{Wang-et-al-ICCV-19}
Y.~Wang, D.~J. Tan, N.~Navab, and F.~Tombari.
\newblock Forknet: Multi-branch volumetric semantic completion from a single
  depth image.
\newblock In {\em Proceedings of the IEEE International Conference on Computer
  Vision}, pages 8608--8617, 2019.

\bibitem{Wen-et-al-CVPR-20}
X.~{Wen}, T.~{Li}, Z.~{Han}, and Y.~S. {Liu}.
\newblock Point cloud completion by skip-attention network with hierarchical
  folding.
\newblock In {\em 2020 IEEE/CVF Conference on Computer Vision and Pattern
  Recognition (CVPR)}, pages 1936--1945, 2020.

\bibitem{Wu-et-al-CVPR-2019}
W.~Wu, Z.~Qi, and F.~Li.
\newblock {PointConv}: Deep convolutional networks on 3d point clouds.
\newblock In {\em CVPR}, 2019.

\bibitem{Xu-et-al-NIPS-2019}
Q.~Xu, W.~Wang, D.~Ceylan, R.~Mech, and U.~Neumann.
\newblock {DISN:} deep implicit surface network for high-quality single-view 3d
  reconstruction.
\newblock In {\em NeurIPS}, 2019.

\bibitem{Yuan-et-al-3DV-2018}
W.~Yuan, T.~Khot, D.~Held, C.~Mertz, and M.~Hebert.
\newblock {PCN:} point completion network.
\newblock In {\em 3DV}, 2018.

\bibitem{Zhang-et-al-ECCV-2018}
J.~Zhang, H.~Zhao, A.~Yao, Y.~Chen, L.~Zhang, and H.~Liao.
\newblock Efficient semantic scene completion network with spatial group
  convolution.
\newblock In {\em ECCV}, 2018.

\end{thebibliography}
}

\end{document}